# PLURALITY AND QUANTIFICATION IN GRAPH REPRESENTATION OF MEANING

By

YU CAO

A dissertation submitted to the

School of Graduate Studies

Rutgers, The State University of New Jersey

In partial fulfillment of the requirements

For the degree of

Doctor of Philosophy

Graduate Program in Linguistics

Written under the direction of

Simon Charlow

And approved by

New Brunswick, New Jersey

October, 2021

ABSTRACT OF THE DISSERTATION

Plurality and Quantification in Graph Representation of Meaning

by YU CAO

Dissertation Director:
Simon Charlow

The study of meaning is inseparable from that of semantic representation, as design efforts in the latter exert far-reaching implications for linguistics and related computation. In this thesis, we present a representation formalism based on directed graphs and explore its explanatory benefits in application to classic issues in plurality and quantification, two aspects of natural language semantics treated in previous graph formalisms with varied linguistic adequacy.

Our graph language (Chapter 2) covers the essentials of natural language semantics (thematic relations, modification, co-reference, intensionality, plurality, quantification, and coordination) while using only monadic second-order variables. We show that the model-theoretical interpretation of this language can be defined in terms of graph traversal, where the relative scope of variables arises from their order of valuation.

We present a unification-based mechanism for constructing semantic graphs at a simple syntax-semantics interface (Chapter 3), whose task is to decide equivalence among discourse referents introduced by linguistic tokens, through syntax and through non-syntactic resolutions. Syntax is then formulated as a deterministic partition function on discourse referents. By establishing a partly deterministic relation between semantics and syntactic distribution, we show that this function finds a natural implementation in categorial grammars, owing to the way they manipulate syntactic resources. The syntax-semantics interface described here is automated to facilitate future exploration.

In applying the present graph formalism to selected topics in plurality and quantification (Chapters 4–5), we show that distributive predication of various forms (and even lack thereof) can be attributed to variants of a graph motif that performs quantification, and the partial determinism between semantics and syntactic distribution allows these variants to share roughly the same syntax. Our syntax-semantics interface offers streamlined solutions to compositional problems in cross-categorial conjunction and scope permutation of quantificational expressions. A scope taking strategy analogous to co-reference resolution is shown to simplify the treatment of exceptional scoping behaviors of indefinites.

# Acknowledgments

I would like to thank my thesis committee members, Simon Charlow, Adam Jardine, Paul Pietroski, Bruce Tesar, and Chris Barker, for their guidance and encouragement, before and throughout these unusual times. To Simon, my advisor, the only linguist I know who chose linguistics upon leaving no other profession, I am particularly grateful, for providing me with freedom to explore my research interests, referring me to the resources that benefited their development, dedicating his time in countless in-depth conversations and correspondence, and constantly reminding me of the standard of presentable writing.

Outside the committee, I wish to give thanks to Michael White, whose feedback helped this work through its formative stage; to Sorcha Gilroy and Adam Lopez, from whose NASSLLI course this work took its origin; to Percy Liang, who shared his thoughts and manuscript on semantic representation; and to Alexander Shen, who answered my mathematical questions.

Over the past few years, I am more than fortunate to have met and learnt from many people. Besides those already noted, I have Akinbiyi Akinlabi, Mark Baker, Veneeta Dayal, Ken Safir, Maria Bittner, Jane Grimshaw, Kristen Syrett, Troy Messick, and Viviane Déprez to thank for my linguistics education. Collaborations with Sam Bowman, Elliott Ash, and Daniel Chen introduced me to computational linguistics and machine learning, changed forever the career to which I aspire, and prepared me for the pursuit. Luca Iacoponi, Ümit Atlamaz, Scott Meredith, and Yimei Xiang have been excellent mentors when I faced the trials in transitioning into the industry.

As it draws to a close, I fondly remember two among those who sent me on this journey: Gladys Tang, who hired a research assistant mainly for him to prepare for PhD applications; and Yusuke Kubota, who wrote a student's recommendations after teaching him for only two weeks at the LSA Summer Institute. And I will cherish my memories with all those who journeyed with me, my brilliant fellow linguists and friends at Rutgers and elsewhere. I mention especially my cohort, Chris Oakden, Shiori Ikawa, and Nate Koser, who made one's graduate school days bearable when they seemed not; and mi colega, Ben Kinsella, who shows that beyond the graduate school, one certainly has a life to look forward to.

Finally, how can I thank you, mom and dad, who have always been there for me, across all lands and seas.

# Contents

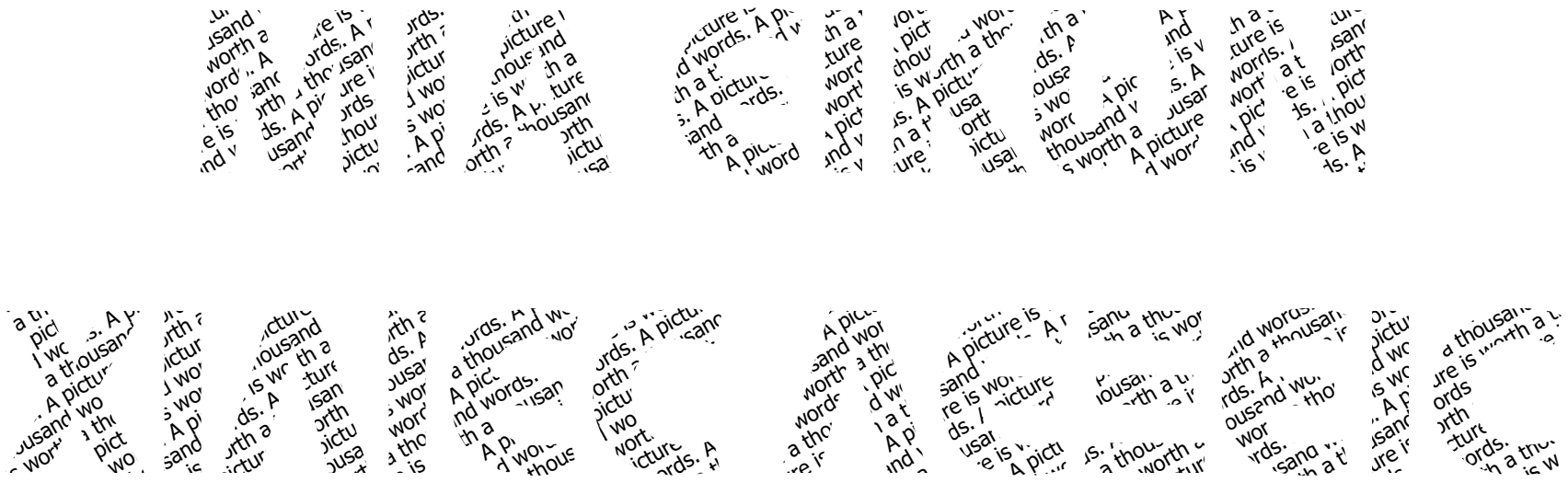

# Chapter 1

# Introduction

This thesis embarks on a systematic investigation of empirical applications of graph representation of meaning. Graphs, a structure dedicated to relational data, are well suited to the relational nature of natural language semantics. Presented here is a graph formalism that possesses useful complexity properties while explicitly representing quantification and coordination, together with its model-theoretical interpreter that takes into account plurality and intensionality. A unification-based mechanism is responsible for constructing semantic graphs at an innovative syntax-semantics interface.

We obtain interesting results in applying this formalism to classic topics in plurality and quantification. The semantic construction mechanism is automated to support future exploration. As we take a less trodden path, this introductory chapter provides the background in which our endeavor unfolds.

## 1.1 Representation matters

Proper representation of meaning lies at the heart of formal semantics, natural language processing, and every discipline that studies meaning. Of course, one's perspective on what counts towards proper representations hinges on the purpose one wants them to serve, but there can still be universal desiderata. Let us review what those could be by considering two major representation paradigms, one symbolic and the other numeric.

Higher-order logic (*first-order* restricts variables to those over entities; where *monadic second-order* adds variables over sets of entities, minus "monadic" one has variables over sets of tuples; with *higher-order* go variables of any finite order) became the lingua franca of formal semantics since Montague (1973). In Heim and Kratzer's (1998) influential textbook (and the literature following its convention), which directly takes often transformed constituency parses (known as *Logic Forms* or LFs) as semantic representations, higher-order logic still serves to record the denotations of (abstract) constituents.

What syntactic properties of this language justify its use is not usually questioned, but its expressive power and compositional transparency do work well towards the research goals of theoretical investigations. For example, (1.1) pairs a sentence with its neo-Davidsonian (Davidson, 1967) translation.

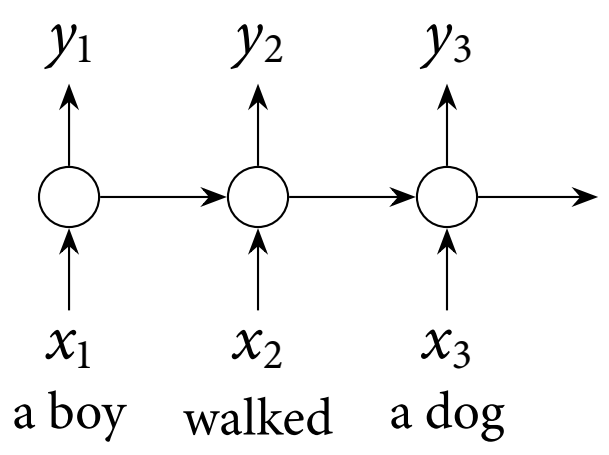

**Figure 1.1:** RNN encoder.

(1.1) A boy walked a dog.

$$\exists xyv \text{ boy}(x) \wedge \text{ag}(v, x) \wedge \text{walk}(v) \wedge \text{th}(v, y) \wedge \text{dog}(y)$$

The formula here can be built from parts in $\lambda$-calculus terms:

(1.2) a. *a boy* $\lambda P\, \exists x \text{ boy}(x) \wedge P(x)$
b. *walk* $\lambda xy\, \exists v \text{ ag}(v, x) \wedge \text{walk}(v) \wedge \text{th}(v, y)$

This is surely an easy example. In theory and in practice, logical expressions assigned to pieces of meaning can be arbitrarily complex in terms of the types of their own and of the variables involved. But the expressivity comes with a computational challenge.

In the context of semantic parsing, inducing token representations like (1.2) from those of sentences like (1.1) requires solving higher-order unification (see Section 1.5), a problem undecidable in general (Huet, 1973), even when restricted to second-order logic (Goldfarb, 1981). If any decomposition hypothesis about sentence semantics is nonviable, no algorithm is to tell that within finite time. One can rephrase this concern in the context of language acquisition, assuming that semantics observable to children is carried by sentences (or for that matter, standalone phrases). If mental representations of semantics are formally expressible in higher-order logic, one should wonder how lexicon acquisition is possible, unless the unification problem is restricted to a computationally tractable subform. However, formally characterizing the latter with linguistic justification is far from trivial. To date we have only seen heuristics guided by practical concerns (see Kwiatkowski et al., 2010) or restrictions identified within a particular linguistic phenomenon (see Dalrymple et al., 1991).

On the other hand, engineering often asks for coverage broader than the language fragments for which higher-order logic is employed. Scalability concerns lead modern NLP to neural network based sentence encoding (Devlin et al., 2019; Kiros et al., 2015; Peters et al., 2018) that stems from distributional semantics (Firth, 1968; Harris, 1954). This approach obviates the need of semantic parsing but builds on word embedding (Deerwester et al., 1990; Mikolov et al., 2013; Rumelhart et al., 1986), which effectively maps words to points in a real-valued vector space, such that similar words as measured by similar distributions settle on proximate points. Sentential semantics, again a vector, is then computed as a combination of word embeddings according to a particular network architecture.

Figure 1.1 illustrates a recurrent neural network (Elman, 1991), where the same computation unit is shared across time steps. With arrows indicating the flow of information, $x_t$ denotes the embedding vector that serves as the input at the step $t$, and $y_t$ denotes the output at the step $t$ computed as a function of $x_t$ and the state information of the computation unit from the previous step $t-1$. The output at the final step (or a function of outputs of all steps) is then taken as the semantic representation of the input sentence.

For all the success of neural sentence encoders in semantic tasks (see Bender and Koller, 2020 for a critical review), this approach does not lend to interpretable compositionality. Even as a sentence vector is composed from those of its tokens, there is a clear sense that their relation is not as transparent as that between (1.1) and (1.2).

Less obvious is how such vectors relate to extra-linguistic models of state-of-affairs. Whatever they implicitly encode, their interpretation has to be retrieved in a *specific* machine learning task, where the sentence encoder feeds the corresponding learning model. To show modern sentence encoders can capture information needed for natural logic inference (MacCartney and Manning, 2009), for example, Bowman (2016) trained a neural classifier that takes two sentence encodings and returns their semantic relation. The situation is quite different from that of logic, which is a self-contained system equipped with model-theoretical interpretation of its formulas and the mechanism for inferring their semantic relations, both among the core explananda of theoretical semantics.

## 1.2 Why graphs

The previous section shows how expressive power, compositional transparency, computational tractability, and formal interpretability can be reasonable desiderata. As an effort to reconcile the latter two, directed graph-based formalisms are gaining increasing attention in recent computational linguistics studies (Baldridge and Kruijff, 2002; Banarescu et al., 2013; Berglund et al., 2017; Bos, 2016; Bos et al., 2017; Kalouli and Crouch, 2018; Kruijff, 2001; Kuhlmann and Oepen, 2016; Liang et al., 2013; Stabler, 2018; White, 2006).

But in search of (symbolic) representation formalisms with sufficient expressive power and useful complexity properties, one is brought to study the "syntax of semantics". To the field of formal semantics, if discourse representation theory (DRT; Kamp, 1981; see for 2.3.3 a review) and dynamic semantics ever since (see Dekker, 2011 for a review) have taught any lesson beyond those about themselves, one would be how fruitful this course of study can be.

So, what of graphs? Graph formalisms can be as expressive as higher-order logic, since higher-order logic and $\lambda$-calculus have their graph representations (e.g. Buliga, 2013; Paliwal et al., 2020). However, graphs are flexible enough to accept complexity restrictions while preserving expressivity:

i) *Flexible interpretation*. Graphs only encode the relations between objects. The interpretation of those objects and the relations between is subject to each use case;
ii) *Flexible construction*. Graph construction is independent of the interpretation of graphs.

Levy et al. (2004) proved that monadic second-order unification is decidable. It turns out that with the flexibility of graphs, one can describe a good amount of natural language semantics using only monadic second-order variables. By contrast, when posed on logic, the same restriction would extensively affect the major existing theories of syntax-semantics interface, where prevalence of higher-order variables follows immediately from a functional design of semantic construction.

Graph formalisms can be of interest for their explanatory benefits. Graphs naturally represent the relational network of discourse referents that characterizes natural language semantics. Their model-theoretical interpretation can be defined analogously to that of logic or DRT, and they can be constructed using a unification-based mechanism. We will see how these ideas provide new perspectives on familiar issues from an empirical domain whose choice is motivated in Section 1.3, how they deal with compositional challenges that would be difficult to handle otherwise.

For the future, once graph representation of meaning reaches a reasonable level of empirical adequacy, we may study their structural properties from graph- and language-theoretical points of view. Given the results in graph grammars achieved in the past few decades (Courcelle and Engelfriet, 2012, a.o.), we may examine the complexity properties of the grammars that generate semantic graphs and the empirical implications thereof. Such inquiries have been around in works on computational phonology and syntax (Heinz, 2010; Heinz and Idsardi, 2013; De Santo and Graf, 2019, e.g.), and traces its root to the beginning of the generative enterprise (see Chomsky, 1956, 1959).

## 1.3 Empirical domain

We claim that natural language semantics encodes the relations or dependencies between discourse referents. Less prefixes the formula in (1.1) describes a relational structure also describable by a graph. Compare (1.3a) with (1.3b): the correspondence between variables and vertices, between predicates and edges is self-explanatory (see Section 1.6 for terminology).

(1.3) a. $\text{boy}(x) \wedge \text{ag}(v, x) \wedge \text{walk}(v) \wedge \text{th}(v, y) \wedge \text{dog}(y)$

b. 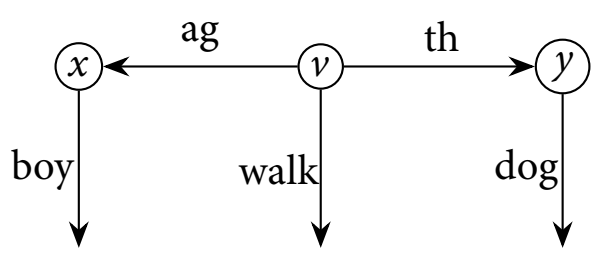

But it is not immediately clear how plurality and quantification fit into this simple relational picture.

Consider how we want to represent the semantics of the following sentence, which has a plural subject. It is imaginable to duplicate (1.3b) as follows.

(1.4) Two boys (each) walked a dog.

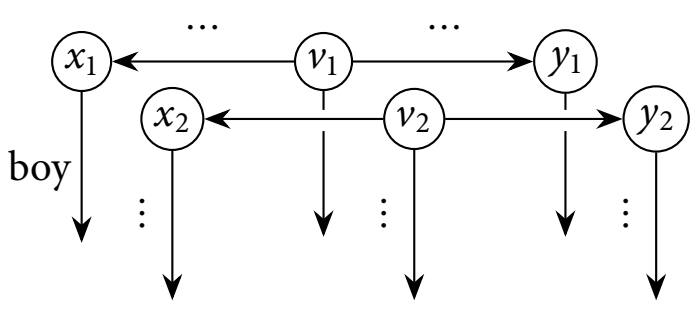

This is analogous to squeezing plurality into first-order logic — besides overheads needed to ensure the desired interpretation (e.g. $x_1 \neq x_2$), structure multiplication does not make a good generalization of plurality; it cannot represent plurals of an unknown number (e.g. *many boys*). And we can say exactly the same thing about quantification. For (1.5a), a representation like (1.4) is inconceivable because the number of boys may vary from situation to situation. For (1.5b), as there are infinitely many even numbers, a representation of an infinite size would be intractable.

(1.5) a. Every boy walked a dog.
b. Every even number is divisable by 2.

Plurality and quantification are indeed two crucial aspects of natural language semantics, calling for a generic treatment if they are to be contained in finite-size representations. This seems one of the rationales behind the development of plural logical semantics (Boolos, 1984; Link, 1983; Simons, 1982, a.o.) and generalized quantifiers (Barwise and Cooper, 1981; Montague, 1973; Mostowski, 1957, a.o.). However, given their respective research goals, existing graph formalisms differ considerably in the linguistic adequacy of their treatments of plurality and quantification. It is not uncommon to find

i) plurality or quantification left out of representation entirely (e.g. Banarescu et al., 2013);
ii) *or* quantification scope underspecified in representation, awaiting an interpretation procedure to freely resolve scope ambiguity (e.g. Bos, 2016; Lai et al., 2020; Stabler, 2018);
iii) *or* quantification scope resolved in representation, but less a syntax-semantics interface that constructs those resolutions compositionally (e.g. Liang et al., 2013; Pustejovsky et al., 2019; Schuler and Wheeler, 2014).

Therefore this thesis will develop a graph formalism suitable for linguistic investigations of plurality and quantification, while not losing sight of other essentials of natural language semantics. Before giving a high-level sketch of this formalism's interpretation and construction modules in Sections 1.4–1.5, we want to say more about the specific phenomena that we will discuss within the broader domain of plurality and quantification to establish the descriptive and explanatory adequacy of our formalism.

These include, for plurality, the ways of distributing predication over it and the ways of creating it with conjunction, and for quantification, the scope taking behaviors of quantifiers and indefinites. For the most part, what ties this selection together are two central themes in working with graphs to approach compositional problems:

i) a *partly* deterministic procedure for aligning semantics resources with syntactic ones;
ii) the *same* format for recording the outputs of syntactic derivations and non-syntactic resolutions.

The first allows not only the semantics of a linguistic expression to be partly inferred from its syntactic distribution, but also variants of a graph motif of similar semantic import to share roughly the same syntax. These ideas are particularly useful in studying distributive predication of various forms (and even lack thereof), which all perform quantification in some way. The second, on the other hand, abstracts semantic composition away from the details of syntactic derivations, and thereby simplifies the treatment of compositional problems in cross-categorial conjunction and scope permutation of quantificational expressions. Feeding semantic composition with separate but combinable outputs from syntax and non-syntactic resolutions further simplifies the treatment of exceptional scope of indefinites.

## 1.4 Interpretation

Semantic representations are, by definition, subject to interpretation. This can be done directly, by defining an interpreter that executes a representation to perform actions or return values; or indirectly, by translating representations of one formalism to those of another for which a direct interpreter is available. For example, higher-order logic formulas receive a direct interpretation, either for a model-theoretical truth value or for an update on information states, while DRSs (of DRT) can be interpreted directly or translated to first-order logic formulas (Kamp and Reyle, 1993).

Instead of working with representations of representations , or in favor of parsimony, we will build a direct model-theoretical interpreter for the graph language to be developed in Chapter 2, while leaving its capacity to update discourse for the future (c.f. Bonial et al., 2020; Kruijff, 2001). For a simple graph like (1.3b), its interpretation is essentially about valuating vertices to satisfy the constraints given by labeled edges, just as the interpretation of the formula (1.3a) is about valuating variables to satisfy the constraints given by predicates. But since graphs are *flat*, by which we mean they do not display explicit hierarchical structures as trees do, it is nontrivial how they may give rise to scope-sensitive semantics like quantification. As we will see, by carefully designing the graph language itself and a graph traversal procedure that valuates vertices and picks up edge constraints in due order, we can nonetheless delimit substructures in graphs, talk about their valuation dependency, and thereby retrieve scope-sensitive semantics.

Because of the difficulty implied by flatness, graph languages in previous works have been encoded in term languages, i.e. trees, and it is for such tree encodings that direct or indirect interpreters are defined. For example, Bos (2016); Stabler (2018) translate the tree encoding of augmented abstract meaning representation (AMR; Banarescu et al., 2013) to logic. Kruijff (2001) provides a direct interpreter for hybrid logic dependency semantics (HLDS) formulas, which are a tree encoding of graphs, while White (2006) translates them to DRSs.

## 1.5 Construction

Graphs belong to a broader family of constraint-satisfaction formalisms (Copestake et al., 2001; Baldridge and Kruijff, 2002; Moore, 1989; Shieber, 1986; Zeevat, 1988), where a common approach to semantic construction is that of unification.

In its original sense, *unification* refers to the process of equating symbolic formulas by finding a system of substitutions of their free variables. For example, the equation

$$f(g) \equiv \exists x y v\ \mathrm{boy}(x) \wedge \mathrm{ag}(v, x) \wedge \mathrm{walk}(v) \wedge \mathrm{th}(v, y) \wedge \mathrm{dog}(y)$$

can be solved up to $\beta$-reduction by

$$\begin{aligned} f &= \lambda P\ \exists x\ \mathrm{boy}(x) \wedge P(x), \\ g &= \lambda x\ \exists y v\ \mathrm{ag}(v, x) \wedge \mathrm{walk}(v) \wedge \mathrm{th}(v, y) \wedge \mathrm{dog}(y). \end{aligned}$$

Unification as such is one of the key ideas behind, say, logic programming and type inference in functional programming.

In unification-based grammars, variable substitutions are determined not to equate but to combine semantic representations. A desirable combination often unifies variables separately named and constrained. For example, the constraints in (1.3a) are a union of (1.6a-c) with $x = x'$ and $y = y'$:

(1.6) a. $\text{boy}(x)$ b. $\text{ag}(v, x') \wedge \text{walk}(v) \wedge \text{th}(v, y')$ c. $\text{dog}(y)$

In the language of graphs, the same process that produces (1.3b) can be phrased as gluing (1.7a-c) together by fusing $x$ with $x'$ and $y$ with $y'$, now as vertices:

(1.7) a. 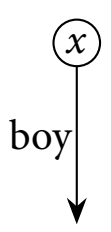

b. 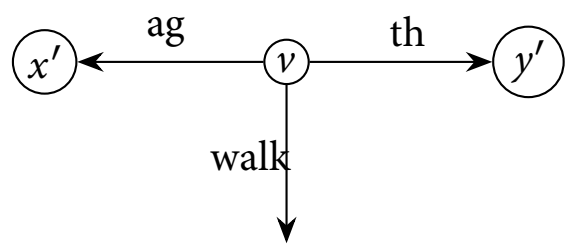

c. 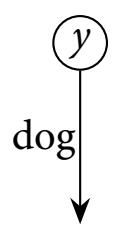

Thus semantic construction reduces to unifying equivalent discourse referents. In Section 1.6.3, graph-gluing is implemented as an operation known as parallel composition from hyperedge replacement (HR) algebra (Courcelle, 1993).

The task of syntax-semantics interface is then to decide equivalence among discourse referents — which vertices to fuse with which. In Chapter 3, we will see that non-syntactic resolutions *and* syntactic type reductions in categorial grammars (CGs; Ajdukiewicz, 1935; Bar-Hillel, 1953) provide exactly the kind of information this decision needs (cf. Bittner, 2001, where non-syntactic resolutions are solely responsible). In fact, the bulk of this thesis adopts the combinatory branch of CGs (Barker and Shan, 2014; Steedman, 1992, 1996, 2011; Szabolcsi, 1992) for the sake of presentation, though the type-logical branch (Barker and Shan, 2014; Kubota and Levine, 2020; Moortgat, 2011; Morrill, 2011; Moot and Retoré, 2012) is no less applicable. The reason for CGs to be such an apt choice consists in the way they manipulate lexical resources coded as syntactic types, whose makeup can be put in natural correspondence with graph vertices.

Since graph unification is ignorant about interpretation of the vertices, we can freely restrict their valuation to lower-order types. The outcome of graph unification shows an accumulative view of compositionality, even more transparent than the case of function application. Whereas (1.7a-c) are each a subgraph of (1.3b) up to vertex names, neither (1.2a) nor (1.2b) is a substring of (1.1) due to $\beta$-reduction.

When converted to some form of symbolic encodings, graph(-like) representations can be constructed with $\lambda$-calculus along the functional paradigm, as showcased by Muskens (1996) for DRT and Artzi et al. (2015) for AMR. This approach not only reintroduces higher-order variables; to translate subsentential expressions, it further creates $\lambda$-calculus terms that are alien to the syntax of DRT or AMR proper. By contrast, with unification, all linguistic expressions consistently denote graphs.

## 1.6 Graph theory background

This chapter closes with an introduction to the language of graph theory, including concepts of graphs in themselves, those relevant to semgraph interpretation, and those relevant to semantic construction. We recommend Section 1.6.1 to start the next chapter. Sections 1.6.2 and 1.6.3 can wait till Section 2.2 and Chapter 3.

Except for Definitions 1.2, 1.3 and 1.6, the rest are largely standard graph-theoretic constructs. The presentation follows Courcelle and Engelfriet (2012).

### 1.6.1 Semgraphs

Semantic graphs, or semgraphs for short, are directed simple graphs with labeled edges.

**Definition 1.1.** A *semgraph* $G$ is a triple $(V, E, \gamma)$ where

i) $V$ is a set of vertices;
ii) $E$ is a set of (directed) unary and binary edges;
iii) $\gamma$ assigns each $e$ in $E$ an edge label.

Whenever necessary any $G$-specific construct can be indexed, e.g. $(V_G, E_G, \gamma_G)$.

A *unary edge* consists of an arrow off a vertex $u$, which is its *tail* and of which it is an *out-edge*. A *binary edge* consists of an arrow from a vertex $u$ to a vertex $v$. In that case the binary edge is an *out-edge* of $u$ and an *in-edge* of $v$, and $u$ and $v$ are its *tail* and *head*. Tails and heads are collectively called *endpoints* or *ends*. Obviously, the ends of any edge in $E$ are to be found in $V$. We write $T(u)$ for the subset of edges of $E$ tailed by $u$, that is, all the out-edges of $u$.

$G$ being *simple* means that there is no *loop*, a binary edge whose tail and head coincides. It also means that there is at most one unary edge off any vertex and there is at most one binary edge per direction between any pair of vertices. Since each edge is uniquely identifiable by its ends, we may write, for example, $\overrightarrow{u}$ for a unary edge and $\overrightarrow{uv}$ for a binary edge.

Edge labels include unary predicate symbols *boy*, *walk*, ∀, ... for unary edges and binary predicate symbols like *ag*, *th*, =, ... for binary edges. In some usage we prefix an edge with its edge label, e.g. *boy* $\overrightarrow{u}$ or *ag* $\overrightarrow{uv}$, so as to avoid periphrasis like "$\overrightarrow{u}$ such that $\gamma(\overrightarrow{u}) = boy$" or "$\overrightarrow{uv}$ such that $\gamma(\overrightarrow{uv}) = ag$". A labeled unary edge is in a sense equivalent to a vertex label. Thus we call vertices with a labeled unary edge *named* and those without *anonymous*.

### 1.6.2 For interpretation

Semgraph interpretation happens as we traverse a graph. One vertex *reaches* another through a *path*: a path from $u$ to $v$ of length $n \geq 1$ is a sequence of binary edges $\overrightarrow{a_0 a_1}, \ldots, \overrightarrow{a_{n-1} a_n}$, where $a_0 = u$, $a_n = v$, and $a_i = a_{i+1}$ for all $\overrightarrow{a_{i-1} a_i}$ next to $\overrightarrow{a_{i+1} a_{i+2}}$. The same path can be consisely denoted by $a_0 \to \cdots \to a_n$.

A path from $u$ to $v$ makes $u$ an *ancestor* of $v$ and $v$ a *descendant* of $u$. The set of vertices reachable from $u$ by a path of length 1 are the *successors* of $u$. Conversely, the set of vertices that reach $u$ by a path of length 1 are the *predecessors* of $u$.

Two paths are *parallel* if they start and end with the same vertices. Except when $n = 1$, we do not require all vertices $a_0, \ldots, a_n$ in a path be distinct.[1] A *cycle* is then a path whose starting vertex $a_0$ and ending vertex $a_n$ coincide. All cycles in semgraphs, we will see in Section 2.1.2, contain at least one $\lambda$ (labeled) edge (whose function need not concern us for now).

[1]Some authors require "paths" consist of distinct vertices and use "walks" for what we call "paths".

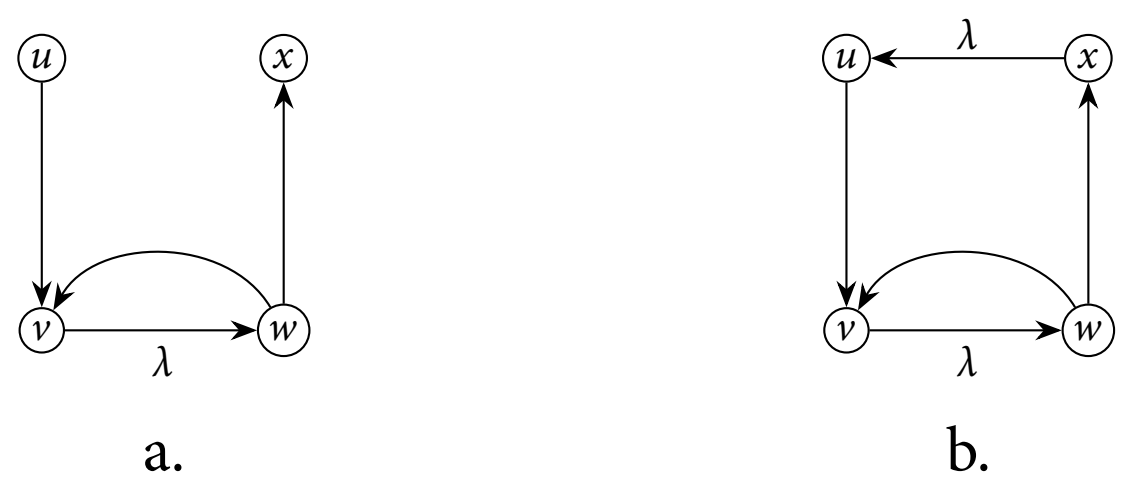

**Figure 1.2:** (a) Rooted at $u$. (b) Rooted at $x$.

**Definition 1.2.** We say $e$ *encycles* $e'$ if both edges lie in a common cycle and $e'$ would not be in any cycle without $e$ and, if $\gamma(e)$ is equality, any equality edge sharing its head with $e$.

A *rooted graph* has one or more vertices that are somehow distinguished from other vertices in the graph. For semgraphs we make the distinction structurally.

**Definition 1.3.** A *root* is a vertex $v$ such that either $v$ has no in-edge; or every in-edge $\overrightarrow{uv}$ encycles some out-edge $\lambda\,\overrightarrow{vw}$.

As we will see in Section 2.2, a semgraph is interpretable only if it has a unique root that reaches every other vertex. The rationale for defining roots in cyclic graphs will be given in Section 2.1.2. To illustrate, $u$ is the root of Figure 1.2a (the only vertex that has no in-edges is $u$; the in-edge $\overrightarrow{uv}$ of $v$ does not encycle $\lambda\overrightarrow{vw}$ because having $\overrightarrow{uv}$ removed, we still find $v \to w \to v$), whereas the added $\lambda\overrightarrow{xu}$ shifts the rootship to $x$ in Figure 1.2b (while every vertex has in-edges, $u$ and $w$ cannot be roots since they have no outgoing $\lambda$-edges; $v$ is not a root as $\overrightarrow{uv}$ does not encycle $\lambda\overrightarrow{vw}$ for the reasons just explained; the rootship of $x$ is given by the fact that $\overrightarrow{wx}$ encycles $\lambda\overrightarrow{xu}$, that is, without $\overrightarrow{wx}$, $\lambda\overrightarrow{xu}$ would not be in any cycle).

Semgraphs are constructed with an important property: every vertex reached by the root of $G$ is in turn the root of some induced subgraph of $G$.

**Definition 1.4.** $H$ is an (*induced*) *subgraph* of $G$, written as $H \subseteq G$, iff

i) $V_H \subseteq V_G$;
ii) $E_H$ is the set of edges of $E_G$that have their ends in $V_H$;
iii) $\gamma_H$ is restriction of $\gamma_G$ to $V_H$.

To induce a subgraph we may simply remove some edges.

**Definition 1.5.** $G - F$ where $F \subseteq E_G$ is defined by $V_{G-F} = V_G$ and $E_{G-F} = E_G - F$.

Another major induction is to find the subgraph reachable from some vertex (see Courcelle and Engelfriet, 2012 for a similar construct defined on trees).

**Definition 1.6.** $G/u$ is defined by $V_{G/u}$: the set of vertices of $V_G$ that contains $u$ and any $v$ reachable from $u$.

The two inductions can be combined. For example, Figure 1.2a is the subgraph reachable from $u$ in Figure 1.2b having $\overrightarrow{xu}$ removed.

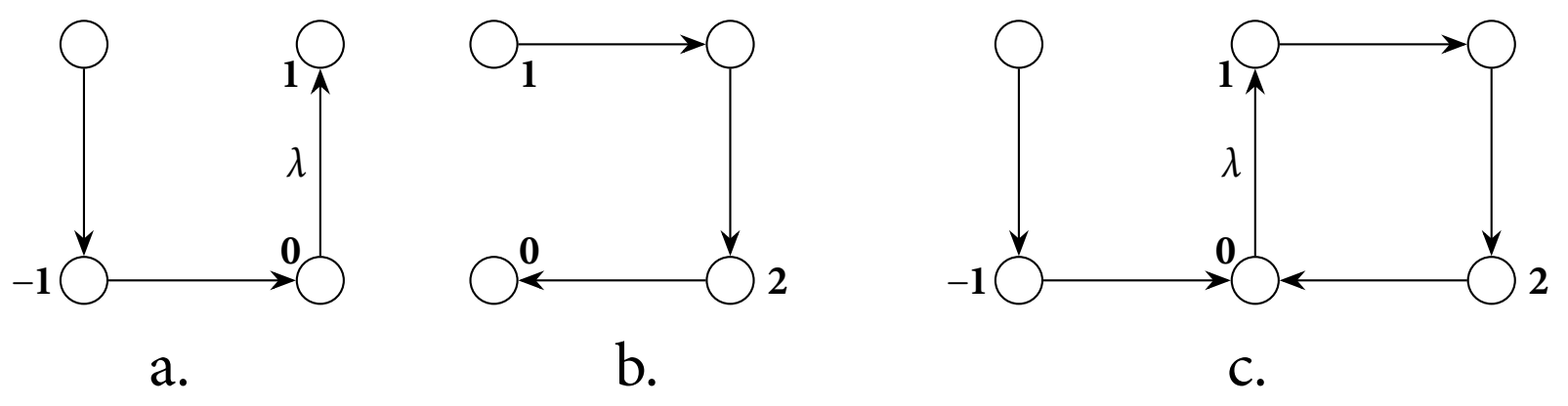

**Figure 1.3:** (a) $G$. (b) $H$. (c) $G \mathbin{/\!/} H$.

### 1.6.3 For construction

As said, semantic construction amounts to gluing disjoint graphs at distinguished vertices. The operation is managed by HR algebra.

We extend $G$ to a quadruple $(V, E, \gamma, \mathit{slab})$, where the fourth component *slab* assigns a subset of vertices of $V$ distinct *source labels*. We will see in later chapters that it suffices to use integers $(\mathbb{Z} = \{0, 1, -1, ...\})$ . We write $\mathit{Src}(G)$ for the domain of *slab*, calling these label-bearing vertices *sources*. If $\mathit{slab}(u) = n$, $u$ is the $n$-source of $G$. The number of non-negative sources of $G$ is denoted by $\sigma(G)$ and called the *sort* of $G$.

HR algebra performs source *renaming* (changing the source label of a vertex) and *forgetting* (removing a vertex from the domain of *slab*). It is nonetheless strictly resource consuming: what has been forgotten cannot be revived. Whatever sources left form the anchors for graph unification.

**Definition 1.7.** Let $G$ and $H$ be semgraphs with sources such that $V_G \cap V_H = \mathit{Src}(G) \cap \mathit{Src}(H)$ and $E_G \cap E_H = \varnothing$. The *parallel composition* of $G$ and $H$, written as $G \mathbin{/\!/} H$, is their union:

i) $V_{G/\!/H} = V_G \cup V_H$;
ii) $E_{G/\!/H} = E_G \cup E_H$;
iii) $\gamma_{G/\!/H} = \gamma_G \cup \gamma_H$;
iv) $\mathit{slab}_{G/\!/H} = \mathit{slab}_G \cup \mathit{slab}_H$.

The last condition implies that $\mathit{slab}_G$ and $\mathit{slab}_H$ agree on $\mathit{Src}(G) \cap \mathit{Src}(H)$, whereas the labels assigned by $\mathit{slab}_G$ to $\mathit{Src}(G) - V_H$ are disjoint from those assigned by $\mathit{slab}_H$ to $\mathit{Src}(H) - V_G$.

Parallel composition generalizes to any pair of $G$ and $H$ as we consider their isomorphic copies to which the above definition applies: the edge sets remain disjoint, but namesake sources coincide.

**Definition 1.8.** $G$ is *isomorphic* to $H$, written as $G \simeq H$, iff there are one-to-one correspondences (bijections) $f_V$ between vertex sets and $f_E$ between edge sets, such that for all $\vec{u}$, $\overrightarrow{uv}$ in $E_G$ and for all $w$ in $\mathit{Src}(G)$,

i) $f_E(\vec{u}) = \overrightarrow{f_V(u)}$;
ii) $f_E(\overrightarrow{uv}) = \overrightarrow{f_V(u) f_V(v)}$;
iii) $\gamma_G(\vec{u}) = \gamma_H(f_E(\vec{u}))$;
iv) $\gamma_G(\overrightarrow{uv}) = \gamma_H(f_E(\overrightarrow{uv}))$;
v) $\mathit{slab}_G(w) = \mathit{slab}_H(f_V(w))$.

In particular, with disjoint $G$ and $H$ we simply take their union and fuse vertices bearing the same source labels; Figure 1.3 gives an illustration. It is easy to see that parallel composition is associative, commutative, and therefore is well-defined for more than two graphs.

# Chapter 2

# Semgraphs

In this chapter, we present a graph language, built on abstract meaning representation (AMR; Banarescu et al., 2013) and elements of hybrid logic dependency semantics (HLDS; Baldridge and Kruijff, 2002; Kruijff, 2001; White, 2006), that covers thematic relations, modification, co-reference, intensionality, plurality, quantification, conjunction, and disjunction. Besides plurality and quantification, the central themes of this thesis, the other aspects of meaning seem too essential to be missed in any semantic formalism (where event semantics is not used, "thematic relations" can be replaced by "predicate-argument structure"), to say nothing of the linguistic practice of relating conjunction to plurality, and intensionality to quantification.

After presenting semgraphs and their model-theoretical interpretation, we will compare this language with AMR, HLDS, and DRT to highlight its characteristic properties.

## 2.1 Elements of semgraphs

We begin with an informal introduction to the building blocks of semgraphs we will be working with. This is to provide some intuitions before we take up in Section 2.2 a formal interpreter that delivers the semantics here attributed to semgraphs.

### 2.1.1 Thematic relation

As in most graph formalisms, we take a neo-Davidsonian view of predicate-argument structure, representing thematic relations as relating events, or eventualities, with their participants:

(2.1) Joe walked a dog.

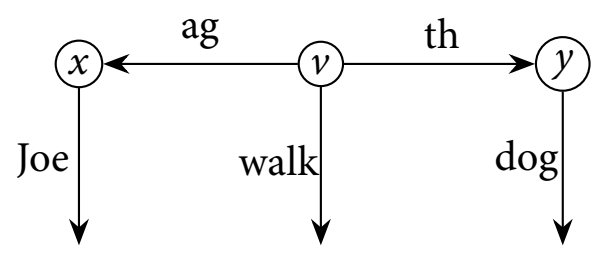

A similar semgraph (1.3) was likened to its formula counterpart. To read (2.1) directly, take vertices as discourse referents, or variables, and binary edges with thematic labels as thematic relations predicated of them. Usage of unary edges is due to Gilroy and Lopez (2018); their lexical labels are predicated of the variables. Thus (2.1) declares $x$ as Joe, $v$ a walking event, $y$ a dog, $v$ has $x$ as its agent and $y$ as its theme.

The above description might leave an impression that variables or vertices range over singular values like entities, but we actually want them to take plural values like sets of entities (see Link, 1983; Kamp and Reyle, 1993; Schwarzschild, 1996; Winter, 2001). Indeed, a claim made in the previous chapter is that it suffices to use monadic second-order variables to model the semantics discussed in this thesis. Given their singleton correspondents, singular values can be seen as a special case of plural ones. The distinction between singular and plural valuations is invisible in representation, so (2.1) passes for a semgraph of (2.2):

(2.2) Joe walked dogs.

The job is then to give sensible interpretation of thematic relations, no matter whether a set of events or a set of participants is a singleton or not. We will discuss plurality in more detail in Section 2.2.1.

As in DRT, our semgraphs indicate existence and logical conjunction implicitly: existence of variable $x$ is conveyed by presence of vertex $x$; co-occurrence of the edges implies the logical conjunction of the constraints they impose on $x$, $y$, and $v$. These two aspects of meaning seem fundamental and stand out from their counterparts: across languages it is much easier to find sentences made of a nominal being indefinites, as opposed to genuine quantifiers, and it is much readier to interpret asyndetic listing as conjunction, as opposed to disjunction:

(2.3) a. A dog (= *there is* a dog).
b. Veni, vidi, vici (= veni, *et* vidi, *et* vici).

The moral of such observations need not be taken seriously, but they should reduce our surprise that DRT has specific devices for genuine quantification and logical disjunction but not for existence or logical conjunction. The rationale is that meaning that can be implicitly indicated in natural language need not be explicitly encoded in representation.

Our semgraphs also conflate proper and common nouns, indefinite and definite descriptions as in DRT (Kamp et al., 2011, secs. 2.3, 4.2). There the assumption is that their distinctions need not concern this level of semantic representation, but lies in presuppositions about contexts, e.g. uniqueness of or familiarity with the target referent.

Before proceeding to semantic relations beyond thematic ones, it is worth asking why for the latter a neo-Davidsonian view is taken. There is nothing in semgraphs per se that necessitates this choice; for (2.1) one may devise (2.4b) in analogy with (2.4a):

(2.4) a. $\text{Joe}(x) \wedge \text{walk}(x, y) \wedge \text{dog}(y)$
b.

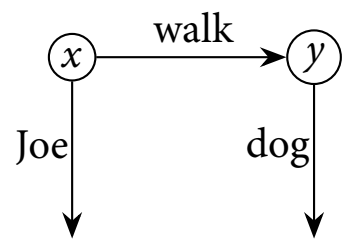

We will not review the arguments in favor of event semantics, but as we will see in Chapter 3, the structure of (2.1), not (2.4), fits better into our syntax-semantics interface.

### 2.1.2 Lambda edge

One event's participant may partake another event through modification, manifesting in its most general form as relative clauses. (2.5), for example, illustrates a situation where the walking and the seeing share the same dog as their themes. The semgraph does capture all the thematic relations involved.

(2.5) Ben saw a dog that Joe walked.

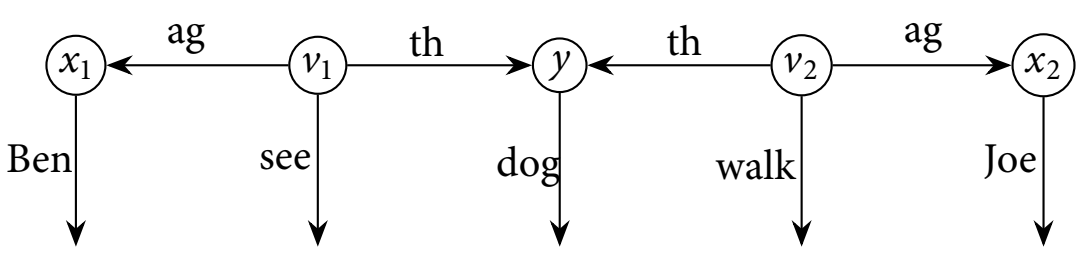

However, (2.5) features a structural peculiarity that comes with participant sharing: the semgraph here is multi-rooted, as both $v_1$ and $v_2$ are roots according to Definition 1.3. But graph interpretation, which proceeds with graph traversal (see Section 2.2.3), requires a semgraph be uniquely rooted. Starting from neither $v_1$ nor $v_2$ can we traverse the whole semgraph.

The problem can be fixed by introducing a vacuous semantic relation, represented as a $\lambda$-edge, directing from the modifyee $y$ to the modifier $v_2$. Now $v_1$ is *the* root of (2.6) that reaches every other vertex:

(2.6)

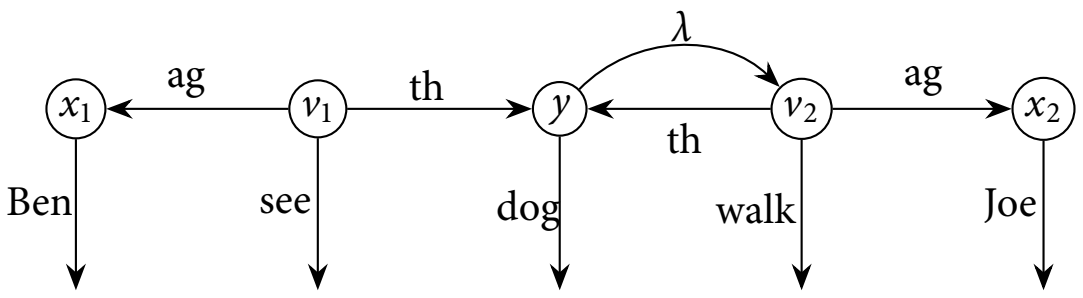

We may read the $\lambda$-edge as *such that*.

Observe that a $\lambda$-edge creates a $\lambda$-cycle, so named for its reminiscence of Buliga's (2013) graph representation of $\lambda$-calculus. Such cyclic structure of modification is due to White's (2006) adaption of Kruijff's (2001) *general relation*. While Kruijff and White treat the general relation on par with thematic relations, we take $\lambda$-edges as contributing no semantic constraint. Rather, a $\lambda$-edge has the effect of guiding graph traversal by bridging vertices and shifting vertex valuation order by shifting rootship (see Section 1.6.2). As we will see in later in Section 2.2.6, by reducing scope effects to the order of valuation, we can also use a $\lambda$-edge for scoping indefinites. It may be more than a coincidence that wide scope indefinites are often paraphrased with relative clauses:

(2.7) Every boy walked a dog. $(\exists\ \textit{dog} > \forall\ \textit{boy})$

= (There is) a dog that every boy walked.

Similarly $\lambda$-edges can be applied to adjectival and prepositional modifiers. We may assume that adjectives denote stative events or states and as in AMR, distinguish between a state itself vs. what is *in* a state. The following example thus represents the dog as the theme in luck.

(2.8) A lucky dog.

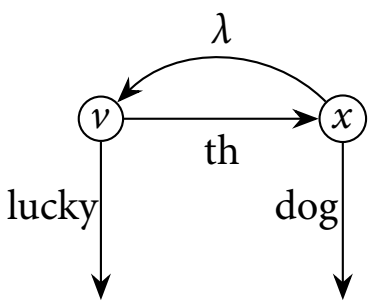

We may think of a circumstantial preposition too as denoting a state that thematically relates a location to a theme, which can be either an individual as in (2.9a) or an event as in (2.9b). The preposition itself specifies the nature of that relation.

(2.9) a. A dog in a park.

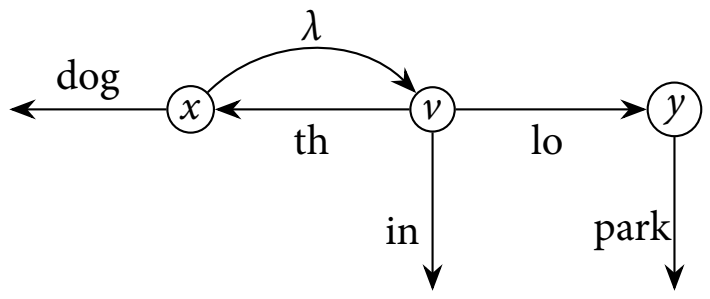

b. Joe walked in a park.

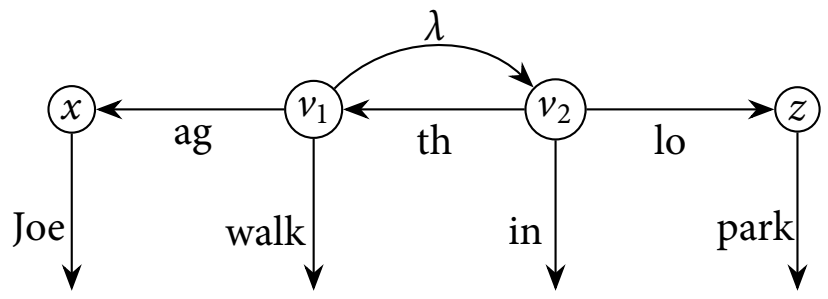

Possessive *of* can be rendered as introducing a state of owning:

(2.10) A dog of Joe.

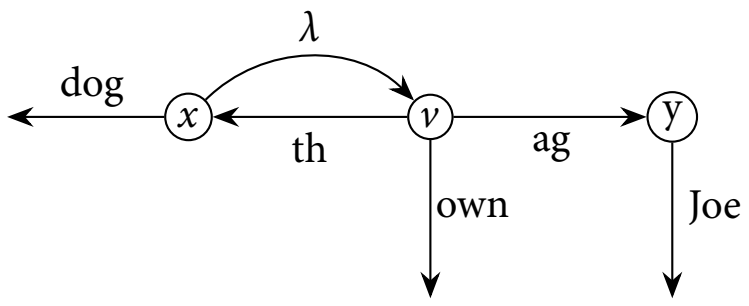

One may change the lexical or thematic labels here for possessives not necessarily indicating an ownership (e.g. *oracles of Delphi*).

### 2.1.3 Co-reference

Co-reference can be represented as equality between anaphors (including pronouns and reflexives) and their antecedents, following Kamp and Reyle (1993) and Liang (2012). Whereas Liang explicitly puts equality as an edge from an anaphor to its antecedent, we turn it the other way around. This is shown by the example below, where *himself* refers to Joe while $x$ points to $y$:

(2.11) Joe washed himself.

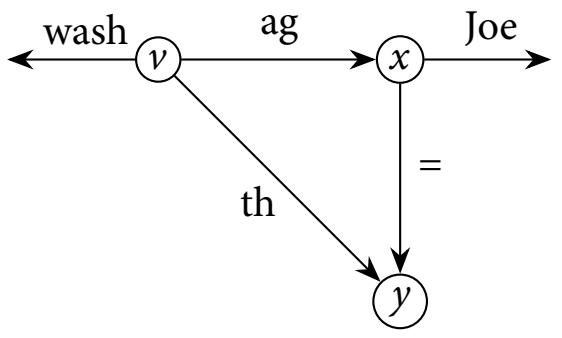

Possessive pronouns then combines co-reference and the owning state also appearing in (2.10):

(2.12) Joe walked his dog.

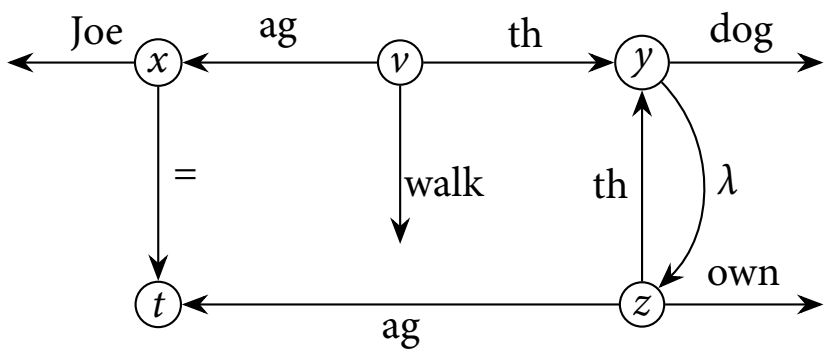

While equality is a symmetric relation, the equality edge is directed as above to deal with split anaphors as in the following example, where = $\overrightarrow{x_1z}$ and = $\overrightarrow{x_2z}$ are contributed by the anaphor (see Section 3.2.1).

(2.13) Joe washed and Ben shaved himself. (Alex Warstadt p.c.)

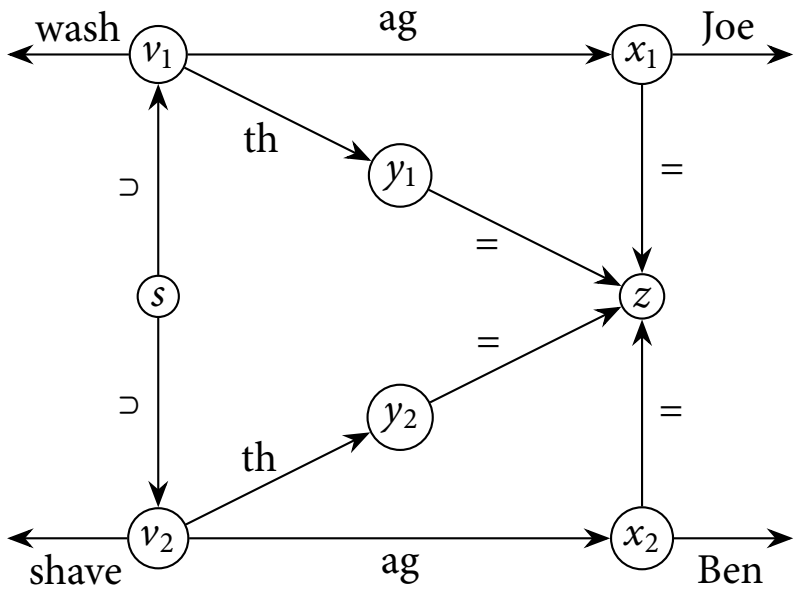

The coordination structure will be introduced in Section 2.1.7. When interpreting (2.13), we split it into the subgraph reachable from $v_1$ (2.14a) and the subgraph reachable from $v_2$ (2.14b) and read them separately, so valuation of $z$ in each may differ.

(2.14) a.

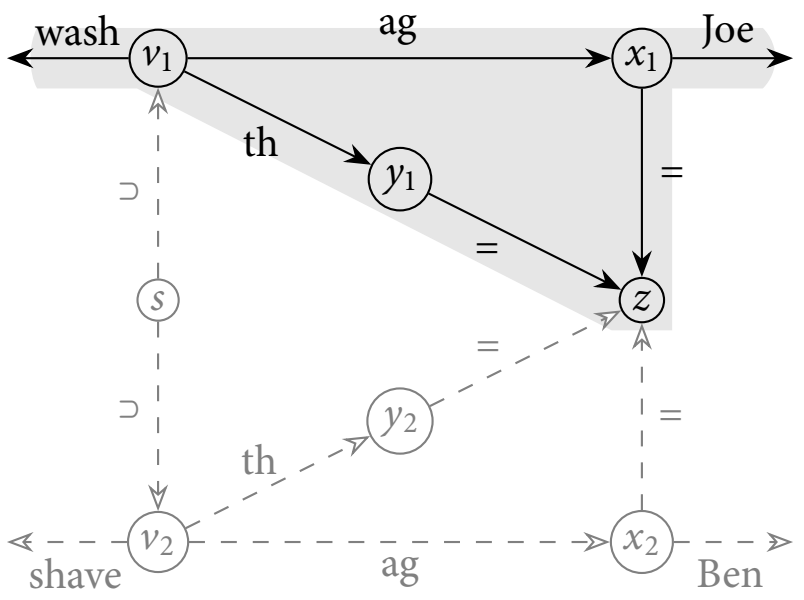

b.

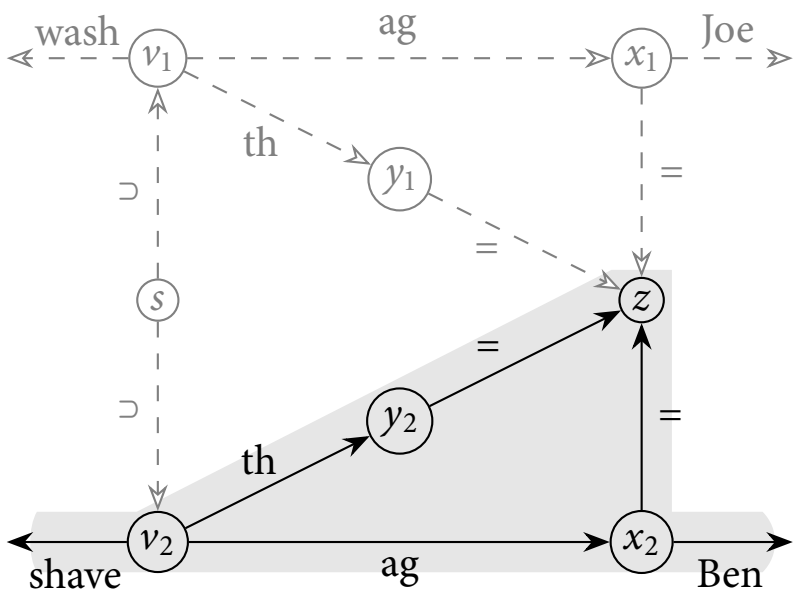

(2.14a&b) thus mean "Joe washed himself" and "Ben shaved himself" respectively. Were the equality edges contributed by the anaphor reversed, however, we would get these subgraphs by following the direction of the edges:

(2.15) a.

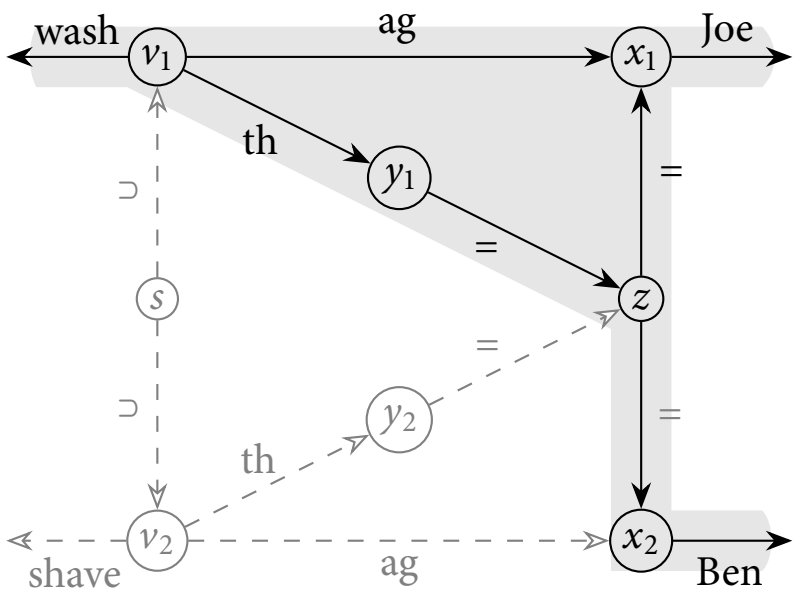

b.

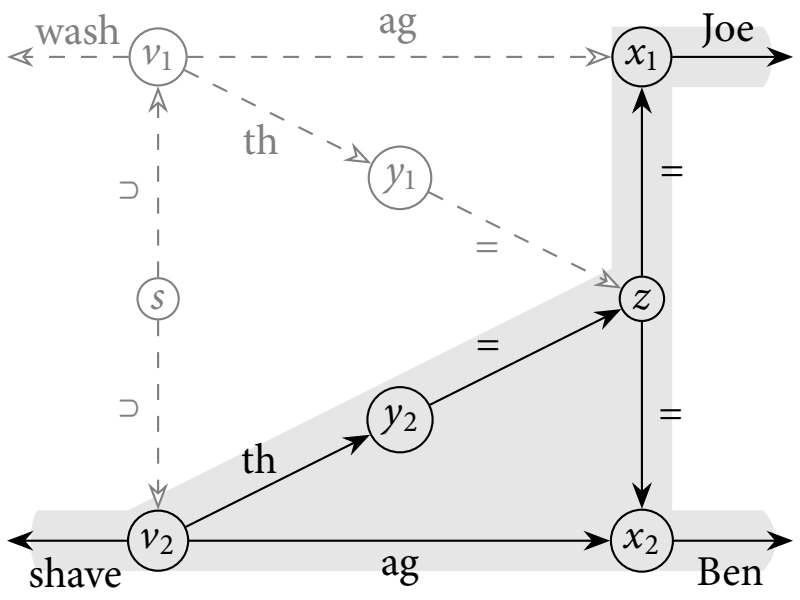

(2.15a&b) have the problem of forcing $z$ to equal Joe and Ben simultaneously.

A conceptual way to understand the equality edge's direction is to note that anaphors depend on their antecedents for values. As far as event participants depend on events (so phrased in dependency grammars; see Mel'čuk, 1988), and it is the dependent that is being pointed to (thematic relations direct from events to participants), then analogously, anaphora equality should direct from antecedents to anaphors.

### 2.1.4 Kappa edge

We introduce $\kappa$-labeled edges or *$\kappa$-edges* ($\kappa$ for "content") to represent intensionality, including modalities and attitudes. Being non-thematic, a $\kappa$-edge relates an intensional state (e.g. a possibility, a necessity, a belief, a promise, etc.) to a content as a subgraph in such a way that the latter is to be evaluated over the set of state-of-affairs or possible worlds compatible with that intensional state (see Section 2.2.3). (2.16) gives an example.

(2.16) Joe can walk a dog.

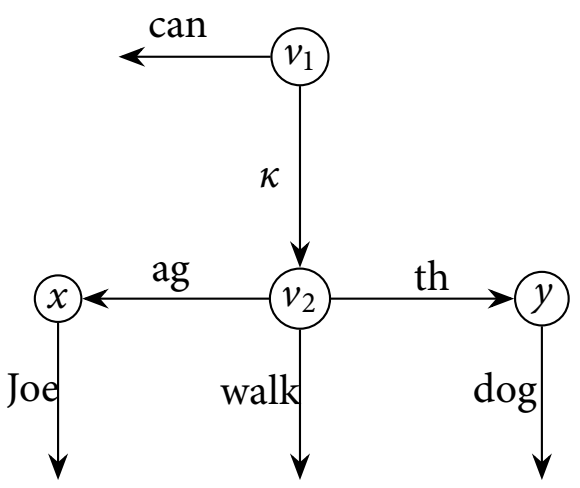

We evaluate the content, i.e. the subgraph reachable from $v_2$ ("Joe walked a dog"), at possible worlds compatible with the possibility $v_1$; its satisfaction at some world verifies (2.16). This quantificational view of intensionality expresses the logical tradition of possible world semantics (Carnap, 1946; Hintikka, 1961; Kripke, 1963).

The representation (2.16) also serves intensionality encoded with a nominal, adjectival, and adverbial syntax. The following examples roughly share a representation with (2.16).

(2.17)
a. A possibility that Joe walked a dog.
b. It is possible that Joe walked a dog.
c. Joe possibly walked a dog.

Intentional states like attitudes can take additional thematic dependents. A thought in (2.18a) has its holder, that is, an agent in a sense.

(2.18) a. Ben thought that Joe walked a dog. b. Ben saw Joe walk a dog.

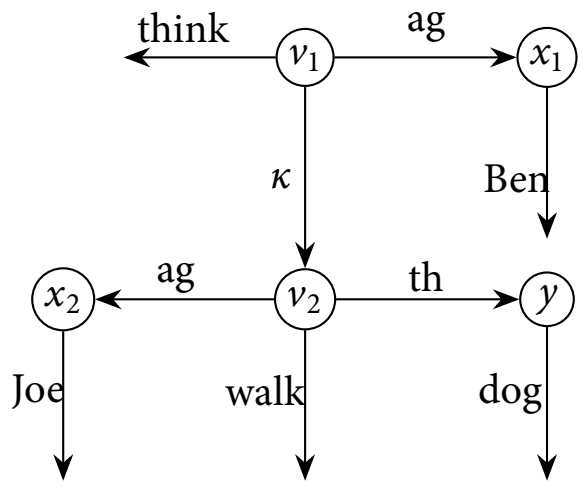

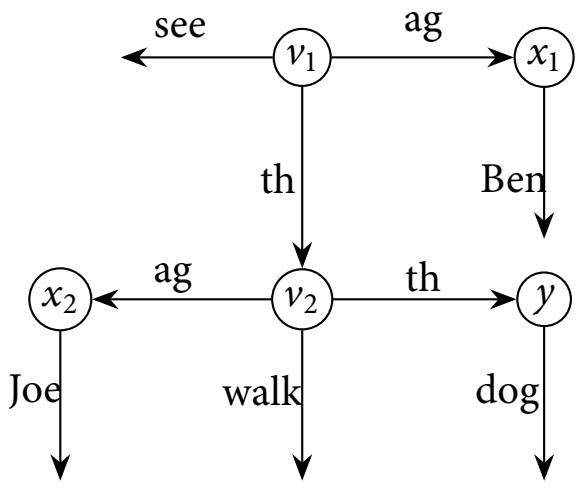

A perception report made with a small clause is usually distinguished from an attitude (see Higginbotham, 1983; Parsons, 1990). So, structurally identical to (2.18a), (2.18b) has $v_2$ related to $v_1$ as a theme, not a content.

### 2.1.5 Cardinality

Plurality can be explicitly marked by cardinality. In the following example, the cardinality edge adapted from AMR's *quantity relation* relates a set of boys to a (singleton of a) number.

(2.19) Two boys sailed.

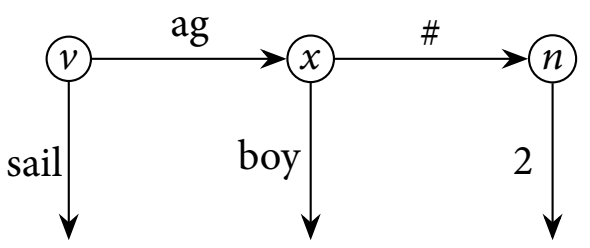

Numerals therefore pattern with the indefinite article *a* in that existence is still implicitly indicated; a number merely counts a plurality. To be precise, we might want to represent the cardinality constraints of indefinites and plurals of an indefinite number; they are omitted for the sake of presentation.

The class of determiners implicitly expressible are upward monotone in their generalized quantifier characterization. Their truth conditions take the form of a constraint that the intersection of the noun's and the predicate's denotations has a cardinality bounded from below by a constant:

(2.20) a/some N VP iff $|[\![N]\!] \cap [\![VP]\!]| \geq 1$
two N VP iff $|[\![N]\!] \cap [\![VP]\!]| \geq 2$

But for downward monotone determiners (in either N or VP), that constraint refers to an unknown or an upper bound:

(2.21) every N VP iff $|[\![N]\!] \cap [\![VP]\!]| = |[\![N]\!]|$
no N VP iff $|[\![N]\!] \cap [\![VP]\!]| \leq 0$

It is easy to see that none of these suits a representation like (2.19). We thus consider them to be genuine quantification that will be discussed in the next section.

### 2.1.6 Quantification

There seems no implicit representation for genuine quantification as there is for existence. However, we may think of quantification as iteration over a plurality. This iterative view underlies van Benthem's (1986) application of automata theory to generalized quantifiers and is comparable with *for*-loops in programming languages. A Python example:

```
for x in X:
    statements about x
```

As $x$ iterates over the objects in $X$, the statements about $x$ mark the scope in which $x$'s value varies.

Similarly in (2.22), for example, $y$ iterates over a set of boys and states that each (or none) of them sailed.

(2.22) a. Every boy sailed. b. No boy sailed.

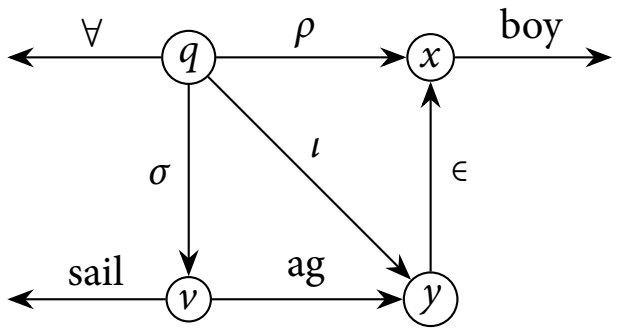

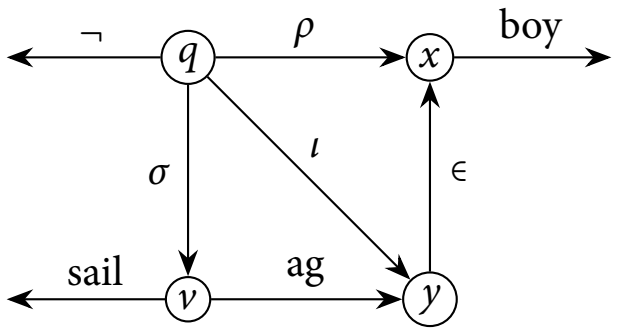

To render iteration structurally, we use a dummy variable $q$ whose unary out-edge labels the force of quantification ($\forall$ or $\neg$). The set to be iterated over is called the *restrictor* $x$, heading the $\rho$-edge from $q$. From the restrictor we draw each element as a singleton ($y \in x$ abbreviates $y \subset x$ and $|y| = 1$), the *iterator* $y$, heading the $\iota$-edge from $q$. A statement about $y$ is given by the *scope* $v$, heading the $\sigma$-edge from $q$. Thus we may read the semgraph as follows: for a contextually salient (maximized by default; see Section 2.2.4) set of boys $x$, each (or no) $y$ in $x$ is the agent of some sailing $v$.

In *quantification structures* illustrated by (2.22), the dummy variable's valuation does not matter; with $\rho$-, $\iota$- and $\sigma$- edges it serves to distinguish the restrictor, the iterator, and the scope from each other. The subgraph reachable from the restrictor is called the *restrictor subgraph* (2.23a). Less the out-edges of the restrictor, the subgraph reachable from the iterator is called the *iterator subgraph* (2.23b). Less the out-edges of the iterator, the subgraph reachable from the scope is called the *scope subgraph* (2.23c).

(2.23) a.

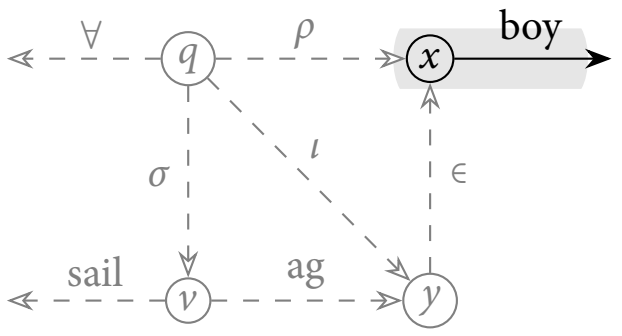

b.

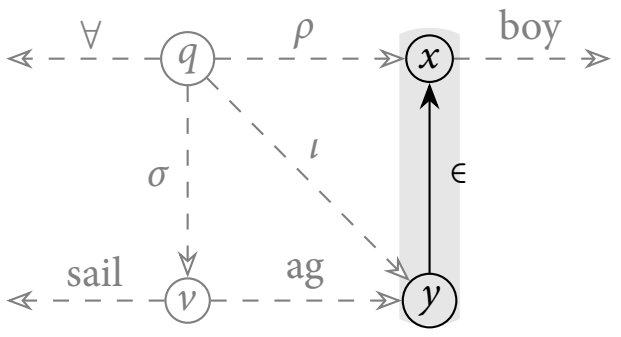

c.

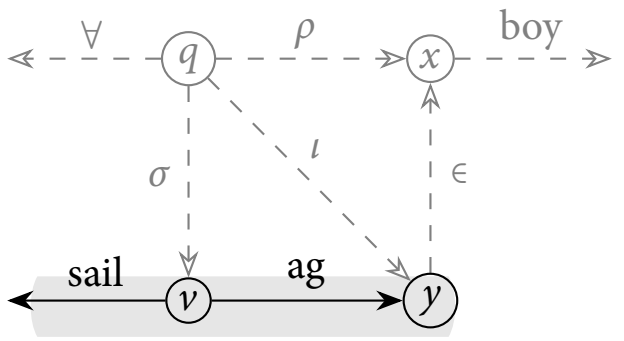

In fact, the iterative evaluation described above is equivalent to this: fixing a value for $x$ that satisfies (2.23a), for each (or no) value of $y$ satisfying (2.23b), there is some value for $v$ that satisfies (2.23c).

Quantification structures are used also for verbal negation and implication (we limit ourselves to "indicative conditionals", whose antecedents can be hypothetically true, and following DRT, treat them as material implication, without reference to possible worlds; see Kamp and Reyle, 1993, pp. 160ff for discussion and von Fintel, 2011 for a general review of conditionals). An example of each kind is shown in (2.24).[1] Being anonymous and thus unconstrained in both cases, $z$ is maximized to the set of all entities of the situation modeled, so we assume $z \neq \varnothing$.

(2.24) a. Joe didn't sail.

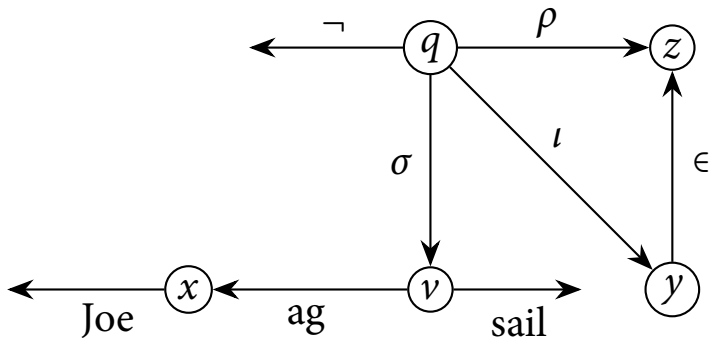

b. If a boy walks a dog, he feeds it.

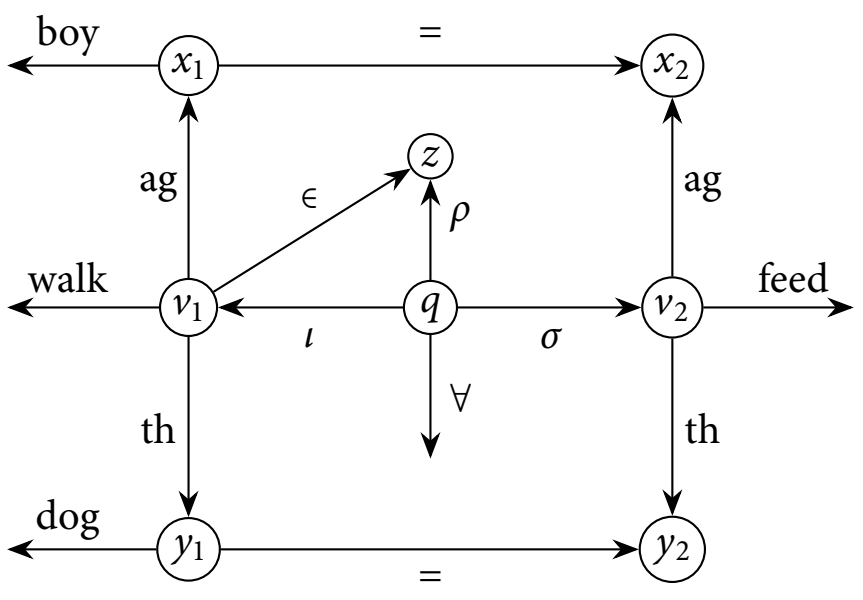

Then we read (2.24a) after (2.22b): for no $y$ drawn from $z$, there is some sailing whose agent is Joe. That simply means there is no sailing whose agent is Joe. And we read (2.24b) after (2.22a): whenever from $z$ we draw a walking $v_1$ whose agent is a boy $x_1 = x_2$ and whose theme is a dog $y_1 = y_2$, there is a feeding $v_2$ of the theme $y_2$ by the agent $x_2$. Dropping the assumption that $z \neq \varnothing$, readers can verify that what we said still holds.

[1] Example (2.24b) illustrates what is known as *donkey anaphora* in linguistic literature (Geach, 1962).

In the examples above, we see that a quantification structure explicitly refers to a plurality, i.e. the restrictor set to be iterated over. Such a representation will make more sense as it is applied to the semantics of distributivity in Chapter 4.

### 2.1.7 Coordination

By coordination we mean overt conjunction (*and*) and disjunction (*or*) in natural language, which convey but do not necessarily equal logical conjunction and disjunction.

*2.1.7.1 Conjunction* So far logical conjunction, like existence, is implicitly indicated in semgraphs. Given our earlier hypothesis that what can be implicitly indicated need not, or perhaps should not, be made explicit, natural language conjunction might better receive a non-Boolean treatment.

A natural proposal is for *and* to create plurality by denoting summation or set union (Chaves, 2007; Heycock and Zamparelli, 2005; Hoeksema, 1983; Lasersohn, 1995; Schwarzschild, 1996), which we label with set inclusion for the following example of nominal conjunction.

(2.25) Joe and Ben sailed.

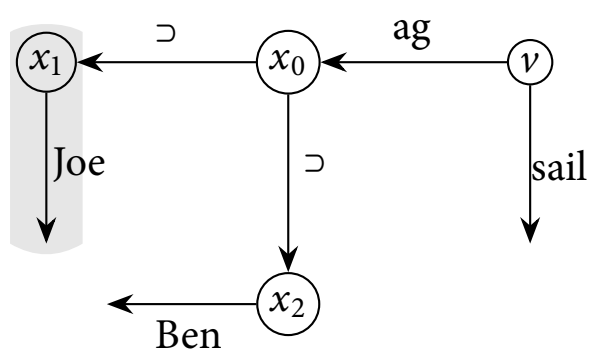

Semgraph (2.25) represents an underspecified reading of the sentence, one or more sailing that has Joe and Ben as the agent(s). We read the plural agent $x_0$ as the least set that includes $x_1$ and $x_2$, $x_1 \cup x_2$, and then evaluate the subgraph corresponding to each conjunct separately. The subgraph reachable from $x_1$ is shaded, declaring $x_1$ as Joe.

Verbal conjunction can be seen as set union of events. For example, $v_0$ in (2.26) sums up $v_1$ and $v_2$, each of which roots a subgraph to be evaluated separately.

(2.26) Joe sailed and Ben surfed.

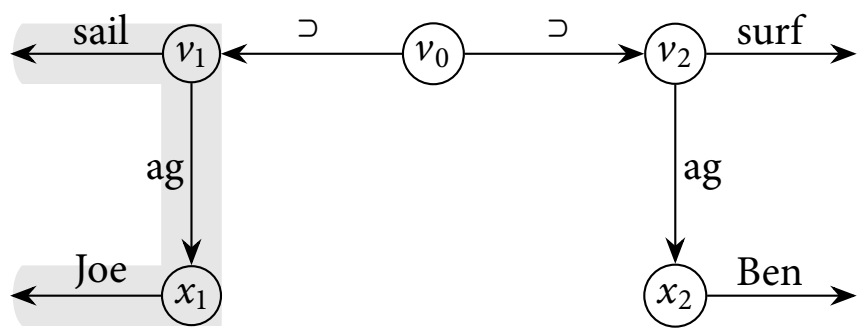

One of these subgraphs is shaded, representing the proposition "Joe sailed". The logical aspect of conjunction arises as a side effect as we interpret *both* conjunct subgraphs. There is actually a logical aspect to conjunction in (2.25), too. The existential proposition represented by each conjunct (e.g. "Joe") happens to be made of a nominal as in (2.3a) (recall that we conflate proper and common nouns).

Verbal conjunction beneath the sentence level involves argument sharing in representation, for which we have met an example (2.13). Here we consider another example.

(2.27) Joe rented and Ben sank a boat.

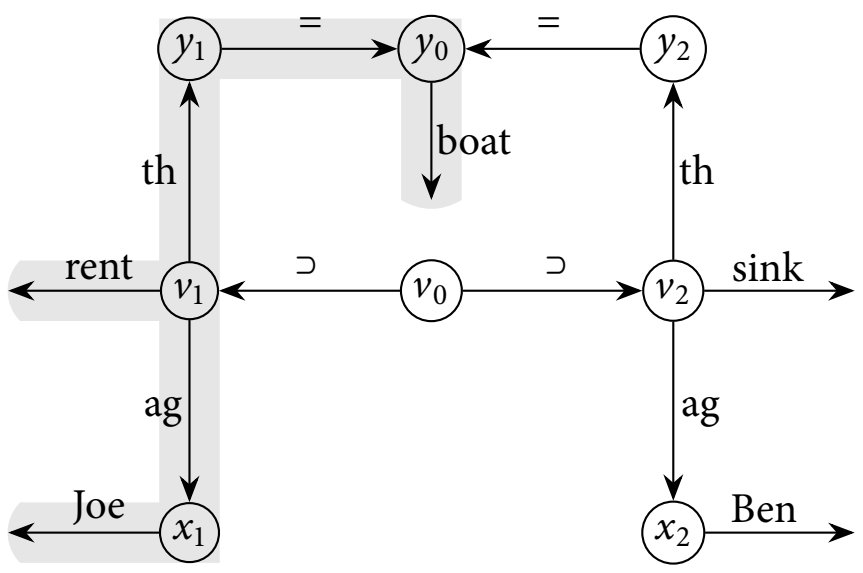

In (2.27) a boat is the theme of a renting and a sinking. This is represented by the equality between $y_1$ and $y_0$ and between $y_2$ and $y_0$.

We emphasize that argument sharing does not fix a value for the shared argument and distribute it among the conjuncts. Indeed, the subgraph reachable from $v_1$ (the shaded area) contains $= \overrightarrow{y_1 y_0}$ but not $= \overrightarrow{y_2 y_0}$, and the dual holds for the subgraph reachable from $v_2$. Separate evaluation of the conjunct subgraphs gives (2.27) the following reading.

(2.28) Joe rented a boat and Ben sank a boat.

So the boat rented and the boat sank can be at variance as desired. To be sure, the mechanism just explained also generates the reading (2.29b) for a shared indefinite subject as illustrated by (2.29a).

(2.29) a. A boy sailed and surfed.
b. A boy sailed and a boy surfed.

There seems a strong preference for the sailor to coincide with the surfer (see Moltmann, 1994, p. 113), though this preference might relate to a more general pattern for subject indefinites to take wide scope (see Section 4.2.2 for discussion).

In Section 3.3.2, we will attribute the equality edges used by argument sharing to the semantics of coordination; it turns out that their presence is conditioned on conjunct categories. For now one may wonder if they are redundant, since (2.27) can be contracted as follows while preserving its semantics.

(2.30)

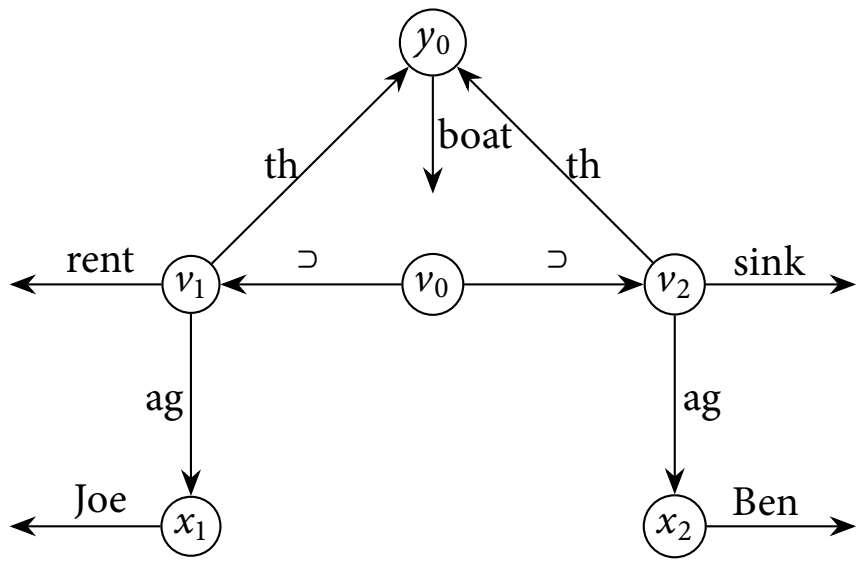

Justification for equality edges comes from coordination of quantifiers. (2.31) gives an example and its intended reading. Compare argument sharing with equality (2.31a) and without it (2.31b):

(2.31) Every dog and no cat barked (= every dog barked and no cat barked).

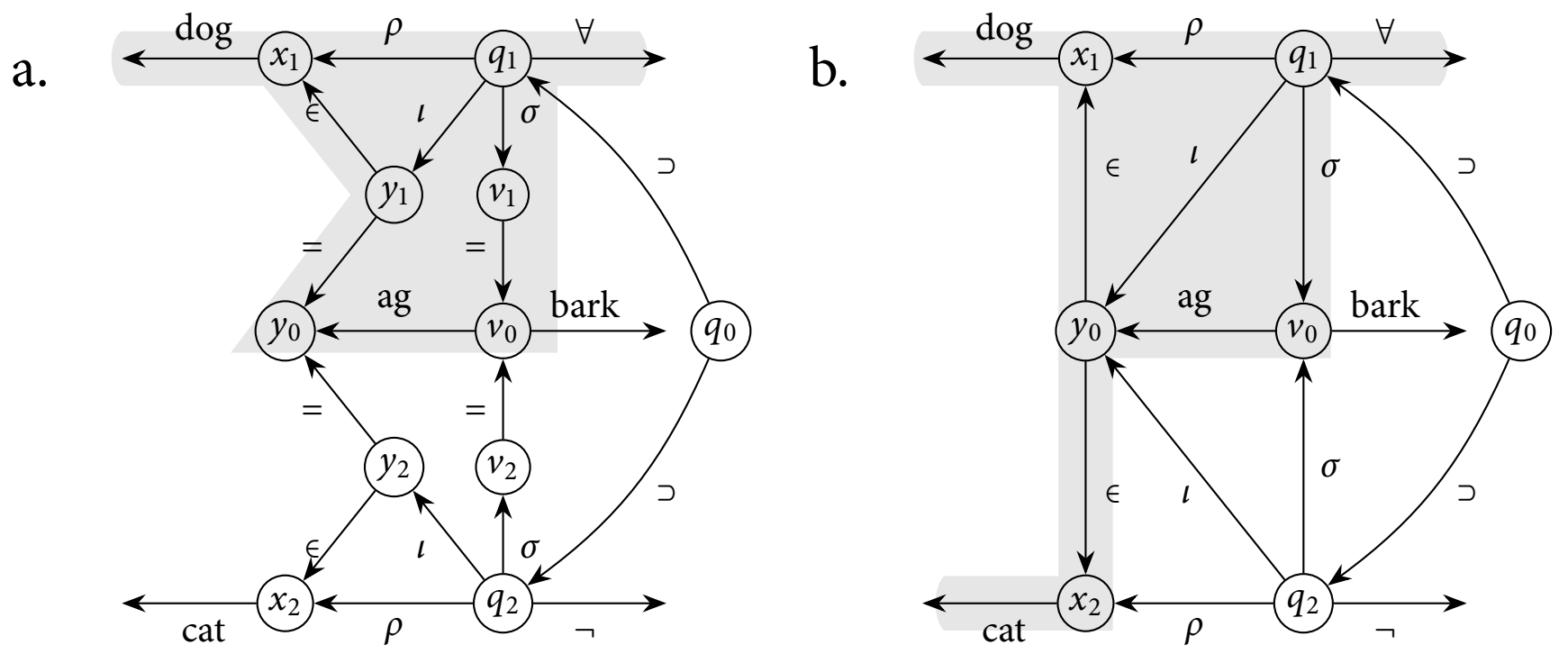

Dummy variables $q_1$ and $q_2$ and their union $q_0$ are not subject to any constraint, but $q_1$ and $q_2$ each roots a quantification structure to be evaluated. For example, the subgraphs reachable from $q_1$ in (2.31a&b) are shaded.

Whereas in (2.31a) $y_1$ iterates over a set of dogs, in (2.31b) $y_0$ incorrectly iterates over the intersection of a set of dogs and a set of cats. The ill-delimited subgraph in (2.31b) is similar to (2.15), but follows from directly sharing the iterator of quantification structures. A symmetric remark goes to the subgraph reachable from $q_2$. Thus (2.31a) alone represents the intended reading.

*2.1.7.2 Disjunction* It would be curious, Winter (2001) remarks, if conjunction is the only natural language coordinator befitting a non-Boolean semantics. We thus introduce a representation of disjunction whose structure is isomorphic to that of conjunction, but whose interpretation differs.

The basic idea is this: where conjunction sums up its operands, disjunction chooses one of them. We label this special relation with ⊒ as if it is an "optional inclusion". For the following example of nominal disjunction, we read $x_0$ as a choice between $x_1$ and $x_2$, and then evaluate only the subgraph corresponding to the chosen disjunct.

(2.32) A boy or a dog swam.

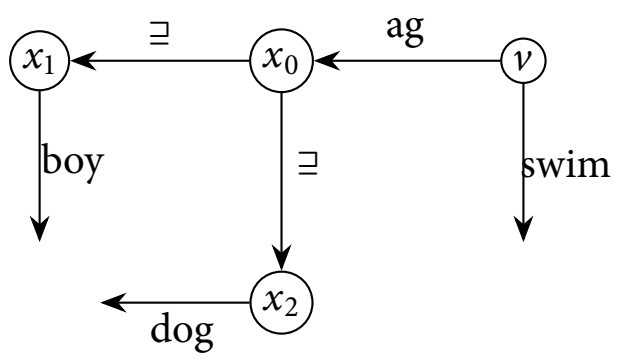

The reason for a single choice ($x_0 = x_1$ or $x_0 = x_2$) but not a multiple choice ($x_0 = x_1 \cup x_2$) is that plural predicates do not take disjunction of singulars:

(2.33) *A boy or a dog hugged.

Verbal disjunction is likewise represented. For example, $v_0$ in (2.34) is chosen from $v_1$ and $v_2$, one of which roots a subgraph to be evaluated.

(2.34) Joe sailed or Ben surfed.

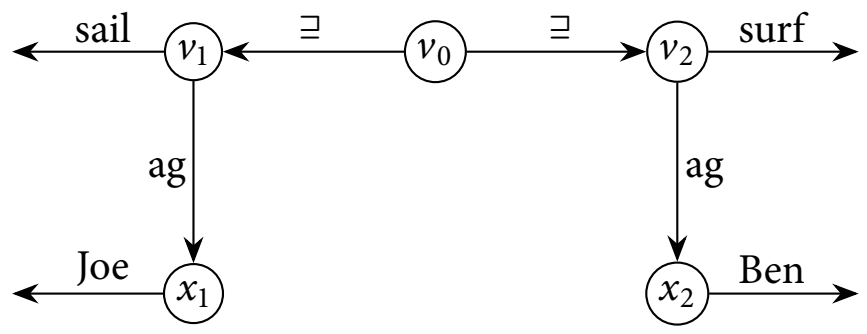

The logical aspect of disjunction arises as a side effect as we choose one of the disjunct subgraphs to verify while ignoring whether the other can be satisfied.

Since argument sharing works the same way for disjunction as for conjunction, we omit further illustration.

## 2.2 Model-theoretical semantics

In this section, we build a model-theoretical interpreter for the semgraphs we have been reading informally so far. The interpreter evaluates semgraphs with respect to models of state-of-affairs and returns a truth value.

### 2.2.1 Plural values

In Chapter 1, we motivated the monadic second-order restriction on computational complexity grounds. Before defining models, we take a further look at why sets of entities can be a reasonable choice for plural values. For a thorough background, one is referred to the literature on plural semantics (Lasersohn, 1995; Link, 1983; Scha, 1981; Schwarzschild, 1996; Simons, 1982; Winter, 2001).

*2.2.1.1 Flat vs. nested* Sets of first-order objects represent a *flat* view of plurality, as opposed to a *nested* view where such sets can be contained in other sets (see Winter, 2001, pp. 38ff). The difference can be illustrated with an example adapted from Hoeksema (1983, p. 75).

(2.35) Blücher and Wellington and Napoleon fought each other.

a. [Blücher and Wellington] and Napoleon fought each other.
b. Blücher and [Wellington and Napoleon] fought each other.

The subject of (2.35) denotes a set of three entities, say $\{b, w, n\}$ under the flat view, but may denote either $\{\{b, w\}, n\}$ or $\{b, \{w, n\}\}$ under the nested view, mirroring the two syntactic parses.

Examples like (2.35) involving reciprocal predicates are sometimes cited in favor of nested plurality. The argument proceeds as follows. If *each other* refers only to the subject's flat denotation $\{b, w, n\}$, drawing each element thereof, and pairing the draw with the rest in a fight (see Heim et al., 1991), we obtain a truth condition the world history does not satisfy. Since (2.35) can describe a historical truth, something like $\{\{b, w\}, n\}$ should be referable by *each other*.

However, flat plurality is maintainable if we work with a different set of assumptions. For example, what *each other* does can be generalized; it may iterate over a partition of $\{b, w, n\}$ like $\{\{b, w\}, \{n\}\}$ and pair each draw with the rest. The idea involves non-atomic distribution (Schwarzschild, 1996), assumed in the previous argument anyway. Alternatively, we may construct a set of two "group entities" for *each other* to iterate over, such that one of them would bear a *comprise* relation with $\{b, w\}$ and the other with $\{n\}$ (see Section 2.2.2).

In either case, the subject retains a flat denotation while *each other* operates on a two-element collection. We will not delve into the semantics of reciprocity, but it does not seem to compel enrichment of the structure of plurality. See Winter (2001, secs. 2.2.3, 6.2) for more discussion.

*2.2.1.2 Set theory vs. mereology* When modeling the ontology of state-of-affairs, if besides entities one introduces also sets as "plural entities", there arises a type distinction between singulars' and plurals' denotations. Those who find this distinction suspect but replacing entities with singletons counterintuitive (Link, 1983) turn to mereology (originally proposed as an alternative to set theory to be the foundation of mathematics; see Gruszczynski and Varzi, 2015 for a historical note), which is quite popular in semantic literature. Following Link, we present its basics in terms of lattice theory.

Recall that a set $X$ is *partially ordered* if some pairs of its elements are comparable under a relation $\leq$, such that

i) (*reflexivity*) $x \leq x$ for all $x \in X$;
ii) (*antisymmetry*) if $x \leq y$ and $y \leq x$ then $x = y$ for all $x, y \in X$;
iii) (*transitivity*) if $x \leq y$ and $y \leq z$ then $x \leq z$ for all $x, y, z \in X$.

Suppose a partially ordered set $X$ further has a property: every pair $x, y \in X$ has a *least upper bound* $z \in X$, that is, $x \leq z$, $y \leq z$, and whenever $u \in X$ satisfies $x \leq u$, $y \leq u$ we have $z \leq u$. Then $X$ is called an *upper semilattice* (henceforth "semilattice"); $z$ is called the *sum* or the *join* of $x$ and $y$, and we write $z = x + y = \sum\{x, y\}$.

Thus in a mereological narrative, singular and "plural entities" constitute a semilattice, where the order $\leq$ is understood as a *part-whole* relation, and "plural entities" sum up singular ones. For such a semilattice to be linguistically useful, it is often assumed to be complete and free (Landman, 1991; Kamp and Reyle, 1993).

A semilattice $X$ is *complete* if each of its nonempty subset has a least upper bound, generalizing the sum of a pair from above. (It is common to define the sum over $\varnothing$ as the least element, which is typically removed from linguistic applications, however.) A finite semilattice is obviously complete.

Suppose we have a *trivial* partially ordered set $P$, that is, its order contains all and only pairs of the form $x \leq x$. A complete semilattice $X$ is *generated by* $P$ if $X$ is the smallest complete semilattice including $P$. Then $X$ is *free* if it satisfies the so-called *universal property*:

> for any complete semilattice $L$, every map $\varphi : P \to L$ can be extended to a *homomorphism* $\psi : X \to L$ (that is, $\varphi(x) = \psi(x)$ for all $x \in P$, and for each nonempty $A \subseteq X$, $\psi(\sum A) = \sum\{\psi(a) \mid a \in A\}$).

In this case we also say $X$ is *freely generated by* $P$, and the elements of $P$ are called *generators* or *atoms*. The universal property holds exactly when $X$ does not imposes nontrivial relations on its elements; otherwise not every $L$ would ensure the extendability of $\varphi$ to $\psi$ (see Grätzer, 2011, p. 81 for a precise formulation).

As Landman (1991, pp. 256ff) explains, freeness rules out mereological structures with weird semantic implications. For example, one may check that (a) alone is free in Figure 2.1. In (b) $x + y$ coincides with $z + u$ even if $\{x, y\}$ and $\{z, u\}$ are disjoint (but an object should be uniquely decomposed into atomic parts). In (c) $z$ has only $x < z$ and $u$ has only $y < u$ (but an object cannot only have one part smaller than itself).

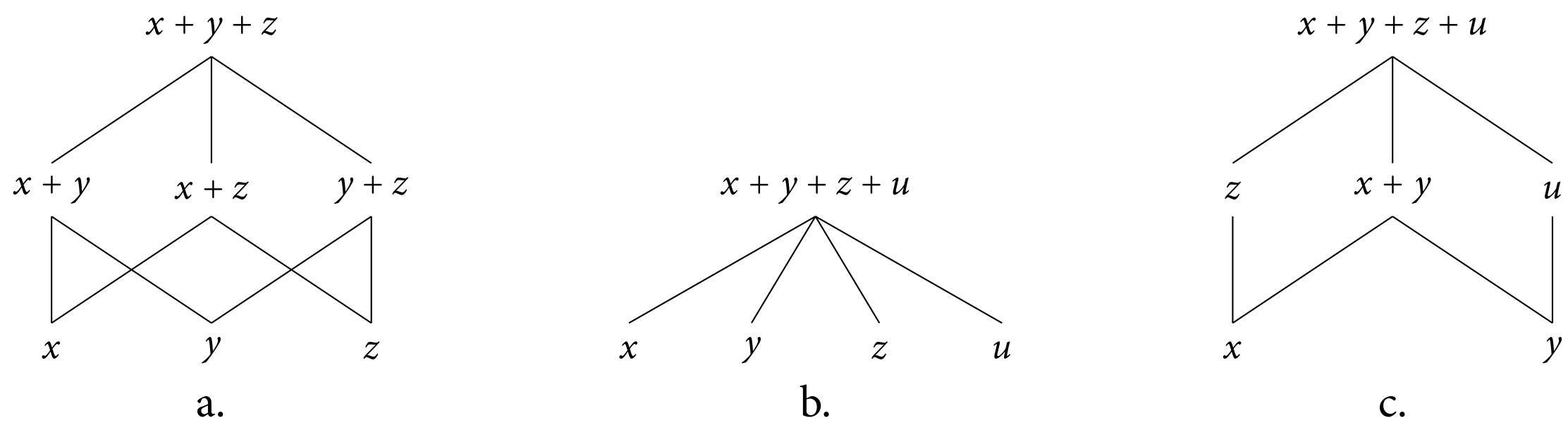

**Figure 2.1:** Examples of finite semilattices visualized in Hasse diagrams.

From the definition of freeness it follows immediately that

**Corollary.** *If $X_1$ is freely generated by $P_1$ and $X_2$ is freely generated by $P_2$ and $|P_1| = |P_2|$, then $X_1$ and $X_2$ are* isomorphic *(that is, there is a homomorphism from $X_1$ to $X_2$ that is also a one-to-one correspondence (bijection)).*

A proof can be gleaned from any formal text on lattice theory (e.g. Freese et al., 1995; Grätzer, 2011). We are interested in this since it establishes that to each complete semilattice $X_1$ freely generated by $P_1$, there corresponds an isomorphic semilattice built on the power set of $P_1$ less $\varnothing$, complete and free.

Let $P_2 = \{\{x\} \mid x \in P_1\}$, which is of the same cardinality as $P_1$. Consider $X_2 = \wp(P_1) - \{\varnothing\}$ ordered by set inclusion. We can show that $X_2$ is a complete semilattice freely generated by $P_2$. If so, the above corollary guarantees that $X_1$ and $X_2$ are isomorphic.

Indeed, the sum of any nonempty $A \subseteq X_2$ is given by $\bigcup A$. Thus $X_2$ is a complete semilattice, since the union of subsets of $P_1$ is again a subset of $P_1$. By construction $P_2$ is a subset of $X_2$. To show $P_2$ generates $X_2$, we show that any $Y \subset X_2$ including $P_2$ is incomplete. Let $y \in X_2 - Y$ and $B = \{\{x\} \mid x \in y\}$. Hence $B \subseteq P_2$ but $y = \bigcup B$ is not in $Y$. Finally, for any complete semilattice $L$ and any map $\varphi : P_2 \to L$, we define $\psi : X_2 \to L$ by $\psi(z) = \sum\{\varphi(\{x\}) \mid x \in z\}$ for each $z \in X_2$. It is easy to verify that $\psi$ is a homomorphism as required.

Thus mereology-based plurality is essentially the same as one based on sets. Champollion (2017, p. 18) mentions another concern: when modeling "mass entities" presumably denoted by mass nouns like *water*, we might need an infinite decreasing part-whole sequence $x_1 > x_2 > \cdots$ of such entities (see Kamp and Reyle, 1993, p. 400 for comments). Yet we see that a power set semilattice $X$ less $\varnothing$ (or any free complete semilattice) generated by $P$ is bounded from below by the atoms; for each $x \in P$ no $y \in X$ satisfies $y \subset x$. This would be a problem, however, only if we limit $P$ to finite sets. If we were to accept an ontology where "mass entities" can be infinitely decomposed, an infinite $P$ would generate $X$ from which an infinite sequence $x_1 \supset x_2 \supset \cdots$ can be formed (e.g. taking $P = \mathbb{N}$ the set of natural numbers, we get $\mathbb{N} \supset \mathbb{N} - \{0\} \supset \mathbb{N} - \{0, 1\} \supset \cdots$).

Our discussion so far should have justified using sets for plural values. Doing so does not necessarily commit oneself to an ontology where sets replace entities, however. As we will see below, we do not have to introduce "plural entities" into the domain of a model of state-of-affairs.

### 2.2.2 Models

Now we may define models with respect to which semgraphs are interpreted. In the current context, a model consists of three components familiar from model theory.

In the first place, we have a *domain* or *universe*, that is, a nonempty set $D$ of first-order entities, including individuals, events, and natural numbers (since we take into account numeral expressions for a finite cardinality). No order is assumed on this set.

Then there is a set $W$ of *worlds*, each of which is an *interpretation* of how things can be, by mapping predicate symbols to objects built of the values in $D$ and entities in $D$ to worlds in $W$.

In specific, a lexical unary predicate (e.g. *dog*, *walk*, ...) is interpreted as a set of entities that *instantiate* that predicate. We assume that predicates for proper names (e.g. *Joe*, *Ben*, ...) have a unique instance. In particular, a numeral has a unique instance that equals its face value. We also take the simplistic view that restricts the subject of instantiation to entities, which means that nominals like *group*, *band*, ... known as "group denoting", are nonetheless instantiated by entities, not sets (see Link, 1983; Landman, 1989). Group entities help to distinguish distinct groups made of the same members or track the same group as its membership changes. The contingency of the relation between a group and its members can be captured by a constituting (or comprising) event for which a set of those members is the agent (or theme) and a singleton of that group is the the theme (or agent).

A thematic binary predicate (e.g. *ag*, *th*, ...) is interpreted as a set of pairs relating an event to a nonempty set of entities. With Carlson (1984); Dowty (1989), we assume that events have at most one participant for each thematic relation, so a binary predicate defines a partial function on $D$. We treat event participants as plural values, i.e. sets, to model joint participation, which can be distinguished from individual participation (Scha, 1981). For example, a joint invitation by or of two people is not equivalent to two individual invitations by or of each, so in that case, we pair one invitation event with a set of two entities in an agent or theme relation. Some events may make sense only with joint participation or non-singleton participants, like meeting or hugging. Such are the idiosyncrasies of lexical semantics (see Winter, 2001, sec. 2.3.1). (We are not in a position to discuss an alternative ontological setup: abandoning the assumption of thematic uniqueness, one can keep event participants singular and model joint participation by pairing an event with each individual involved, so a joint invitation by two people makes each one an agent; see Schein, 1993.)

To model intensionality, we also want to interpret an intensional state (e.g. a possibility, a belief, ...) as a set of worlds. This way an intensional state defines an *accessibility relation* between worlds (see Kripke, 1963): $w'$ is accessible from $w$ according to, say, a belief, whenever $w'$ is among the worlds assigned to that belief by $w$. Metaphorically we say $w'$ is among the worlds compatible with that belief at $w$. There is no point, but again, no harm, in likewise interpreting all entities in $D$, so we need not explicitly divide $D$ into all intensional states and the rest.

Therefore, a world $w$ maps

i) a unary predicate $P$ to $P_w \subseteq D$,
ii) a binary predicate $R$ to $R_w \subseteq D \times (\wp(D) - \{\varnothing\})$,
iii) an entity $a \in D$ to $a_w \subseteq W$.

At last, there is a *valuation* $g$. Whereas worlds interpret lexically or thematically labeled edges (edges with special labels have a world-independent semantics to be discussed later), a valuation is a partial function interpreting variables, that is, semgraph vertices, as possibly empty sets of entities. Thus $x^g \subseteq D$, where $x^g$ denotes the value of $x$ assigned by $g$.

Hence the following definition.

**Definition 2.1.** A model $\mathcal{M}$ is a triple $(D, W, g)$ where

i) $D$ is a domain of entities;
ii) $W$ is a set of worlds;
iii) $g$ is a valuation.

Whenever necessary any $\mathcal{M}$-specific construct can be indexed, e.g. $(D_\mathcal{M}, W_\mathcal{M}, g_\mathcal{M})$.

The models we just defined contrast with most models used in the plural semantics literature (a.o. Link, 1983). We have not added "plural entities" to the domain, nor have we closed the interpretation of any predicate under summation. (Otherwise, this would mean that if $x_1, \dots, x_n \in P_w$, then $\{x_1, \dots, x_n\} \in P_w$; if $(v_1, X_1), \dots, (v_n, X_n) \in R_w$,then $(\{v_1, \dots, v_n\}, \bigcup_{i=1,\dots,n} X_i) \in R_w$.) The reason is that instead of pluralizing models more than we have done, we can pluralize the interpreter, which we discuss below (cf. Brasoveanu, 2013, sec. 4.1; Chaves, 2007, chap. 4).

### 2.2.3 Basics

Given a model $\mathcal{M}$, we may evaluate a semgraph $G$ by checking whether the valuation of vertices satisfy the constraints of edges according to some world. Evaluation happens as we traverse $G$, whose unique root becomes the entrance of traversal. We track the vertices deemed visited so that the scope of variables may arise from the order of their valuation (see Section 2.2.6).

We now define a relation $\mathcal{M}, w \vDash_\eta G$ with $w \in W_\mathcal{M}$, or simply $w, g \vDash_\eta G$ when the model in question is clear. This relation is read as "given the (vertex) visiting history $\eta$, $\mathcal{M}$ satisfies $G$ at $w$", or "$G$ is true at $w$ in $\mathcal{M}$", or simply "$g$ satisfies $G$ at $w$". In a family of all models differing only in valuations, if $w, g \vDash_\varnothing G$ for some $g$, we write $w \vDash G$ and say $G$ is true or satisfied at $w$.

Let $x$ be the root of $G$. $x$ roots a quantification structure introduced in Section 2.1.6 if it has an out-edge labeled as $\rho$; it roots a coordination structure introduced in Section 2.1.7 if it has an out-edge labeled as $\supset$ or $\sqsupseteq$. In both cases, the out-edges of $x$ shall be considered jointly to define the operations involved in quantification and coordination, which we will discuss in Sections 2.2.4 and 2.2.5. But if $x$ does not root a special structure of those kinds, each of its out-edges, that is, each member of $T(x)$ can be considered independently. The following definition starts with this case, implementing graph traversal by induction on the structure of $G$.

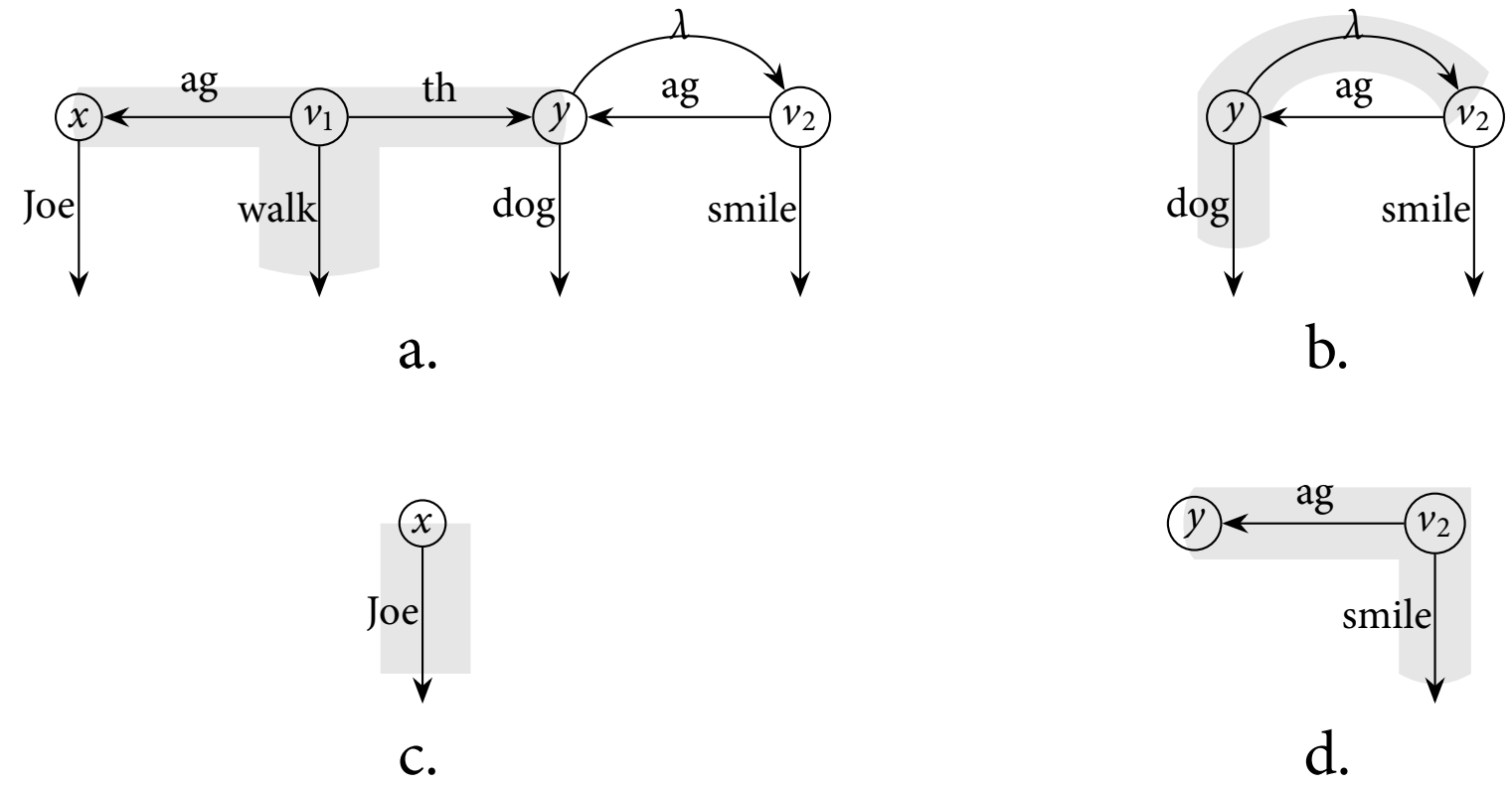

**Figure 2.2:** An example of semgraph interpretation by traversal/induction, glossing over edge constraint evaluation. Out-edges to be evaluated in each subgraph are shaded. (a) Rooted at $v_1$. Evaluate edges $walk\overrightarrow{v_1}$, $ag\overrightarrow{v_1 x}$, $th\overrightarrow{v_1 y}$. (b) Subgraph of (a) reachable from $y$ less out-edges of $v_1$, rooted at $y$. Evaluate edges $dog\overrightarrow{y}$, $\lambda\overrightarrow{y v_2}$. (c) Subgraph of (a) reachable from $x$ less out-edges of $v_1$, rooted at $x$. Evaluate edge $Joe\overrightarrow{x}$. (d) Subgraph of (b) reachable from $v_2$ less out-edges of $y$. Evaluate edges $smile\overrightarrow{v_1}$, $ag\overrightarrow{v_2 y}$.

**Definition 2.2.** Let $\mathcal{M} = (D, W, g)$ be a model, $w \in W$ and $G$ a semgraph rooted at $x$.

i) If $x$ does not root a special structure then $w, g \vDash_\eta G$ iff

- i.a) when $P\overrightarrow{x} \in T(x)$,
  $w, g \vDash_\eta P\overrightarrow{x}$;
- i.b) when $R\overrightarrow{xy} \in T(x)$ and $R \neq \kappa$,
  $w, g \vDash_\eta R\overrightarrow{xy}$ and $w, g \vDash_{\eta \cup [x]} G - T(x)/y$;
- i.c) when $\kappa\overrightarrow{xy} \in T(x)$ and $a \in x^g$,
  for all/some $w' \in a_w$ there is $h \supseteq g$ s.t. $w', h \vDash_{\eta \cup [x]} G - T(x)/y$.

Given $G$ rooted at $x$, (i) evaluates each out-edge of $x$ and then, with $\eta$ updated by $[x]$ — containing $x$ and any vertex in $G$ reachable from $x$ via a path free of $\rho, \iota, \sigma$ and ending in $\lambda$ or equality — the subgraphs reachable from each of its successors. Note how we remove all out-edges of $x$ when inducing the subgraph reachable from a successor (see Section 1.6.2). This way each edge is interpreted only once, and the interpreter thereby avoids an infinite loop when $x$ is in a cycle. To give a worked example, Figure 2.2 shows how the semgraph of the following sentence is broken down into edge constraints in the process.

(2.36) Joe walked a dog which smiled.

In the above definition (i.a) and (i.b) depend on the following rules for evaluating constraints given by edges.

**Definition 2.3.** (Continuing Definition 2.2).

ii) $w, g \vDash_\eta \lambda\overrightarrow{xy}$;
iii) $w, g \vDash_\eta = \overrightarrow{xy}$ iff $y^g = x^g$;
iv) $w, g \vDash_\eta \#\overrightarrow{xy}$ iff $y^g = \{|x^g|\}$;
v) $w, g \vDash_\eta \in \overrightarrow{xy}$ iff $x^g \subseteq y^g$ and $|x^g| = 1$;
vi) $w, g \vDash_\eta P\overrightarrow{x}$ iff $x^g \subseteq P_w$ and $x^g \neq \varnothing$;
vii) $w, g \vDash_\eta R\overrightarrow{xy}$ iff $y^g = \bigcup_{a \in x^g} R_w(a)$.

Here (ii)–(vii) each handles an edge constraint. (ii) shows that a $\lambda$-edge is trivially satisfied; it does not constrain its endpoints. (iii)–(v) give the self-explanatory constraints for equality, cardinality, and membership. These four rules show that edges with a special label have a world-independent semantics.

(vi) and (vii) show how evaluation of predicates takes plurality into account. (vi) says that for $x^g$ to instantiate a lexical predicate $P$ (e.g. *dog*, *walk*, ...), each element of $x^g$ should instantiate $P$. The rule expresses the intuition that predicative plural nouns are typically *distributive* (Link, 1983). As we allow the empty set $\varnothing$ in vertex valuation (see Sections 2.2.4 and 2.2.5), however, it needs to be explicitly stated that $\varnothing$ does not instantiate a predicate (a price for omitting the cardinality constraints of indefinites, which also deprives us of Bylinina and Nouwen's (2018) treatment of the semantics of *zero*); otherwise, for example, *a dog* would be satisfied in a situation where there is no dog, since $\varnothing$ is a subset of any set.

To specify what it takes for $x^g$ to bear a thematic relation $R$ (e.g. *ag*, *th*, ...) with $y^g$, (vii) uses the function notation that exploits the assumption of thematic uniqueness: with $R_w(a)$ denoting the unique $R$-participant of $a$ at $w$, if defined, (vii) asserts that $y^g$ is the union of the family of the $R$-participants of the events in $x^g$. The rule expresses the intuition that thematic relations are *cumulative* (Krifka, 1992; Landman, 2000).

One may apply these definitions to the constraints of edges in Figure 2.2. Here we consider another example that highlights the distributivity of (vi) and cumulativity of (vii).

(2.37) Five boys adopted six dogs.

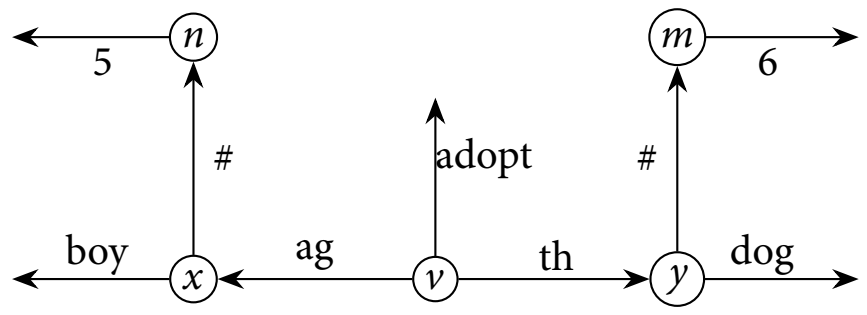

Through a similar process as illustrated in Figure 2.2, the truth condition of (2.37) at some world can be gathered as follows, with cardinality constraints simplified on the assumption that numeral predicates are always interpreted according to their face values (e.g. $5_w = \{5\}$).

(2.38) There is some $g$ such that

a. $v^g \subseteq adopt_w$, $x^g \subseteq boy_w$, $y^g \subseteq dog_w$;
b. $|x^g| = 5, |y^g| = 6$;
c. $x^g = \bigcup_{a \in v^g} ag_w(a)$, $y^g = \bigcup_{a \in v^g} th_w(a)$.

In other words, we have a set $x^g$ of five boys, a set $y^g$ of six dogs, and a set $v^g$ of adoptions; the agents of these adoptions sum up to $x^g$, and their themes to $y^g$. (2.38) can be satisfied by a variety of situations, from four boys each adopting a dog and the fifth boy adopting two, to five boys jointly adopting a litter of six. The number of adoptions does not matter, as far as each of the five boys engaged in an adoption by himself or with others, and each of the six dogs was adopted by itself or with others.

What we illustrate is the so-called *cumulative reading* of the sentence, an example of underspecification of plural predication (Scha, 1981). A special case of this reading, where there is only a single adoption, is sometimes identified as a distinct reading called the *collective reading*. There might be arguments for not reducing cumulativity to collectivity (Landman, 2000, sec. 5.4.3), but the evidence against the opposite is less clear. Following (Link, 1998, p. 180), we do not make such a distinction but use "cumulative reading" as subsuming "collective reading".

But there is a distinctive class of readings of (2.37) that can be paraphrased as follows.

(2.39) a. Five boys each adopted six dogs. ($\geq$ 5 adoptions)
b. Six dogs were each adopted by five boys. (6 adoptions)

These are known as *distributive readings*. The desired semantics of such examples does not follow from semgraphs like (2.37) but requires quantification structures to perform distribution, as we will see in Section 2.3.2.

The last clause (i.c) of Definition 2.2 handles the interpretation of intensionality. Recall that an intensional state at some world $w$ is linked to a set of worlds accessible from $w$. For each intensional state in $x^g$, (i.c) says the subgraph reachable from the head of a $\kappa$-edge is true at all or some of its accessible worlds. The force of quantification depends on the lexical semantics of the intensional state in question: *for all* if a necessity (or an attitude of the same nature, like a belief), *for some* if a possibility (or an attitude of the same nature, like a doubt), as commonly treated in modal logic.

In the following example, the shaded area shows the subgraph to be evaluated over the worlds compatible with a possibility. For (2.40) to be true, it should be satisfied at one of those worlds at least.

(2.40) Joe can walk a dog. (2.16)

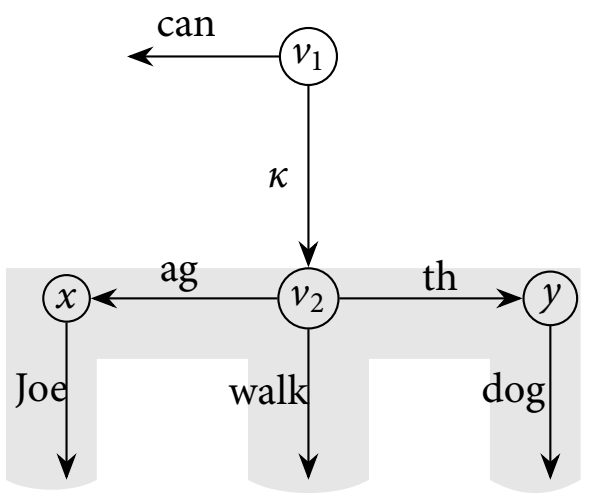

According to (i.c), at each accessible world we are looking for some valuation $h$ extending the valuation $g$ (an *extension* $h \supseteq g$ satisfies the condition $x^h = x^g$ for all $x^g$ defined). Therefore valuations considered at different accessible worlds are independent from each other, but all preserve the assignments of $g$. This is desired, since interpretations of lexical and thematic predicates may vary across worlds and satisfactory valuations thus need not stay the same. On the other hand, as we will see later, assignment preservation is what accounts for *de re* readings of attitudes.

### 2.2.4 Quantification

To develop the subgraph induction story, we now consider how to proceed if the root $x$ of a semgraph $G$ is the dummy variable of a quantification structure, schematized as follows.

(2.41)

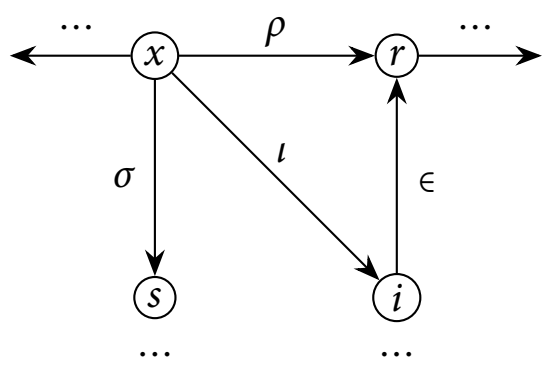

As a convention of this section, we write $r$, $i$, and $s$ for the restrictor, the iterator, and the scope (see Section 2.1.6). We understand quantification as iteratively making statements for an amount of entities in the restrictor, a contextually salient plurality.

As an approximation, we may think of contextual salience as maximality. With *every boy*, for example, we would normally want to go over the set of all boys in the domain.

**Definition 2.4.** $x^g$ is *maximal at* $w$ w.r.t. $\eta$ if $x \in \eta$ or given $G$ reachable from $x$, $x^g \not\subset x^h$ whenever $w, h \vDash_\eta G$.

This definition takes the value of $x$ to be maximal if $x$ has been visited, or it cannot be enlarged and still satisfies $G$ at $w$. It follows that $x^g$ can be empty while maximal at $w$ if no $h$ satisfies $G$ at $w$.

We may then iterate over the elements in a salient restrictor, stating constraints for $x$-*many*, by which we mean the quantification force given by the unary out-edge of $x$. For the universal and the negative quantifiers, the membership edge $\in \overrightarrow{ir}$ allows this to be done by iterating over valuations satisfying the iterator subgraph and checking their extendability. The idea is comparable with DRT's treatment of quantification (see Section 2.3.3).

**Definition 2.5.** (Continuing Definition 2.2).

viii) If $x$ roots a quantification structure then $w, g \vDash_\eta G$ iff

- viii.a) $r^g$ is maximal at $w$ w.r.t. $\eta$ and $r^g \neq \varnothing \Rightarrow w, g \vDash_\eta G/r$;
- viii.b) $g$ is undefined on $V_{G-T(r)/i} \cup V_{G-T(i)/s} - \eta - [r]$;
- viii.c) for $x$-many minimal $h \supseteq g$ s.t. $w, h \vDash_{\eta\cup[r]} G - T(r)/i$, there is $k \supseteq h$ s.t. $w, k \vDash_{\eta\cup[i]} G - T(i)/s$.

Here (viii.a) fixes a salient value for $r$, that is, a maximum constrained by the restrictor subgraph. As (2.23a) shows, what constitutes a restrictor graph can be as simple as a unary edge contributed by nouns. Further constraints, often referencing a representation of contexts (von Fintel, 1994; Stanley and Szabó, 2000), can be incorporated to limit the elements to be iterated over. We will meet examples of such kind in Section 2.3.2.

We then iterate over *minimal* extensions of $g$ that satisfy the iterator subgraph, counting $x$-many extendable to satisfy the scope subgraph. (viii.b) requires $g$ be undefined on any vertex of the iterator subgraph or of the scope subgraph if it is neither in the visiting history nor in what the restrictor subgraph will add to that. Otherwise iteration would be nullified; were $g$ defined on $i$, for example, we would have $i^h = i^g$ for any $h$ extending $g$.

Similarly, (viii.c) counts only $h$ minimally extending $g$ to satisfy the iterator subgraph (that is, no proper subset of $h$ extends $g$ while satisfying the iterator subgraph). Otherwise extendability could be falsely denied. For example, suppose $h$ satisfies the iterator subgraph and is defined on $s$. If $h$ cannot be extended to satisfy the scope subgraph, but if $h' \subset h$ undefined on $s$ can, then it would be wrong to say that the valuation of the iterator subgraph by $h$ cannot be extended to satisfy the scope subgraph.

We show how (viii) works with two earlier examples, gathering their truth conditions at some world in (2.43). One can verify that they are equivalent to our informal description in Section 2.1.6.

(2.42) a. Every boy sailed.

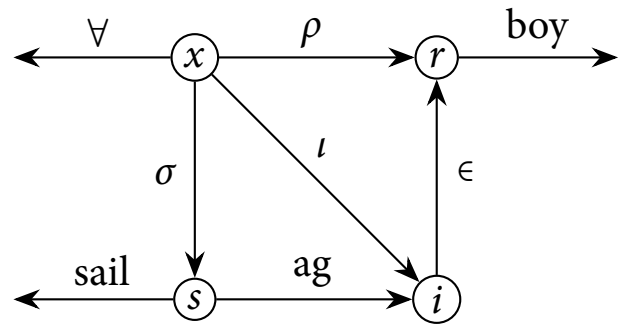

b. If a boy walks a dog, he feeds it.

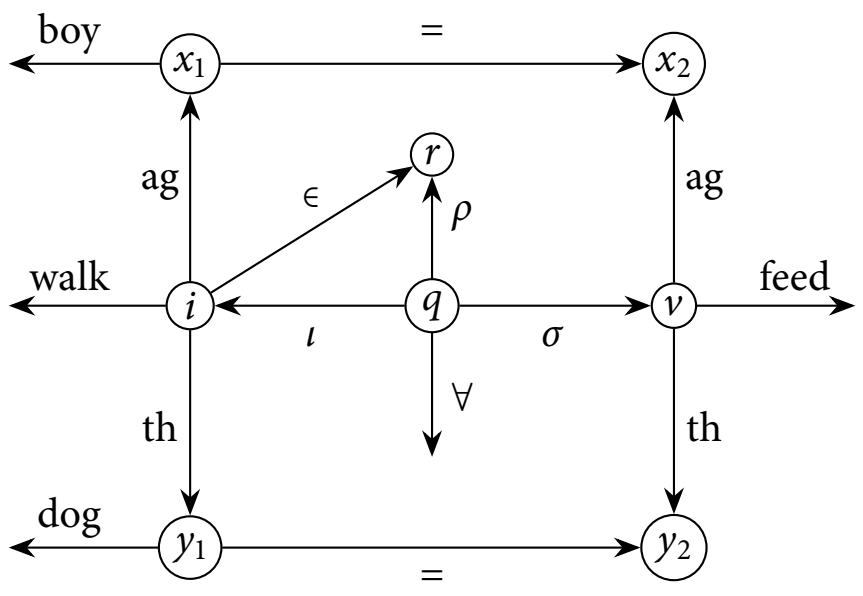

(2.43) a. There is some $g$ such that
  i. $r^g = boy_w$;
  ii. $i^g$ is undefined;
  iii. for each $h \supseteq g$ s.t. $|i^h| = 1$ and $s^h$ is undefined,
  there is $k \supseteq h$ s.t. $s^k \subseteq sail_w$, $i^k = \bigcup_{a \in s^k} ag_w(a)$.

b. There is some $g$ such that
  i. $r^g =$ the domain $D$;
  ii. $i^g, x_1^g, y_1^g, x_2^g, y_2^g$ are undefined;
  iii. for each $h \supseteq g$ s.t. $|i^h| = 1$, $v^h$ is undefined,
  $i^h \subseteq walk_w$, $ag_w(i^h) = x_1^h = x_2^h \subseteq boy_w$, $th_w(i^h) = y_1^h = y_2^h \subseteq dog_w$,
  there is $k \supseteq h$ s.t. $v^k \subseteq feed_w$, $x_2^k = \bigcup_{a \in v^k} ag_w(a)$, $y_2^k = \bigcup_{a \in v^k} th_w(a)$.

On the other hand, proportional determiners like *most* need a more sophisticated treatment than iterating over valuations. It turns out that in such cases, counting valuations is not exactly the same as counting elements in the restrictor. The so-called *proportion problem* (Kadmon, 1987) can be illustrated with a concrete example:

(2.44) Most boys who gazed-at a star sailed.

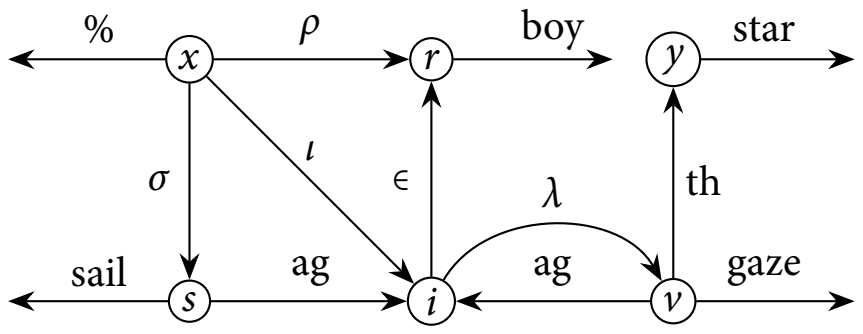

Imagine that there were a hundred boys; one of them gazed at a thousand stars and sailed, and the rest gazed at one star but never sailed. Here the sentence sounds false, but most valuations of $i, v, y$ that satisfy the iterator subgraph, precisely those assigning $i$ the sailing boy, do find some extension that satisfies the scope subgraph.

To actually count star-gazing boys, we may rather iterate over *classes* of all valuations satisfying the iterator subgraph, such that two valuations belong to the same class if they agree on $i$. For the situation above, valuations from most classes cannot be extended to satisfy the scope subgraph. Generalizing this scenario, we can revise Definition 2.5 as follows.

**Definition 2.6.** (Revising Definition 2.5).

viii) If $x$ roots a quantification structure then $w, g \vDash_\eta G$ iff

- viii.a) $r^g$ is maximal at $w$ w.r.t. $\eta$ and $r^g \neq \varnothing \Rightarrow w, g \vDash_\eta G/r$;
- viii.b) $g$ is undefined on $V_{G-T(r)/i} \cup V_{G-T(i)/s} - \eta - [r]$;
- viii.c) classify minimal $h \supseteq g$ s.t. $w, h \vDash_{\eta \cup [r]} G - T(r)/i$
  according to the value of $i$,
  for all/some $h$ from $x$-many classes, there is $k \supseteq h$ s.t. $w, k \vDash_{\eta \cup [i]} G - T(i)/s$.

This revision changes only (viii.c). Relevant valuations are divided into classes according to their assignment to the root of the iterator subgraph, which conveniently coincides with the iterator itself.

We leave it open, for a class to count, whether we want to verify the extendability of all or some of its valuations. Consider the following example that illustrates the difference.

(2.45) Most boys who gazed-at a star named it. (% *boy* > ∃ *star*)

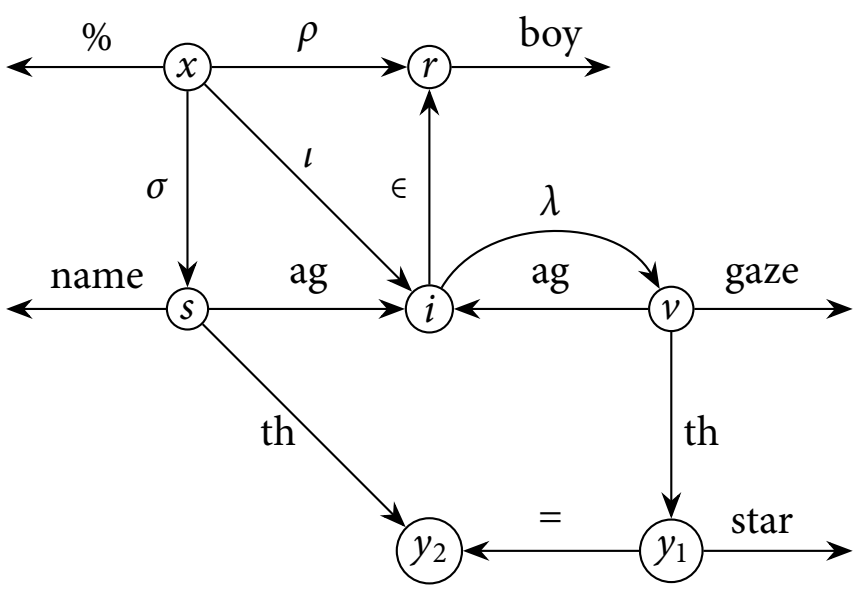

Each valuation class checked by (viii.c) now corresponds to a star-gazing boy with a list of stars he gazed at. For such a boy to count, should he name all listed stars or just some? The reading of (2.45) that requires *all* and the one that requires *some* are respectively known as the *strong* and *weak* readings of donkey sentences (Chierchia, 1992; Heim, 1990; Kanazawa, 1994).

Whether we face a genuine ambiguity here turns out to be debatable. When the iterator of a quantificational structure is constrained only by a membership edge as in (2.42), the two readings coincide. Sometimes one of them vanishes: even if the semantics of (2.45) might not be clear, the following example seems to have only the weak reading, which requires that no boy name *any* star he gazed at and happens to be the reading given by Definition 2.5.

(2.46) No boy who gazed-at a star named it.

Note, in examples like (2.44) and (2.45), that the subgraph contributed by a relative clause is imposed as a constraint on the iterator, not the restrictor. The reason is that the constraint in question seems strictly distributive over the elements of the restrictor. To see what this means, let us compare what happens when the iterator and the restrictor are constrained, respectively as in (2.47a&b).

(2.47) Every boy who washed a dog sighed.

a.

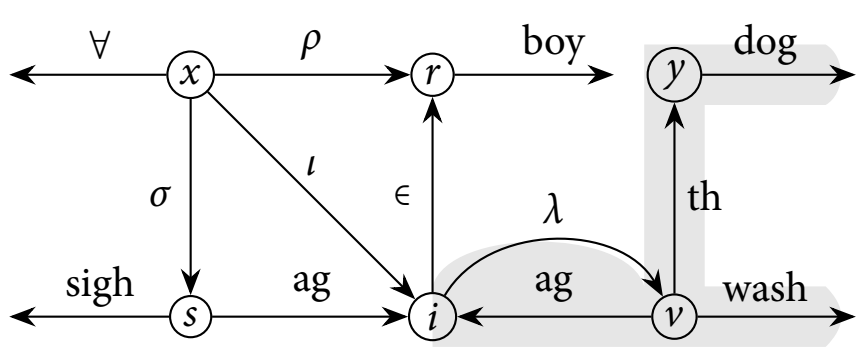

b.

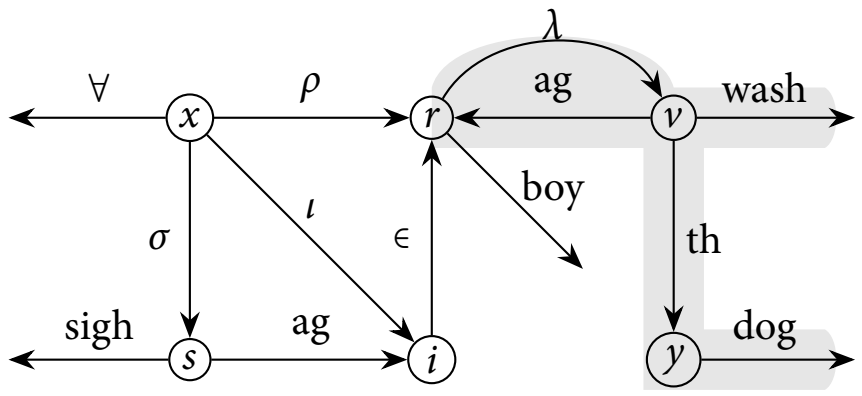

(2.47a) iterates over the maximal set of boys and then verifies that each one that washed a dog sighed. By contrast, (2.47b) iterates over the maximal set of boys that washed a dog and then verifies that each one sighed. Thus the relative clause applies distributively in the former, but as plural predication in the latter.

Now consider a situation where among five boys, one washed a dog and sighed, while the other four jointly washed that dog but none sighed. Here (2.47b) is not satisfiable but (2.47a) is, agreeing with our intuition that (2.47) holds in this situation. If the relative clause were to constrain the restrictor, we should expect otherwise.

Under a quantifier, modifiers like relative clauses are often treated as a predicate on par with the noun they modify (e.g. Heim and Kratzer, 1998), and thereby are said to be *explicit domain restriction*. The above discussion suggests they might be better treated as constraints on the iterator, not the restrictor. For this reason, we might call these modifiers *iterator filters*.

### 2.2.5 Coordination

To complete the subgraph induction story, below we consider what if the root $x$ of a semgraph $G$ roots a coordination structure.

A coordination structure can specifically be a conjunction structure, which represents a semantics of summation; or a disjunction structure, which represents a semantics of choice. Writing $y, z$ for the successors of $x$, the two cases can be implemented as follows.

**Definition 2.7.** (Continuing Definition 2.2).

ix) If $x$ roots a conjunction structure then $w, g \vDash_\eta G$ iff

ix.a) $x^g = y^g \cup z^g$;

ix.b) there are $h, k \supseteq g$ s.t. $w, h \vDash_{\eta \cup [x]} G - T(x)/y$ and $w, k \vDash_{\eta \cup [x]} G - T(x)/z$.

x) If $x$ roots a disjunction structure then $w, g \vDash_\eta G$ iff

x.a) $x^g = y^g$ and $w, g \vDash_{\eta \cup [x]} G - T(x)/y$;

x.b) *or* $x^g = z^g$ and $w, g \vDash_{\eta \cup [x]} G - T(x)/z$.

The summation aspect of conjunction is given by (ix.a) and its logical aspect by (ix.b), where the conjunct subgraphs reachable from $y$ and $z$ are evaluated by independent extensions $h, k$ of $g$. On the other hand, in (x) the logical aspect of disjunction is implied by the semantics of choice: one of the successors is chosen and the disjunct subgraph therefrom is evaluated, irrespective of however the evaluation of the other would be. Actually, the successor left not chosen can even be valuated as an empty set.

To see (ix) and (x) in action, consider these examples:

(2.48) Joe rented and/or Ben sank a boat.

a.

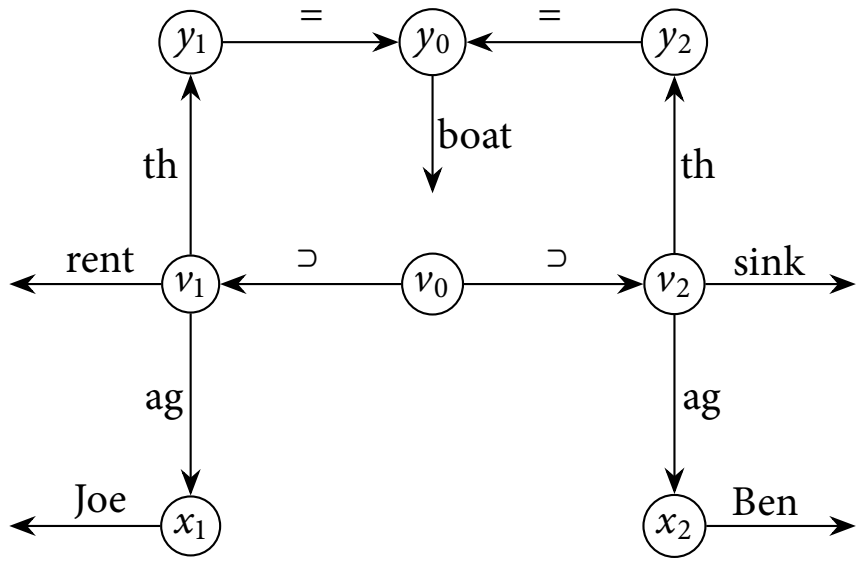

b.

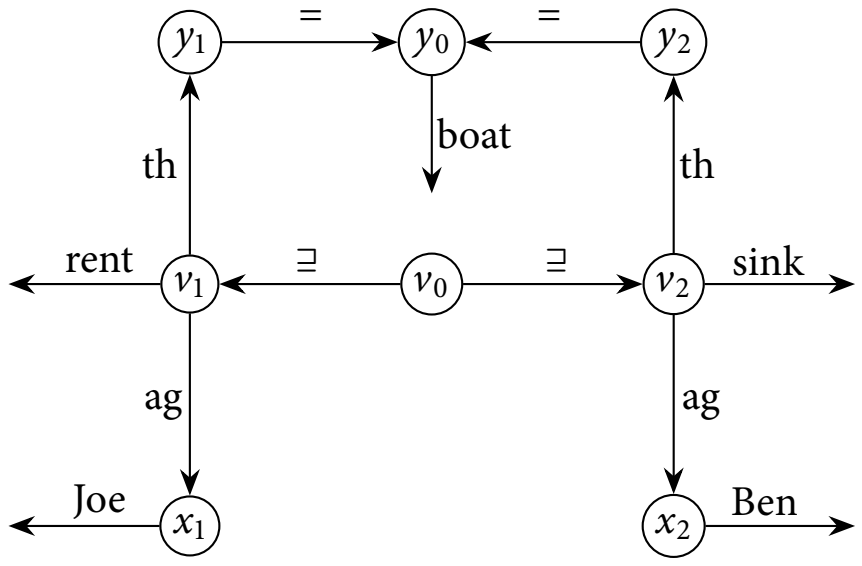

(2.48a&b) share the same subgraphs reachable from $v_1$ and $v_2$:

(2.49) a. Joe rented a boat. b. Ben sank a boat.

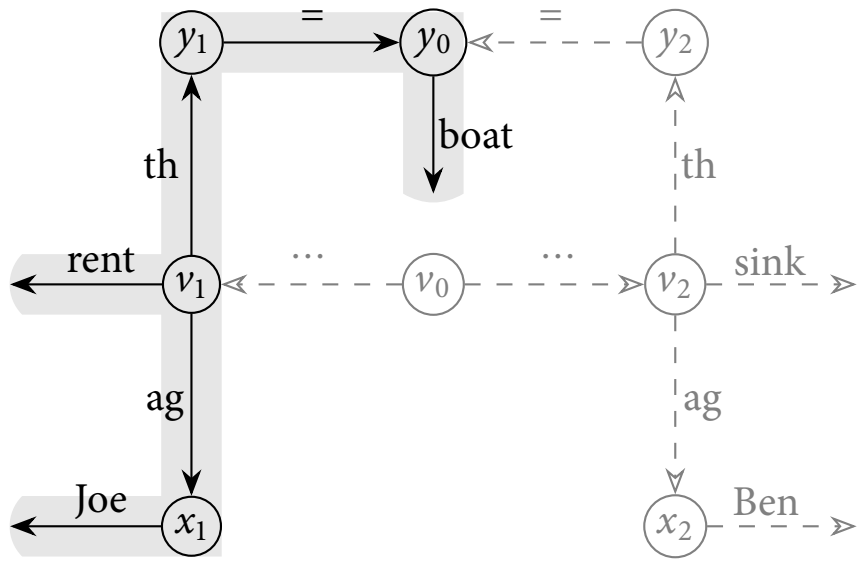

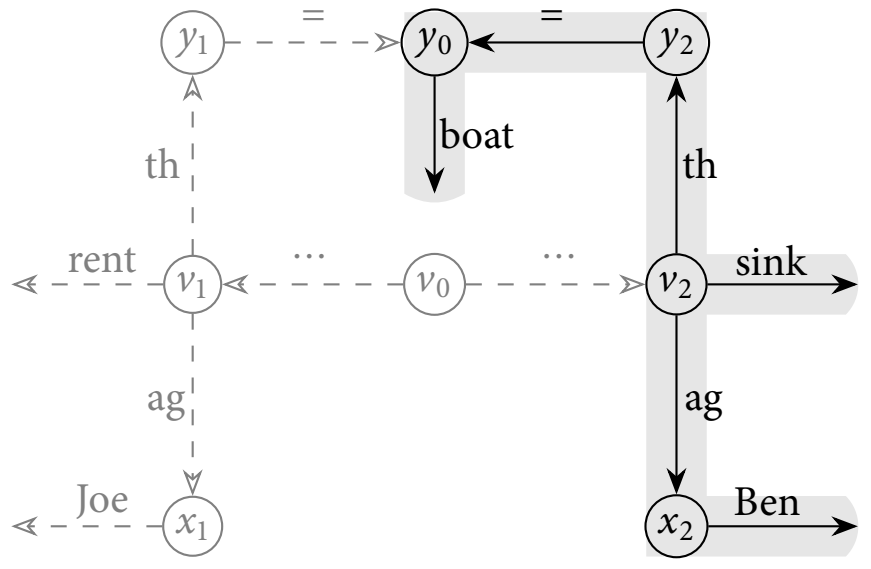

The truth condition of (2.48a) is then gathered as follows.

(2.50) There are $h, k$ such that

a. $h, k$ agree on $v_0, v_1, v_2$;
b. $v_0^h = v_1^h \cup v_2^h$;
c. $h$ satisfies (2.49a);
d. $k$ satisfies (2.49b).

Since $h$ and $k$ may or may not agree on $y_0$, the boat Ben sank may or may not be the same boat Joe rented.

The truth condition of (2.48b) is given as follows.

(2.51) There is $g$ such that

a. $v_0^g = v_1^g$ and $g$ satisfies (2.49a);
b. *or* $v_0^g = v_2^g$ and $g$ satisfies (2.49b).

The semantics of choice thus evaluates a part of the semgraph and ignores the rest. When (2.51) holds because of (2.51a), for example, (2.51b) may or may not hold; $g$ may assign the empty set to $v_2$ or even be undefined on $v_2$.

### 2.2.6 Order of Valuation

The semantics of intensionality, quantification, and coordination introduces dependencies between valuations as one extends another. When evaluating a semgraph, a chain of valuations $g_1 \subseteq \cdots \subseteq g_n$ induces an order on valuation of vertices. For any two vertices $x$ and $y$, if both are in the domain of $g_n$, we may find in the chain the first valuation $g_i$ that must be defined on $x$ and the first valuation $g_j$ that must be defined on $y$: we say that *$x$ is valuated before $y$* if $i < j$. (The relations *valuated-simultaneously* and *valuated-after* can be analogously defined.)

Here are a few examples.

(2.52) a. Ben thought that Joe walked a dog. (*think* > ∃ *dog*)

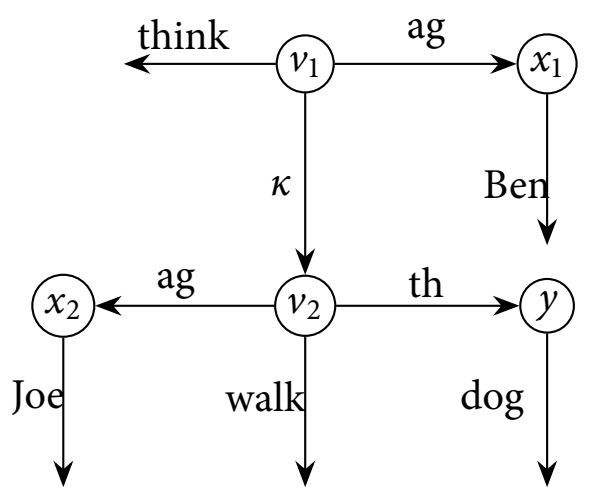

b. Every boy walked a dog. (∀ *boy* > ∃ *dog*)

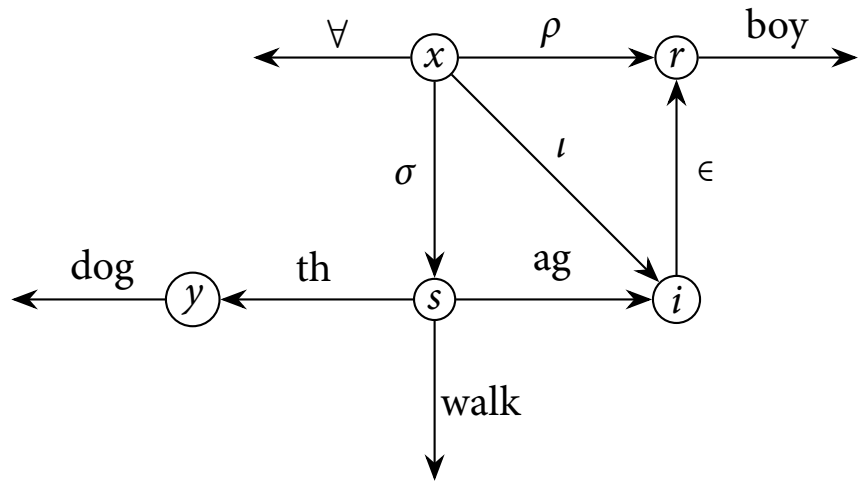

c. Joe walked and Ben fed a dog. (*and* > ∃ *dog*)

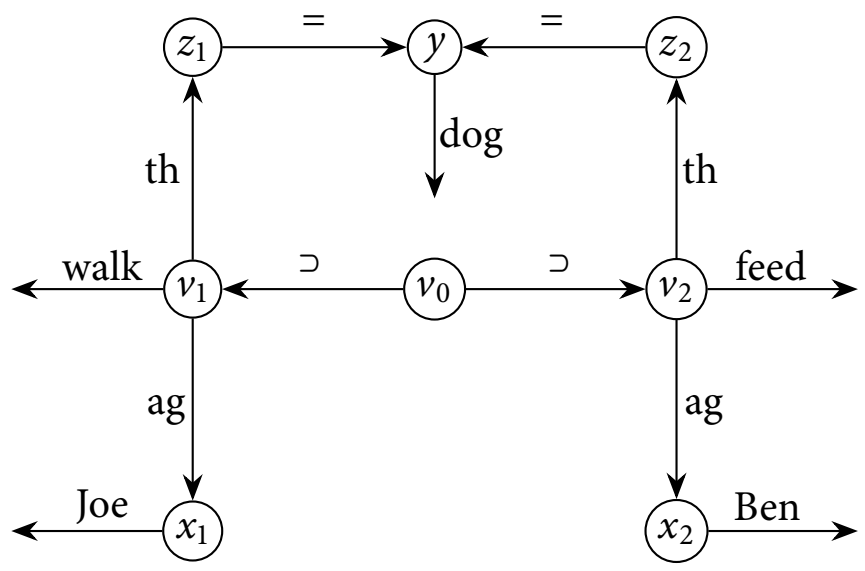

According to the interpretation rules discussed so far, we can find $v_1$ valuated before $y$ in (2.52a); $r$ before $i$ and $i$ before $y$ in (2.52b); $v_0$ before $y$ in (2.52c).

This order of valuation reflects precisely how one variable varies in relation to another. Having $v_1$ fixed in (2.52a), $y$ can vary as the subgraph reachable from $v_2$ is verified at each world compatible with the thought in $v_1$. Having a set $r$ of boys fixed in (2.52b), the dog in $y$ can vary across each $i$ drawn from $r$ (by the same token, walking events $s$ can vary with $i$ as well). Having $v_0$ fixed in (2.52c), the two conjunct subgraphs can be satisfied by different values of $y$.

But that one variable's variation depends on the other is but another way of saying that the former takes scope under the latter. The relative scope of variables thus arises from the order of their valuation. Now that this order is determined by semgraph structure, for variables to take scope we can shift their order of valuation with a $\lambda$-edge, which does the job by shifting rootship.

Take the examples from above. Pointing to the original roots in (2.52), below the $\lambda$-edges make the referents $y$ of the indefinites the new roots, according to Definition 1.3.

(2.53) a. Ben thought that Joe walked **a dog**. ($\exists$ *dog* > *think*)

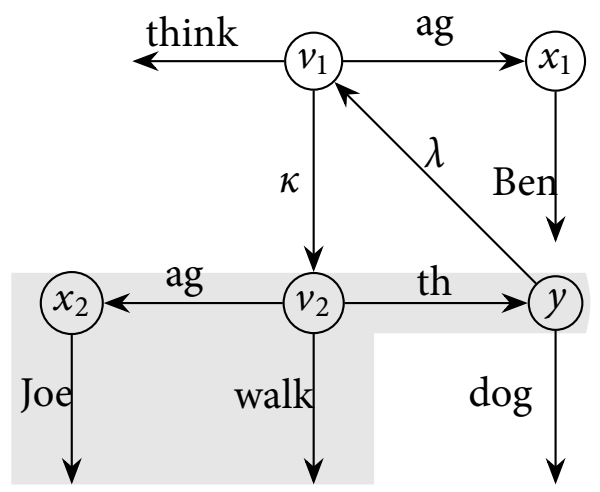

b. Every boy walked **a dog**. ($\exists$ *dog* > $\forall$ *boy*)

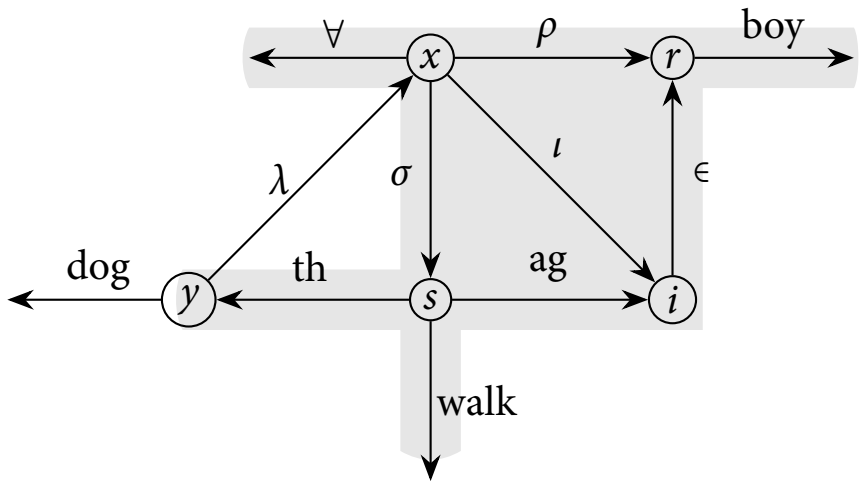

c. Joe walked and Ben fed **a dog**. ($\exists$ *dog* > *and*)

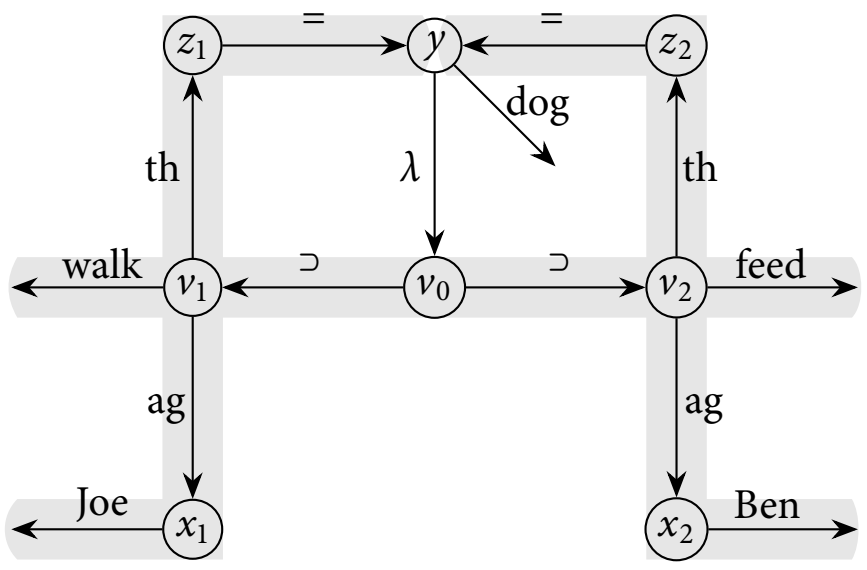

One may verify that $y$ is valuated at the same time with $v_1$ in (2.53a); before $i$ and $s$ in (2.53b); at the same time with $v_0$ in (2.53c).

Therefore, in (2.53a) the value of $y$ stays constant as we evaluate the subgraph reachable from $v_2$ (the shaded area) at different worlds compatible with Ben's thought. Particularly, when inducing the subgraph in question $dog\overrightarrow{y}$ is excluded by Definition 2.2; it is a constraint checked once at the reference world. Since the fixed $y$ may or may not be a dog at all those accessible worlds, that is, in Ben's thought, we get the so-called *de re* reading of the indefinite, as opposed to its *de dicto* reading shown by (2.52a).

Similarly, in (2.53b) the value of $y$ is fixed in the valuation $g$ satisfying the restrictor subgraph, and stays constant across the valuations $h \supseteq g$ satisfying the iterator subgraph and the valuations $k \supseteq h$ satisfying the scope subgraph. As $y$ is the root, the subgraph reachable from $x$ (the shaded area) is evaluated with respect to a visiting history $\eta$ that contains $y$. So the requirements on the domains of $g$ and $h$ by (viii) in Definition 2.5/2.6 are met.

We can likewise scope a variable in the iterator subgraph. Compare the following example, which is about a specific star, with (2.45):

(2.54) Most boys who gazed-at **a star** named **it**. ($\exists$ *star* > % *boy*)

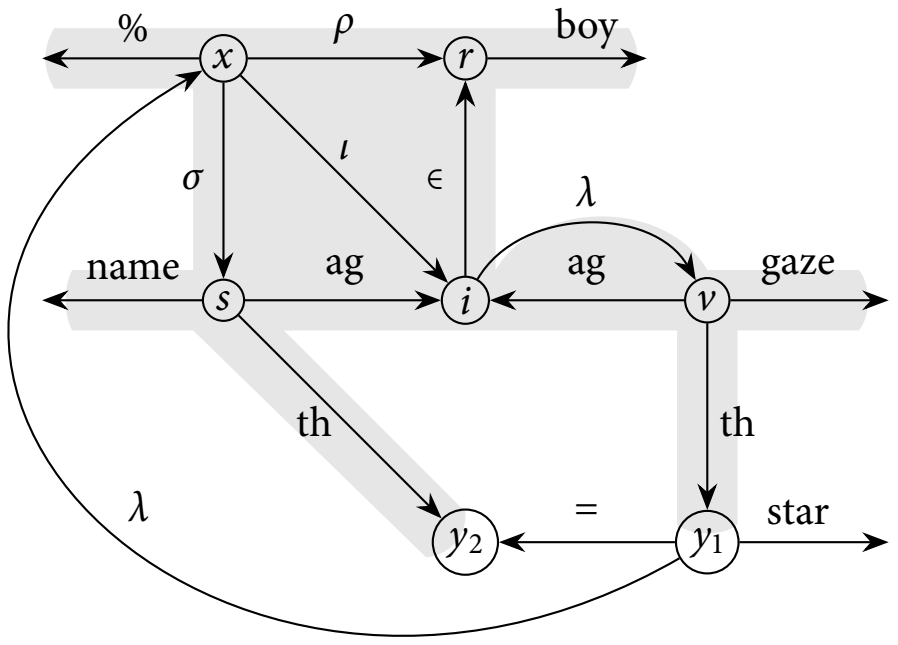

(viii) in Definition 2.5/2.6 requires evaluating the subgraph reachable from $x$ (the shaded area) with respect to $\eta$ that contains $y_1$ and $y_2$. That is why Definition 2.2 updates $\eta$ with more than the local root.

A variable in the iterator subgraph can become the latter's root by taking "local scope". Since in proportional quantification, it is precisely the root of the iterator subgraph that gives the criterion for classifying valuations (see Definition 2.6), we can use scope taking to handle a subtlety in adverbial proportional quantification. Kadmon (1987) mentions that sentences like (2.55) have three readings, describing a tendency of general state-of-affairs, of dog-walking boys, and of boy-walked dogs. They will arise, respectively, when $v_1$ stays the root of the iterator subgraph in (2.55a), when $x_1$ takes over rootship in (2.55b), when $y_1$ does so in (2.55c).

(2.55) Mostly if a boy walked a dog, he fed it.

a. (count walks)

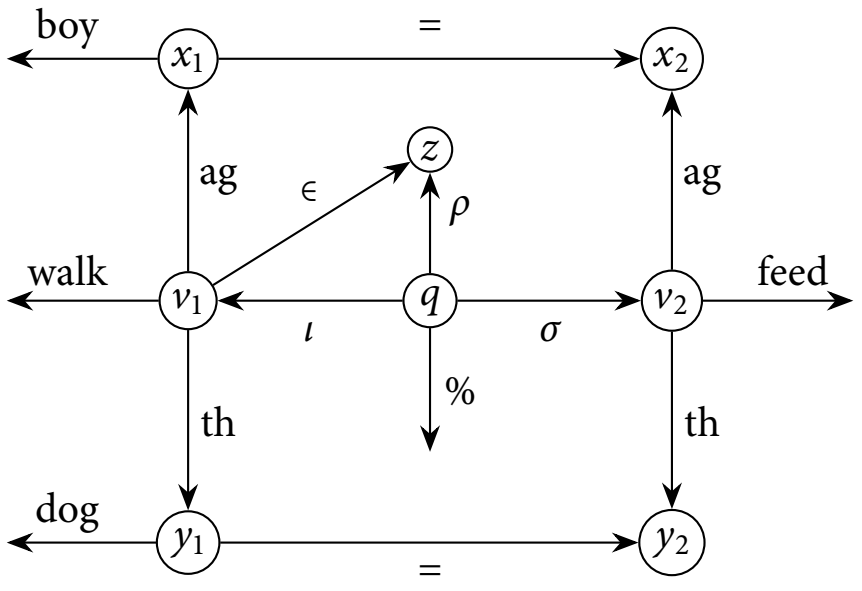

b. (count boys)

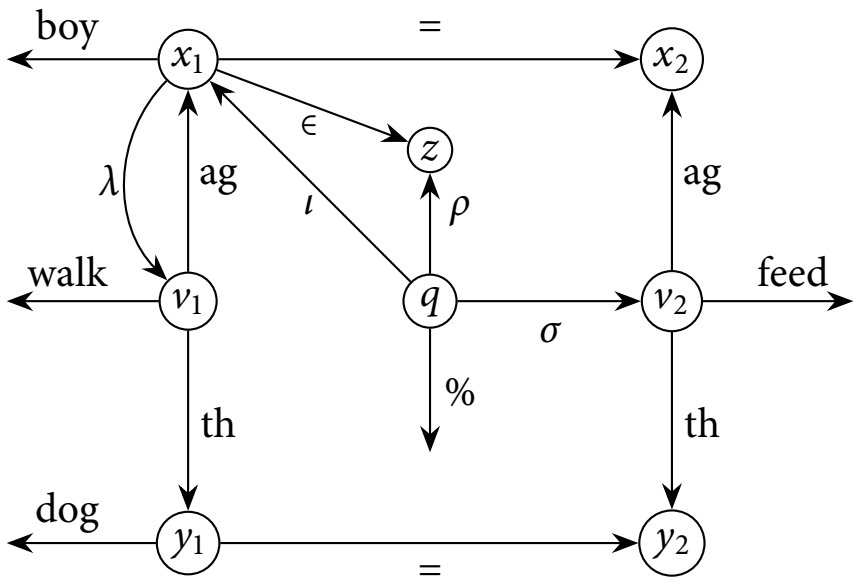

c. 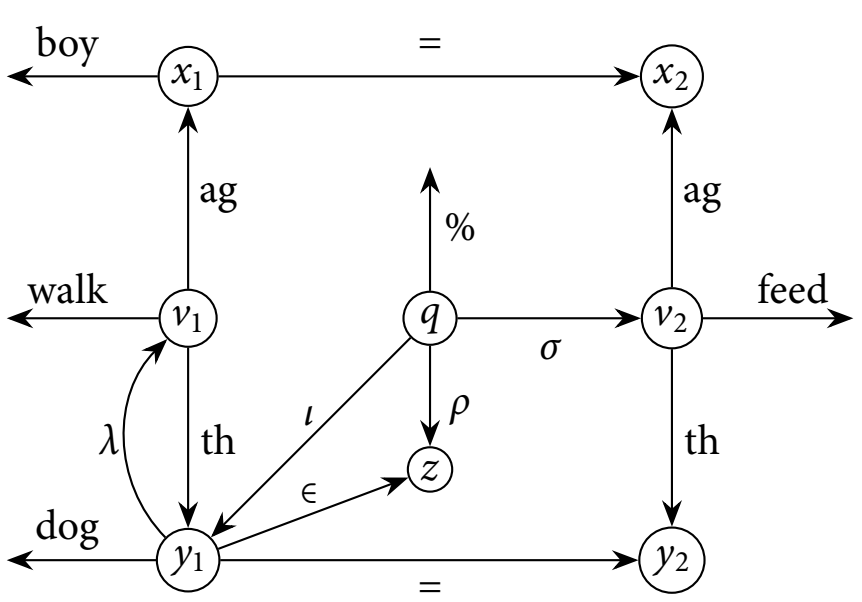

(count dogs)

We notice that local-scope taking as such does not alter the semantics of a local environment; the iterator subgraphs of (2.55a-c) indeed share the same truth condition.

Turning to (2.53c), the value of $y$ is fixed along with $v_0$ and thereby preserved in evaluation of either conjunct subgraph. Scope taking also ensures co-reference between coordinate subgraphs. For example, $x_1$ being the root in (2.56a), the disjunction structure is evaluated with the edge constraint $= \overrightarrow{x_1 x_2}$ already satisfied.

(2.56) Joe sailed or he surfed.

a. 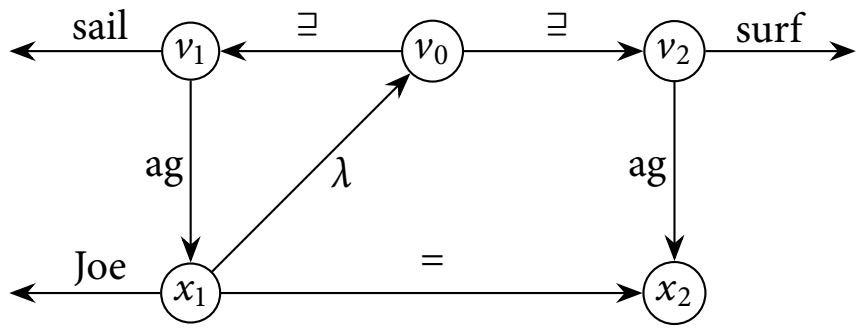

b. 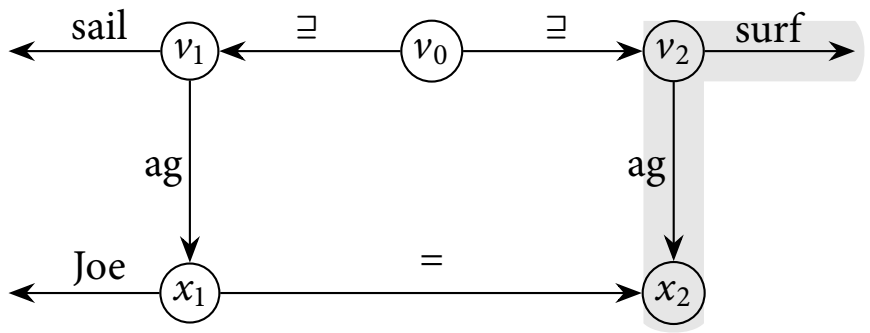

In (2.56b), however, the disjunction structure is satisfiable when the shaded disjunct subgraph alone is satisfied, which means that $x_2$ may or may not be Joe.

Scoping a disjunct is worth special attention. When $z$ takes scope in (2.57), the existence of a dog is asserted before $y$ chooses between $z$ and $s$. In other words, the (shaded) subgraph off $z$ must be satisfied, whether $z$ is later chosen or not. This differs from the non-scoping case; were $\lambda\overrightarrow{zv}$ removed, the subgraph off $z$ need not be satisfied when $y$ chooses $s$.

(2.57) Joe saw **a dog** or a cat.

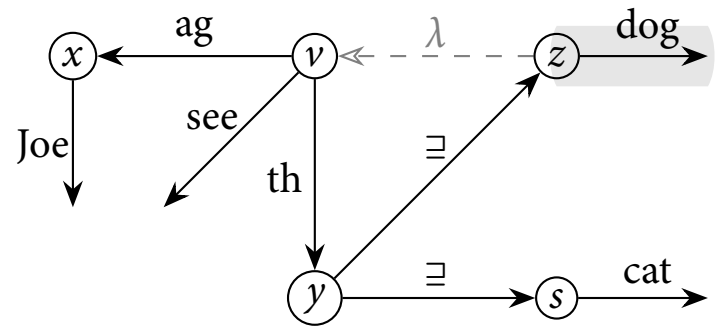

There seems no reason to satisfy an unchosen disjunct subgraph merely for the reason of scoping, so we need to suspend the evaluation of the subgraph off a scoping disjunct (e.g. $z$), until it is later chosen by disjunction (e.g. $y$).

To this end, we need to revise (i) of Definition 2.2, which so far evaluates all the out-edges of a root. When a root is a scoping disjunct, all its out-edges can be ignored but for the $\lambda$-edge by which it takes scope. We define

$$\overline{T}(x) = \begin{cases} \{e \in T(x) \mid f \text{ encycles } e\} & \text{if } x \text{ heads some } \sqsupseteq\text{-edge } f, \\ T(x) & \text{otherwise,} \end{cases}$$

and replace $T(x)$ in (i) of Definition 2.2 with $\overline{T}(x)$. We will see in Section 5.2.3 how this consideration allows us to approximate exceptionally scoped disjunction, while not over-generating exceptionally scoped conjunction.

To conclude this section, we collect all pieces of the semgraph interpreter in one place.

**Definition 2.8.** Let $\mathcal{M} = (D, W, g)$ be a model, $w \in W$ and $G$ a semgraph rooted at $x$.

i) If $x$ does not root a special structure then $w, g \vDash_\eta G$ iff

- i.a) when $P\overrightarrow{x} \in \overline{T}(x)$,
  $w, g \vDash_\eta P\overrightarrow{x}$;
- i.b) when $R\overrightarrow{xy} \in \overline{T}(x)$ and $R \neq \kappa$,
  $w, g \vDash_\eta R\overrightarrow{xy}$ and $w, g \vDash_{\eta \cup [x]} G - \overline{T}(x)/y$;
- i.c) when $\kappa\overrightarrow{xy} \in \overline{T}(x)$ and $a \in x^g$,
  for all/some $w' \in a_w$ there is $h \supseteq g$ s.t. $w', h \vDash_{\eta \cup [x]} G - \overline{T}(x)/y$.

ii) $w, g \vDash_\eta \lambda\overrightarrow{xy}$;
iii) $w, g \vDash_\eta = \overrightarrow{xy}$ iff $y^g = x^g$;
iv) $w, g \vDash_\eta \#\overrightarrow{xy}$ iff $y^g = \{|x^g|\}$;
v) $w, g \vDash_\eta \in \overrightarrow{xy}$ iff $x^g \subseteq y^g$ and $|x^g| = 1$;
vi) $w, g \vDash_\eta P\overrightarrow{x}$ iff $x^g \subseteq P_w$ and $x^g \neq \varnothing$;
vii) $w, g \vDash_\eta R\overrightarrow{xy}$ iff $y^g = \bigcup_{a \in x^g} R_w(a)$.

viii) If $x$ roots a quantification structure then $w, g \vDash_\eta G$ iff

- viii.a) $r^g$ is maximal at $w$ w.r.t. $\eta$ and $r^g \neq \varnothing \Rightarrow w, g \vDash_\eta G/r$;
- viii.b) $g$ is undefined on $V_{G-T(r)/i} \cup V_{G-T(i)/s} - \eta - [r]$;
- viii.c) classify minimal $h \supseteq g$ s.t. $w, h \vDash_{\eta \cup [r]} G - T(r)/i$
  according to the value of $i$,
  for all/some $h$ from $x$-many classes, there is $k \supseteq h$ s.t. $w, k \vDash_{\eta \cup [i]} G - T(i)/s$.

ix) If $x$ roots a conjunction structure then $w, g \vDash_\eta G$ iff

- ix.a) $x^g = y^g \cup z^g$;
- ix.b) there are $h, k \supseteq g$ s.t. $w, h \vDash_{\eta \cup [x]} G - T(x)/y$ and $w, k \vDash_{\eta \cup [x]} G - T(x)/z$.

x) If $x$ roots a disjunction structure then $w, g \vDash_\eta G$ iff

- x.a) $x^g = y^g$ and $w, g \vDash_{\eta \cup [x]} G - T(x)/y$;
- x.b) *or* $x^g = z^g$ and $w, g \vDash_{\eta \cup [x]} G - T(x)/z$.

## 2.3 Related work

Having semgraphs and their interpreter in place, this section reviews the previous graph(-like) formalisms from which the current work draws its inspirations. The aim is not to be comprehensive, but to provide comparisons that further illustrate some distinctive features of our semgraphs.

### 2.3.1 Abstract Meaning Representation

Our graph language is built on Abstract Meaning Representation (AMR; Banarescu et al., 2013; Bonial et al., 2018), a semantic annotation scheme proposed to facilitate machine learning-based natural language understanding and generation.

AMR aims to cover a wide range of phenomena found in engineering applications, but not always with enough linguistic adequacy to serve theoretical investigations. It does not concern itself with their model-theoretical semantics, and it is agnostic about how they can be constructed from natural language sentences. Full documentation of AMR is given in its annotation guideline.[2] Below we sketch the basics of this language at a high level.

Adopting neo-Davidsonian semantics, AMR renders linguistic expressions as uniquely rooted directed acyclic graphs (here a root is a vertex with no in-edges), with vertex labels indicating concepts and edge labels semantic relations. For example, (2.58a) gives an AMR graph, where we replace PropBank (Palmer et al., 2005) relation labels used by AMR (e.g. ARG0, ARG1, ...) in keeping with the conventions of this thesis (e.g. *ag*, *th*, ...). (2.58b) gives the induced tree encoding of (2.58a), which is more prevalent in the literature ("*v* / *walk*" means *v* is an instance of *walk*; "*v* ... :*ag* (*x* ...)" indicates the *ag* relation between *v* and *x*).

(2.58) A boy walked a dog.

a. 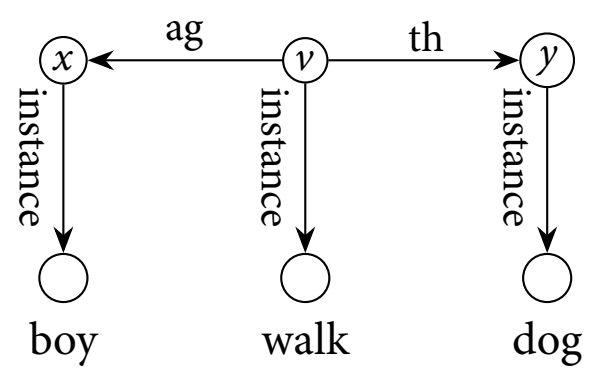

b.
```
(v / walk
   :ag (x / boy)
   :th (y / dog))
```

One may compare (2.58) with (1.3b). In AMR graphs, vertices can denote either variables ($x, v, \ldots$) or concepts (*boy*, *walk*, ...) while edges are consistently binary. By contrast, we use vertices consistently for variables and labeled unary edges for instantiation of concepts (i.e. unary predicates). These are merely notational variants, but we do find our choice making the structure of semgraphs simpler and description of semgraph semantics easier. For one thing, in Section 2.2.2, worlds interpret lexical and thematic edges on the one hand and valuations handle all vertices on the other. With the original notation of AMR we would not have this clear division.

Thus far AMR graphs and ours are not different in essentials. We may consider a few cases where they are. To begin with, AMR uses inverse relations like *th-of* to avoid multi-rootedness, a problem that would otherwise come with modification and we solve with $\lambda$-edges (see Section 2.1.2). Compare (2.59a&b):

(2.59) Ben saw a dog that Joe walked.

a. 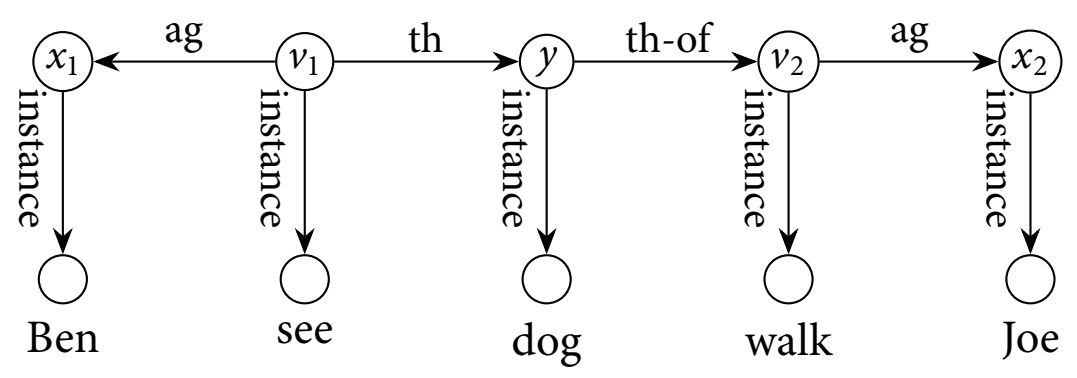

[2] https://github.com/amrisi/amr-guidelines/blob/master/amr.md.

b. 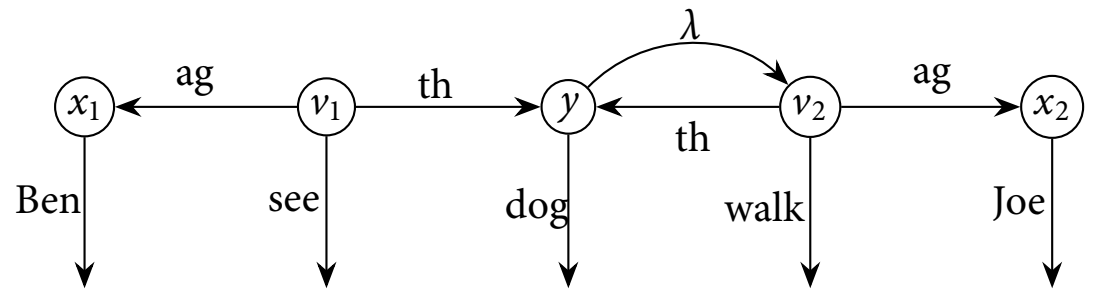

(2.6)

The *th-of* edge in (2.59a) can be seen as the result of compressing the $\lambda$-cycle in (2.59b). This simple solution may work for the majority of practical data, but does not lend to long-distance dependencies. As the following example shows, there is no one-edge substitute for a $\lambda$-cycle of length 3 or above.

(2.60) A dog that Ben thought Joe walked. (2.53a)

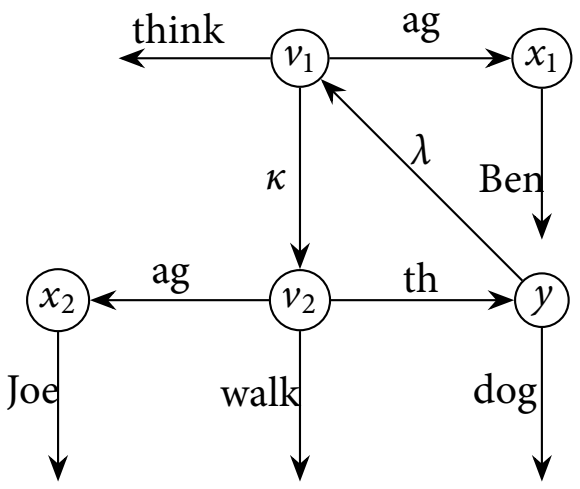

Inverse relations are also incompatible with the principle of compositionality. Suppose that *walk* contributes its *th* edge in (2.59) much as it does in (2.58). Its presence in a relative clause in (2.59) should not alter that contribution.

AMR and its recent reforms represent scope-related phenomena, if ever, in a syntactic style where relative scope is ignored. Banarescu et al. (2013) did not cover intensionality and quantification. Bos (2016) proposes a representation for quantification like this:

(2.61) Every boy walked most dogs.

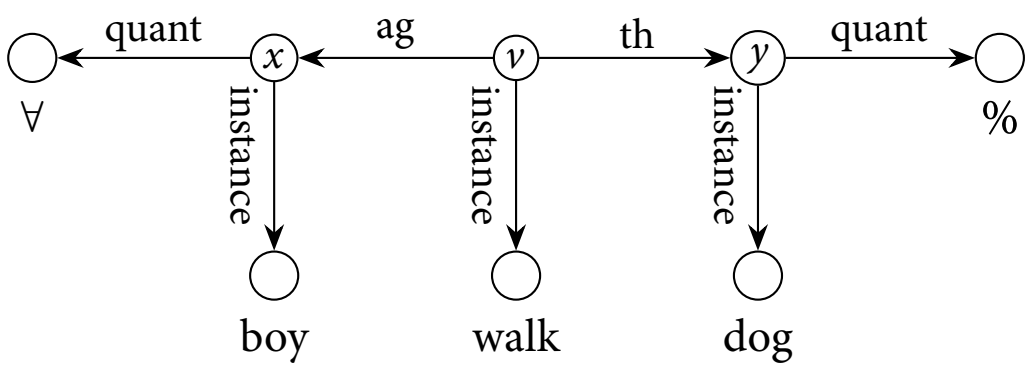

As (2.61) shows, the variables for *boy* and *dog* is each connected to a determiner by a quantity relation. The structure captures the syntactic dependency between a noun and a determiner along the tradition of dependency grammars (e.g. Mel'cuk, 1988), but leaves the scope ambiguity unresolved.

To interpret such representations, Bos proposed a non-deterministic translation from tree encodings of AMR to first-order logic (an extension seems possible if higher-order logic is the preferred target language). The result can give any scope permutation in case of multiple quantifiers. Alternatively, Stabler (2018) translates normalized tree encodings of AMR to higher-order logic deterministically to derive the surface scope, and then generates other permutations using a non-deterministic finite state mechanism.

But if the scope ambiguity is indeed a case of ambiguity rather than underspecification, as often assumed by theorists (but cf. Poesio, 1996), it might be better to pair each scope permutation with its own semantic representation. For the previous example we will construct in Chapter 5 the following, where one quantificational structure nests under the scope subgraph of another:

(2.62) a.

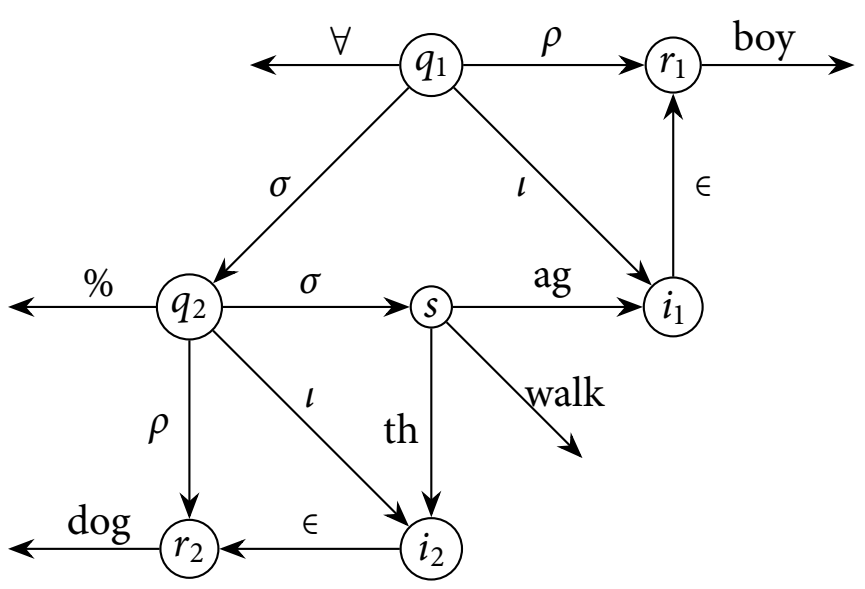

b.

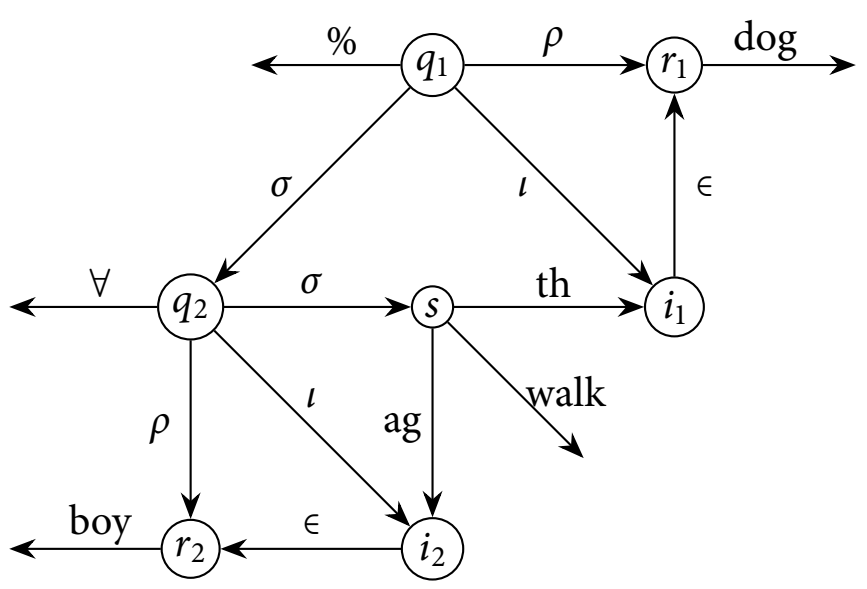

Section 2.2.6 derives scope from the order of valuation. Following the same principle, one can verify that (2.62a&b) represent the surface and the inverse scope readings of (2.61), respectively: since the values of $r_2, i_2, s$ may vary with the value of $i_1$, in (2.62a) we consider an independent majority of dogs for each boy, and in (2.62b) we go through each boy for each one of a fixed majority of dogs.

### 2.3.2 Hybrid Logic Dependency Semantics

The $\lambda$-cycles and quantificational structures of our graph language are adapted from Hybrid Logic Dependency Semantics (HLDS; Baldridge and Kruijff, 2002; Kruijff, 2001; White, 2006), a representation formalism based on hybrid modal logic (Blackburn, 2000).

Modal logic originates in discussion of temporality and intensionality, how propositions hold under different modalities; it talks about the "(possible) worlds" at which a proposition holds and models modalities with accessibility relations between "worlds". But modal logicians recommend a more general perspective: if we think of "worlds" as entities, then propositions and accessibility relations become unary and binary predicates of first-order logic. In a sense, modal logic would be a natural means for describing relational structures, if it had variables for explicitly referencing "worlds", or rather, entities. This extension is fulfilled in hybrid (modal) logic by introducing *nominals*.

Above is the background in which Kruijff (2001) develops HLDS for sentential and discourse meaning. At its basics HLDS expresses semantic dependencies or relations just like AMR. For example, (2.63a&b) show a HLDS formula and its underlying graph, omitting the tense. Here we replace the dependency labels used by HLDS with familiar ones:

(2.63) A boy walked a dog.

a. $v \wedge \textbf{walk}$
$\wedge \langle\textsc{ag}\rangle\,(x \wedge \textbf{boy})$
$\wedge \langle\textsc{th}\rangle\,(y \wedge \textbf{dog})$

b.

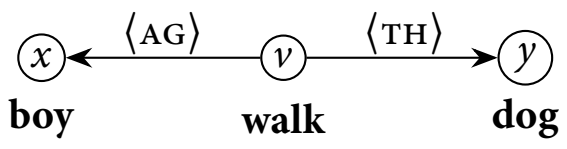

Technically, $v$, $x$, $y$ are nominals; **walk**, **boy**, **dog** propositional symbols; and $\langle\textsc{ag}\rangle$, $\langle\textsc{th}\rangle$ existential modals. Despite the terminology (2.63a) simply means this: there are a walking event $v$, a boy $x$, and a dog $y$ such that $v$ has $x$ as its agent and $y$ as its theme. The formula can be readily seen as a notational variant of (2.58b). Likewise (2.63b) is a notational variant of (1.3b), with unary predicates indicated by vertex labels as in White (2006).

A contribution of White (2006) is the introduction of a cyclic structure for modification and a quantification structure. Kruijff represents adjectival modification with the *general relation* $\langle\textsc{gr}\rangle$ as in (2.64a), which resembles inverse thematic relations of AMR. Extending its use to relative clauses, White places $\langle\textsc{gr}\rangle$ in a cycle as in (2.64b):

(2.64) a. A long book.

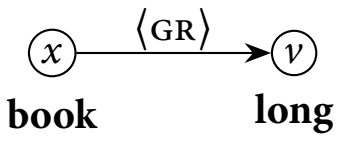

b. A book Joe read.

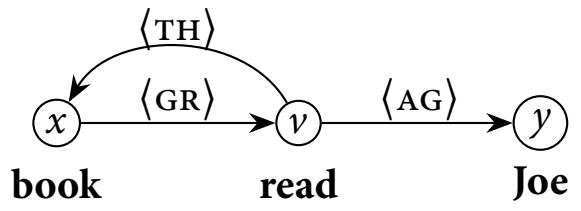

In appearance the cycle here differs from $\lambda$-cycles only by a label. Interpretation-wise, whereas White keeps Kruijff's treatment of ⟨GR⟩ as a thematic relation, in Sections 2.2.3 and 2.2.6, we saw that other than potentially shifting the order of valuation $\lambda$-edges are semantically vacuous. This allows them to serve as a device for scoping indefinites.

The quantification structure proposed by White is quite similar to ours. The example below illustrates both of them side by side:

(2.65) Every boy sailed.

a.

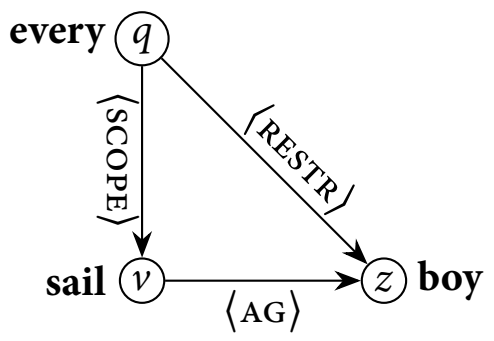

b.

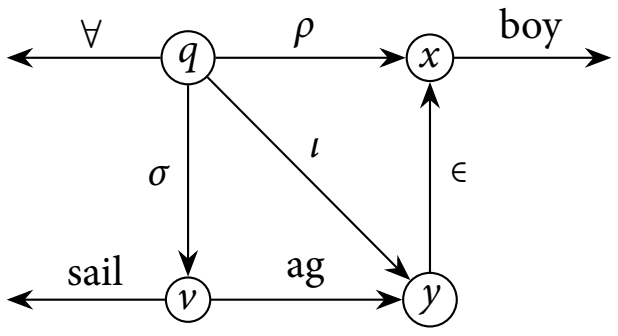

(2.65a) relates a dummy nominal $q$ to a ⟨SCOPE⟩ dependent $v$ and a ⟨RESTR⟩ dependent $z$, the latter of which actually corresponds to the iterator $y$ of (2.65b). The intended semantics of (2.65a) can be given in an iterative style: for each $z$, if $z$ is a boy then there is some sailing $v$ for which $z$ is the agent. Without an explicit restrictor, the domain of quantification in (2.65a) is set to the domain of all entities. Thus the effect is the same as (2.65b) when contextual salience maximizes $x$ to the set of all boys.

We do find evidence, however, in favor of representing the restrictor explicitly. For one thing, it may serve as the antecedent of a pronoun:

(2.66) Every boy sailed and they returned.

Here it is most natural to have *they* co-refer with a plurality of boys, the domain of quantification.

An explicit restrictor can also be helpful if we need maximization subject to further constraints. For example, consider (2.67) in a context that gives salience to all the boys who jointly washed a dog. The restrictor subgraph off $r$ may include the shaded constraint, so $r$ is maximized to the relevant set of boys.

(2.67) Every boy sighed.

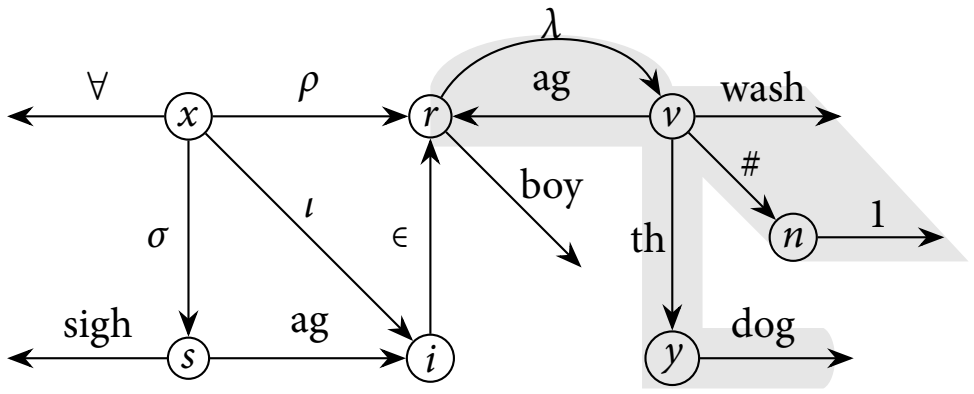

The same cannot be reproduced for the quantification structure of White, for its ⟨RESTR⟩ dependent plays the role of our iterator, but according to the discussion of (2.47), a constraint on the iterator applies to each single element in the domain of quantification. So if we were to incorporate the shaded constraint into a structure like (2.65a), we would obtain this truth condition: for each $z$ in the domain, if $z$ is a boy who washed a dog then $z$ sighed. In a situation where all the boys who jointly washed a dog did sigh and another boy washed a dog without sighing, the latter condition will be falsified, but (2.67) stays true.

As the restrictor opens up a slot to be filled by a plurality, we can reuse the quantification structure for distributive predication. In Section 2.2.3 we mentioned that (2.68) has a reading where a property of "adopting six dogs" is asserted of each boy. Just as in the case of a quantificational determiner, the restrictor is specified by a common noun, in the case of distributive predication, it can be specified by the plural nominal *five boys*:

(2.68) Five boys (*each*) adopted six dogs.

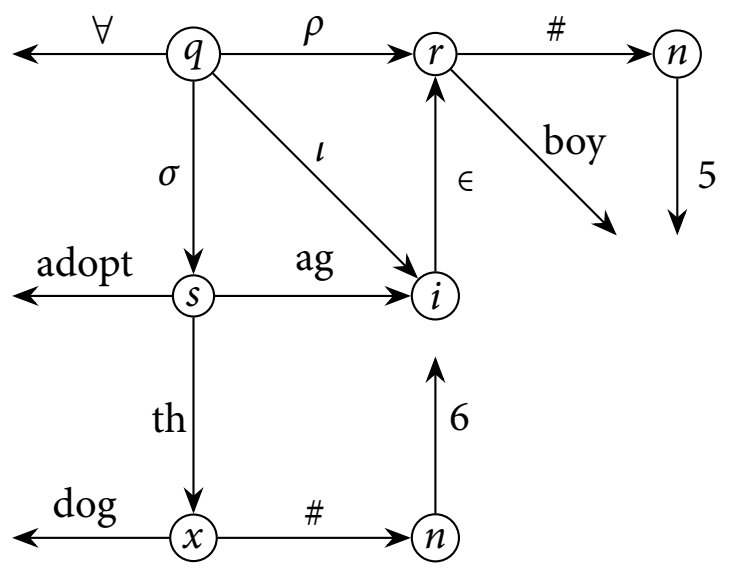

This reading cannot be expressed without a restrictor. We will make more use of our quantification structure when discussing distributivity of various forms in Chapter 4.

Another benefit of an explicit restrictor comes with the distinction of the so-called "existential scope" vs. "distributive scope", a terminology due to Szabolcsi (2010, chap. 7) that describes how an existentially quantified plurality can take wider scope than its distributive quantificational force. We have to defer examples and discussion until Section 5.1.3, but we already have sufficient empirical grounds for adapting the quantification structure of HLDS.

### 2.3.3 Discourse Representation Theory

Discourse Representation Theory (DRT; Kamp, 1981; see also Heim, 1982) is a representation formalism designed to address issues in semantics of anaphora and tense, and more broadly, in how interpretation of linguistic expressions affects and is affected by contexts. Well-established in theoretical and computational linguistics communities, DRT covers a wide range of linguistic phenomena in depth. For a detailed documentation one may refer to Kamp and Reyle (1993); Kamp et al. (2011). In what follows we will focus on the structural properties of this graph-like language and their implications.

The basic units of DRT are *discourse representation structures* (DRSs), which comprise a declaration of variables, that is, the *universe*, and a series of *conditions*. As an example (2.69) shows a DRS in the iconic box notation:

(2.69) A boy walked a dog.

| $x, y, v$ | |
|---|---|
| boy($x$) | walk($v$) |
| dog($y$) | ag($v, x$) |
| | th($v, y$) |

A *simple* DRS begins with a set of variables (possibly empty) followed by *simple conditions*, that is, unary and binary relations holding thereof. An insight of DRT is that existence and logical conjunction can be implicitly indicated. While this example is self-explanatory, its model-theoretical interpretation consists in finding a valuation of the universe to satisfy the conditions.

One can immediately see that a simple DRS describes a directed graph, with the universe being the vertex set and the conditions being the labeled edges, although the graph can be multi-rooted as in (2.70a) (as both $v_1$ and $v_2$ have no in-edges), or even disconnected as in (2.70b) (as $y$ is isolated from $x$ and $v$).

(2.70) a. Joe walked a dog that barked.

| $x, y, v_1, v_2$ | | |
|---|---|---|
| Joe($x$) | walk($v_1$) | bark($v_2$) |
| dog($y$) | ag($v_1, x$) | ag($v_2, y$) |
| | th($v_1, y$) | |

b. A dog barked and there was a cat.

| $x, y, v$ | |
|---|---|
| dog($x$) | bark($v$) |
| cat($y$) | ag($v, x$) |

If $K_1, K_2$ are DRSs, then $\neg K_1$, $K_1 \vee K_2$, $K_1 \Rightarrow K_2$ make *complex conditions* that can be embedded in *complex* DRSs to represent negation, disjunction, and implication or universal quantification. For the sake of illustration, the outer DRSs below contain no variables or simple conditions of their own.

(2.71) a. There was no dog.

$K_0$: $\neg$ [ $x$ | dog($x$) ]$_{K_1}$

b. There was a dog or a cat.

$K_0$: [ $x$ | dog($x$) ]$_{K_1}$ $\vee$ [ $y$ | cat($y$) ]$_{K_2}$

c. Every boy sailed.

$K_0$: [ $x$ | boy($x$) ]$_{K_1}$ $\Rightarrow$ [ $v$ | sail($v$), ag($v, x$) ]$_{K_2}$

To read these, (2.71a) holds whenever there is no $x$ satisfying $K_1$; (2.71b) holds whenever some $x$ satisfies $K_1$ or some $y$ satisfies $K_2$; (2.71c) holds whenever each $x$ satisfying $K_1$ finds some $v$ to satisfy $K_2$ together.

Complex DRSs are not exactly graphs; through nesting they form trees whose vertices are graphs. (A simple DRS is also a tree of graphs, that is, a tree of a single vertex.) For a complex condition in $K_0$, we take its first operand $K_1$ as a child of $K_0$ and its second operand $K_2$, if any, as a child of $K_1$. Meanwhile, a unary operator labels the edge from $K_0$ to $K_1$, and a binary operator labels the edge from $K_1$ to $K_2$. For example, (2.71) can be recast as follows:

(2.72) a.

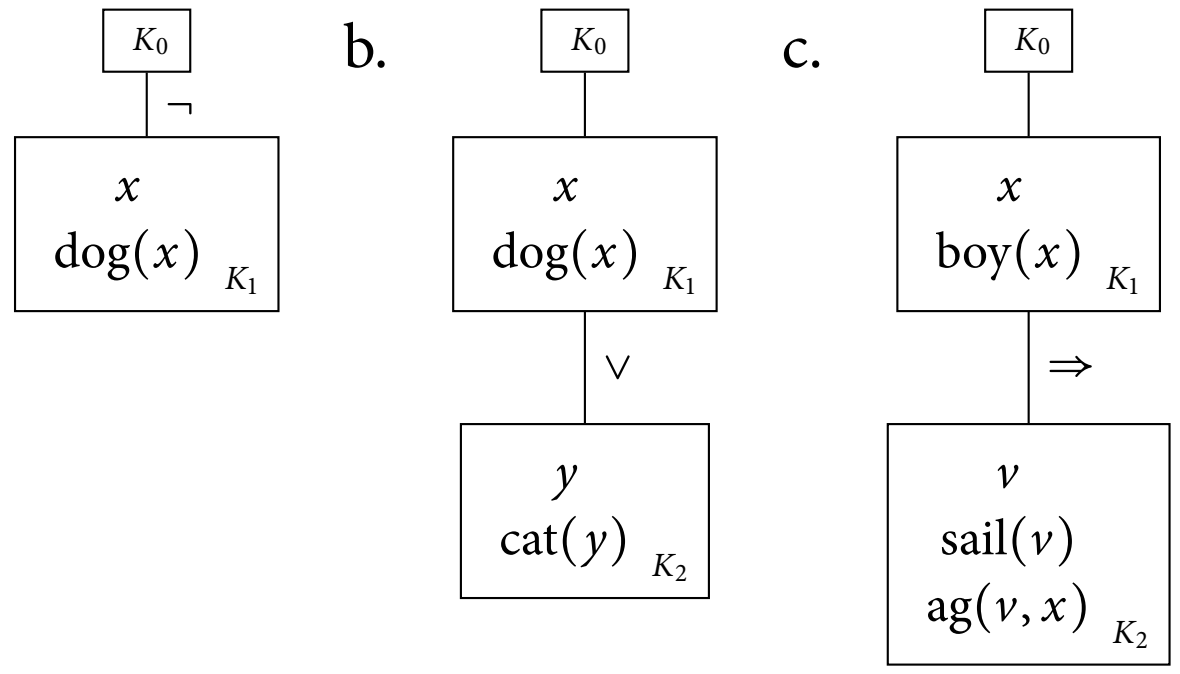

Note that a graph may inherit vertices from its ancestors (e.g. $K_2$ refers to $x$ in (2.72c)), if its valuation depends on theirs in the sense given below.

Viewing the DRT language this way facilitates comparing it with a semantic representation made homogeneously of graphs like ours. A DRS essentially rolls out a decomposition of an underlying graph into subgraphs, structuring their valuation dependency as a tree. When $K_i$ has a descendant $K_j$ in a DRS tree and $K_j$ is not a child of $K_i$ through disjunction, valuation of $K_j$ depends on valuation of $K_i$. This can be illustrated with a previous example (2.52/2.53b):

(2.73) Every boy walked a dog.

Consider the narrow scope reading of the indefinite for the moment. In DRT we may also put a restrictor variable that makes the domain of quantification explicit (maximized by default), as shown in (2.74). We may think of the variables as ranging over sets and interpret the edges (i.e. predicates) the way we did in Section 2.2.3. To verify (2.74b), find a valuation $g$ defined on $r$ that satisfies $K_0$, such that whenever $h \supseteq g$ defined on $r, i$ satisfies $K_1$, there is $k \supseteq h$ defined on $r, i, s, y$ that satisfies $K_2$.

(2.74) a.

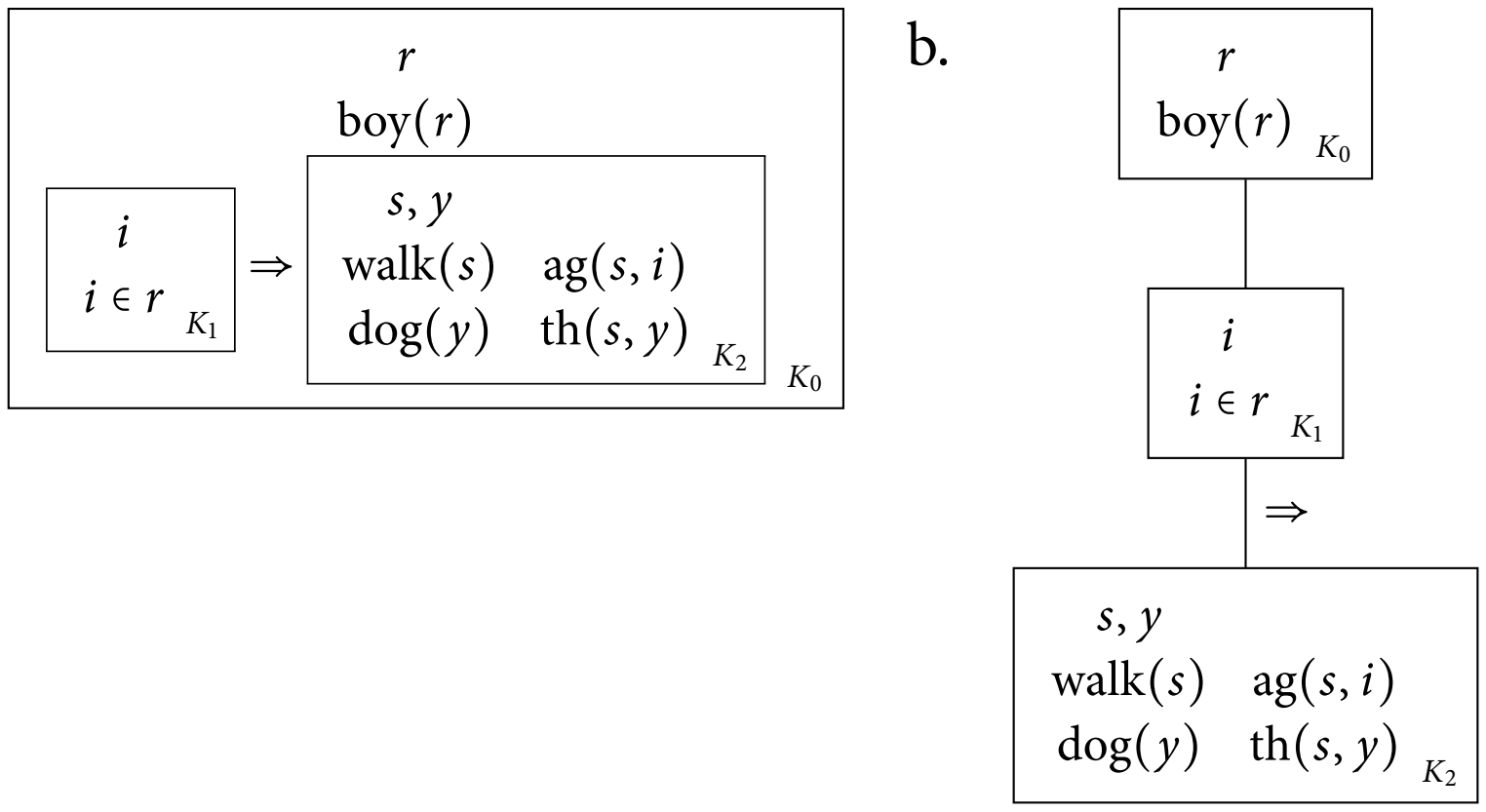

Thus (2.74) is nothing but a tree decomposition of the semgraph (2.75a) less $x$ and its out-edges, with $K_0, K_1, K_2$ being the restrictor subgraph, the iterator subgraph, and the scope subgraph.[3] The tree structure encodes the intention that valuation of $K_2$ depends on that of $K_1$, which in turn depends on that of $K_0$. Because such subgraphs and their valuation dependency arise automatically as semgraphs are traversed and interpreted, a DRS, in a sense, is a transcription of that process.

[3]We use "tree decomposition" loosely, though (2.74b) qualifies for the technical sense (see Courcelle and Engelfriet, 2012, p. 122).

(2.75) a.

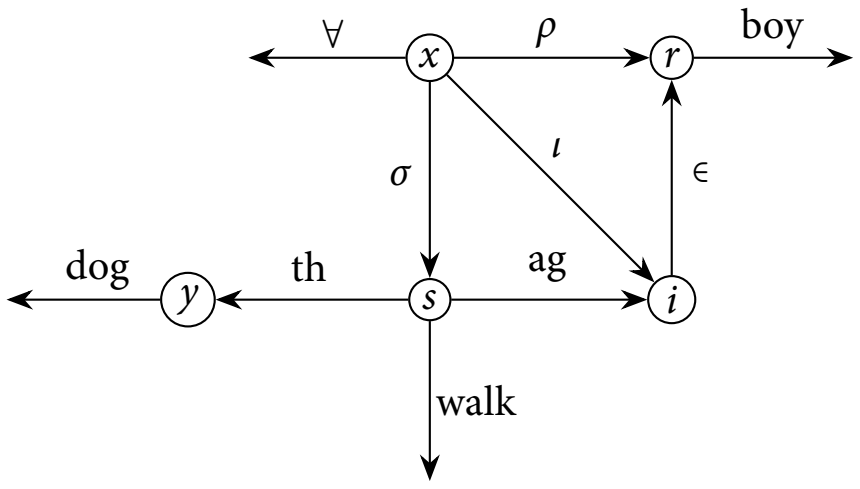

b.

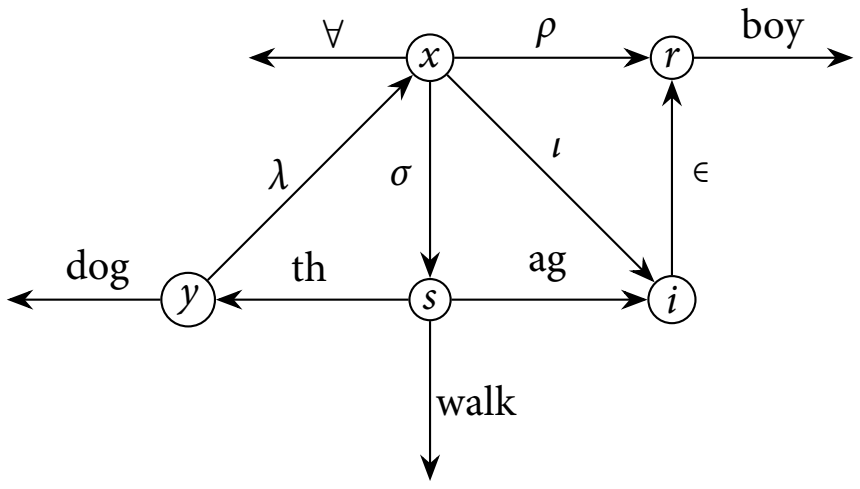

As a consequence, variables' order of valuation cannot be shifted in a DRS the way it can be in its underlying graph. With a $\lambda$-edge (2.75b) gets the wide scoping indefinite of (2.73). But (2.74) imposes a top-down order that requires moving the declaration of $y$ and the condition $dog(y)$ from $K_2$ to $K_0$ to have $y$ valuated before $i$.

(2.76) a.

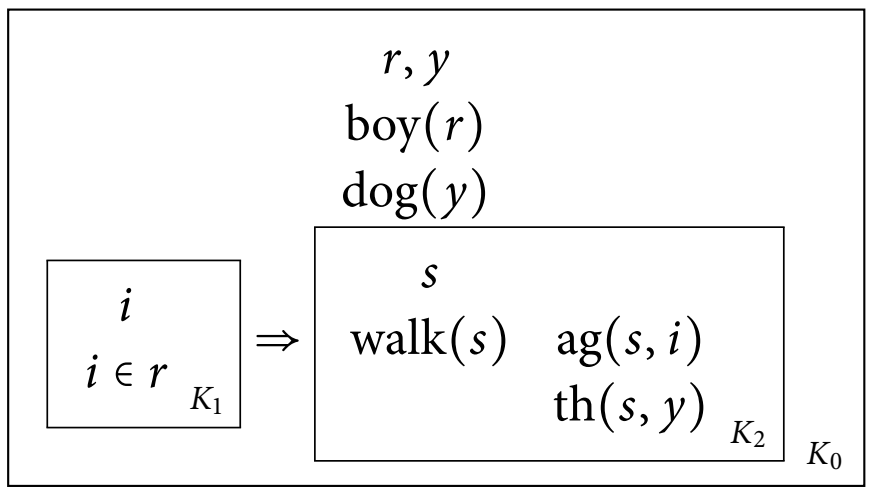

b.

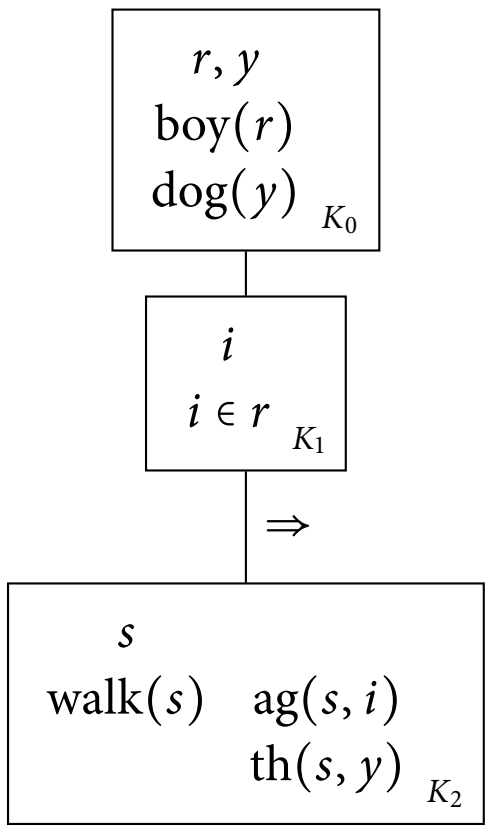

Note that $K_0$ in (2.76b) is no longer a connected subgraph of (2.75b). Whereas (2.75a&b) can be constructed from the same syntax in roughly the same way, as we will discuss later, to construct (2.76) one needs to make adjustments to either the syntax (e.g. via quantifier raising; Muskens, 1996) or the semantic construction mechanism (Kamp and Reyle, 1993, pp. 282ff) assumed for (2.74).

Specifying variables' order of valuation along DRS trees also avoids the need to refer to graph rootedness. The graphs in a DRS tree can, as we have seen, be multi-rooted and disconnected. But rootship is nonetheless a useful property of vertices: in dealing with the proportion problem in Section 2.2.4, the root of an iterator subgraph gives the criterion for classifying valuations. (2.55a-c) show that the three readings of (2.77) (i.e. about state-of-affairs in general, about dog-walking boys, about boy-walked dogs) are a matter of which variable becomes the root of an iterator subgraph by taking "local scope".

(2.77) Mostly if a boy walked a dog, he fed it.

By contrast, to mark criterion variables DRT introduces *duplex conditions*, where a variable-bounding quantifier in a diamond box generalizes the arrow in implicative conditions ($K_1 \Rightarrow K_2$). For example, (2.78) represents the reading of (2.77) about dog-walking boys, by marking $x_1$ ranging over boys as the target of counting.

(2.78)

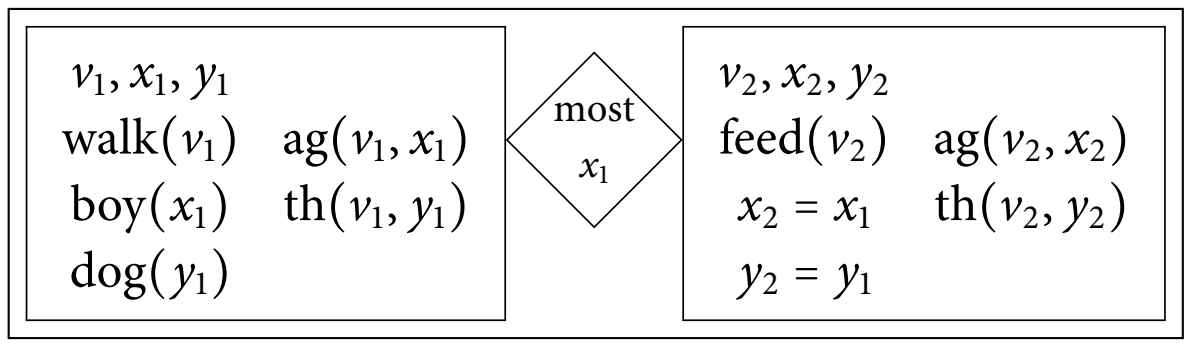

This is actually the sole purpose duplex conditions serve. If we are to distinguish certain vertices in graphs, however, it seems a natural choice to do so with rootship, which is inherent in graph structure, and a rootship shifting device whose use can be justified elsewhere.

Finally, discussion of (2.74) shows how growth of valuation domains is explicitly managed in DRS interpretation. More precisely, in a DRS tree the root $K_0$ is to be satisfied by a valuation whose domain equals the universe of $K_0$; with $K_i$ being the first ancestor of $K_j$ that is not a parent of $K_j$ through disjunction, the valuation of $K_j$ extends the domain of the valuation of $K_i$ by the universe of $K_j$. Since an edge of any graph $K_j$ can only have ends on which the valuation of $K_j$ is defined, it follows that an edge of $K_j$ always finds its ends in the universe of $K_j$ or the universe of any $K_i$ that $K_j$ depends on in valuation. For (2.74), that means $K_2$ may refer to $r, i$ but neither $K_0$ nor $K_1$ may refer to $s, y$. Valuation dependency thus defines a relation that characterizes when one graph may reference the variables declared in another. This relation is known as *DRS accessibility* and is used to account for co-reference constraints. For example, as $K_0$ in (2.79) has no access to $x$ declared in $K_1$, *he* cannot be resolved to co-vary with *every boy*.

(2.79) Every boy sailed and he returned.

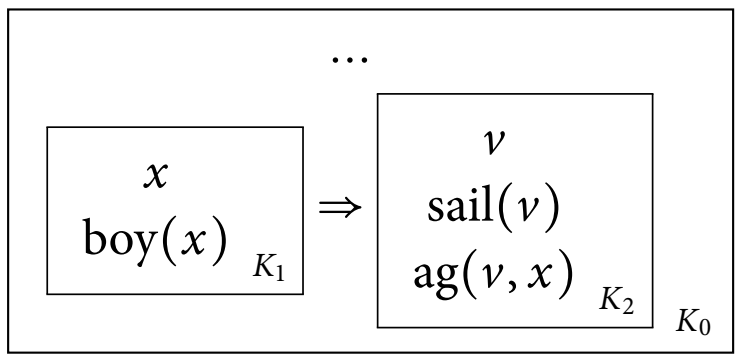

However, the motivation for managing growth of valuation domains, which induces DRS accessibility, is to ensure a correct semantics of quantification. Take (2.74) for example. A valuation $g$ satisfying $K_0$ can only be defined on $r$; were $g$ be allowed to valuate $i$, in a situation where among a few boys only Joe walked a dog, $g$ could satisfy $K_0$ by assigning Joe to $i$. Similarly, any valuation $h \supseteq g$ enumerated to satisfy $K_1$ can only be defined on $r, i$; were $h$ be allowed to valuate $y$, in a situation where every boy did walk a dog, the extendability of $h$ could be denied for not assigning $y$ a dog walked by the boy $i$.

It is these considerations that have led our semgraph interpreter (see Definitions 2.5/2.6) to similar requirements on the domains of valuations. The difference is that where we specify what should be excluded from them, DRT specifies what should be admitted to them. The two approaches then differ in how they deal with interaction between co-reference resolution and scope taking.

A relevant example concerns the scope cap of indefinites (see Brasoveanu and Farkas, 2011; Schwarz, 2001). Below even as *he* is resolved to *boy* (the two co-vary), *a dog* cannot take wide scope over *every boy*, that is, (2.80) does not have a reading that requires there be a dog that every boy walked and fed.

(2.80) Every boy walked a dog (that) he fed.

Now the task for DRT lies in semantic construction or syntax-semantics interface, namely, to explain why a DRS cannot be constructed to the effect of moving the declaration of $y$ and the condition $dog(y)$ from $K_2$ to $K_0$ in (2.81), while (2.76) is fine.

(2.81)

$$\boxed{\boxed{\begin{array}{c} x \\ \text{boy}(x) \end{array}}_{K_1} \Rightarrow \boxed{\begin{array}{lll} v_1, v_2, y, z & & \\ \text{dog}(y) & \text{walk}(v_1) & \text{feed}(v_2) \\ z = x & \text{ag}(v_1, x) & \text{ag}(v_2, z) \\ & \text{th}(v_1, y) & \text{th}(v_2, y) \end{array}}_{K_2}}_{K_0}$$

By contrast, as we will see later, there is no difficulty for our semantic construction mechanism to scope $x$ with a $\lambda$-edge as follows:

(2.82)

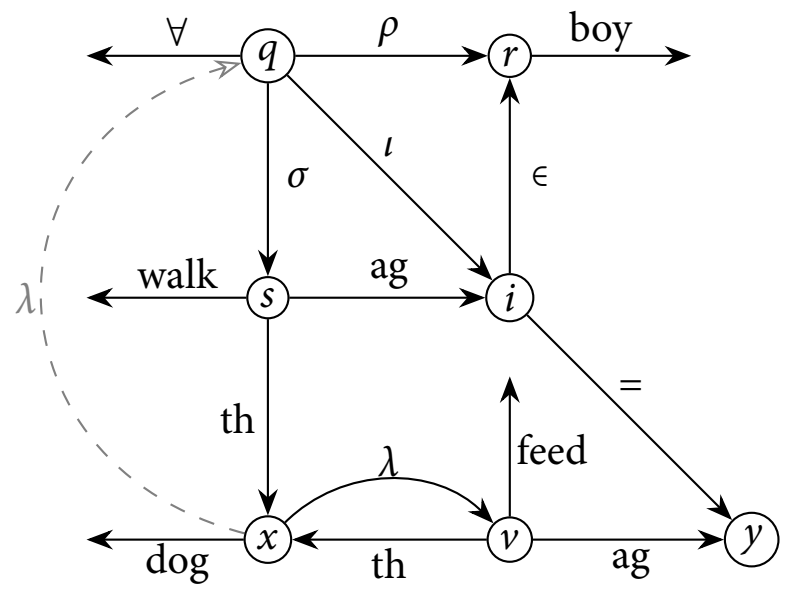

The task for us is rather to explain why the dashed addition is problematic for the semgraph interpreter developed here. We will revisit this example in Section 5.2.4.

# Chapter 3

# Syntax-Semantics Interface

In this chapter, we provide a unification-based semantic construction mechanism that builds semgraphs from smaller ones at the syntax-semantics interface. Graph unification, implemented as parallel composition, amounts to gluing disjoint graphs by fusing equivalent vertices, or rather, equivalent discourse referents. The task of the syntax-semantics interface is therefore to decide equivalence among referents introduced by linguistic tokens, through syntax and non-syntactic resolutions. While there are two sources of information, the outputs we expect from them are nonetheless of the same format — pairs of equivalent vertices. This allows for a simple presentation of syntax and an easy combination of the information needed for semantic construction.

### 3.1 Bigger picture

To clarify our expectation for the syntax-semantics interface, consider a concrete example:

(3.1) Joe saw himself.

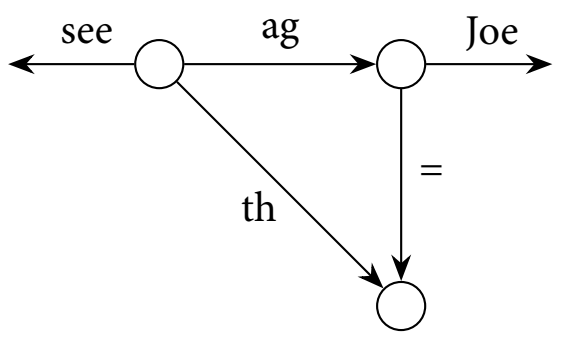

Here we have three tokens whose semantics is given in Figure 3.1. Recall from Section 1.6.3 that semgraph vertices that are assigned an integer label are called sources. In the context of semantic construction, we will see, it is convenient to refer to vertices by their source labels instead of their names, which are usually omitted.

To compose the semgraph in (3.1), we need to find the fusing equivalence given by coloring in Figure 3.1. Once the latter is in place, the rest is simply to so rename sources before parallel composition, that vertices shall bear the same label if and only if they are equivalent (of the same color). The question is how that equivalence arises.

We take both syntax and non-syntactic resolutions as responsible for the desired equivalence. We expect syntax to produce (3.2a), that is, the 1-source of *see* equals the 0-source of *himself*, and the 2-source of *see* equals the 0-source of *Joe*. We expect co-reference resolution to decide, as in (3.2b), that the –1-source of *himself* equals the 0-source of *Joe*.

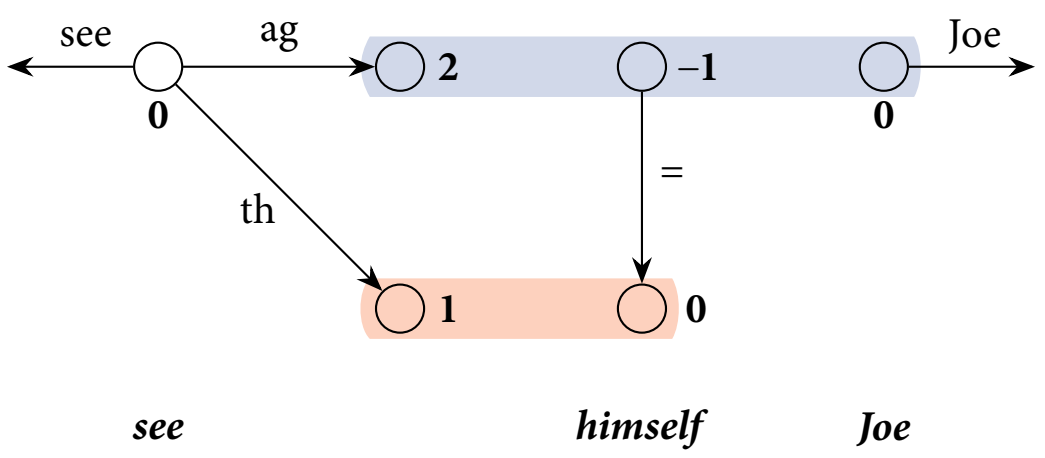

**Figure 3.1:** Fusing equivalence for *Joe saw himself.*

(3.2) a. $(\mathbf{1}_{see}, \mathbf{0}_{himself}), (\mathbf{2}_{see}, \mathbf{0}_{Joe})$
b. $(-\mathbf{1}_{himself}, \mathbf{0}_{Joe})$

Together (3.2a&b) yield the equivalence in Figure 3.1. Strictly speaking, these pairs are not yet an equivalence relation, which should rather be *reflexive* ($x \equiv x$ for all $x$), *transitive* ($x \equiv y$ and $y \equiv z$ imply $x \equiv z$), and *symmetric* ($x \equiv y$ implies $y \equiv x$). But they uniquely *generate* one — the smallest of all equivalence relations including them. The partitions corresponding to this equivalence can be computed thus: processing one vertex pair at a time, we either unite it and all existing partitions it overlaps with, or make it a new partition if it overlaps with none (we omit the proof of correctness and refer readers to Cormen et al. (2009, chap. 21) for efficient implementations).

This is how syntax and non-syntactic resolutions work in tandem. Separating the two sources of referent equivalence keeps syntax simple and clean, and in principle, allows their respective computations to proceed in parallel. In the following sections, we will describe how syntax as a deterministic process comes to its outputs, and where non-syntactic resolutions as non-deterministic processes can be useful. We start with the latter.

## 3.2 Non-syntactic resolutions

By non-syntactic resolutions we mean any non-deterministic computations, though not necessarily unconstrained, that feed semantic construction. For the purpose of this thesis, we will discuss co-reference resolution, an established process for identifying the antecedent of an anaphor, and precedence resolution, a novel process we propose for scoping indefinites.

### 3.2.1 Co-reference resolution

Anaphor tokens (including pronouns and reflexives) *co-refer* with other linguistic tokens when the referents they introduce coincide. This relation is often indicated with co-indexation. Take the previous example:

(3.3) **Joe**$^{i}$ saw **himself**$_{i}$.

In the linguistic literature, the anaphoric relation of the following kind is often called "variable-binding" (in that *himself* serves as a variable bound by the quantifier *every boy*) instead of co-referential.

(3.4) **Every boy**$^i$ saw **himself**$_i$.

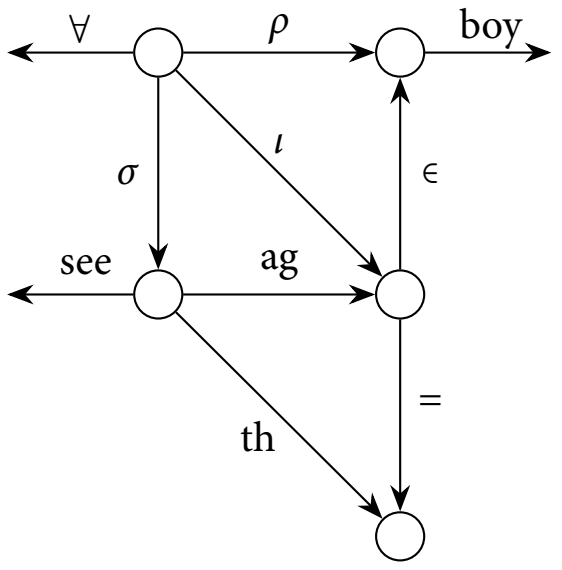

We nonetheless consider this an instance of co-reference, since its semgraph features the same equality edge as in (3.1). The only difference is that the co-reference in (3.1)/(3.3) is in a sense static, whereas that in (3.4) is fluid, living in co-variation.

*Co-reference resolution* is then the task of identifying the antecedent of an anaphor. As a convention, we use negative sources to indicate equivalence to be resolved non-syntactically. With the following entries for anaphors and possessive pronouns (note that the latter's entry contains an owning event),

(3.5) a. it/itself b. its

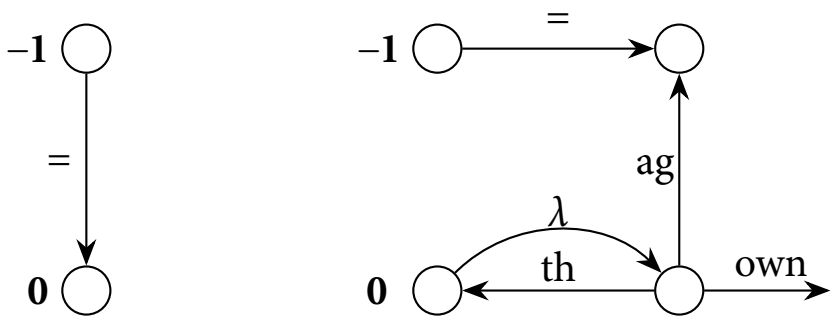

we may more precisely say that co-reference resolution finds in another linguistic token the equivalent of a negative source that tails an equality edge. Here is an example apart from (3.3). Given the semgraph

(3.6) every boy

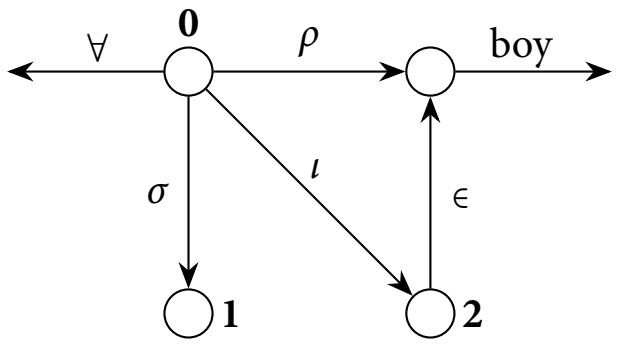

co-reference resolution in (3.4) yields $(-\mathbf{1}_{himself}, \mathbf{2}_{every\ boy})$, that is, the −1-source of *himself* equals the iterator of the quantification structure.

Our approach becomes quite handy as we deal with split and donkey anaphora later in this chapter. The reason we put co-reference resolution outside syntax is that it is non-deterministic, as *his* can mean Joe's or Ben's or someone else's in (3.7a), and may work across sentences, as *he* does in (3.7b).

(3.7) a. **Joe**$^i$ saw **Ben**$^j$ walk **his**$_{i/j}$ dog.
b. **Joe**$^i$ sailed. Later **he**$_i$ returned.

We are agnostic about how co-reference resolution is carried out (see Jurafsky and Martin, 2020, chap. 21 for a computational introduction), but we know what kind of outputs to expect. At any rate, resolved co-reference is usually taken alongside syntax as input to semantic interpretation (e.g. Heim and Kratzer, 1998) or construction (e.g. Groenendijk and Stokhof, 1991; Jacobson, 1999; Jäger, 2005; Muskens, 1996). This is also the approach taken here; see Beaver (2002); Muskens (2011) for a different view.

*Remark.* Co-reference does not account for Karttunen's (1969) "paycheck pronouns", e.g. *Joe deposited his paycheck, but Ben spent it*, where *it* refers to Ben's paycheck. Comparing (3.8a) and (3.8b), we obtain desired semantics if *it* makes a copy of the shaded subgraph in (3.8a) and anchors that to the thematic dependents of *spent*.

(3.8) a. Joe deposited his paycheck. b. Ben spent it.

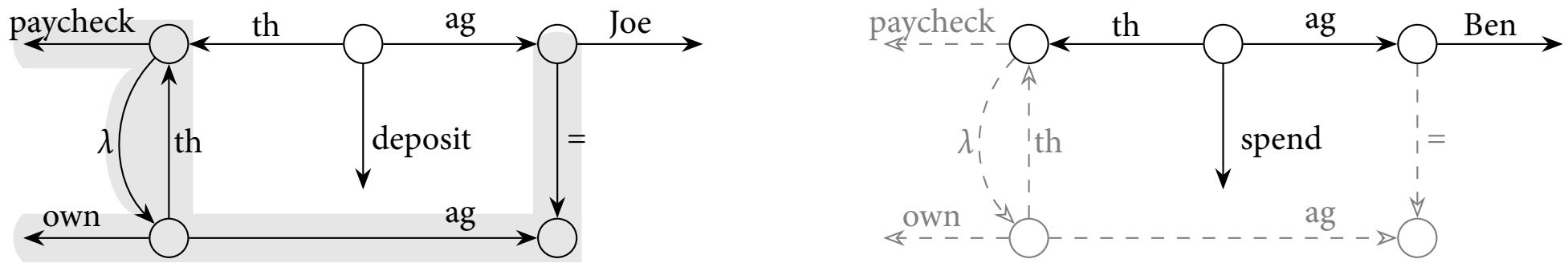

Thus, it is natural to say that the semantics of paycheck pronouns consists in (structural) antecedent resolution, which relies on the discourse to complete ongoing semantic construction. In this regard our formulation does not differ much from that of Jacobson (1999). Antecedent resolution arguably also happens in verbal ellipsis (e.g. *Joe spent his paycheck, and Ben did too*) and in anaphors bearing a discourse dependency:

(3.9) a. Every boy walked a dog. b. Most boys fed it (= a dog he walked).

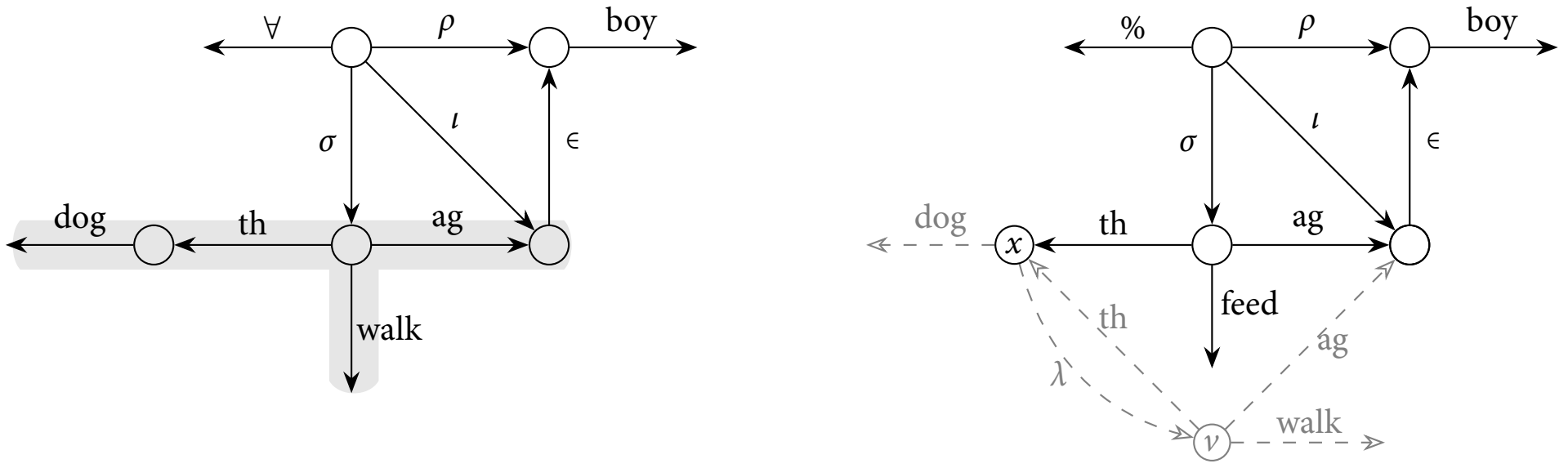

Again we may copy the shaded subgraph in (3.9a) into (3.9b), though $\lambda\overrightarrow{xv}$ has to be somehow added to ensure the correct semantics.

### 3.2.2 Constraints on co-reference resolution

We now turn to constraints on co-reference resolution, some of which are preferable and others inviolable. Examples of the first kind include recency (a more recently mentioned antecedent is preferred), grammatical function parallelism (e.g. a subject antecedent is preferred for a subject anaphor), and reasoning based on lexical semantics and common sense. The complexity and diversity of such factors again justify the non-determinism of co-reference resolution.

Examples of the second kind, on the other hand, include agreement morphology, accessibility in dynamic semantic theories (e.g. Groenendijk and Stokhof, 1991; Kamp and Reyle, 1993), and locality in binding theory (Chomsky, 1981). Of these our graph formalism is able to say something interesting about the latter two. Accessibility was mentioned towards the end of Section 2.3.3 and will be explored more in Section 5.2.4. Binding theory constraints, originally stated with reference to syntax, can be recovered with reference to semgraph structure in a nontrivial way.

A classic observation is that reflexives are resolved in a "local domain" that pronoun resolution avoids. The locality in question may refer to clausal boundaries as in (3.10a&b), or something more restrictive as in (3.10c&d) (see Lasnik and Uriagereka, 1988 for an exact characterization).

(3.10) a. **Joe**$^i$ saw ***him**$_i$/**himself**$_i$.
b. **Joe**$^i$ thought Ben saw **him**$_i$/***himself**$_i$.
c. **Joe**$^i$ saw **his**$_i$/***himself**$_i$'s dog.
d. A dog of **Joe**$^i$ saw **him**$_i$/***himself**$_i$.

The valid co-references in (3.10a-d) are represented in that order as follows:

(3.11) a. $v \to y \to x \parallel v \to x$

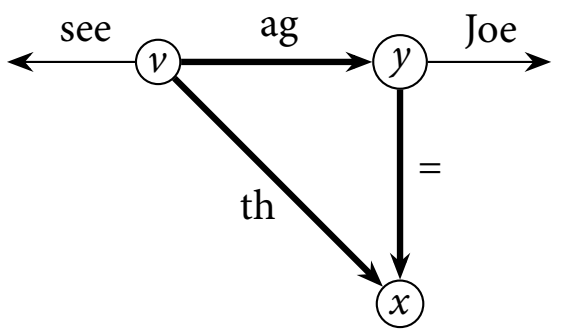

b. $v_1 \to y \to x \parallel v_1 \to v_2 \to x$

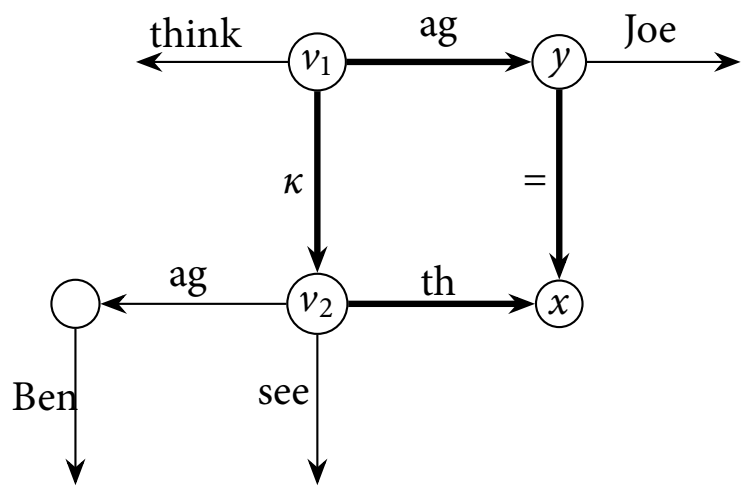

c. $v_1 \to y \to x \parallel v_1 \to z \to v_2 \to x$

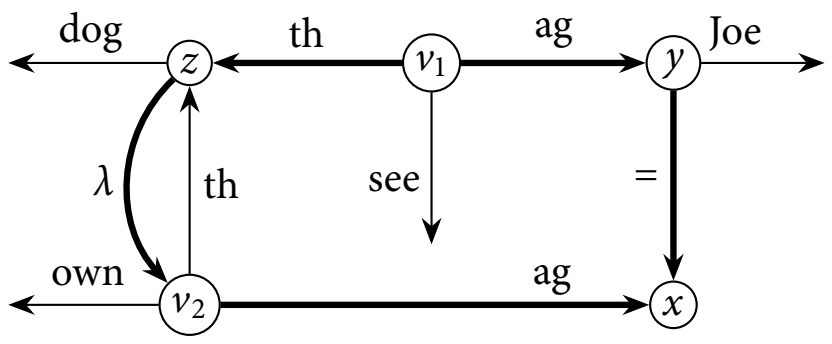

d. $v_1 \to z \to v_2 \to y \to x \parallel v_1 \to x$

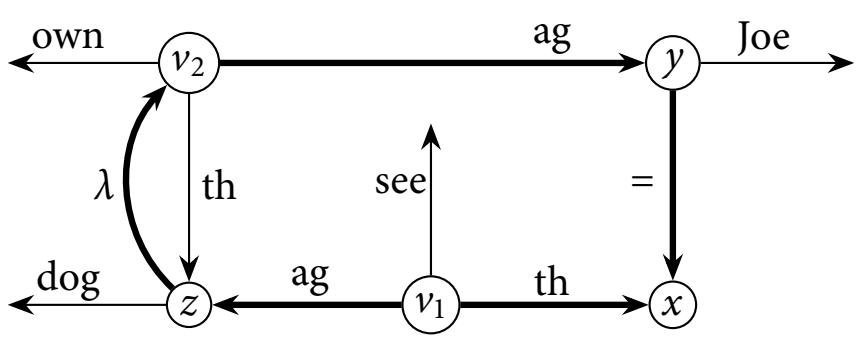

Crucially, co-reference creates such a pair of parallel paths: exactly one of them ends in equality ($= \overrightarrow{yx}$ above) introduced by anaphors, and they have no subpaths forming a pair of parallel paths with the latter property. Let $P_e$ be the path ending in equality, $P_f$ the one without, $u$ the starting point, and $v$ the ending point. We notice that

i) (*locality*) $P_e \parallel P_f$ created by reflexives contains at most one event distinct from $v$;
ii) (*antilocality*) $P_e \parallel P_f$ created by pronouns contains at least two events distinct from $v$;
iii) (*obliqueness*) the first edge of $P_e$ is less oblique (see below) than the first edge of $P_f$, provided $u$ is not an individual.

Statements (i) and (ii) express the requirements for local and nonlocal antecedents by reflexives and pronouns, with locality measured by the lengths of the parallel paths. Put differently, when retracing from the head of the equality edge $v$ via different paths, it takes shorter distance for them to reunite in the case of reflexives than it does in the case of pronouns.

Statement (iii) applies to reflexives and pronouns alike. Relative obliqueness of semantic dependents typically reflects the order in which a functor receives its syntactic arguments (e.g. the agent combines *later* with the verb than the theme and is thus *less oblique* than the latter; see Section 3.3.2), as discussed in Bach and Partee (1980); Chierchia (1988); Dowty (1992); Hepple (1990); Pollard and Sag (1987). The obliqueness statement plays a role similar to that of the c-command constraint in binding theory, but not exactly the same. For example, the parallel paths in (3.11d) satisfy (iii), but *Joe* does not c-command *him* in (3.10d).

A main effect of statement (iii) is to avoid cataphora in examples like these:

(3.12) a. ***Himself**$_i$ saw **Joe**$^i$.
b. ***He**$_i$ thought Ben saw **Joe**$^i$.

(3.11a&b) can be easily modified to illustrate the problem. The pattern of cataphora avoidance also generalizes to the well-known *crossover* phenomenon (Postal, 1971; see Safir, 2004, chap. 3 for a review), if the gap in a relative clause can be seen as an antecedent besides the actual one, i.e. the noun phrase modified by that relative clause:

(3.13) a. **A boy**$^i$ who __$^i$ thought **he**$_i$ saw Ben.
b. ***A boy**$^i$ who **he**$_i$ thought __$^i$ saw Ben.

The semgraphs of (3.13a&b) are given by (3.14a&b), respectively:

(3.14) a.

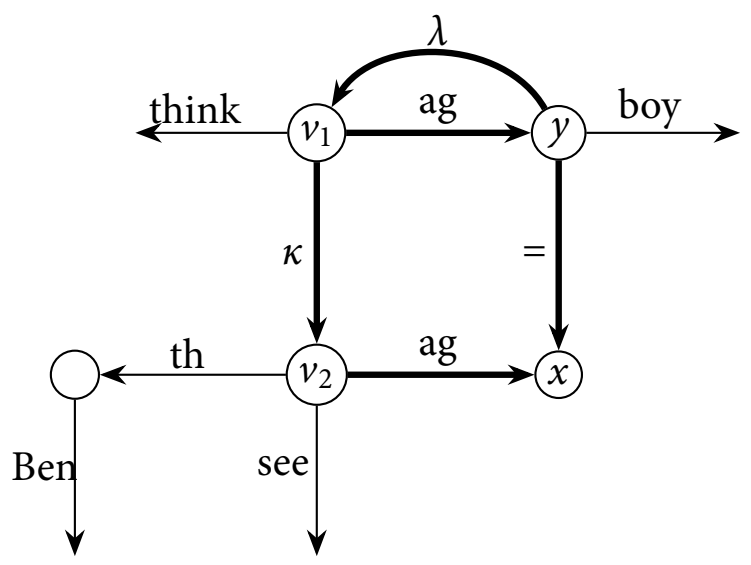

b.

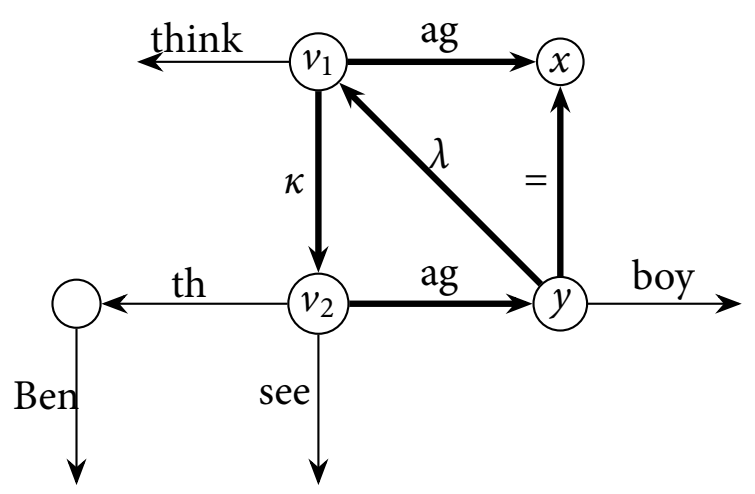

While co-reference creates multiple pairs of parallel paths in (3.14a), we can see that for any of them, the equality-free path ($P_f$) satisfies antilocality and the equality-ended path ($P_e$) ensures relative obliqueness (e.g. one such pair also appears in (3.11b); $y \to x \parallel y \to v_1 \to v_2 \to x$ gives another pair, as $y$ is not an event). In (3.14b), however, we have $y \to x \parallel y \to v_1 \to x$ against antilocality and $v_1 \to v_2 \to y \to x \parallel v_1 \to x$ against the obliqueness statement.

A special class of crossover examples, known as *weak* crossover, differs from *strong* crossover by having an embedded cataphor that does not c-command the co-referring gap. Compare (3.15b) with (3.13b):

(3.15) a. **A boy**$^i$ who __$^i$ saw **his**$_i$ dog.

b. *A **boy**$^i$ who **his**$_i$ dog saw __$^i$.

Inspecting the semgraphs of (3.15a&b), again, we find in (3.16a) any pair of parallel paths of interest satisfying antilocality and relative obliqueness. But unlike (3.14b), (3.16b) only has parallel paths that violate relative obliqueness (e.g. $v_1 \to y \to x \parallel v_1 \to z \to v_2 \to x$). The lack of antilocality violation in weak crossover examples might be responsible for their sometimes reported lesser ungrammaticality (Lasnik and Stowell, 1991).

(3.16) a.

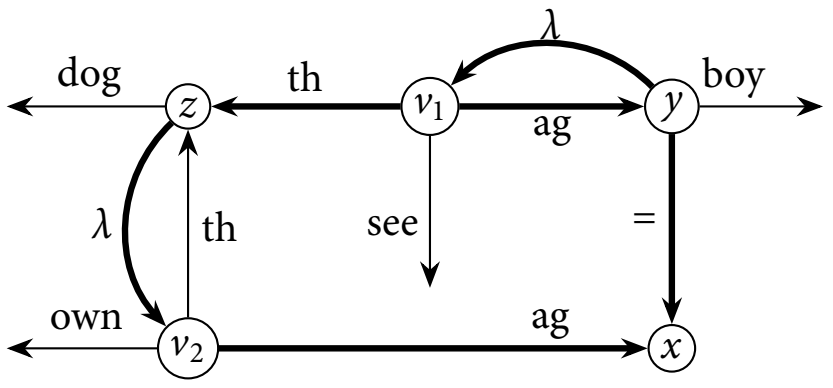

b.

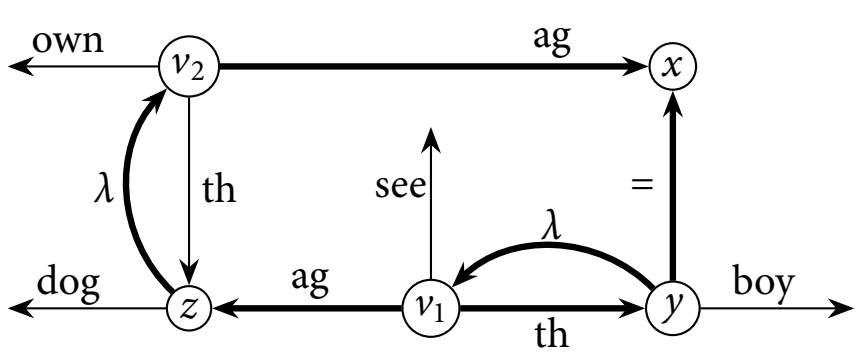

We should add that discussion of (weak) crossover includes examples where the surface syntax bears no gaps:

(3.17) a. **Every boy**$^i$ saw **his**$_i$ dog.
b. ***His**$_i$ dog saw **every boy**$^i$.

These yield the following semgraphs, to which the comment on (3.16) also applies.

(3.18) a.

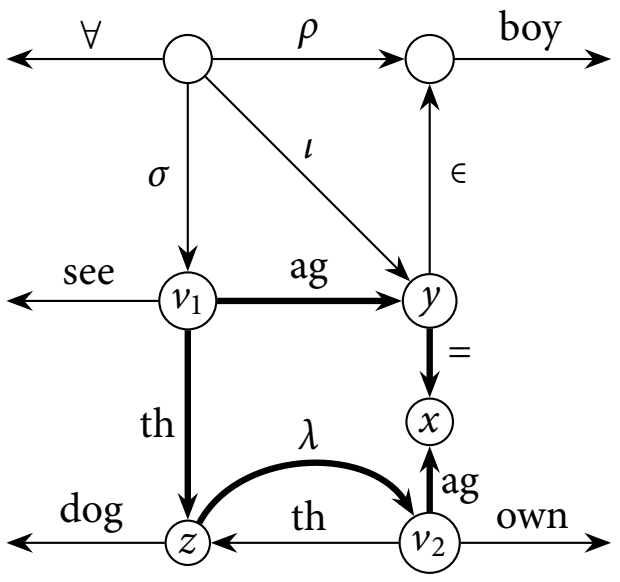

b.

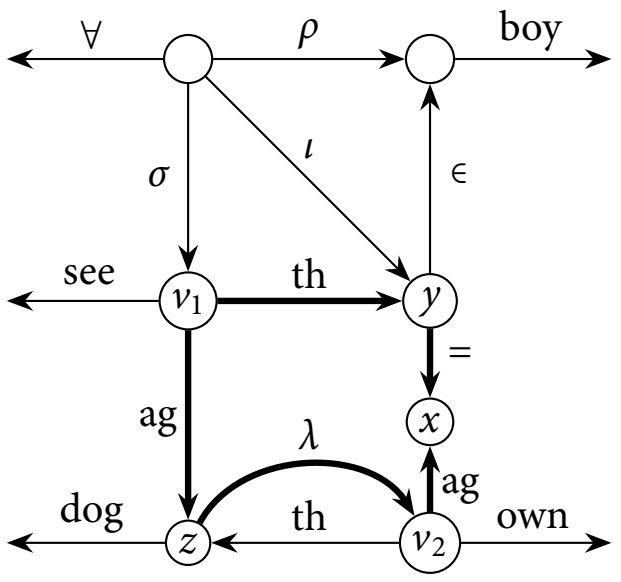

The discussion above has shown how co-reference resolution may be sensitive to the properties of the semantic representation it implies. From the perspective of perception and semantic parsing, at the very least, we may think of statements (i)-(iii) as lexical requirements that limit the number of resolution candidates to be checked against other constraints (e.g. common sense), whose evaluation cost could be expensive.

### 3.2.3 Precedence resolution

With the discussion of the order of valuation in mind (Section 2.2.6), we say that a linguistic token *precedes* another if some referents of the former are valuated before some of the latter. The task of determining certain linguistic tokens' precedence over others is called *precedence resolution*. We are again agnostic about how this process is carried out, but will use its outputs to scope indefinites.

Consider the following example. The narrow- and wide-scope readings of *a dog*, shown by (3.19a&b), arise respectively from having $x$ valuated after and before each boy $i$ (ignore $\lambda\overrightarrow{xy}$ in (3.19a) for a moment).

(3.19) Every boy saw a dog.

a. ∀ boy > ∃ dog

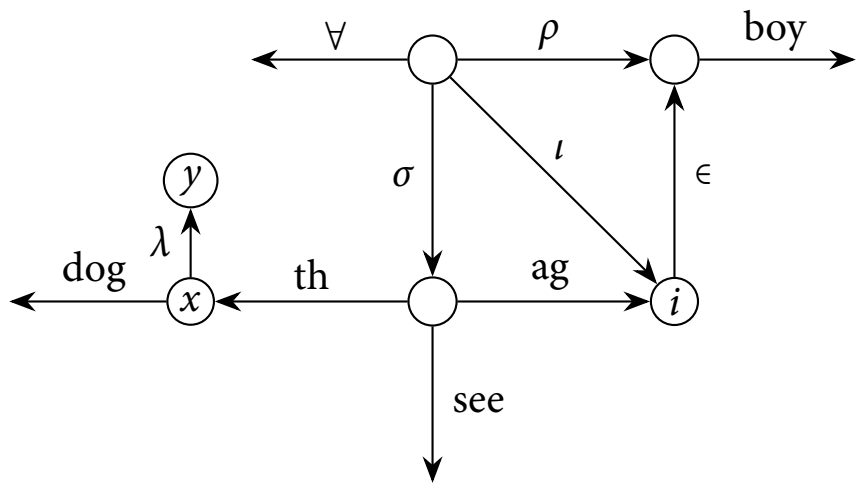

b. ∃ dog > ∀ boy

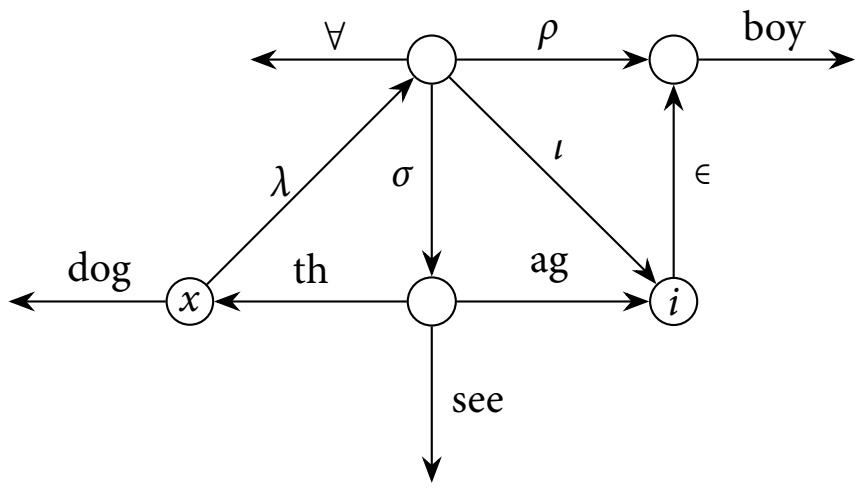

Omitting the cardinality, we introduce the entry for the indefinite article *a* as a $\lambda$-edge:

(3.20) a

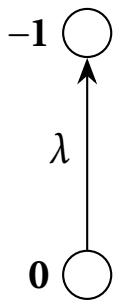

Precedence resolution then determines an indefinite's precedence by finding the equivalent of the negative source in its article. With the semgraph (3.6) for *every boy*, precedence resolution finds no equivalent for the −1- source of *a* in (3.19a), but yields $(-\mathbf{1}_a, \mathbf{0}_{every\ boy})$ in (3.19b), that is, the −1- source of *a* equals the dummy of the quantification structure.

The "unresolved precedence" in (3.19a) leaves a $\lambda$-edge, like $\lambda\overrightarrow{xy}$ in (3.19a), out of any cycle. Such an edge can be dropped without affecting anything, for it is semantically vacuous and does not shift the order of valuation. Thus the indefinite article can be treated as lexically ambiguous between (3.20) and (3.21),

(3.21) a

and we can have (3.22) in place of (3.19a).

(3.22)

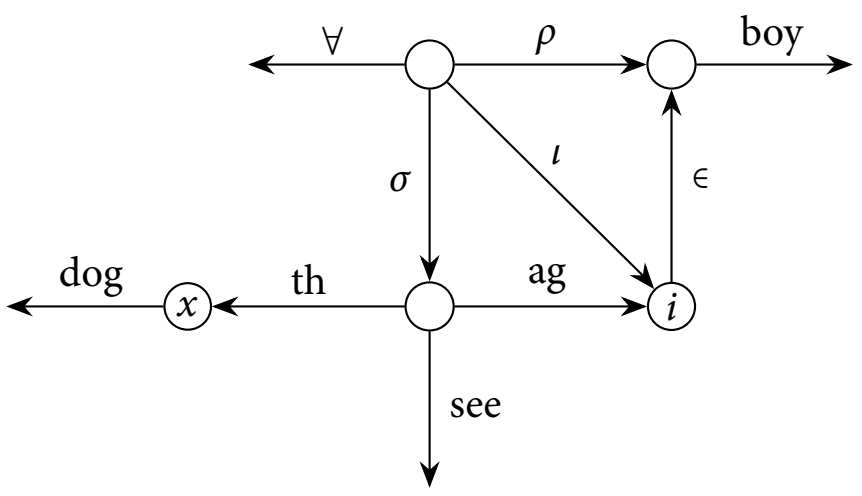

This simpler representation is exactly what we have been using for non-scoping variables. The lexical ambiguity around an optional $\lambda$-edge pending resolution extends to proper names:

(3.23) Joe

a.

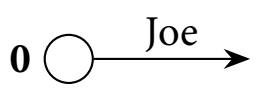

b.

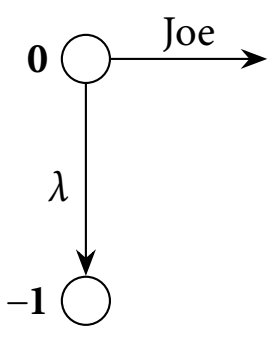

This is because they may also take scope via a $\lambda$-edge, as discussed in Section 2.2.6.

We treat precedence resolution as a non-syntactic process because of its non-determinism, well exemplified by the almost unrestricted scoping ability of indefinites (see Section 5.2 for discussion and citations). For example, even as embedded in a relative clause, usually a scope island, *a toy* may take scope in all three following ways (see Farkas, 1981):

(3.24) Every boy fed every dog which caught a toy.

a. $\forall$ *boy* > $\forall$ *dog* > $\exists$ *toy*
b. $\forall$ *boy* > $\exists$ *toy* > $\forall$ *dog*
c. $\exists$ *toy* > $\forall$ *boy* > $\forall$ *dog*

The same consideration leads Jäger (2005, chap. 6) to attribute essentially equal semantics to anaphors and indefinites. Comparing (3.5a) with (3.20), we can say that for us, the commonality between anaphors and indefinites lies not in semantics, but in the way they participate in semantic construction.

That said, precedence resolution is constrained in how it may interact with co-reference resolution, a topic we mentioned in Section 2.3.3 and will revisit in Section 5.2.4. There might also be (language-specific) lexical requirements on where the negative head of a $\lambda$-edge may find its equivalent, possibly of a similar nature to the locality and antilocality requirements of anaphors discussed in Section 3.2.2. In a Tibeto-Burman language Tiwa, Dawson (2020, pp. 148ff) reported, indefinites marked by a particular particle must take widest scope to within the immediate finite clause that contains them.

*Remark.* We will see later that the semantics of relativizers like *that*, *which*, and *who* also consists of a $\lambda$-edge. That they and the indefinite article share the same semantics is motivated by the fact that cross-linguistically, it is quite common for *wh*-interrogatives to morphologically relate to indefinites. The two differ in how they participate in semantic construction: it is through syntax that both ends of a relativizer's $\lambda$-edge find their equivalents.

## 3.3 Syntactic type reduction

As seen from Section 3.1, syntax is essentially a function that sends sequences of tokens to valid equivalence relations, i.e. partitions on the referents thereof introduced.

We treat this function as a rule-based procedure because at least how arguments bind into thematic roles is pretty deterministic. Consider a familiar example.

(3.25) A boy walked a dog.

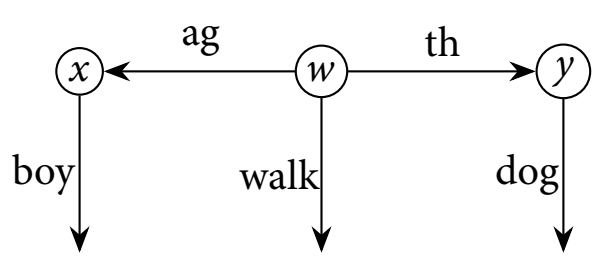

The desired fusing equivalence is shown in Figure 3.2 as coloring: the 0-source of *boy* must equal the agent of *walk*, and the 0-source of *dog* its theme. English syntax rejects swapping the thematic roles of *a boy* and *a dog*.

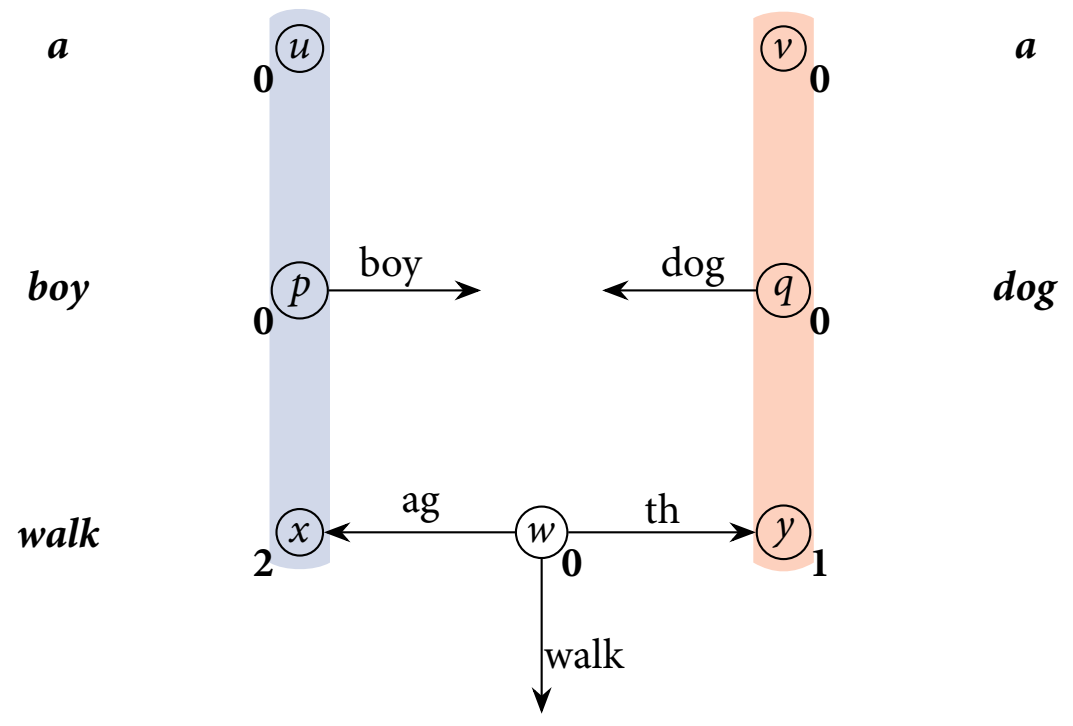

**Figure 3.2:** Fusing equivalence for *a boy walked a dog*.

The function of syntax as we think of it finds a natural implementation in categorial grammars (CGs; Ajdukiewicz, 1935; Bar-Hillel, 1953). CGs manipulate a kind of syntactic resources known as syntactic types or categories, and syntactic derivations in CGs, known as type reductions, boil down to matching and canceling identical atomic types or atoms. If we can establish a proper correspondence between syntactic resources and semantic resources, that is, between atoms and vertices, then atom matches computed by syntax translate immediately to desired pairs of equivalent vertices.

The idea of coupling CGs and unification-based semantic construction goes back to Calder et al. (1988); Zeevat (1988), and finds use particularly in HLDS in Baldridge and Kruijff (2002); White (2006). The present work sets itself apart in two ways:

i) showing that a semgraph relates to its syntactic type in a non-arbitrary way, which allows us to partly infer one from the other;
ii) formulating syntax as a partition function, which allows us to abstract away from derivation details and to some extent, stay neutral to syntactic theories.

We can add that having uniformed outputs of syntax and non-syntactic resolutions enables their easy combination. In semantic construction, one may wonder where lies the fine line between the two. On the one hand, one can push all decision of equivalence to co-reference resolution and like mechanisms (see Bittner, 2001), if they plausibly explain how word salad can be made sense of at all, but at the cost of losing much observable determinism; on the other, syntax can be made to absorb the work of co-reference resolution (e.g. Jacobson, 1999; Jäger, 2005) and of precedence resolution (e.g. Charlow, 2019), but at the cost of increasing the intricacy of a rule-based system. It is not our purpose to evaluate the trade-offs, so we proceed to describe how syntax plays its assigned part.

### 3.3.1 Syntactic type

The counterparts of parts of speech and phrase structure labels in CGs are *syntactic types*, built of arbitrarily chosen *atomic types*, or simply *atoms*, and arbitrarily chosen *connectives*. We define the set

$\mathsf{T}$ of types as follows:

$$\mathsf{T} \quad ::= \quad \mathsf{A} \quad | \quad \mathsf{T}\backslash\mathsf{T} \quad | \quad \mathsf{T}/\mathsf{T} \quad | \quad \mathsf{T}{\downarrow}\mathsf{T} \quad | \quad \mathsf{T}{\uparrow}\mathsf{T}$$

That is, $\mathsf{T}$ is the least set that includes a set $\mathsf{A}$ of atoms and is closed under the binary connectives $\backslash, /, \downarrow$, and $\uparrow$. As a notional convention, the letters $p, q, \ldots$ range over atoms, the capitals $A, B, \ldots$ range over types, and all connectives associate to the left (e.g. $A/B/C$ means $(A/B)/C$). In this thesis we use three atoms, $s$ (sentence), $np$ (noun phrase), and $n$ (common noun). Non-atomic types are also called *fractions*, and therefore $A$ is said to be the *denominator* and $B$ the *numerator* in $A\backslash B$, $B/A$, $A{\downarrow}B$, and $B{\uparrow}A$.

We assign syntactic types to linguistic tokens, and by reducing a sequence of types to a single one (see Section 3.3.4), we find the types of linguistic expressions of multiple tokens. It is important to recognize that syntactic types are an encoding of syntactic distribution, not only because they are distributionally definable (a.o. Clark, 2013) and their assignment distributionally learnable (e.g. Buszkowski and Penn, 1990; Bisk and Hockenmaier, 2012), but also because they are thus interpreted.

In CGs following Lambek (1958), an atomic type is assigned to all linguistic expressions that share the same distribution. Thus besides noun phrases, proper names and pronouns are also assigned $np$ by occurring in all contexts noun phrases do.

An expression is of type $A\backslash B$ (resp. $B/A$) if and only if it yields an expression of type $B$ when concatenated with an expression of type $A$ on its left (resp. right). For example, *see* has type $np\backslash s/np$, since it takes an object noun phrase on the right and then a subject noun phrase on the left to yield a sentence. Forward and backward slashes used this way belong to the so-called Lambek's notation. Readers familiar with Steedman's (1996) notation may recall that $B/A$ means the same thing, but $B\backslash A$ replaces Lambek's $A\backslash B$, so the type of a transitive verb is written as $s\backslash np/np$.

On the other hand, an expression is of type $A{\downarrow}B$ (resp. $B{\uparrow}A$) if and only if it yields an expression of type $B$ when filling a distinguished gap in (resp. having a distinguished gap filled by) an expression of type $A$. For example, the expression in (3.26a) has type $s{\uparrow}np$, since filling its gap with a noun phrase yields a sentence. In (3.26b), *every dog* has type $s{\uparrow}np{\downarrow}s$, since it fills the gap of an expression of type $s{\uparrow}np$ to yield a sentence.

(3.26) a. **A boy saw __ in a park**.
b. **A boy saw** every dog **in a park**.

Thus, in addition to slashes for concatenation, vertical arrows situate a filler in a context, or as Barker and Shan (2014) put it in terms of programming languages, a *continuation*. CGs using such connectives are developed by Barker (2019); Barker and Shan (2014); Morrill (2011); Morrill et al. (2010) to deal with scope-taking, displacement, and other discontinuity phenomena in natural language. Here we adopt Morrill et al.'s notation as downward and upward arrows signify respectively insertion and extraction. The equivalents in Barker and Shan's notation are $B/\!\!/A$ for $A{\downarrow}B$ and $A\backslash\!\!\backslash B$ for $B{\uparrow}A$. For readability, later we will often write a type of the form $B{\uparrow}A{\downarrow}C$ in Barker and Shan's "tower notation" as

$$\frac{C \mid B}{A}.$$

One can think of the horizontal and vertical bars as the upward and downward arrows, respectively.

### 3.3.2 Atom-vertex correspondence

CGs output atom matches in the process of inferring the type of a linguistic expression as a sequence of typed tokens. To translate atom matches to pairs of equivalent vertices as seen in Section 3.1, we need to know how atom occurrences correspond to vertices.

We propose a partly deterministic way to set up desired atom-vertex correspondence, with the deterministic part being this: for any two atoms in a type, whether or not they correspond to the same vertex is determined by a lexical property of the typed expression in question. We encode this lexical property with three possible *tones* of type connectives: the *applicative* tone (unmarked), the *modificative* tone (marked by $), and the *coordinative* tone (marked by &). Connective tonality as such elaborates on the casual distinction in the CG literature between functor and modifier types.

Let us explain what this would mean for the following tokens (with atoms numbered for ease of reference).

(3.27) a. sail $\to np_1 \backslash s_2$
b. big $\to np_3 /_{\$} np_4$
c. and $\to s_5 \backslash_{\&} s_6 /_{\&} s_7$

For reasons that will later become apparent, we want $np_1$ and $s_2$ in (3.27a) to correspond to different vertices, $np_3$ and $np_4$ in (3.27b) to correspond to the same vertex, and all three of $s_5$, $s_6$, $s_7$ in (3.27c) to correspond to different vertices. According to our plan, these should result from the fact that the connective of (3.27a) is applicative, that of (3.27b) modificative, and those of (3.27c) coordinative.

We will capture the effect of tones with *depth calculation*, a deterministic procedure that assigns each atom of a type a non-negative integer called its *depth*, so two atoms will have the same correspondent if and only if they have the same depth. Atom-vertex correspondence can then be obtained by putting certain vertices of a token in one-to-one correspondence with the depths in its type. This is exactly the purpose served by the source label assignment. We will shortly provide the details and examples for depth calculation in each tone.

The idea of atom-vertex correspondence, though formulated differently, is due to Calder et al. (1988); Zeevat (1988) and has been applied in Baldridge and Kruijff (2002); White (2006) to construction of HLDS formulas. In these works, the correspondence between atoms and "vertices" is entirely arbitrary, directly specified for each linguistic token. So why do we introduce the determinism of depth calculation and limit lexical arbitrariness to connective tonality and the source label assignment? The reason is primarily methodological. The account above implies that a token should have (at least) as many non-negative sources as there are depths in its type. Since, tonality aside, types encode syntactic distribution, depth calculation effectively allows us to go back and forth between the syntactic distribution and semantics of a linguistics expression, partly inferring one based on independent knowledge about the other.

The success of distributional semantics demonstrated by modern language models (see Section 1.1) should have convinced us of a strong connection between syntactic distribution and semantics. Expressing it in symbolic terms, depth calculation narrows down the hypothesis space for aligning syntactic and semantic resources and makes testable predictions. For example, consider the following type of *every* (which we will replace with a continuized analogy, but that does not matter here).

(3.28) every $\to s_1 \,/\, np_2 \setminus s_3 \,/\, n_4$

a.

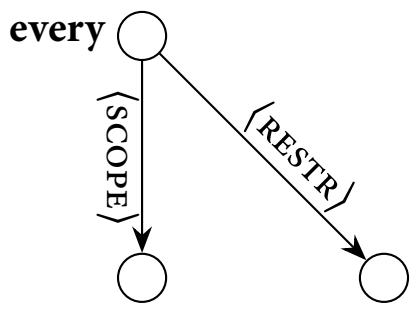

b.

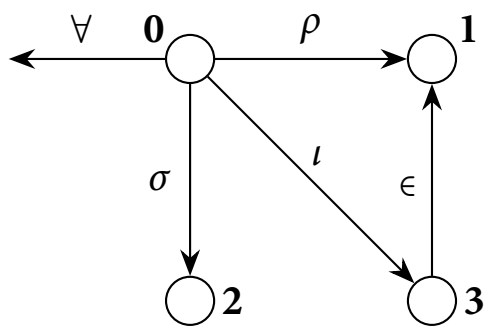

HLDS (White, 2006) pairs with this type a quantification structure (3.28a) of three vertices by associating $np_2$ and $n_4$ with the same vertex. According to depth calculation, however, all four atoms will correspond to distinct vertices, so we expect a quantification structure containing at least four vertices as in (3.28b). Section 2.3.2 has gathered the empirical considerations in favor of the latter.

### 3.3.3 Depth calculation

We introduce a parameterized function $\mathsf{d}_{i,j}$ that tags each atom of a type with a superscript. The depths in a type $A$ are given by $\mathsf{d}_{0,0}(A)$. Readers may jump to Section 3.3.4 if they wish to see established atom-vertex correspondence in use, and return to this section when they want to work out the details for themselves.

*3.3.3.1 Applicative tone* Let | stands for any applicative connective, $B$ the numerator, and $A$ the denominator. Define

$$\mathsf{d}_{i,j}(p) = p^i,$$
$$\mathsf{d}_{i,j}(B \,|\, A) = \mathsf{d}_{i,j+1}(B) \,|\, \mathsf{d}_{i+j+1,0}(A).$$

We show how this works with the types of a transitive and a quantificational determiner:

(3.29) a. $\mathsf{d}_{0,0}(np\backslash s/np) = \mathsf{d}_{0,1}(np\backslash s) \,/\, \mathsf{d}_{1,0}(np)$
$= \mathsf{d}_{2,0}(np) \setminus \mathsf{d}_{0,2}(s) \,/\, \mathsf{d}_{1,0}(np)$
$= np^2 \setminus s^0 \,/\, np^1$

b. $\mathsf{d}_{0,0}(s{\uparrow}np{\downarrow}s/n) = \mathsf{d}_{0,1}(s{\uparrow}np{\downarrow}s) \,/\, \mathsf{d}_{1,0}(n)$
$= \mathsf{d}_{2,0}(s{\uparrow}np) \downarrow \mathsf{d}_{0,2}(s) \,/\, \mathsf{d}_{1,0}(n)$
$= \mathsf{d}_{2,1}(s) \uparrow \mathsf{d}_{3,0}(np) \downarrow \mathsf{d}_{0,2}(s) \,/\, \mathsf{d}_{1,0}(n)$
$= s^2 \uparrow np^3 \downarrow s^0 \,/\, n^1$

(Check the source labels of *walk* in Figure 3.2 and *every* in (3.28b) for the corresponding vertices of these atoms.)

The intuition behind this calculation is that in the applicative tone, depths indicate the order in which arguments are received by a functor. Since that order is usually a reflection of relative obliqueness, depth calculation generalizes the latter to cases where a comparison of edge labels is not conventionally defined. For example, from the depths in (3.29b) we find that the $\iota$-edge is less oblique than the $\sigma$-edge and thus count donkey cataphora (see Barker and Shan, 2014, p. 94) among the weak crossover examples in Section 3.2.2:

(3.30) *If **he**$_i$ walks **it**$_j$, **a boy**$^i$ feeds **a dog**$^j$.

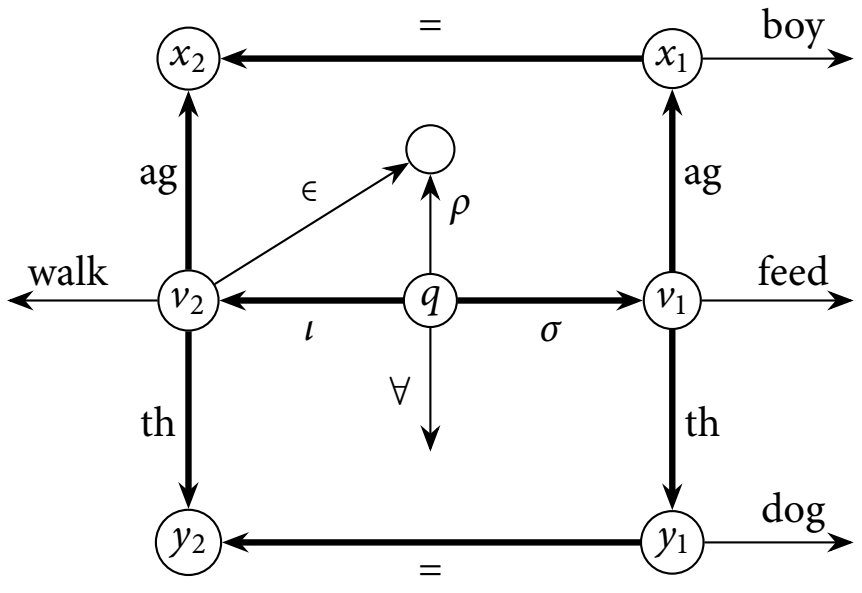

(The problematic parallel paths are $q \to v_1 \to x_1 \to x_2 \parallel q \to v_2 \to x_2$ and $q \to v_1 \to y_1 \to y_2 \parallel q \to v_2 \to y_2$.)

*3.3.3.2 Modificative tone* The action of $\mathsf{d}_{i,j}$ on a modificative connective $|_\$$ is as follows, where $B$ denotes the numerator and $A$ the denominator as before:

$$\mathsf{d}_{i,j}(B\,|_\$\,A) = \mathsf{d}_{i,j}(B)\,|_\$\,\mathsf{d}_{i,j}(A)$$

In particular, the modificative tone produces duplicate depths for the numerator and denominator when they are identical, as shown by the lexical entry of an adjective:

(3.31) lucky $\to n^0 /_\$ n^0$

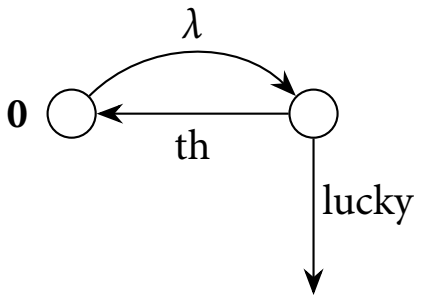


Having the numerator and denominator of a modifier share corresponding vertices ensures the same atom matches for a modifiable type, regardless of the presence of that modifier (see (3.41) for example).

We give the lexical entries of another two tokens closely related to modification. Whereas in (3.31) an adjective carries a $\lambda$-cycle of its own, a relativizer contributes a $\lambda$-edge to be contained in a cycle once its ends find equivalents through type reductions:

(3.32) which $\to np_1^0 \backslash_\$ np_2^0 / (s_3^1 \uparrow np_4^0)$

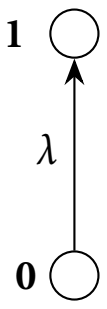

Here it is crucial for the gapped noun phrase, so to speak, to be identified with the noun phrase being modified, hence $np_1$, $np_2$, $np_4$ all correspond to the tail of the $\lambda$-edge. But depth calculation yields instead

$$\mathsf{d}_{0,0}(np_1 \backslash_{\$}\, np_2 \,/\, (s_3 \uparrow np_4)) = np_1^0 \backslash_{\$}\, np_2^0 \,/\, (s_3^1 \uparrow np_4^2),$$

where $np_4$ refers to a vertex not existing. A fix we use is to take modulo by the sort of the semgraph (the number of non-negative sources it actually has). Such a post-processing might seem a setback for using depth calculation to predict the sorts of semgraphs from their types, though fortunately (3.32) is the only case known where it is needed. In case one wonders, a relativizer usually gets typed as $n\backslash_{\$}n/(s{\uparrow}np)$ or so (e.g. Morrill, 2011, pp. 110ff) on the assumption that relative clauses are noun modifiers. But if we want to treat them as iterator filters following the remark of Section 2.2.4, they should serve as noun phrase modifiers in accordance with (3.32). The two views can also be differentiated by plural predication:

(3.33) [Every boy and every dog] who hugged sailed.

Given the semantics of *hug*, the relative clause above has to sit external to either conjunct (see Section 4.2.3.4 for discussion).

As for the gap in a relative clause, a somewhat traditional view treats it as a silent linguistic token to be typed as usual. The following lexical entry uses the type given in Barker and Shan (2014, p. 39):

(3.34) __ $\rightarrow (s^0 \uparrow np^1) \downarrow_{\$} (s^0 \uparrow np^1)$

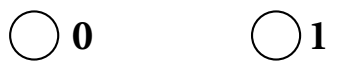


A gap is so typed because when it fills the gap in a sentence missing a noun phrase (e.g. (3.26a)), the latter remains unchanged. On the other hand, its semgraph has no edge but two vertices that are both sources. It is easy to see that an all-source *edgeless* graph is an identity element with respect to parallel composition, provided all its source labels are also found in the other graph. Actually we have already encountered an example of this kind, the indefinite article (3.21).

*3.3.3.3 Coordinative tone* Coordinators can assume any type of the form $A \backslash_{\&} A /_{\&} A$ in CGs (e.g. Steedman, 2011, p. 91), so as to capture the fact that coordination is polymorphic, able to join any expressions identically typed to yield one typed alike. Depth calculation in the coordinative tone is tailored specifically to such types:

$$\mathsf{d}_{i,j}(A \backslash_{\&} A /_{\&} A) = \mathsf{d}_{i+j+2,|A|-1}(A) \backslash_{\&} \mathsf{d}_{i,j+2|A|}(A) /_{\&} \mathsf{d}_{i+j+1,1}(A),$$

where $|A|$, called the *size* of $A$, is the number of distinct depths given by $\mathsf{d}_{0,0}(A)$.

The effect is for the left and right denominators and the last numerator to each retain a disjoint set of depths, with those of the left denominator greater than those of the right denominator, as shown by the following examples:

(3.35) a. $\mathsf{d}_{0,0}(s \backslash_{\&} s /_{\&} s)$
$= \mathsf{d}_{2,0}(s) \backslash_{\&} \mathsf{d}_{0,2}(s) /_{\&} \mathsf{d}_{1,1}(s)$
$= s^2 \backslash_{\&} s^0 /_{\&} s^1$

b. $\mathsf{d}_{0,0}((np\backslash s) \backslash_{\&} (np\backslash s) /_{\&} (np\backslash s))$
$= \mathsf{d}_{2,1}(np\backslash s) \backslash_{\&} \mathsf{d}_{0,4}(np\backslash s) /_{\&} \mathsf{d}_{1,1}(np\backslash s)$
$= (np^4\backslash s^2) \backslash_{\&} (np^5\backslash s^0) /_{\&} (np^3\backslash s^1)$

c. $\mathsf{d}_{0,0}((np\backslash s/np) \backslash_{\&} (np\backslash s/np) /_{\&} (np\backslash s/np))$
$= \mathsf{d}_{2,2}(np\backslash s/np) \backslash_{\&} \mathsf{d}_{0,6}(np\backslash s/np) /_{\&} \mathsf{d}_{1,1}(np\backslash s/np)$
$= (np^6\backslash s^2/np^5) \backslash_{\&} (np^7\backslash s^0/np^8) /_{\&} (np^4\backslash s^1/np^3)$

To justify this calculation, below we provide a lexical entry of each type alongside an example of its use. As a matter of notation, we abbreviate $np\backslash s$ and $np\backslash s/np$ by $vp$ and $vt$.

(3.36) a. and → $s^2 \backslash_{\&} s^0 /_{\&} s^1$

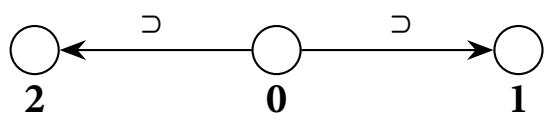


Joe surfed *and* Ben sailed.

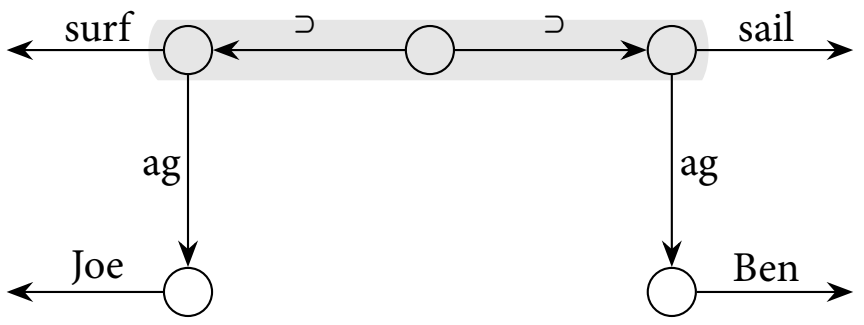


b. and → $vp \backslash_{\&} vp /_{\&} vp$

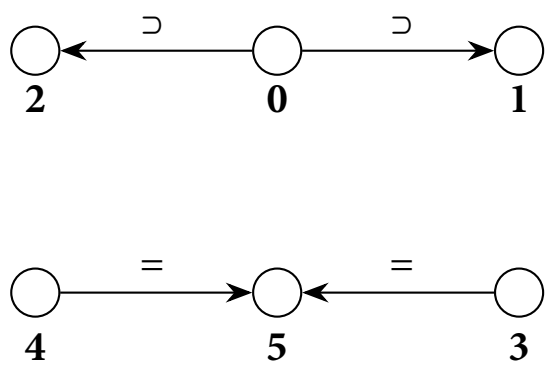


Joe surfed *and* sailed.

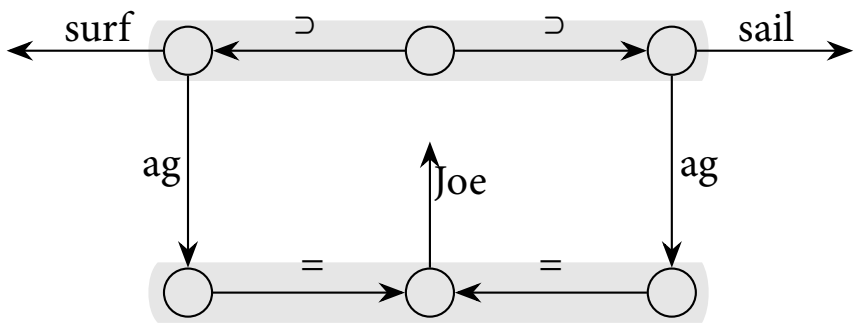


c. and → $vt \backslash_{\&} vt /_{\&} vt$

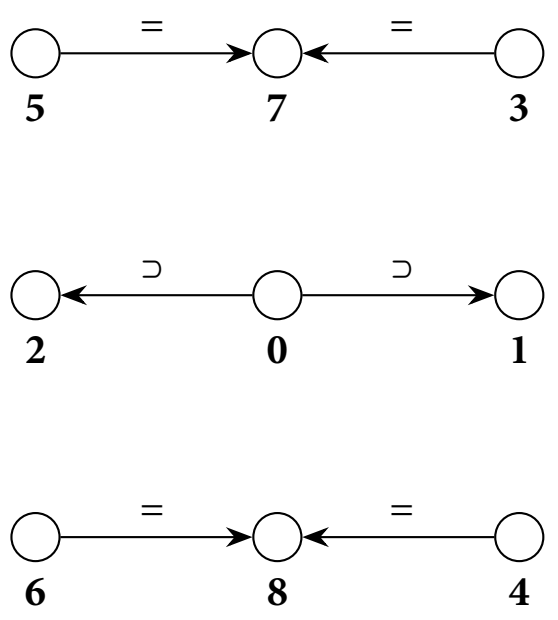


Joe rented *and* sank a boat.

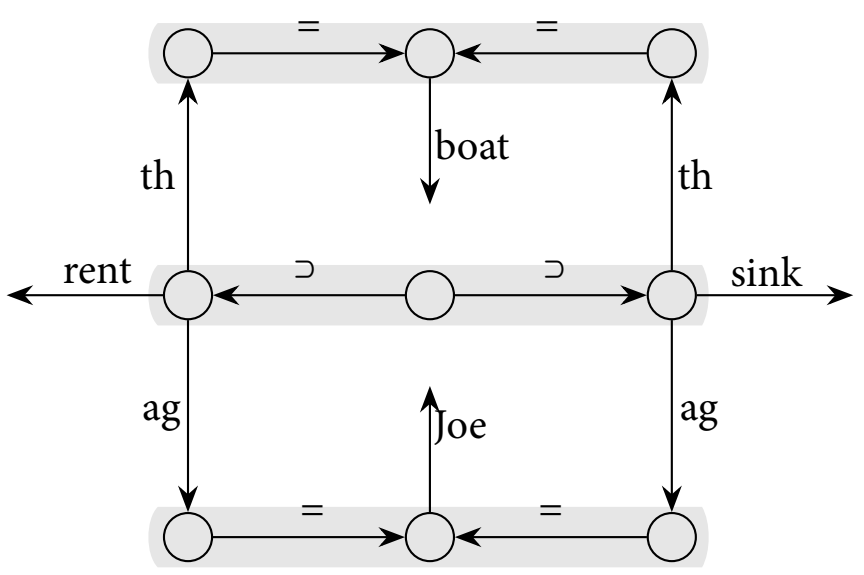


Now one can easily relate depth calculation in the coordinative tone to the way we represent coordination in semgraphs. As shown in (3.36a-c), to each coordinator type $A \backslash_{\&} A /_{\&} A$ there corresponds a token whose semgraph consists of

i) a unique triple of the form $\bullet_2 \leftarrow \bullet_0 \rightarrow \bullet_1$ for the semantics of coordination (a union if a conjunction, a choice if a disjunction),
ii) and as many as $|A| - 1$ triples of the form $\bullet_i \rightarrow \bullet_{i+|A|-1} \leftarrow \bullet_{i-|A|+1}$ for argument sharing.

How these triples are divided evenly among the left and right denominators and the last numerator should be clear from a comparison of (3.35a-c) with (3.36a-c).

A nice consequence of the depths we found is that one coordinate of the coordination structure is made less oblique than the other. For example, $v_1$ is less oblique than $v_2$ in (3.36) by being associated with a greater depth. We can thus follow up on the discussion of Section 3.3.3.1 with cataphora in coordination, as illustrated by the following example:

(3.37) ***He**$_i$ sailed and **a boy**$^i$ returned.

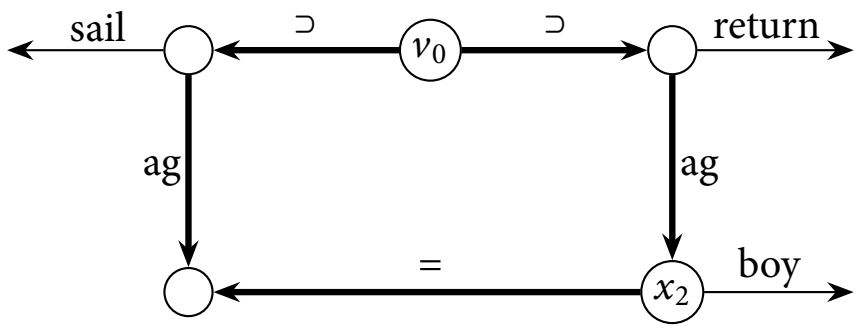


$(v_0 \to v_2 \to x_2 \to x_1 \parallel v_0 \to v_1 \to x_1)$

(Also, the returning boy need not be the sailor unless $x_2$ take scope over $v_0$, as discussed in Section 2.2.6.)

In conclusion, it should be mentioned that connective tonality as discussed here is to be distinguished from connective modality as discussed in Baldridge and Kruijff (2003); Hepple (1990); Jacobson (1990); Moortgat (1997). Whereas modes refer to different strategies for grammatical composition and relate to type reductions (e.g. slashes for concatenation vs. arrows for continuation), tones play no roles beyond those in atom-vertex correspondence.

### 3.3.4 Type reduction

We now turn to type reductions to compute atom matches from a sequence of typed tokens. With atom-vertex correspondence in place, such matches immediately translate to pairs of equivalent vertices.

*3.3.4.1 Rules* A *type reduction* $\Gamma \to A$ relates a finite sequence $\Gamma$ of types, denoted by capital Greek letters, to a single type $A$. A type reduction is also called a *sequent*, where $\Gamma$ is said to be the *antecedent* and $A$ the *succedent*. We read it as a statement that an antecedent is *reducible to* a succedent, or conversely, the latter is *provable from* the former.

CGs specify rules for deriving valid type reductions. There are mainly two ways to do this and present derivations correspondingly. In the first place, one can curate a system of $n$-ary reduction rules (i.e. templates) and apply them successively to reduce a sequence. This approach leads to combinatory CGs (CCGs; e.g. Steedman, 1992, 1996, 2011; Szabolcsi, 1992), an immediate extension of the original works by Ajdukiewicz (1935); Bar-Hillel (1953). The bulk of this thesis takes the combinatory approach for its simplicity: it organizes syntactic derivations in tree structures that mirror the phrase structures familiar from constituency grammars. Thus combining types can be thought of as merging constituents, where reductions rules serve as a phrase labeling method in the sense of Chomsky (2013).

Alternatively there is the perspective of logical calculi. Following Lambek (1958), type logical grammars (TLGs) are the branch of CGs where rules of inference operate directly with sequents,

forwarding premises to a conclusion at each step (e.g. Barker and Shan, 2014; Kubota and Levine, 2020; Moortgat, 2011; Morrill, 2011; Moot and Retoré, 2012). In Appendix 3.A we present a TLG implementation of type reductions that produces atom matches nonetheless. This is to demonstrate that our view of syntax-semantics interface detaches one from the details of syntactic theories, as far as semantic construction is informed of clues to unification.

Now we introduce a CCG based on Barker and Shan (2014, sec. 12.2) and White et al. (2017) with the reduction rules given below (recall that connective tonality is irrelevant here).

**Definition 3.1.** *Continuized CCG.*

BASE RULES

$$\begin{array}{lll}
\textit{function application}: & A,\ A\backslash B \to B & B/A,\ A \to B \\
\textit{switched application}: & A,\ A\backslash B/C \to B/C & C\backslash(B/A),\ A \to C\backslash B
\end{array}$$

TOWER RULES

$$\begin{array}{rlll}
\textit{lowering}: & \dfrac{B \mid A}{A} \to B & & \\
\textit{lifted lowering}: & \dfrac{E \mid D}{B} \to \dfrac{E \mid D}{C} & & \textit{where } B \to C \textit{ is a (lifted) lowering} \\
\textit{lifted application}: & B,\ \dfrac{F \mid E}{C} \to \dfrac{F \mid E}{D} & \dfrac{F \mid E}{B},\ C \to \dfrac{F \mid E}{D} & \textit{where } B,\ C \to D \textit{ is an application}
\end{array}$$

Of these unary rules are conventionally called *type-shifters* and binary rules *combinators*. Let us walk through them in action.

All CGs evolve from function application, which, like all other combinators, comes in a pair. The first rule applies $A\backslash B$ to $A$ on its left, and the second $B/A$ to $A$ on its right, both resulting in type $B$. The following derivation of a familiar sentence illustrates function application in both directions.

(3.38)

$$\dfrac{\dfrac{\overset{\text{a}}{np_1/_{\$}n_2} \quad \overset{\text{boy}}{n_3}}{np_1} \qquad \dfrac{\overset{\text{walked}}{np_4\backslash s_5/np_6} \quad \dfrac{\overset{\text{a}}{np_7/_{\$}n_8} \quad \overset{\text{dog}}{n_9}}{np_7}}{np_4\backslash s_5}}{s_5}$$

In CCG derivations, each rule instance is indicated by a horizontal bar, separating the antecedent on the top and the succedent on the bottom. In (3.38), indefinite articles apply to nouns on the right. The transitive applies to the object on the right, resulting in a verb phrase that applies to the subject on the left.

By function application $A\backslash B/C$ would combine with $C$ on its right, and $A\backslash(B/C)$ with $A$ on its left. Switched application, however, allows the former to combine with $A$ on its left to yield $B/C$ and the latter to combine with $C$ on its right to yield $A\backslash B$. This is useful in handling improper constituents, as illustrated by the following example:

(3.39)

Joe $np_1$   washed $np_2\backslash s_3/np_4$   and $(s_5/np_6)\backslash_{\&}(s_7/np_8)/_{\&}(s_9/np_{10})$   Ben $np_{11}$   shaved $np_{12}\backslash s_{13}/np_{14}$   himself $np_{15}$

$$\dfrac{\dfrac{\dfrac{np_1 \quad np_2\backslash s_3/np_4}{s_3/np_4} \quad \dfrac{(s_5/np_6)\backslash_{\&}(s_7/np_8)/_{\&}(s_9/np_{10}) \quad \dfrac{np_{11} \quad np_{12}\backslash s_{13}/np_{14}}{s_{13}/np_{14}}}{(s_5/np_6)\backslash_{\&}(s_7/np_8)}}{s_7/np_8} \quad np_{15}}{s_7}$$

Here *Joe washed* and *Ben shaved*, though conjoined, are traditionally not considered constituents. Switched application allows a subject to combine with a transitive to yield a conjunct of type $s/np$.

*Remark.* Switched application is a composite of function application and the so-called *switching rule*

$$\textit{switching}: \quad A\backslash B/C \rightleftharpoons A\backslash(B/C).$$

It should be distinguished from *function composition*, a combinator used together with *type-raising* in commonly known CCGs (e.g. Steedman, 1996) to serve improper constituents, scope ambiguity, and displacement:

$$\begin{array}{rll} \textit{function composition}: & C\backslash A,\ A\backslash B \to C\backslash B & B/A,\ A/C \to B/C, \\ \textit{type-raising}: & B \to A/B\backslash A & B \to A/(B\backslash A). \end{array}$$

We are not using these partly because continuation provides a better coverage for scope-taking and displacement (see Barker and Shan, 2014, chap. 11), partly because type-raising is unbounded in applicability, introducing atoms that would break existing atom-vertex correspondence. Following Dowty (1988); Steedman (1991), our take is to allow certain tokens to be lexically ambiguous between plain and raised types (including their continuation analogy, $A{\uparrow}B{\downarrow}A$). This not only facilitates proof search automation, but also saves the need of graph editing (e.g. by adding vertices) to maintain atom-vertex correspondence.

The tower rules are devised by Barker and Shan (2014) for reasoning about continuations, a grammatical composition mode that relates an expression to its wrapping context. In the base case, lowering shifts $A{\uparrow}A{\downarrow}B$ down to $B$ (think of $A{\uparrow}A$ as a gap left by extracting an expression of type $A$ away from itself: a filler of gap yielding $B$ is by definition of type $A{\uparrow}A{\downarrow}B$, but it should also have type $B$). The derivation of a relative clause illustrates its use.

(3.40)

who $np_6\backslash_{\$}np_7/(s_8{\uparrow}np_9)$   —   walked $np_{14}\backslash s_{15}/np_{16}$   a $np_{17}/_{\$}n_{18}$   dog $n_{19}$

$$\dfrac{np_6\backslash_{\$}np_7/(s_8{\uparrow}np_9) \qquad \begin{array}{c} \dfrac{\dfrac{s_{12}{\uparrow}np_{13} \mid s_{10}}{np_{11}} \qquad \begin{array}{c} np_{14}\backslash s_{15}/np_{16} \quad np_{17}/_{\$}n_{18} \quad n_{19} \\ \vdots \qquad \vdots \qquad \vdots \\ np_{14}\backslash s_{15} \end{array}}{\dfrac{s_{12}{\uparrow}np_{13} \mid s_{10}}{s_{15}}} \\ \Downarrow \\ s_{12}{\uparrow}np_{13} \end{array}}{np_6\backslash_{\$}np_7}$$

For readability $\Downarrow$ replaces a horizontal bar to indicate lowering. After combining *walked a dog* with a gap by lifted application (see below), reducing the gapped clause's type to $s{\uparrow}np$ (i.e. a sentence missing a noun phrase) allows it to combine with the relativizer *who*.

Lifted lowering is defined recursively: if $B$ shifts to $C$ by lowering, lifted or not, then $D{\uparrow}B{\downarrow}E$ shifts to $D{\uparrow}C{\downarrow}E$. While the use of this rule is best seen in our treatment of quantifier scope in Section 5.1, the intuition behind is that the local behavior of an expression (as encoded by a tower's lower part) is independent of its continuation (as encoded by a tower's top layer).

The same intuition carries over to lifted application. If combining $B$ and $C$ through plain, switched, or lifted application yields $D$, then putting either $B$ or $C$ under a continuation layer $F \mid E$ merely passes it on to $D$. Using this rule, the following example extends (3.40) to a full sentence.

(3.41)

$$
\dfrac{\dfrac{\dfrac{\overset{\text{every}}{\dfrac{s_3 \mid s_1}{np_2}/n_4} \quad \overset{\text{boy}}{n_5}}{\dfrac{s_3 \mid s_1}{np_2}} \qquad \dfrac{\overset{\text{who}}{np_6\backslash_\$ np_7/(s_8{\uparrow}np_9)} \quad \overset{\text{—}}{\dfrac{s_{12}{\uparrow}np_{13} \mid s_{10}}{np_{11}}} \quad \overset{\text{walked}}{np_{14}\backslash s_{15}/np_{16}} \quad \overset{\text{a}}{np_{17}/_\$ n_{18}} \quad \overset{\text{dog}}{n_{19}}}{\vdots \; np_6\backslash_\$ np_7}}{\dfrac{s_3 \mid s_1}{np_7}} \qquad \dfrac{\overset{\text{fed}}{np_{20}\backslash s_{21}/np_{22}} \quad \overset{\text{it}}{np_{23}}}{np_{20}\backslash s_{21}}}{\dfrac{s_3 \mid s_1}{s_{21}}}
$$

$$
\Downarrow
$$

$$
s_3
$$

Lifted application happens twice in this derivation, first when the relative clause applies to *every boy*, then when *fed it* applies to its subject. The continuation layer introduced by *every* persists until the tower $s{\uparrow}s{\downarrow}s$ is finally lowered. Note the role played by the modifier here, that is, the relative clause. On the one hand, $np_6$ is canceled by $np_2$, which corresponds to the iterator of the quantifier over boys, and on the other, $np_7$ is canceled by $np_{20}$, which corresponds to the agent of feeding. Thus by assigning $np_6$ and $np_7$ the same vertex, $np_2$ and $np_{20}$ are related as they would be were the relative clause not there.

*3.3.4.2 Atom matches* From a derivation or proof tree like (3.38) we can gather the reduction

$$
np_1/_\$ n_2,\; n_3,\; np_4\backslash s_5/np_6,\; np_7/_\$ n_8,\; n_9 \to s_5,
$$

where the antecedent comprises all the terminals, and the succedent equals the root. We can also gather the atom matches that derive this reduction, if we notice that all rules in Definition 3.1 eventually match and cancel a pair of some $A$ occurrences.

Define the set $\mathsf{m}(A, A)$ of *atom matches produced by a rule* as the set of atom pairs (order irrelevant) identically situated therein. That is,

$$
\mathsf{m}(A, A) = \begin{cases} \{(p, p)\} & \text{for any atom } A = p, \\ \mathsf{m}(B, B) \cup \mathsf{m}(C, C) & \text{for any fraction } A = B \mid C. \end{cases}
$$

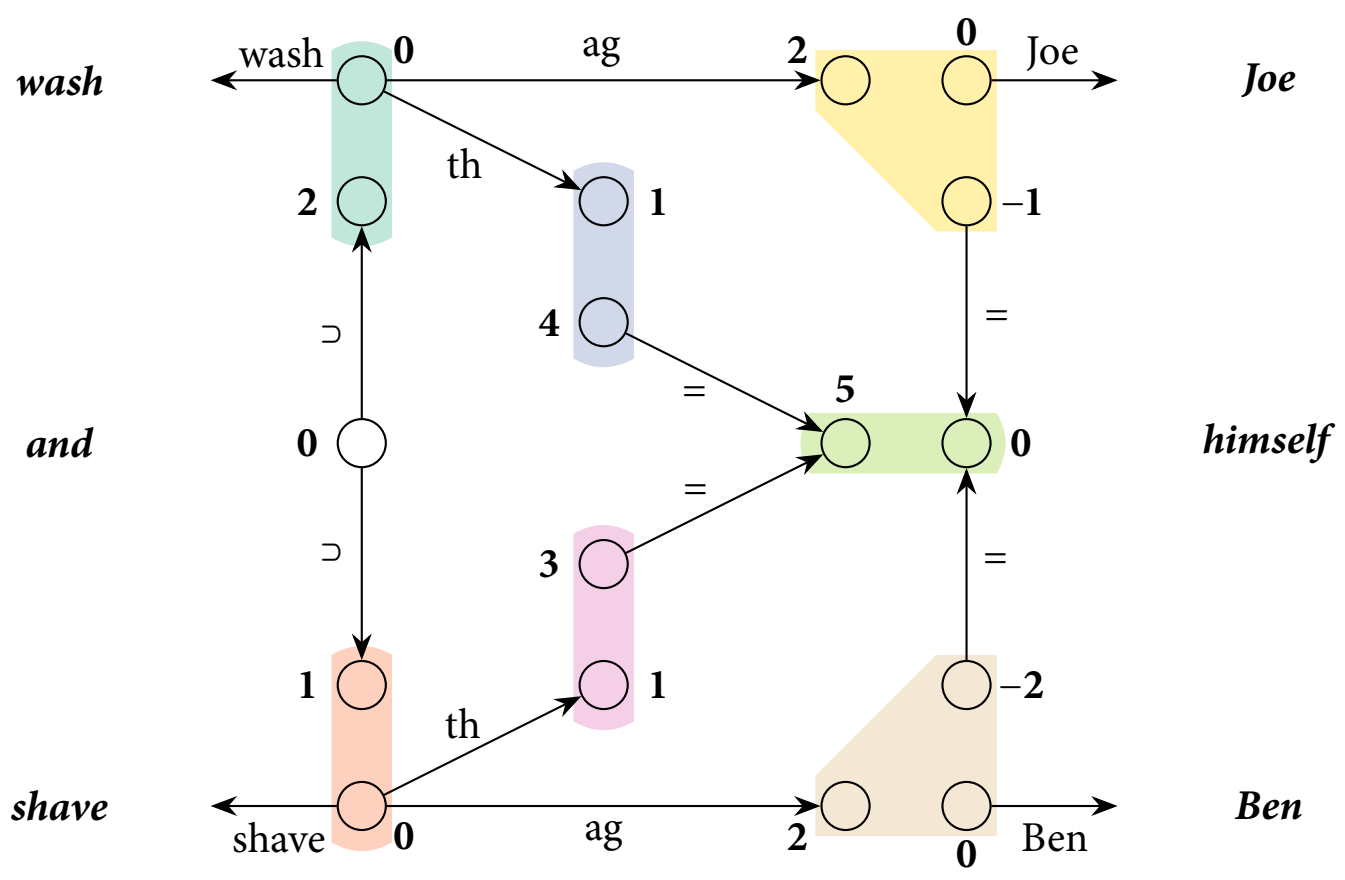


**Figure 3.3:** Fusing equivalence for *Joe washed and Ben shaved himself.*

The union of such sets from all derivation steps gives us the *atom matches produced by a derivation* or *proof.* Given atom-vertex correspondence established in Section 3.3.3, such atom matches translate immediately into equivalence among vertices (more precisely, sources).

For example, (3.38) produces the matches in (3.42a). Using the depths calculated earlier, we obtain (3.42b), which yields exactly the equivalence relation illustrated in Figure 3.2.

(3.42) a boy walked a dog
$np_1^0/_{\$}n_2^0 \quad n_3^0 \quad np_4^2\backslash s_5^0/np_6^1 \quad np_7^0/_{\$}n_8^0 \quad n_9^0 \quad \to s_5^0$

a. $(np_1^0, np_4^2), (n_2^0, n_3^0), (np_6^1, np_7^0), (n_8^0, n_9^0)$

b. $(\mathbf{0}_a, \mathbf{2}_{walk}), (\mathbf{0}_a, \mathbf{0}_{boy}), (\mathbf{1}_{walk}, \mathbf{0}_a), (\mathbf{0}_a, \mathbf{0}_{dog})$

We have arrived at a rather simple presentation of syntax. By listing the ways of matching atoms in a type reduction, we abstract away from the details of syntactic derivations while preserving their essentials. This idea is formally articulated in the studies on proof nets (for Lambek calculus, de Groote, 1999; Lamarche and Retoré, 1996; Pentus, 1998, 2010, a.o.; for continuized TLGs, Moot, 2016, 2020), which show that atom matches characterize what meaningfully distinguishes derivations apart. The point can be intuitively made here as different sets of atom matches imply different partitions of discourse referents.

Now it is easy to combine fusing equivalence decided by syntactic and non-syntactic means. For example, whereas (3.39) produces the matches in (3.43a) and therefore the equivalence in (3.43b), co-reference resolution is responsible for the equivalence in (3.43c). Together, (3.43b&c) yield the partition in Figure 3.3. See (3.36) for the lexical entries for coordinators. The semantics of a split anaphor contains multiple equality edges tailed by negative sources; cf. (3.5a).

(3.43) Joe washed and Ben shaved himself
$np_1^0 \quad np_2^2\backslash s_3^0/np_4^1 \quad (s_5^2/np_6^4)\backslash_{\&}(s_7^0/np_8^5)/_{\&}(s_9^1/np_{10}^3) \quad np_{11}^0 \quad np_{12}^2\backslash s_{13}^0/np_{14}^1 \quad np_{15}^0 \quad \to s_7^0$

a. $(np_1^0, np_2^2), (s_3^0, s_5^2), (np_4^1, np_6^4), (np_8^5, np_{15}^0),$

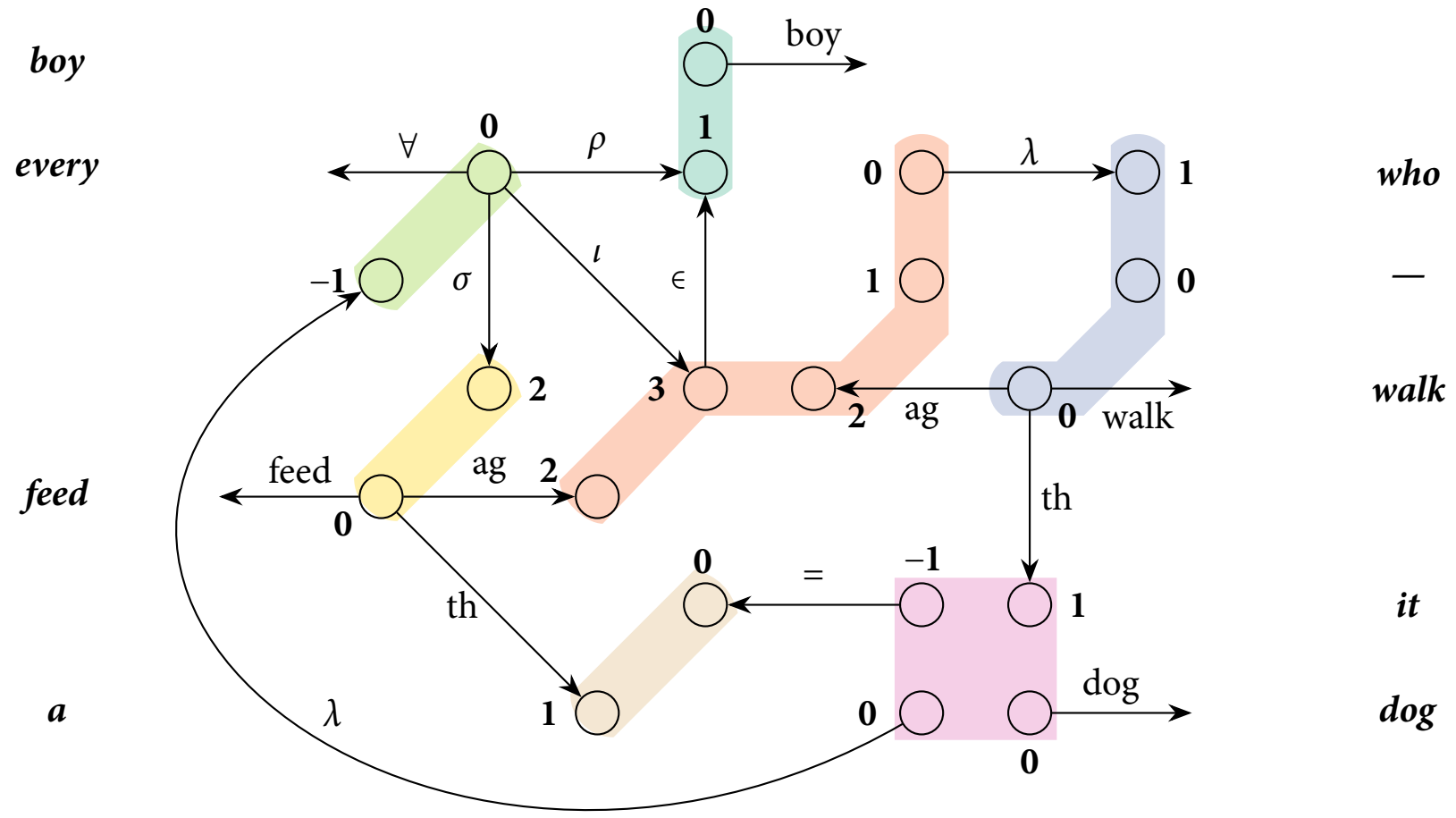


**Figure 3.4:** Fusing equivalence for *every boy who walked a dog fed it.*

$(s^1_9, s^0_{13}), (np^3_{10}, np^1_{14}), (np^0_{11}, np^2_{12})$

b. $(\mathbf{0}_{Joe}, \mathbf{2}_{wash}), (\mathbf{0}_{wash}, \mathbf{2}_{and}), (\mathbf{1}_{wash}, \mathbf{4}_{and}), (\mathbf{5}_{and}, \mathbf{0}_{himself}),$
$(\mathbf{1}_{and}, \mathbf{0}_{shave}), (\mathbf{3}_{and}, \mathbf{1}_{shave}), (\mathbf{0}_{Ben}, \mathbf{2}_{shave})$

c. $(-\mathbf{1}_{himself}, \mathbf{0}_{Joe}), (-\mathbf{2}_{himself}, \mathbf{0}_{Ben})$

To see another example, consider the wide-scope-indefinite reading of (3.44), where we talk about a specific dog always fed by its walkers. The equivalence in Figure 3.4 is due to (3.41) producing (3.44a) and its translation (3.44b), and (3.44c) from co-reference and precedence resolution. Note how the $\lambda$-edges contributed by *a* and *who* find the equivalents of their heads differently, and how donkey anaphora is treated as usual co-reference. See (3.32) and (3.34) for the lexical entries of relativizers and gaps.

(3.44) every $s^2_1{\uparrow}np^3_2{\downarrow}s^0_3/n^1_4$ boy $n^0_5$ who $np^0_6\backslash_\$ np^0_7/(s^1_8{\uparrow}np^0_9)$ — $(s^0_{10}{\uparrow}np^1_{11}){\downarrow}_\$(s^0_{12}{\uparrow}np^1_{13})$

walked $np^2_{14}\backslash s^0_{15}/np^1_{16}$ a $np^0_{17}/_\$ n^0_{18}$ dog $n^0_{19}$ fed $np^2_{20}\backslash s^0_{21}/np^1_{22}$ it $np^0_{23}$ $\to s^0_3$

a. $(s^2_1, s^0_{21}), (np^3_2, np^0_6), (n^1_4, n^0_5), (np^0_7, np^2_{20}),$
$(s^1_8, s^0_{12}), (np^0_9, np^1_{13}), (s^0_{10}, s^0_{15}), (np^1_{11}, np^2_{14}),$
$(np^1_{16}, np^0_{17}), (n^0_{18}, n^0_{19}), (np^1_{22}, np^0_{23})$

b. $(\mathbf{2}_{every}, \mathbf{0}_{feed}), (\mathbf{3}_{every}, \mathbf{0}_{who}), (\mathbf{1}_{every}, \mathbf{0}_{boy}), (\mathbf{0}_{who}, \mathbf{2}_{feed}),$
$(\mathbf{1}_{who}, \mathbf{0}_{-}), (\mathbf{0}_{who}, \mathbf{1}_{-}), (\mathbf{0}_{-}, \mathbf{0}_{walk}), (\mathbf{1}_{-}, \mathbf{2}_{walk}),$
$(\mathbf{1}_{walk}, \mathbf{0}_{a}), (\mathbf{0}_{a}, \mathbf{0}_{dog}), (\mathbf{1}_{feed}, \mathbf{0}_{it})$

c. $(-\mathbf{1}_{it}, \mathbf{0}_{dog}), (-\mathbf{1}_{a}, \mathbf{0}_{every})$

Obviously, if one inspects the subtrees of a syntactic derivation, one obtains the type reductions pertinent to semantic construction up to certain stages. Take the following examples from (3.41) or (3.44), with the source labels adapted to the new depths:

(3.45) a. every boy

$s_1^2{\uparrow}np_2^3{\downarrow}s_3^0/n_4^1$ $\quad n_5^0$ $\quad \rightarrow s_1^1{\uparrow}np_2^2{\downarrow}s_3^0$

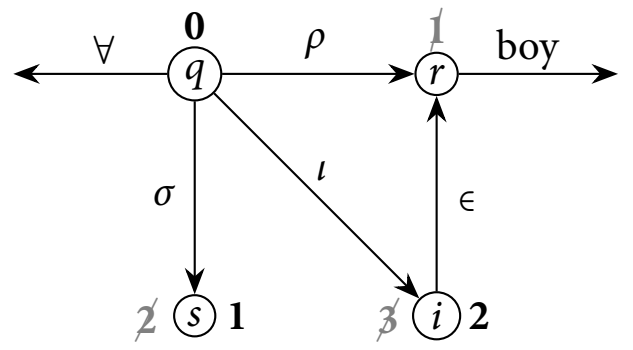


b. a dog

$np_{17}^0/_\$ n_{18}^0$ $\quad n_{19}^0$ $\quad \rightarrow np_{17}^0$

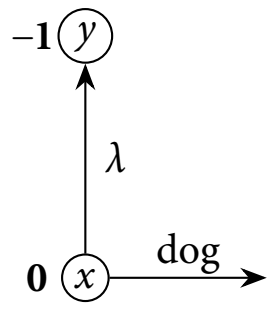


For phrases as such that appear frequently enough, we will often use them as if they were tokens.

We have described all the machinery for constructing semgraphs at the syntax-semantics interface. Upon reflection, recall that syntactic types encode distribution. Given their crucial roles in determining atom matches in type reductions, one may conceive a symbolic argument, if ever so vague, for the learnability of semantic compositionality from distributional cues.

## 3.4 Automation

To facilitate future exploration, we provided a Python implementation of the present syntax-semantics interface, accessible and documented at `https://git.io/JtUTm`.

The automated process of semantic parsing consists of the following steps:

i) take as input a string of tokens with part-of-speech (POS) tags, which are used to look up typed semgraph templates in a JSON lexicon, and a reduction goal (i.e. the succedent type);
ii) search all valid sets of atom matches deriving the type reduction via a specified CG calculus;
iii) perform graph unification based on fusing equivalence given by atom matches as well as non-syntactic resolution.

One can then visualize the composed semgraphs, inspect the underlying atom matches with their proofs, and export these outcomes to LaTeX formats.

Noticeably, our syntax-semantics interface renders semantic construction separable in time from syntactic derivation, even if they can proceed in parallel. This design allows us to pass syntax as a function argument: where we implemented a continuized CCG with the Cocke-Kasami-Younger algorithm (well-known for constituency parsing; see Jurafsky and Martin, 2020, chap. 13), a continuized TLG fragment with naive search and caching (practically efficient), and a product-free Lambek calculus with proof nets in Pentus (2010), users are welcome to substitute in their own CG recipes.

## 3.A Type logical alternative

This appendix demonstrates a logical approach to producing atom matches from type reductions. We present a (product-free associative) continuized TLG based on Barker (2019); Barker and Shan (2014, chap. 13) and Morrill et al. (2010); Morrill (2011, chap. 6) in terms of sequent calculi. To this end, we generalize Γ in sequents of the form $\Gamma \rightarrow A$ to "structures".

Let ∘ denote a constant (that is, a zero-place structural connective) called a *hole*. Define the set B of *terms* and the set S of *structures* as follows:

$$\begin{array}{rcl} \mathsf{B} & ::= & \mathsf{T} \;\mid\; \circ \;\mid\; \mathsf{S}^{\mathsf{S}} \\ \mathsf{S} & ::= & \Lambda \;\mid\; \mathsf{B}\,\mathsf{S} \end{array}$$

That is, a term is either a type, a hole, or a *power*, whose base and exponent are structures, while a structure is a finite sequence of terms ($\Lambda$ stands for the empty sequence). We write $\Gamma[\Delta]$ for a structure where $\Delta$ occurs as a substructure. Our continuized TLG has the following axioms and rules of inference.

**Definition 3.A.1.** *Continuized TLG.*

AXIOM $A \to A$

LOGICAL RULES

$$\frac{A\Phi \to B}{\Phi \to A\backslash B}\,(\backslash R) \qquad \frac{\Phi \to A \quad \Gamma[B] \to C}{\Gamma[\Phi\, A\backslash B] \to C}\,(\backslash L)$$

$$\frac{\Phi A \to B}{\Phi \to B/A}\,(/R) \qquad \frac{\Phi \to A \quad \Gamma[B] \to C}{\Gamma[B/A\,\Phi] \to C}\,(/L)$$

$$\frac{A^{\Phi} \to B}{\Phi \to A{\downarrow}B}\,({\downarrow}R) \qquad \frac{\Phi \to A \quad \Gamma[B] \to C}{\Gamma[\Phi^{A\downarrow B}] \to C}\,({\downarrow}L)$$

$$\frac{\Phi^{A} \to B}{\Phi \to B{\uparrow}A}\,({\uparrow}R) \qquad \frac{\Phi \to A \quad \Gamma[B] \to C}{\Gamma[B{\uparrow}A^{\Phi}] \to C}\,({\uparrow}L)$$

STRUCTURAL RULES

$$\frac{\Gamma[(\Phi \circ \Psi)^{\Delta}] \to A}{\Gamma[\Phi\Delta\Psi] \to A}\,(\vee) \qquad \frac{\Gamma[\Phi\Delta\Psi] \to A}{\Gamma[(\Phi \circ \Psi)^{\Delta}] \to A}\,(\wedge)$$

Starting with axioms, a type reduction is derived by forwarding premises to conclusions using one of these inference rules at each step. Here is a TLG proof of (3.38).

(3.A.1)

$$\cfrac{\cfrac{s_5 \to s_5 \qquad \cfrac{np_1 \to np_4 \qquad n_3 \to n_2}{np_1/_{\$}n_2\;\; n_3 \to np_4}\,(/L)}{np_1/_{\$}n_2\;\; n_3\;\; np_4\backslash s_5 \to s_5}\,(\backslash L) \qquad np_7 \to np_6}{np_1/_{\$}n_2\;\; n_3\;\; np_4\backslash s_5/np_6\;\; np_7 \to s_5}\,(/L) \qquad n_9 \to n_8$$

$$\overline{\begin{array}{cccccc} np_1/_{\$}n_2 & n_3 & np_4\backslash s_5/np_6 & np_7/_{\$}n_8 & n_9 & \to s_5 \\ \text{a} & \text{boy} & \text{walked} & \text{a} & \text{dog} & \end{array}}\,(/L)$$

It is easy to see that any non-atomic axiom is derivable from atomic ones as such. Thus we define the *atom matches produced by a TLG proof* as the set of atom pairs (order irrelevant) given by its axioms and consider two proofs *equivalent* if they produce the same matches. Clearly, then, (3.A.1) yields the same matches as in (3.42).

The following theorem establishes that if we cap the number of holes anywhere, we can compute all valid sets (if any) of atom matches that derive a type reduction.

**Theorem 3.A.1.** *The TLG is decidable if at most $m \geq 0$ holes are allowed in any sequent.*

*Proof.* Consider any sequent $\mathcal{N}_0 = \Gamma \to A$. If $\mathcal{N}_0$ has no (type) connective, then it is derivable if and only if it is an axiom.

Suppose $\mathcal{N}_0$ has connectives. Its finite proof trees (if any) can be built thus: we start with the set of partial proof trees of height 2 obtained by deriving $\mathcal{N}_0$ via some inference rule. Repeatedly, whenever a tree $\mathcal{T}$ of height $n$ has a connective-free terminal that is not an axiom, $\mathcal{T}$ is discarded; otherwise given a terminal $\mathcal{N}_i$ with connectives, we replace $\mathcal{T}$ with trees of height $n+1$ generated by deriving $\mathcal{N}_i$. This process continues until our working set becomes empty or all remaining trees have only axiom terminals. The number of partial proof trees of a finite height is obviously finite, so for the described process to terminate, it suffices to show that trees of an infinite height can be avoided. This in turn follows if we only allow *paths* from $\mathcal{N}_0$ terminating in finite steps.

Consider a path $\mathcal{N}_0, r_1, \mathcal{N}_1, r_2, \mathcal{N}_2, \ldots$ where $r_i$ derives $\mathcal{N}_{i-1}$ from $\mathcal{N}_i$. If $r_i$ is a logical rule, $\mathcal{N}_i$ has less connectives than $\mathcal{N}_{i-1}$. But if $r_i$ is a structural rule, $\mathcal{N}_{i-1}$ and $\mathcal{N}_i$ have an equal number of connectives. Given finitely many connectives in $\mathcal{N}_0$, the number of logical rules is bounded. The path will be finite if the number of structural rules is also be bounded. We show this is the case.

Let $r_{i_1}, r_{i_2}, \ldots$ $(i_1 < i_2 < \ldots)$ be the structural rules in the path. Since any sequent has at most $m$ holes, after at most $m$ consecutive rules of the same direction, the reverse must follow. Whenever a reversal from $r_{i_j} = (\vee)$ to $r_{i_{j+1}} = (\wedge)$ occurs, $r_{i_{j+1}}$ must target a hole in $\mathcal{N}_0$ or one created by some $r_{i_k} = (\vee)$ $(k \leq j)$. In the latter case, we can assume $i_j + 1 < i_{j+1}$ (since if $i_j + 1 = i_{j+1}$, the path can be contracted by removing $r_{i_k}, \mathcal{N}_{i_k}$ and $r_{i_j+1}, \mathcal{N}_{i_j+1}$), which means that $r_{i_j+1}$ is a logical rule.

But we know that the number of logical rules is bounded and $\mathcal{N}_0$ has finitely many holes. Therefore the sequence $r_{i_1}, r_{i_2}, \ldots$ of structural rules must terminate. □

In practice, it is much more convenient to use a fragment of the present TLG, where the following composites replace the structural rules and the logical rules involving continuation arrows (for the first column, cf. Barker and Shan, 2014, chap. 17).

**Definition 3.A.2.** *Fragment of TLG.*

$$\frac{\Phi\ A\ \Psi \to B}{\Phi \circ \Psi \to B\uparrow A}\ (\wedge\!\uparrow\!\mathrm{R}) \qquad \frac{\Phi \to A \quad B \to C}{\Phi \to B\uparrow A\downarrow C}\ (\uparrow\!\mathrm{L}\!\downarrow\!\mathrm{R}) \qquad \frac{A \to C \quad D \to B}{C\downarrow D \to A\downarrow B}\ (\downarrow\!\mathrm{L}\!\downarrow\!\mathrm{R})$$

$$\frac{\Phi \circ \Psi \to A \quad \Gamma[B] \to C}{\Gamma[\Phi\ A\downarrow B\ \Psi] \to C}\ (\downarrow\!\mathrm{L}\vee) \qquad \frac{\Phi \to A \quad B \to C}{\Phi \to C\uparrow(A\downarrow B)}\ (\downarrow\!\mathrm{L}\!\uparrow\!\mathrm{R}) \qquad \frac{A \to C \quad D \to B}{D\uparrow C \to B\uparrow A}\ (\uparrow\!\mathrm{L}\!\uparrow\!\mathrm{R})$$

It is with this fragment that we illustrate the role of holes in the TLG. In the relative clause taken from (3.40), we replace the silent token (3.34) with a hole.

(3.A.2)

$$\dfrac{\dfrac{\dfrac{np_6 \to np_6 \quad np_7 \to np_7}{np_6\ np_6\backslash_{\$}np_7 \to np_7}\,(\backslash\mathrm{L})}{np_6\backslash_{\$}np_7 \to np_6\backslash_{\$}np_7}\,(\backslash\mathrm{R}) \qquad \dfrac{\begin{array}{c} s_{15} \to s_8 \quad np_9 \to np_{14} \quad np_{17} \to s_{16} \quad n_{19} \to n_{18} \\ \vdots \qquad\quad \vdots \qquad\quad \vdots \qquad\quad \vdots \\ np_9\ np_{14}\backslash s_{15}/s_{16}\ np_{17}/_{\$}n_{18}\ n_{19} \to s_8 \end{array}}{\circ\ np_{14}\backslash s_{15}/s_{16}\ np_{17}/_{\$}n_{18}\ n_{19} \to s_8\uparrow np_9}\,(\wedge\!\uparrow\!\mathrm{R})}{\underset{\text{who}}{np_6\backslash_{\$}np_7/(s_8\uparrow np_9)}\ \ \underset{__}{\circ}\ \ \underset{\text{walked}}{np_{14}\backslash s_{15}/s_{16}}\ \ \underset{\text{a}}{np_{17}/_{\$}n_{18}}\ \ \underset{\text{dog}}{n_{19}} \to np_6\backslash_{\$}np_7}\,(/\mathrm{L})$$

The hole serves not so much as a typed token as a "punctuation"; it makes virtually no semantic contribution since there is no atom to correspond to graph vertices. But compare the produce of (3.A.2) and the matches found in (3.44) (which can be reproduced using the TLG):

(3.A.3) a. $np_6\backslash_\$ np_7/(s_8\uparrow np_9),\ \dfrac{s^0_{12}\uparrow np^1_{13} \mid_\$ s^0_{10}}{np^1_{11}},\ np_{14}\backslash s_{15}/np_{16},\ np_{17}/_\$ n_{18},\ n_{19} \to np_6\backslash_\$ np_7$

b. $(s_8, s^0_{12}), (s^0_{10}, s_{15}), (np_9, np^1_{13}), (np^1_{11}, np_{14}), (np_{16}, np_{17}), (n_{18}, n_{19})$

With $s^0_{12}$ coinciding with $s^0_{10}$ and $np^1_{13}$ with $np^1_{11}$, (3.A.3b) eventually equates $s_8$ with $s_{15}$ and $np_9$ with $np_{14}$. Thus holes and typed gap tokens are equivalent for the purpose of deciding fusing equivalence.

Holes can be justified to replace types of the following general form. From ($\wedge\uparrow$R) it immediately follows that

$$\circ \to A \uparrow A$$

holds for any $A$. The so-called *cut theorem* then allows us to substitute $\Delta$ into $\Gamma[A] \to B$ whenever $\Delta \to A$. Given its special status in TLGs and more generally in logical calculi, we show this theorem holds for Definition 3.A.1, simplifying Lambek's (1958) presentation somewhat.

**Lemma 3.A.1.** *(Normalization) For any proof of* $\Gamma \to A$ *where* $A$ *is non-atomic, an equivalent proof that ends with some R-rule exists.*

*Proof.* By induction on the length of derivation. □

**Theorem 3.A.2.** *(Cut)* $\dfrac{\Delta \to A \quad \Gamma[A] \to B}{\Gamma[\Delta] \to B}$

*Proof.* We proceed by induction on the degree of cuts, defined as follows:

> the *degree* $\#(\cdot)$ *of a term, structure, sequent* is the number of type connectives it contains; the *degree of a cut* as above is $\#(\Delta) + \#(\Gamma[A] \to B)$.

Consider the proof tree of $\Gamma[A] \to B$. If $A$ is atomic, we can replace the axiom $A \to A$ with the proof tree of $\Delta \to A$ and continue the derivation as before.

But if $A$ is non-atomic, we can find the step at which its main connective is introduced by some L-rule. With $A = C\backslash D$ (the cases of other connectives are similar), we have

$$\frac{\Phi \to C \quad \Pi[D] \to E}{\Pi[\Phi\, C\backslash D] \to E}\ (\backslash\mathrm{L})$$

By the normalization lemma, $\Delta \to C\backslash D$ is derivable from $C\Delta \to D$. Thus the induction hypothesis implies

$$\frac{\Phi \to C \qquad \dfrac{C\Delta \to D \quad \Pi[D] \to E}{\Pi[C\Delta] \to E}\ (\mathrm{Cut})}{\Pi[\Phi\Delta] \to E}\ (\mathrm{Cut})$$

It can be shown that both cuts are of a smaller degree than the original one. Henceforth we continue the derivation as before. □

# Chapter 4

# Topics in Plurality

We are in a position to consider the empirical applications of the graph formalism developed so far. In this chapter, we discuss a few selected topics in plurality, with a focus on various ways of distributing predication over a plurality and creating a plurality with conjunction.

As we will see, distributivity in various forms can be expressed in terms of subtle variants of quantification structures. The fact that they all share more or less the same syntactic type attests to the convenience of having partial determinism in atom-vertex correspondence. Our syntax-semantics interface allows us to easily capture the long-existing intuition that conjunctions across categories all create a plurality of some kind, while offering new perspectives on some well-known compositional challenges.

## 4.1 Distributivity

We have seen in Section 2.2.3 that plural predication relates events and participants in plurality in an underspecified way (Scha, 1981). Thus the following example describes a variety situations, where a number of adoptions come down to five boys being the agents and six dogs the themes; see (2.37).

(4.1) Five boys adopted six dogs.

This is the so-called cumulative reading of (4.1), of which the collective reading (where there is only one adoption) is a special case.

On the other hand, the distributive readings distribute predication over a plurality (see Nouwen, 2016; Winter and Scha, 2015 for a review). For (4.1), this means attributing a property of "adopting six dogs" to each boy, or a property of "being adopted by five boys" to each dog:

(4.2) a. Five boys each adopted six dogs. (≥ 5 adoptions)
b. Six dogs were each adopted by five boys. (6 adoptions)

Although for sentences like (4.1), distributive predication over a plural subject as in (4.2a) is traditionally considered a more accessible reading than that over a plural object as in (4.2b) (see Reinhart, 2006, sec. 2.7.3), an experimental study by Križ and Maldonado (2018) nonetheless reports the marginal availability of the latter. Since the adopted dogs may vary across the boys (resp. the adopting boys

may vary across the dogs), we can be sure that distributive readings are distinct from cumulative ones.

Treating numerals like (4.3) as indefinite articles specified for cardinality, we can easily construct semgraphs for plural predication or cumulativity, in a manner similar to (3.38) and (3.42).

(4.3) five → $np^0/_{\$}n^0$

a. 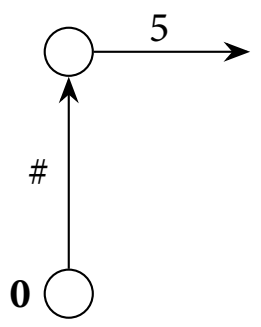


b. 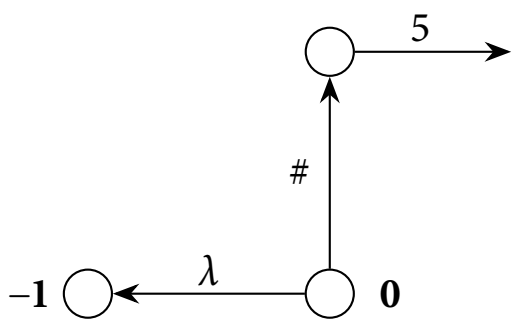


The question we deal with in this section is how distributive predication should be represented and constructed.

### 4.1.1 Via quantification

The characterization of distributivity makes clear its iterative or quantificational nature. To implement distributivity, we can naturally introduce a quantification structure with a silent linguistic token, called the *distributor* or DIST, that takes a plural nominal to specify the restrictor. This is essentially the idea proposed in Kroch (1974); Link (1983).

Starting with the distributivity over subjects, we give the following the lexical entry for DIST:

(4.4) DIST → $np^1\backslash(s^0/(np^3\backslash s^2))$

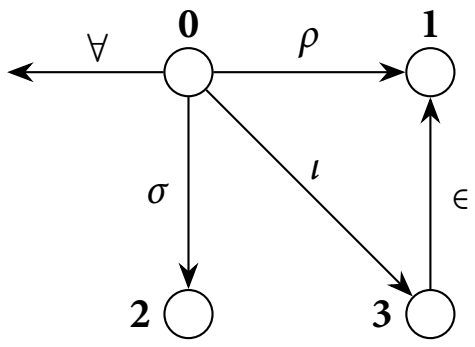


The similarity between (4.4) and the lexical entry (3.27b) for *every* is obvious. In fact, $s/(np\backslash s)$ is a concatenation analogy of $s{\uparrow}np{\downarrow}s$ commonly used to type subject quantifiers in CCGs (e.g. Steedman, 2011, p. 110).

How DIST works can be illustrated with the following reduction.

(4.5) five boys — DIST — adopted six dogs

$np^0_1$ — $np^1_2\backslash(s^0_3/(np^3_4\backslash s^2_5))$ — $np^1_6\backslash s^0_7$ → $s^0_3$

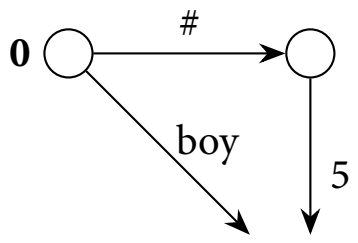


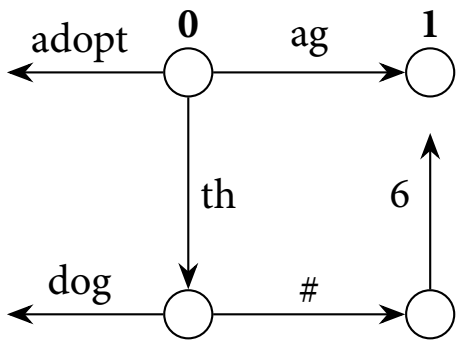


According to (4.4) DIST first combines with a subject and then with a predicate, and the switched application rule (see Definition 3.1) allows that order to be reversed. In either case, we find the same atom matches in (4.6a) and construct (4.6b) as desired.

(4.6) a. $(np_1^0, np_2^1), (np_4^3, np_6^1), (s_5^2, s_7^0)$

b. 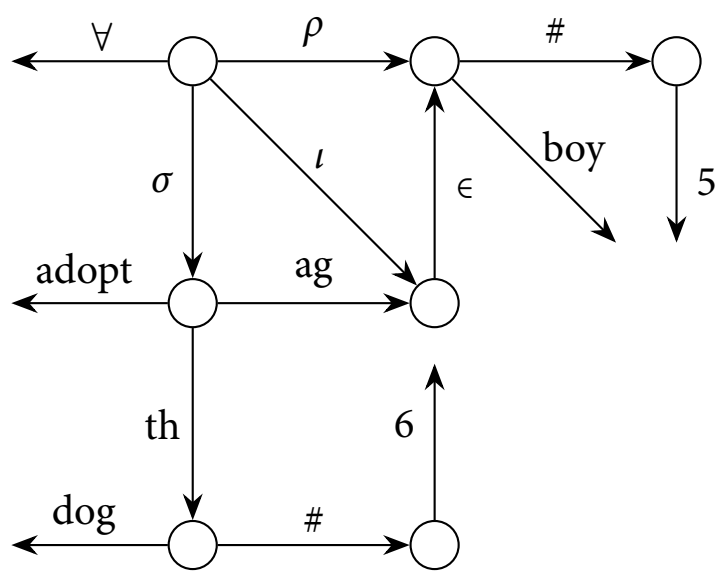


It is convenient for DIST to be able to associate to either side, especially when the syntactic or semantic context seems to favor one way over the other. For example, in (4.7a), DIST sits within one verbal conjunct; we do not want to broadcast the effect of distributivity to the other as in (4.7b), assuming it makes little sense for each single boy to meet.

(4.7) Five boys met and adopted six dogs.

a. Five boys [met and DIST adopted six dogs].
b. *Five boys DIST [met and adopted six dogs].

Similarly, DIST should be confined to one nominal conjunct as in (4.8a), if we are to derive the reading of (4.8) where each dog hugged each boy (see Section 4.2.3.3).

(4.8) Six dogs and every boy hugged.

a. [Six dogs DIST and every boy] hugged.
b. *[Six dogs and every boy] DIST hugged.

In the literature examples of verbal conjunction like (4.7) are often cited to locate DIST exclusively in predicates (a.o. Dowty, 1987; Lasersohn, 1995, chap. 7). Less frequently are examples of nominal conjunction like (4.8) mentioned to motivate the need for a nominal DIST in addition (Winter, 2001, sec. 6.3). Yet it seems equally reasonable to take both as arguing for DIST's versatility, made possible by our unification-based syntax-semantics interface (in the functional paradigm, a nominal decorator has to be of a different semantic type and thus different semantics from a verbal decorator).

Now, to also take distributivity over objects into account, we can continuize the quantifier part of the type of DIST, while leaving the semantics unchanged (it is easy to see that the type in (4.4) would not work here):

(4.9) DIST $\rightarrow np^1\backslash\dfrac{s^0 \mid s^2}{np^3}$

We omit illustrating how this can be used to derive both readings in (4.2) for (4.1), but notice that it also handles distributive predication in ditransitive sentences as such:

(4.10) Five boys gave six dogs DIST a treat.

gave: $np^3\backslash s^0/np^2/np^1$

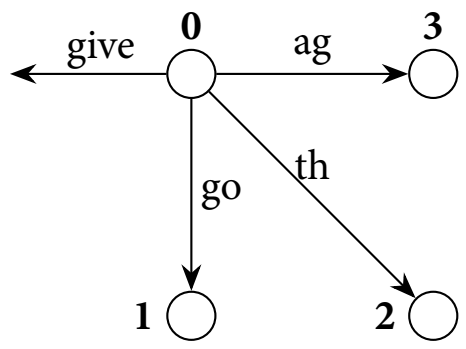

A complication of (4.9), though, is that the CCG in Definition 3.1 does not let it combine with a predicate. To obtain (4.11b) from (4.11a), one may want to employ a "lifted" function composition rule, or consider a continuized TLG instead (see Appendix 3.A).

(4.11) a. $np_1\backslash\frac{s_4 \mid s_2}{np_3},\ np_5\backslash s_6 \rightarrow np_1\backslash s_4$

b. $(s_2, s_6), (np_3, np_5)$

The distributivity in its present form is said to be *atomic* in that the underlying quantification iterates through singletons. There are three kinds of examples sometimes discussed in support of a non-atomic distributivity (a.o. Gillon, 1987; Schwarzschild, 1996). (4.12) gives one of each kind followed by a situation it is meant to describe.

(4.12) a. These eggs cost \$4. (Nouwen, 2016, p. 276)
(*Each dozen costs this much.*)

b. Joe, Ben, and Ed collaborated.
(*Joe and Ben collaborated, so did Ben and Ed.*)

c. Two dogs and two cats met.
(*Two dogs met and two cats met.*)

Hence the argument: quantification over something larger than singletons is needed so as to assert a price of each twelve instead of one in (4.12a), or two collaboration of pairs instead of one of a triple in (4.12b), or two meetings of likes instead one of four animals in (4.12c).

However, each of these cases can be rethought differently. Examples like (4.12a) might illustrate metonymy, Winter and Scha (2015) suspect, for the names of the entities grouped by convention can apply loosely to such groups (a tag that reads "eggs \$4 ea" might as well price each carton). We have already seen that examples like (4.12b) can be interpreted as plural predication. As we will see in Section 4.2.3, examples like (4.12c) can result from distributive coordination. For further discussion, see Winter and Scha (2015) and references therein.

### 4.1.2 Non-scoping distributivity

The distributivity above noticeably takes under its scope the referents inside the material being distributed. Usually this may be what we want, but we also find "distributivity" that appears to have this effect of scoping waived.

Here is an example of the kind discussed by Roberts (1987); Schein (1993):

(4.13) Two boys gave ten dogs a treat.

One reading of (4.13) requires giving of ten treats, one per dog, by two boys. This reading differs from the usual distributive ones by limiting the agents to two boys fixed across the dogs. It is not exactly cumulative either, since each dog got its own treat. Such readings, sometimes described as a mix of distributivity and cumulativity, are better seen with ditransitives (just as in Section 4.1, transitives, not intransitives, distinguish distributivity from cumulativity), but it is both reasonable and helpful to extend our discussion to transitives. That means a reading for (4.14) that requires ten adoptions, one per dog, by two boys:

(4.14) Two boys adopted ten dogs.

To make a similar point, following Schein and Kratzer (2002, chap. 2) we can use explicit quantifiers:

(4.15) Two boys adopted every dog.

The relevant reading of (4.15) not only scopes *two boys*, but allows each dog to be adopted by one of them, as far as both get involved eventually.

We can capture the non-scoping distributivity (NSD) in (4.14) and (4.15) with a semantically more sophisticated distributor or quantificational determiner, whose syntax remains as before:

(4.16) $\text{DIST} \to np^1\backslash\dfrac{s^0 \mid s^2}{np^3}$ $\qquad$ every $\to \dfrac{s^0 \mid s^2}{np^3}/n^1$

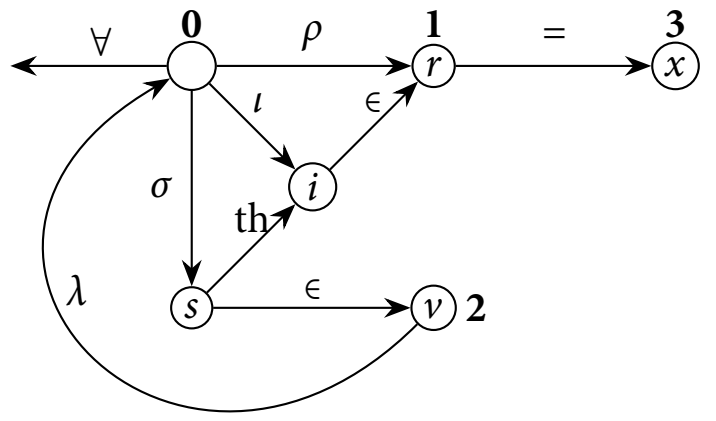


A quantification structure is still recognizable; besides, (4.16) adds $v$ and $x$ that assume the roles of the 2-source and 3-source previously served by $s$ and $i$ (we will address $\overrightarrow{si}$ soon). By making $v$ the root, the built-in $\lambda$-cycle accounts for the non-scoping effect.

Let us clarify how this is so with (4.15), whose semgraph would be as follows:

(4.17)

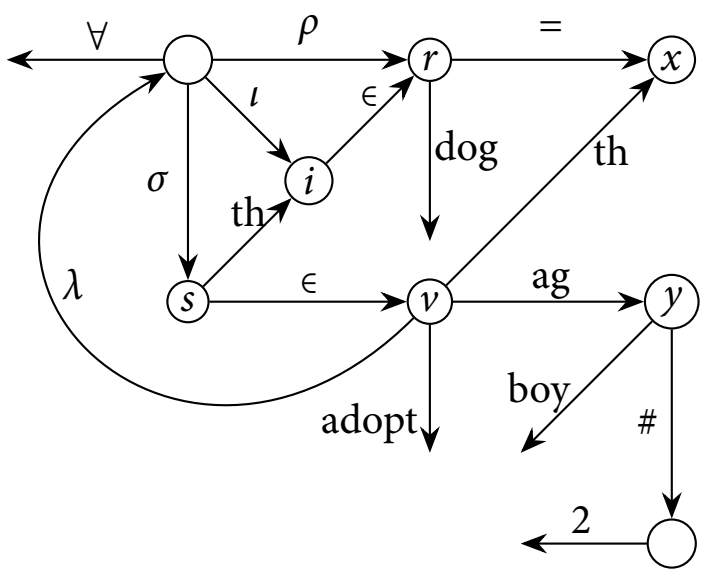


The cumulativity comes from the fact that $v, y, x, r$ are valuated simultaneously: $v$ contains a number of adoptions whose agents sum up to a set $y$ of two boys, whose themes to $x$, while $r = x$ is maximized to the set of all dogs. When these constraints are satisfied, the quantificational structure yields the distributivity: each $i \in r$ is the theme of some $s \in v$ (here, too, $\in$ abbreviates "being a singleton subset of").

From the the label of $\overrightarrow{si}$ one can see that (4.16) specializes in the NSD over themes, but it should be generalized to other thematic roles, like goals:

(4.18) Two boys taught every dog a trick.

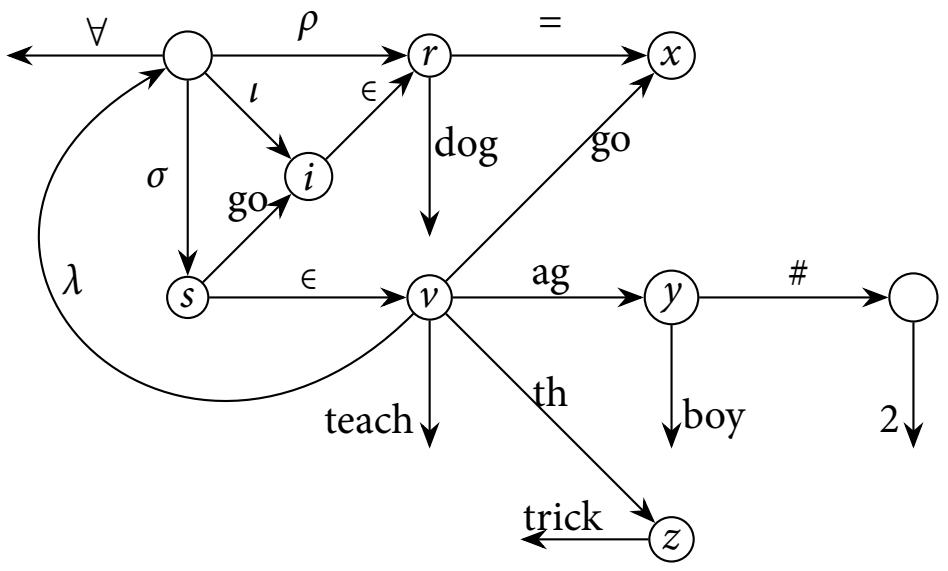


Thus a better solution is for $\overrightarrow{si}$ to "copy" the label from $\overrightarrow{vx}$ — a parallelism characteristic of the NSD sentences. What might be the most natural implementation of "edge-copying", however, would require our graph formalism be extended with *hyperedge replacement* (see Courcelle, 1993; Drewes et al., 1997), which we will not discuss in this thesis.

Note that *a trick* is not distributed over *every dog* in (4.18), since $z$ is valued along with $x$ (all dogs) and $y$ (two boys). While this is a plausible reading of (4.18), it differs from the reading of (4.13) mentioned above, which should rather be rendered as follows:

(4.19)

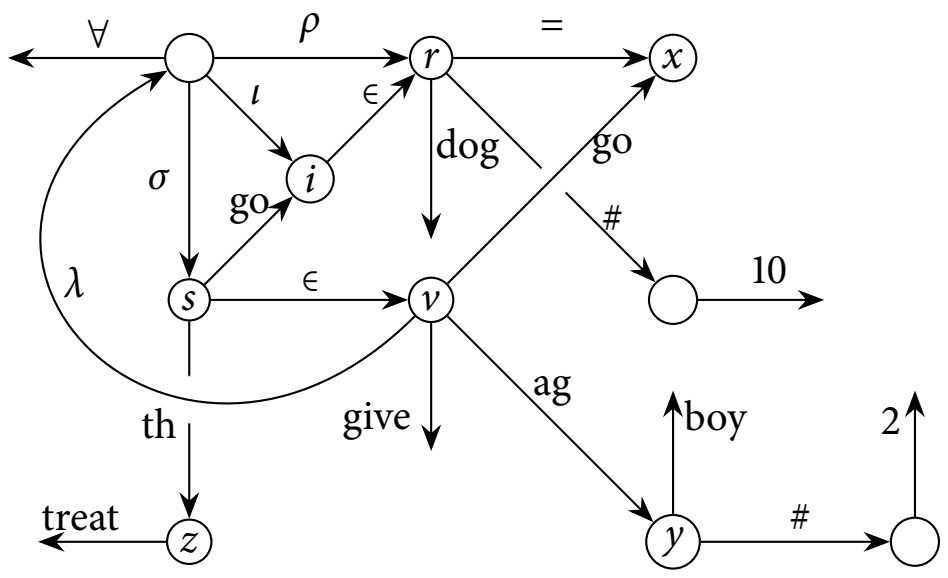


Here $v$ contains a number of givings by $y$ (two boys) to $x = r$ (ten dogs). Since the *th*-dependent (a treat) docks at $s$ instead of $v$, each dog $i \in r$ receives some $z$ (a treat) through some $s \in v$. However, (4.19) breaks the entry for ditransitive verbs given in (4.10). To compose (4.19), one might need a special syntax for introducing thematic roles (see Champollion, 2017, sec. 9.5).

Actually, (4.13) makes sense only if *a treat* is distributed over *every dog*, assuming a treat can be given and consumed only once, whereas in (4.18), one trick can be taught many times to different dogs. This contrast illustrates an inference associated with the NSD: when there are more events than there are non-distributed entities, some of them must participate in multiple events — we just rephrased the famous *pigeonhole principle* (originally stated as packing less holes with more pigeons). This inference can be explicit, as in (4.14) (at least one boy adopted more than one dog), and even if that is not the case, its possibility cannot be denied, as in (4.15) (one boy should be able to adopt more than one dog). Now the question is, whether this multi-participation fits the lexical semantics of the events and entities involved.

One can therefore reason that the NSD over subjects is difficult in (4.20) but obtainable in (4.21).

(4.20) Every copy editor caught 500 mistakes in the manuscript. (Kratzer, 2002, chap. 2)

(4.21) a. [In 1916,] 17,667,827 Americans voted for two leading presidential candidates. (Jim Hargrove, *The Story of Presidential Elections*, p. 17)

b. Yet not everyone made these mistakes.
(Nancy Langston, *Where Land and Water Meet*, p. 10)

The sheer number of mistakes in (4.20) suggests a division of labor, where the editors worked only on their shares, and no mistake could be caught twice. On the other hand, it is only natural for a candidate to receive many votes, or a mistake to be made by many people.

Based on examples like (4.20) alone, there have been generalizations that the NSD with non-distributed themes is impossible (e.g. Kratzer, 2002) or the NSD over subjects is so (e.g. Champollion, 2017), and analyses laid accordingly. Our discussion suggests those examples show more of a problem of clashing inferences than one of grammaticality.

### 4.1.3 Distributivity undone

Downward monotone quantifiers that set a upper bound on counting introduce genuine quantification (see Section 2.1.5), and are thus associated with distributivity automatically. The following example derives from a lexical entry of *less-than-three* almost identical with that of *every*.

(4.22) Less-than-three dogs swam.

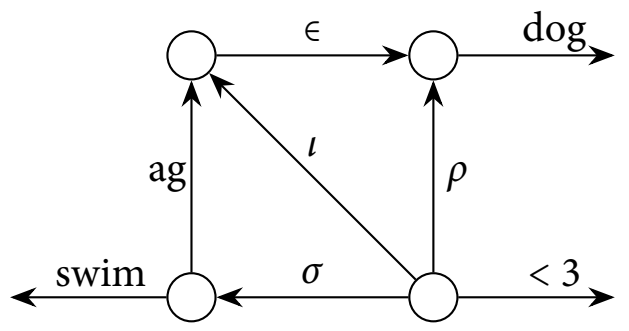


But with plural predication as in (4.23), such quantifiers are known to be able to undo that distributivity (see Winter and Scha, 2015, sec. 5 for a review):

(4.23) Less-than-three boys met.

For (4.23) a semgraph like (4.22) would not represent what it seems to convey: the negation of all statements of the form "$n$ boys met" with $n \geq 3$ made via plural predication. Put differently, (4.23) requires all meetings of boys involve less than three boys in total.

One way to capture this cumulativity with an upper bound is through another variant (4.24a) of quantification structures, which exploits the maximality requirement associated with the restrictor by placing the 2-source and 3-source in the restrictor subgraph:

(4.24) a. less-than-three $\rightarrow \dfrac{s^0 \mid s^2}{np^3}/n^1$ b. meet $\rightarrow np^1\backslash s^0$

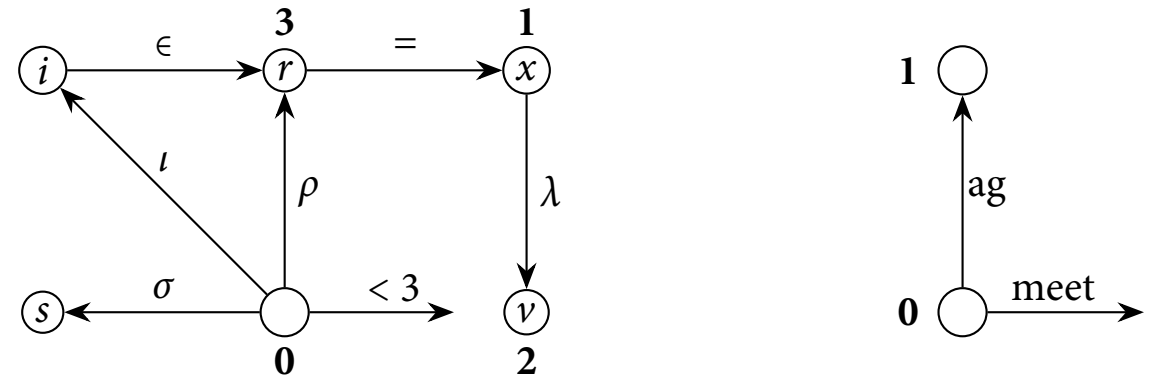


This along with (4.24b) allows us to construct (4.25) for (4.23).

(4.25)

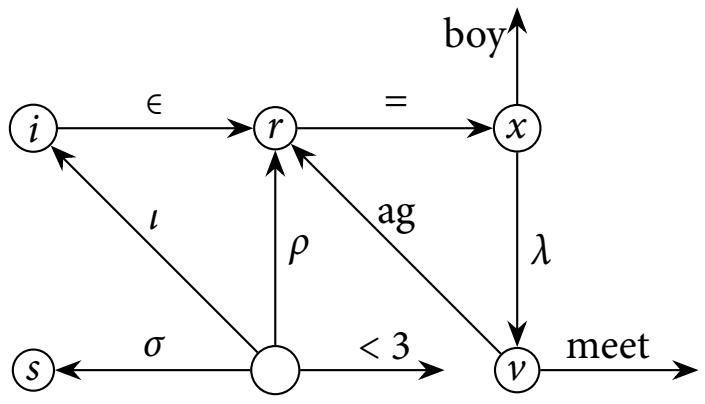


To verify (4.25), $r$ is maximized to the union of all sets of boys that met. The quantificational structure is then satisfied if and only if that union contains less than three elements, as the scope subgraph is always satisfiable for being unconstrained.

The entry (4.24a) has a problem with transitive sentences containing two such quantifiers, neither of which takes scope over the other:

(4.26) Less-than-three boys adopted less-than-three dogs.

The reading under consideration, sometimes known as the *branching reading* (see Barwise, 1979; van Benthem, 1991; Keenan, 1992; Sher, 1990; Westerståhl, 1987), requires all adoptions of dogs by boys jointly involve less than three of either. But (4.24a) leads to (4.27), which subjects the maximization of $r_1$ to the cardinality constraint on $r_2$.

(4.27)

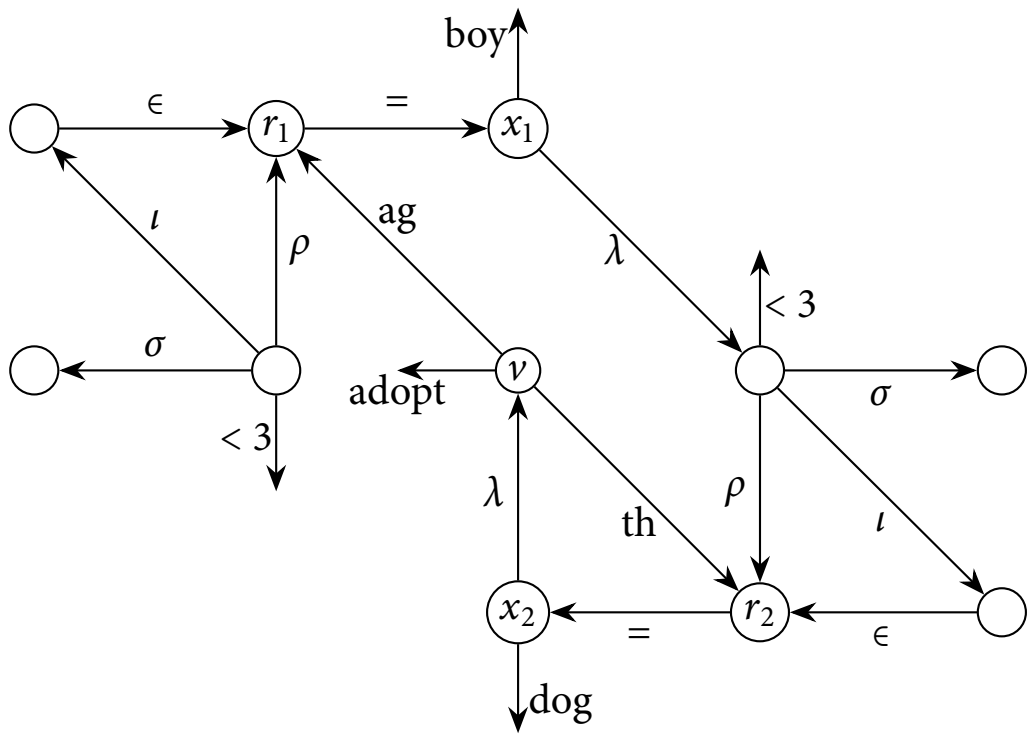


As a consequence, the semgraph can be satisfied with $r_1 = \varnothing$ when any nonempty value of $r_1$ would falsify the restrictor subgraph off $r_1$ (see Definitions 2.4, 2.5/2.6), so it incorrectly holds in a situation, say, where a number of boys each walked three dogs.

Depending on the representation formalism one works with, composing from syntax the branching reading of (4.26) can be a challenge. Since the reading is irreducible to quantifier nesting (which would scope one over another), quantification over tuples (see Westerståhl, 2015, sec. 11 for a review) and dynamic-semantic accounts (Brasoveanu, 2013; Charlow, 2018) have been proposed in the literature. However, the way we align syntactic and semantic resources allows us to approximate a different solution with some tweaking of (4.24).

In (4.28) we only change the tone of the downward arrow, so $s_1$ and $s_3$ correspond to the same vertex:

(4.28) less-than-three → $s^0_1{\uparrow}np^2_2{\downarrow_{\$}}s^0_3/n^1_4$

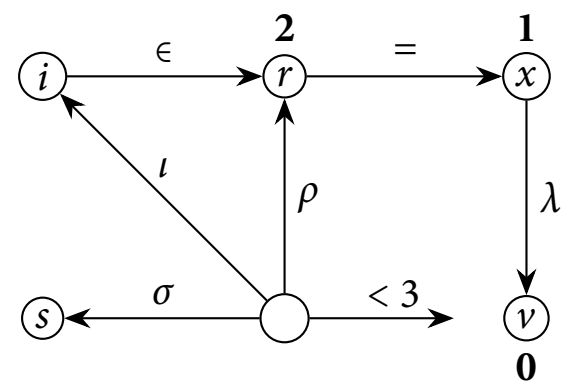


Using this entry we can then construct the following semgraph for (4.26), which is almost what we want, except that it is not yet interpretable due to multi-rootedness.

(4.29)

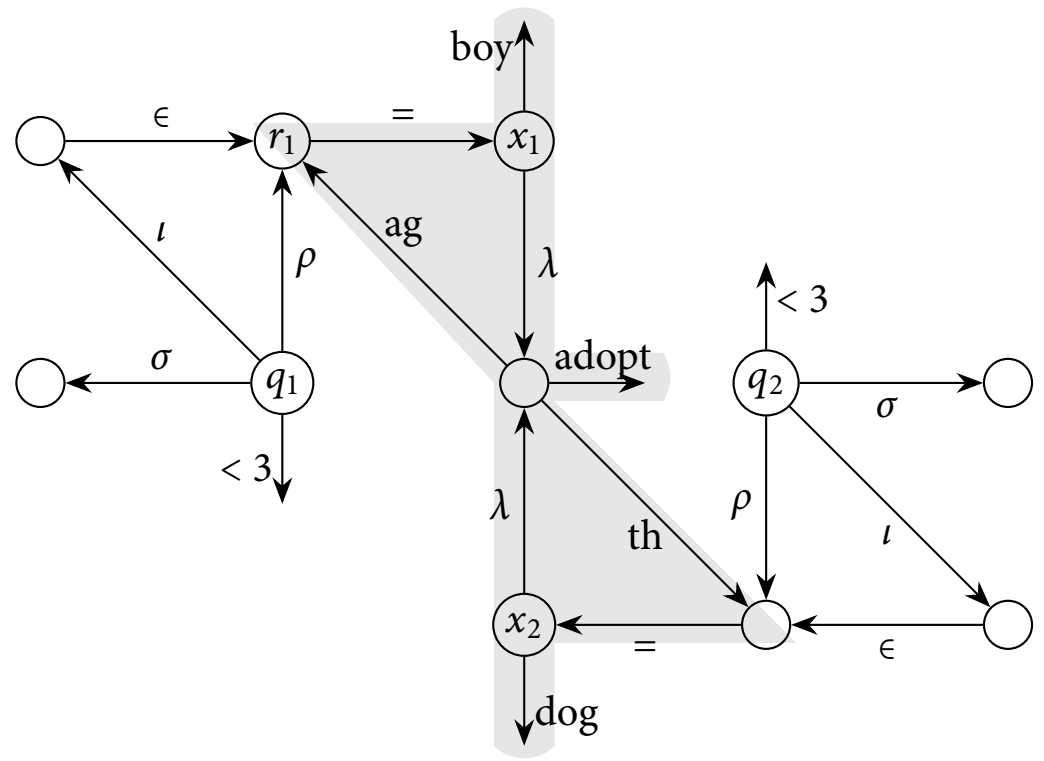


In (4.29) both $q_1$ and $q_2$ are roots, and in the restrictor subgraphs of $q_1$ and $q_2$ (the shaded area), both $x_1$ and $x_2$ are roots. That said, if (4.29) is evaluated with entrance through $q_1$ and then, independently, with entrance through $q_2$, each time taking, say, $x_1$ as the root of the respective restrictor subgraph, we obtain both branches of the required truth condition: less than three boys, and then less than three dogs are involved in all adoptions of dogs by boys.

To restore the unique rootedness of (4.29) and ensure the interpretation described above, one can add $\supset \overrightarrow{zq_1}, \supset \overrightarrow{zq_2}$ to join $q_1, q_2$ in a conjunction structure and break the tie between $x_1, x_2$ by adding $\lambda\overrightarrow{x_1x_2}$. The question then is how these can be done in a principled way. We will not dwell here, but our discussion has shown that a major challenge of branching readings can be reformulated as dealing with multi-rootedness.

## 4.2 Conjunction

Cross-linguistically conjunction might be *the* most available means for creating plurality. While treating it as set union may be a simple idea (see Section 2.1.7), many research efforts have gone into composing desired semantic representations for conjunction across categories. Using the coordinator entries introduced earlier (Section 3.3.3.3), in this section we will discuss examples in both nominal and verbal domains.

### 4.2.1 Simple union

Recall that for conjuncts of atomic types, or more generally types of size one, the conjunction token consists of only a triple for union. Primary examples of this kind are *np*- and *s*-conjunctions.

(4.30) and → $p^2\backslash_{\&}p^0/_{\&}p^1$

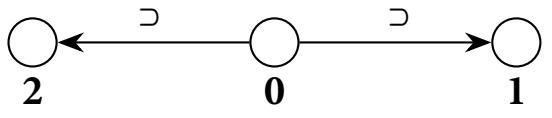


With (4.30) it is fairly easy to conjoin simple noun phrases and sentences in the following examples.

(4.31) a. Joe and Ben

$np_1^0$ $\quad np_2^2\backslash_{\&}np_3^0/_{\&}np_4^1$ $\quad np_5^0$ $\quad \rightarrow np_3^0$

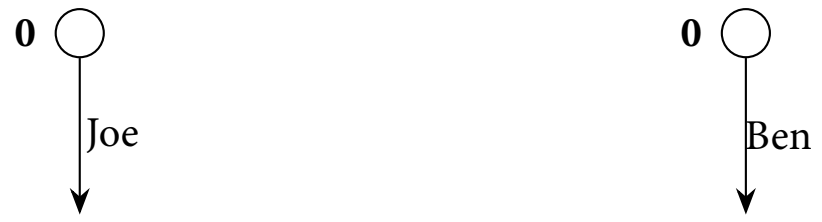


b. Joe sailed and Ben surfed

$s_1^0$ $\quad s_2^2\backslash_{\&}s_3^0/_{\&}s_4^1$ $\quad s_5^0$ $\quad \rightarrow s_3^0$

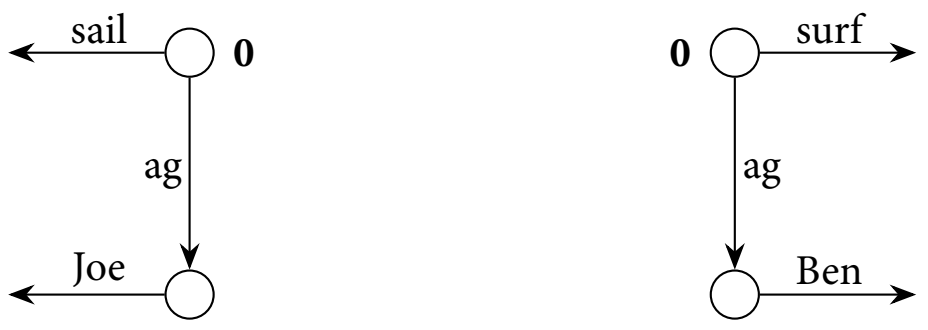


Matching the atoms by $(np_1^0, np_2^2), (np_4^1, np_5^0)$ in (4.31a) and by $(s_1^0, s_2^2), (s_4^1, s_5^0)$ in (4.31b) leads to these:

(4.32) a. Joe and Ben. b. Joe sailed and Ben surfed.

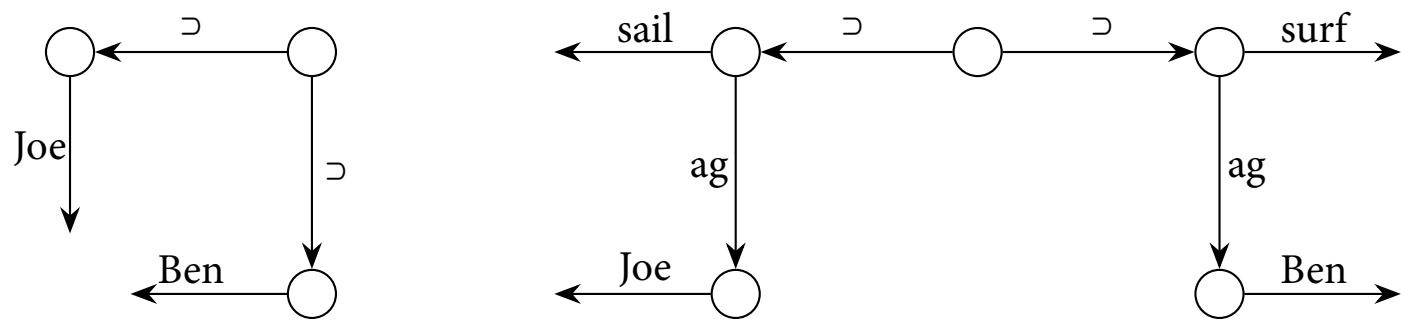


*Remark.* Giving copulas a non-thematic semantics with the following entry,

(4.33) be → $np^2\backslash s^0/np^1$

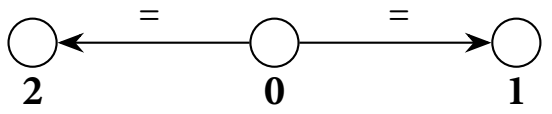


we can maintain that conjoining predicate nominals creates a plurality nonetheless:

(4.34) a. Joe is a sailor and a surfer. b. Joe and Ben are a sailor and a surfer.

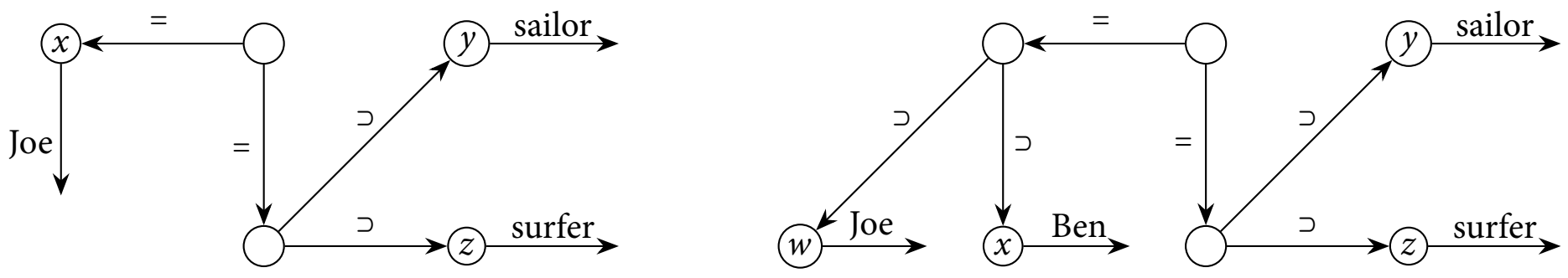


Suffice it to note that (4.34a) holds exactly when $x = y = z$ contains Joe the sailor and surfer. (4.34b) holds exactly when the set $w \cup x$ containing Joe and Ben coincides with a set $y \cup z$ containing a sailor and surfer (the sentence can be followed by *I don't know who is which*).

A nominal plurality created this way functions just like those introduced by numerals. Thus to (4.32) both plural and distributive predications apply, as shown by the following examples, where DIST in (4.35b) is the atomic distributor from Section 4.1.1.

(4.35) a. Joe and Ben sailed.

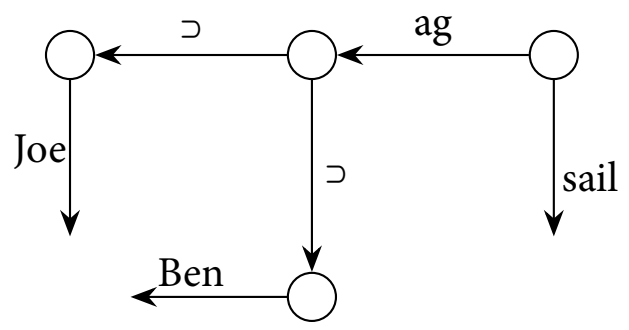


b. [Joe and Ben] DIST walked a dog.

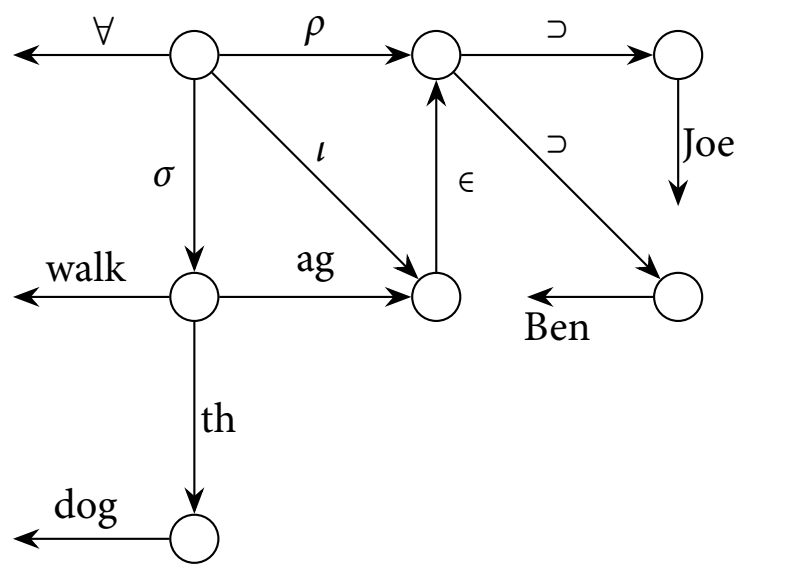


When conjuncts are themselves plural, however, we observe both the atomic distributivity over their union, as in (4.36a), and the non-atomic distributivity over such conjuncts, as in (4.36b). It is clear that the distributor above works for the former but not the latter.

(4.36) a. Two dogs and two cats ate (= each of the four animals ate).
b. Two dogs and two cats met (= two dogs met *and* two cats met).

The intended reading of (4.36b) as two meetings of likes certainly falls into the situations described by plural predication (as does one meeting of four animals), but it also seems a special case that merits a separate treatment, since distributive coordination is seen in examples that has nothing to do with plural predication:

(4.37) Every dog and no cat swam (= every dog swam *and* no cat swam).

We will discuss such examples in Section 4.2.3, where (4.36) is likened to (4.37) in a sense.

Not all seemingly distributive predication needs decoration. With one conjunct being a disjunction, the paraphrased distributivity in (4.38) follows simply from combining the semantics of union and choice. Since the subgraph reachable from $x$ is satisfiable with $x$ being (a singleton of) either Ben or Ed, (4.38) accepts either a meeting of Joe and Ben or one of Joe and Ed.

(4.38) Joe and [Ben or Ed] met (= Joe and Ben met *or* Joe and Ed met).

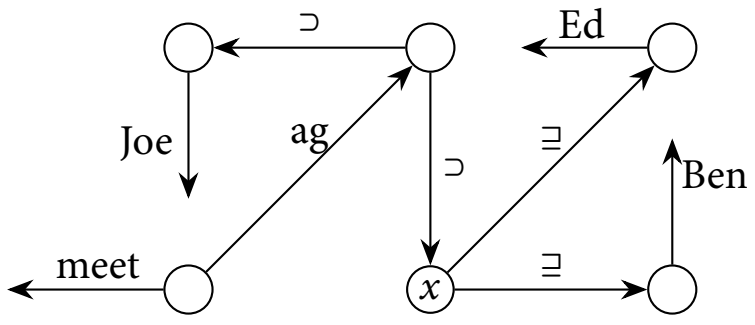


The effect is similar to the distributivity of conjunction over disjunction in Boolean algebra, that is, $p \wedge (q \vee r) = (p \wedge q) \vee (p \wedge r)$. This is how *or* takes scope over *and* in examples like (4.38), though not literally.

### 4.2.2 Argument sharing

The idea of sharing unfilled arguments underlies *generalized conjunction* in Partee and Rooth (1983), according to which the conjunction $h$ of Boolean functions $f$ and $g$ is defined as $h(x_1, ..., x_n) = f(x_1, ..., x_n) \wedge g(x_1, ..., x_n)$, where $f$ and $g$ share the arguments passed to $h$.

Likewise, for conjunct types of size greater than one, the conjunction token includes triples for argument sharing via equality (Section 3.3.3.3). Such is the case of verbal conjunction beneath the sentence level, where earlier works on non-Boolean conjunction often skip the question of compositionality, particularly how arguments are bound to the thematic roles of events (e.g. Lasersohn, 1995; Schein, 1993; but see Chaves, 2007 for a treatment using a constraint-satisfaction grammar).

In Section 3.3.4, we showed in detail how examples like (4.39b) can be constructed with the entry (4.39a).

(4.39) a. and → $(s^2/np^4)\backslash_{\&}(s^0/np^5)/_{\&}(s^1/np^3)$ b. Joe rented and Ben sank a boat.

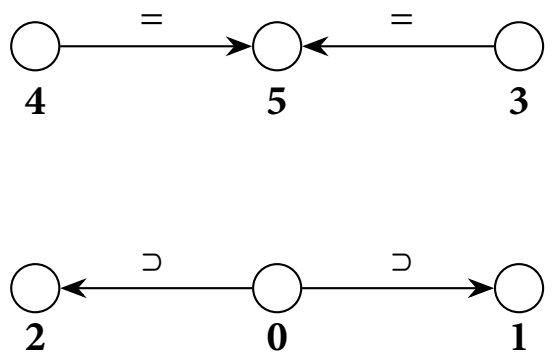


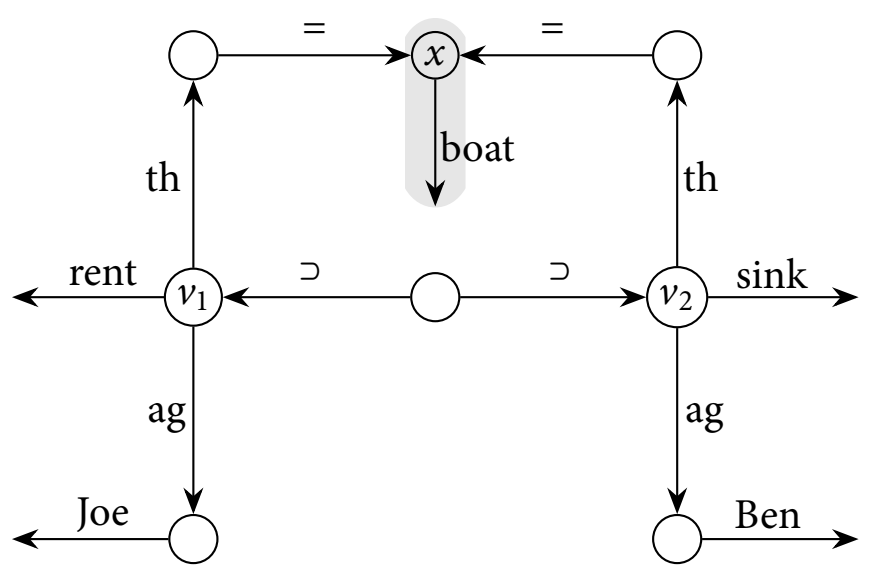


While the shaded area is shared between the conjunct subgraphs off $v_1$ and $v_2$, by evaluating them independently the value of $x$, a boat, need not be shared (see Section 2.2.5). Thus (4.39b) can be paraphrased with *a boat* distributed as follows.

(4.40) Joe rented a boat *and* Ben sank a boat.

This is what we mean by *distributive coordination*, a consequence of argument sharing and the way we interpret the conjunction structure. It does not follow from *conjunction reduction*, a syntactic analysis that derives non-sentential conjunctions from sentential ones (Lakoff and Stanley, 1969), nor is it a result of scoping *and* in any sense.

We may rather say that $x$ is not taking scope in (4.39b), but this could be a problem for indefinites in subject position. Changing the type in (4.39a) to (4.41a), one can easily construct (4.41b), which represents a reading often considered near impossible in the literature (e.g. Moltmann, 1994, p. 113).

(4.41) a. and → $(np^4\backslash s^2)\backslash_{\&}(np^5\backslash s^0)/_{\&}(np^3\backslash s^1)$

b. A boy surfed and sailed (= a boy surfed *and* a boy sailed).

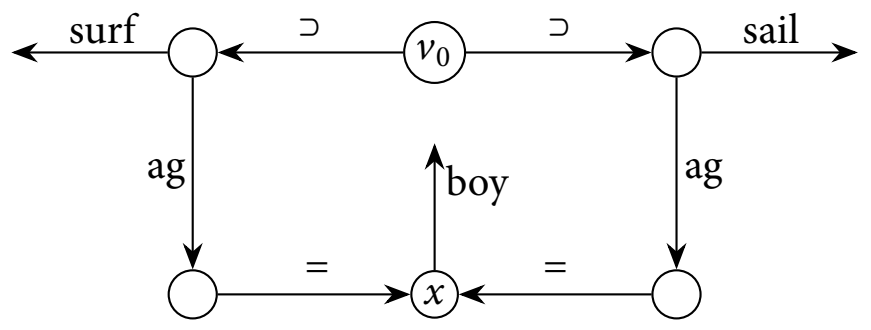


It is not clear whether this judgment is entirely grammatical in nature (see Kubota and Levine, 2020, pp. 79ff), since a narrow-scope-indefinite reading can be explicitly coerced (Simon Charlow p.c.):

(4.42) A different boy surfed and sailed.

Thus, if construction of (4.41b) is not itself undesirable, we can ask why it is almost obligatory for *a boy* to take wide scope, while keeping in mind that this amounts to its precedence over conjunction by $\lambda x \overrightarrow{v_0}$:

(4.43)

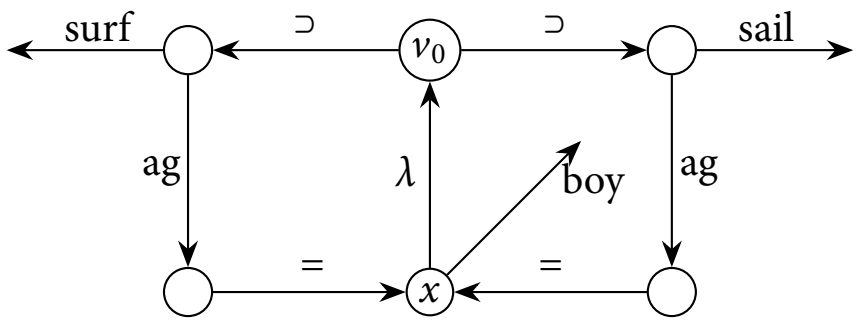


An answer to this question, which we do not have, should take into account the long-known pattern for subject indefinites to outscope subsequent scope-takers, especially if the latter are not quantifiers. For the following example, the first reading is much more accessible than the next.

(4.44) A boy didn't surf.

a. There is a boy who didn't surf. $(\exists\ boy > \neg)$

b. No boy surfed. $(\neg > \exists\ boy)$

It is useful to mention that argument sharing under coordination is flexible in allowing sharing of dissimilar thematic dependents (and even non-thematic dependents; see Section 4.2.3.1). A relative clause, for example, can conjoin a sentence with a gapped object and another with a gapped subject, both of type $s{\uparrow}np$. Thus in (4.45), the theme of *Joe likes* is to be aligned with the agent of *likes Joe*.

(4.45)

$$\begin{array}{llll}
\text{a boy} & \text{who} & \text{Joe likes} & \text{—} \\
np_1^0 & np_2^0\backslash_{\$} np_3^0/(s_4^1{\uparrow}np_5^0) & s_6^0/np_7^1 & \dfrac{s_{10}^0{\uparrow}np_{11}^1\ |_{\$}\ s_8^0}{np_9^1}
\end{array}$$

$$\begin{array}{llll}
\text{and} & \text{—} & \text{likes Joe} & \\
(s_{12}^2{\uparrow}np_{13}^4)\backslash_{\&}(s_{14}^0{\uparrow}np_{15}^5)/_{\&}(s_{16}^1{\uparrow}np_{17}^3) & \dfrac{s_{20}^0{\uparrow}np_{21}^1\ |_{\$}\ s_{18}^0}{np_{19}^1} & np_{22}^1\backslash s_{23}^0 & \rightarrow np_3^0
\end{array}$$

From (4.45) we find the equivalence in (4.46a), collapsing atom matches to shorten presentation, and thereby construct (4.46b).

(4.46) a. $\{np_1^0, np_2^0, np_3^0, np_5^0, np_{15}^5\}, \{s_4^1, s_{14}^0\}, \{s_6^0, s_8^0, s_{10}^0, s_{12}^2\},$
$\{np_7^1, np_9^1, np_{11}^1, np_{13}^4\}, \{s_{16}^1, s_{18}^0, s_{20}^0, s_{23}^0\}, \{np_{17}^3, np_{19}^1, np_{21}^1, np_{22}^1\}$

b.

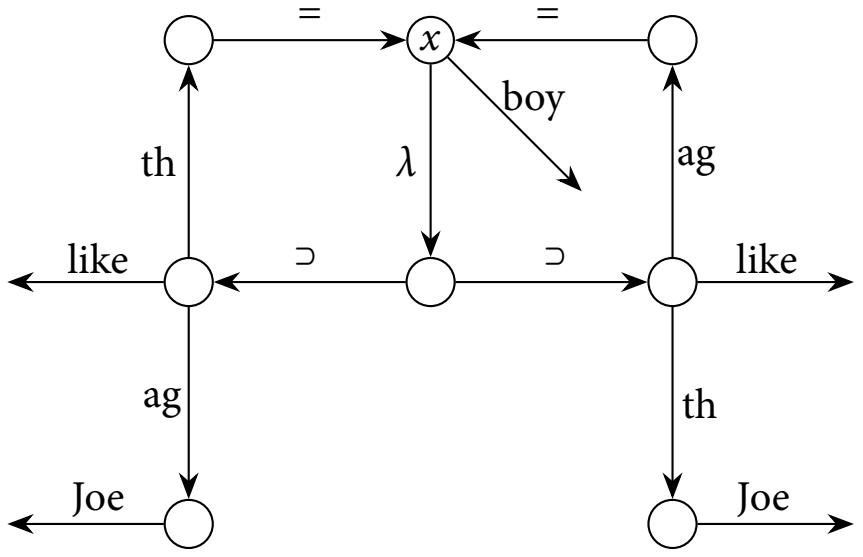

Note how $np_7$, the theme of *Joe likes*, equates with $np_{13}$, the 4-source of (4.39a), and similarly how $np_{22}$, the agent of *likes Joe*, equates with $np_{17}$, the 3-source of (4.39a). Note also that a boy $x$ takes scope, so to speak, over the conjunction by virtue of relativization.

*Remark.* Kubota and Levine (2020, chap. 4) discuss extensively the examples above and the likes, using a TLG in the natural-deduction style (apparently similar to CCGs but essentially equivalent to TLGs in the sequent-calculus style; see Appendix 3.A) and generalized conjunction for subsentential conjunction. The latter differs from our conjunction structure in *how* arguments are shared.

When two verbs share an argument in our representation, the referent and description it introduces do not duplicate. In (4.41b), for example, there is only one copy of $x$ and of $boy\,\vec{x}$. By contrast, generalized conjunction has arguments "shared" through duplication, as applying $h = \lambda x_1, \ldots, x_n\, f(x_1, \ldots, x_n) \wedge g(x_1, \ldots, x_n)$ to its arguments $x_1, \ldots, x_n$ makes two copies of each.

More generally, given multi-occurrence abstraction, the usual $\lambda$-calculus allows $\beta$-reduction to copy an argument as many times. By contrast, our semantic construction mechanism is resource-sensitive in that no lexical resource can be implicitly copied (see the comment on (4.18)). When and how resource-sensitivity plays a role in syntactic and semantic computation is itself a topic of interest; see Barker (2010); Jäger (2005); Oehrle (2003) for discussion.

### 4.2.3 Many facets of quantifiers

Some of the most interesting discussion about conjunction concerns quantifiers, and for that matter, nominals decorated by distributors (see Westerståhl, 2015; Winter, 2001, chap. 2; Zamparelli, 2011 for a review). Among other things, they can be conjoined at two levels due to their continuized typing.

*4.2.3.1 Quantifier level* Quantifiers can be conjoined at the quantifier level with the entry (4.47), along the argument sharing scheme discussed in Section 4.2.2.

(4.47) and $\rightarrow \dfrac{s^2 \mid s^5}{np^6} \backslash_{\&} \dfrac{s^0 \mid s^7}{np^8} /_{\&} \dfrac{s^1 \mid s^3}{np^4}$

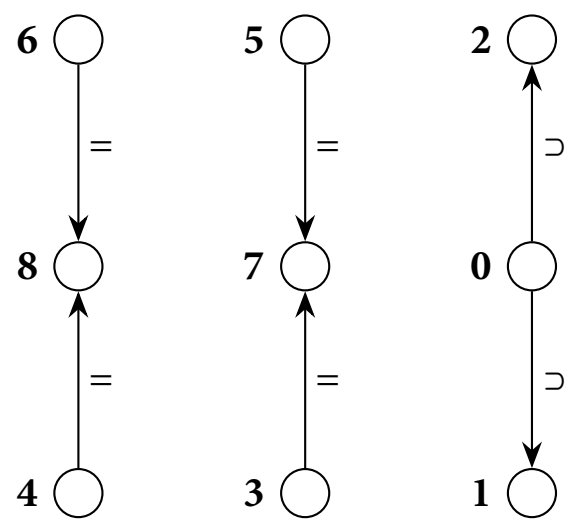


In this case, what is to be shared will no longer be thematic dependents.

For example, in (4.48a) *every dog* and *no cat* share their 1-sources (scopes) and 2-sources (iterators), which will respectively equal some swimming events and their agents, as shown in (4.48b).

(4.48) a. Every dog and no cat swam.

Every dog: $\dfrac{s^0 \mid s^1}{np^2}$ no cat: $\dfrac{s^0 \mid s^1}{np^2}$

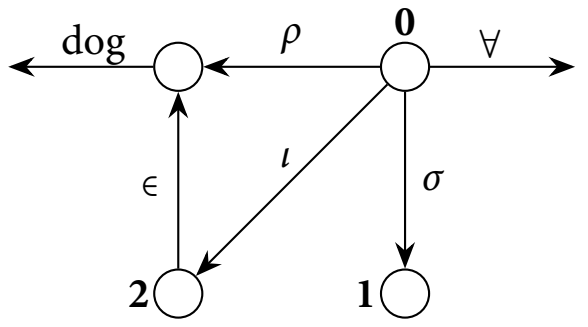


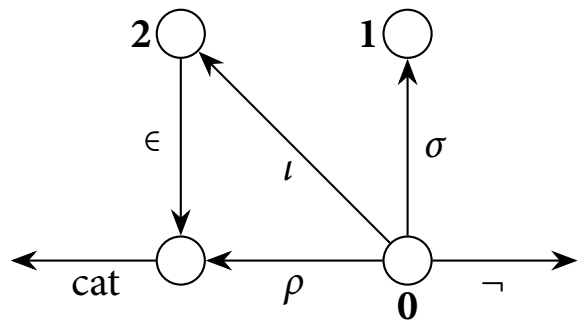


b.

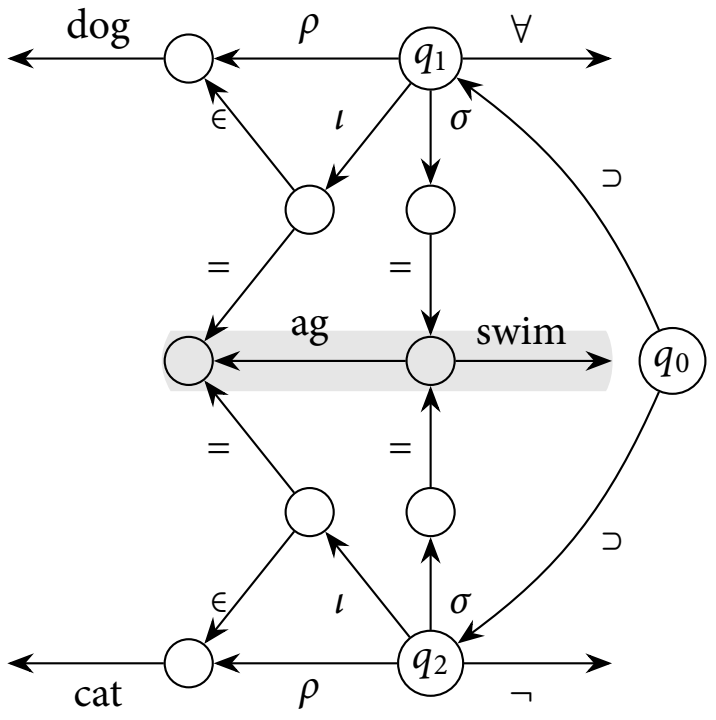


Quantifiers that undo distributivity can be similarly conjoined (see Section 4.1.3).

(4.49) Less-than-three dogs and less-than-three cats met

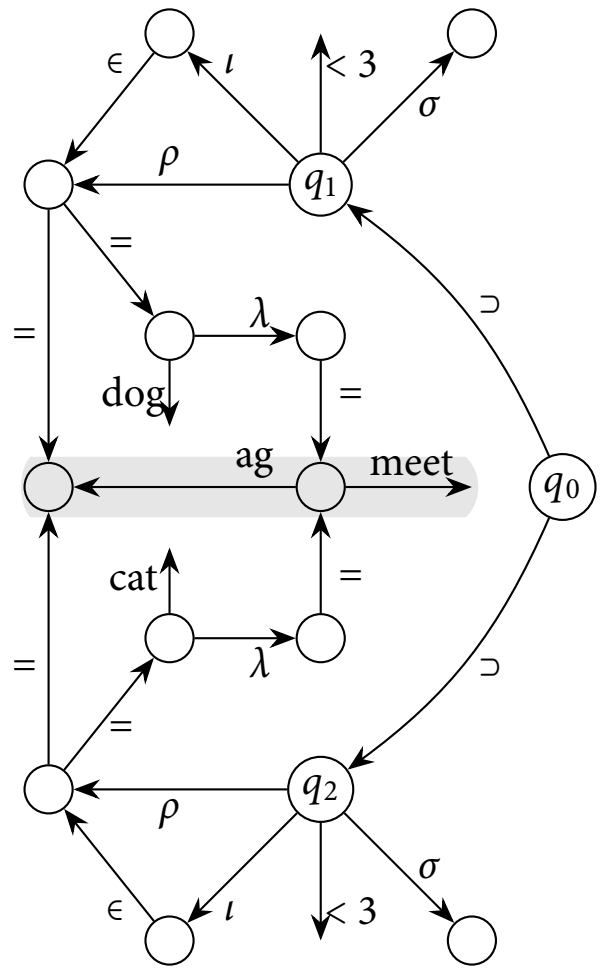


In these examples, the two quantification structures end up sharing the verb (the shaded area). By evaluating them independently, we arrive at the intended readings "every dog swam *and* no cat swam" for (4.48), and "less than three dogs met *and* less than three cats met" for (4.49). The effect of distributive coordination here is the same as what one would obtain from Partee and Rooth's (1983) generalized conjunction (the end remark of Section 4.2.2 still applies).

Our representation of argument sharing clearly gives quantifier-level conjunction a Boolean character, but as shown above, it nonetheless creates a plurality $q_0$, the union of the two quantification structures' dummies $q_1$ and $q_2$, except that the value of this "dummy union" was of no use. This is where our approach differs from previous non-Boolean approaches to quantifier conjunction

(Heycock and Zamparelli, 2005; Hoeksema, 1983), which bear little resemblance to generalized conjunction in terms of argument sharing, and create “meaningful” plural objects in a way sensitive to conjuncts’ monotonicity. Consequently, for such approaches downward monotone conjuncts have been noted as a difficulty; see Champollion (2015, sec. 7.1) and Winter (2001, sec. 2.2.2) for discussion.

*4.2.3.2 Noun phrase level* Continuized typing also allows quantifiers to conjoin at the *np* level with the entry (4.30), that is, without argument sharing. This is useful for dealing with branching quantification, including both cases reducible and those irreducible to quantifier nesting (see Section 4.1.3).

Consider the following example, where the semantics of *hugged* makes it clear that the intended reading of (4.50) is not one of distributive coordination. Rather, it means that every boy hugged every dog.

(4.50) a. Every boy and every dog hugged.

$\frac{s^0 \mid s^1}{np^2}$ $\quad np^2 \backslash_{\&} np^0 /_{\&} np^1 \quad$ $\frac{s^0 \mid s^1}{np^2}$

b.

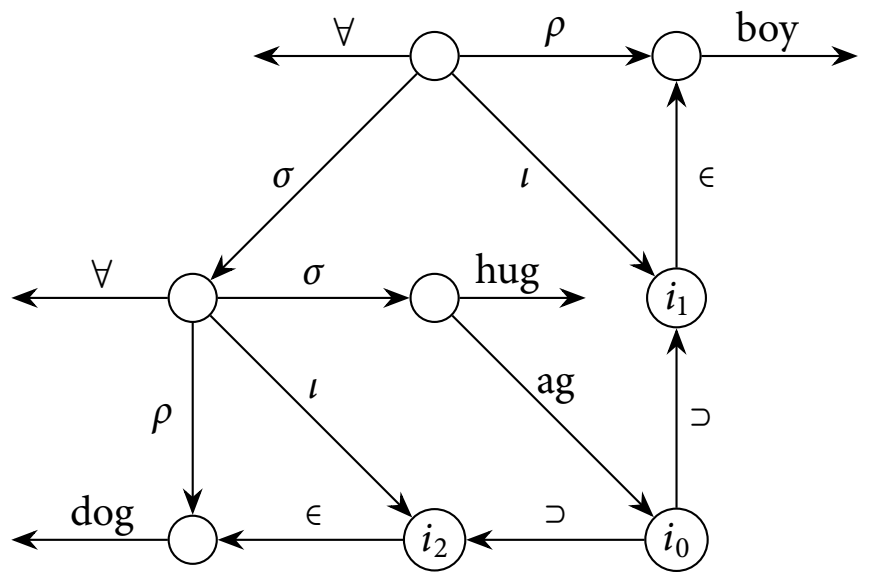


It is easy to work out the atom matches here, which should show that given the usual entry for quantifiers, *np*-conjunction sums up the iterators $i_1$ and $i_2$ while nesting quantifiers (the relative scope of *every boy*, *every dog* is irrelevant here). The effect is as if the quantifiers take scope out of conjunction by means of quantifier-raising (May, 1985), leaving their “traces” to conjoin each other. Actually we can paraphrase (4.50) this way (*every dog* not raised for readability):

(4.51) For every boy $x$, $x$ and every dog hugged.

Thus (4.50) expresses an idea deployable independently of graph formalisms; see Hoeksema (1988) for example, despite its oft-criticized scoping technique (see Winter, 2001, sec. 2.2.2).

This is not the case, however, for branching readings of downward monotone quantifiers under conjunction. For example, (4.52) can be read as (4.52a) or (4.52b) but not (4.52c), with (4.52a) itself being ambiguous between (4.52b) and (4.52c).

(4.52) No dog and no cat hugged.

a. No dog hugged no cat.
b. No dog hugged any cat.
c. For no dog $x$, $x$ and no cat hugged (= every dog hugged some cat).

Now that a treatment like (4.50) would derive exactly (4.52c), let us consider the entry (4.28) used for branching readings with transitives:

(4.53) no $\rightarrow \dfrac{s^0 \mid_\$ s^0}{np^2} / n^1$

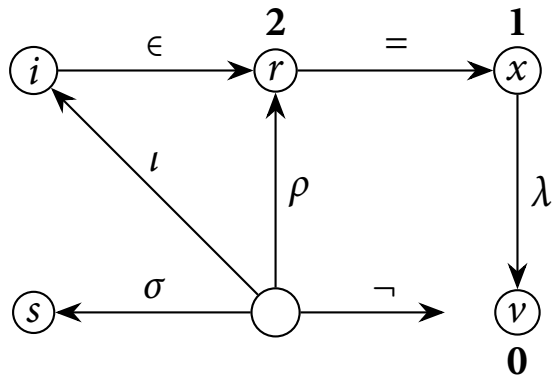


In (4.53) *np* corresponds to the restrictor, so that performing *np*-conjunction here has the effect of summing restrictors. Using the same atom matches from (4.50) we obtain what follows:

(4.54)

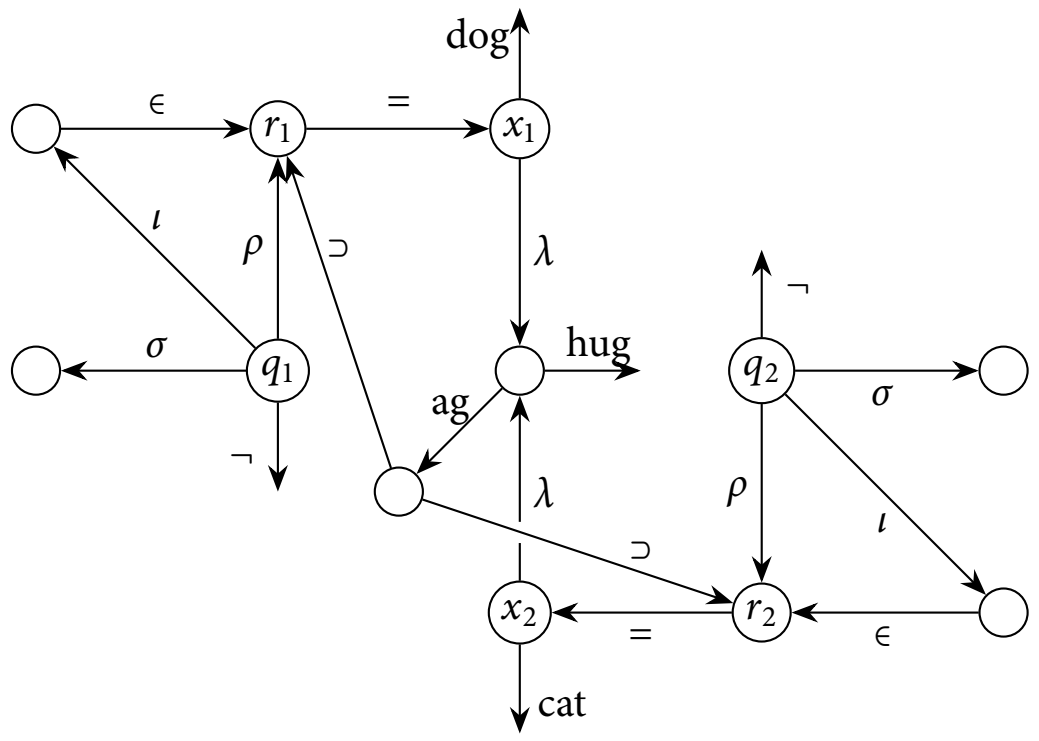


The earlier remark on (4.29) about multi-rootedness applies to (4.54) as well. That aside, the latter would represent a reading that requires zero dogs and zero cats be involved in all hugs of dogs and cats, which *entails* (4.52b), but wrongly excludes hugs among dogs (or cats) themselves.

The same problem would arise in any analysis of (4.52) that requires $r_1, r_2$ be empty when maximized subject to the constraint "$r_1$ is a set of dogs; $r_2$ is a set of cats; $r_1$ and $r_2$ hugged" (e.g. Brasoveanu, 2013; Charlow, 2018). To push any further, one has to answer deeper questions in lexical semantics: should we distinguish between the agent and theme of hugging (or any reciprocation)? If no, how can we restrict the result of conjunction to cat-hugging dogs and dog-hugging cats? There also remains the question why downward monotone quantifiers *have to* undo distributivity under *np*-conjunction: why is a quantifier-nesting reading missing here? We leave these for the future.

*4.2.3.3 Reconciliation* There is a potential type mismatch when conjoining a noun phrase and a quantifier, and there are mainly two ways to reconcile.

The easiest way is to directly conjoin them at the *np* level according to the discussion in the last section. For example, (4.55) illustrates a context where this is useful. The sentence has a reading paraphrased as (4.55a), where the valuation of *a dog* depends on the boy, a situation that can be made explicit with *a dog he walked*. Alternatively the indefinite can take wide scope. The two options follow from the precedence of $y$ with respect to quantification in (4.55b).

(4.55) Every boy and a dog hugged.

a. For every boy $x$, $x$ and a dog hugged.

b. 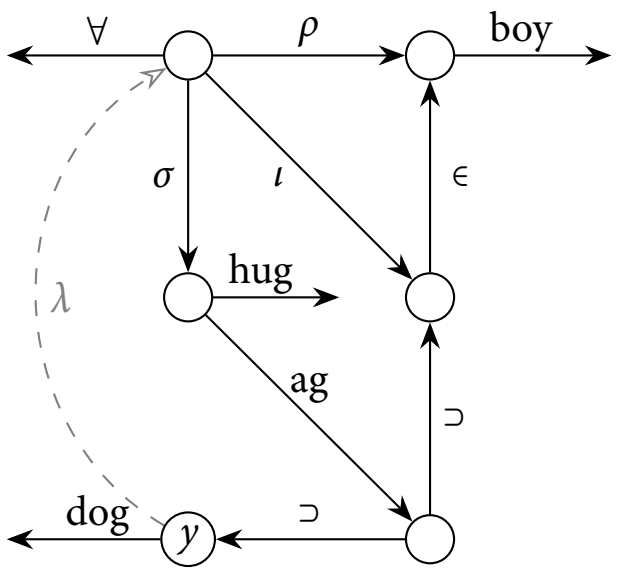


Another way is to perform the quantifier- or *np*-level conjunction after decorating the noun phrase with a distributor. Consider the following example, which is at least three-way ambiguous (assuming dogs can sail on their own):

(4.56) Six dogs DIST and every boy sailed.

a. Six dogs each sailed *and* every boy sailed. (quantifier-level, distributive predication)
b. Six dogs sailed *and* every boy sailed. (quantifier-level, plural predication)
c. Six dogs each sailed with each boy. (*np*-level)

Quantifier-level conjunction yields distributive coordination, in which case *sailed* can be applied to *six dogs* either distributively, as in (4.56a), or by plural predication, as in (4.56b). This distinction is not necessary for *np*-level conjunction; we are only interested in (4.56c), the reading that relates each dog to each boy in sailing. The reading that relates six dogs to each boy is derivable from direct *np*-conjunction, as in (4.55).

Now, we already know how to derive (4.56a) and (4.56c). With the distributor token (4.4) and its continuized typing, they can be treated in a similar manner to (4.48) and (4.50). But none of the distributors (or for that matter, quantificational determiners) discussed in Section 4.1 represents (4.56b) properly. Thus, we can introduce another entry (4.57a), and construct (4.57b) after (4.48):

(4.57) a. DIST $\rightarrow np^1 \backslash \dfrac{s^0 \mid s^2}{np^3}$ b. Six dogs DIST and every boy sailed.

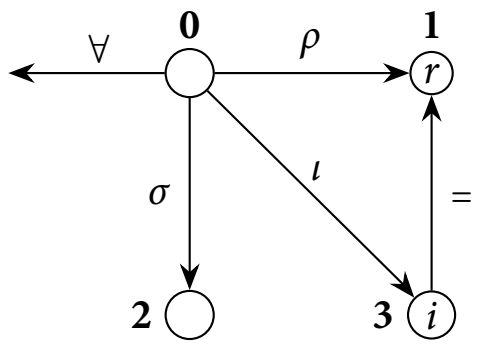


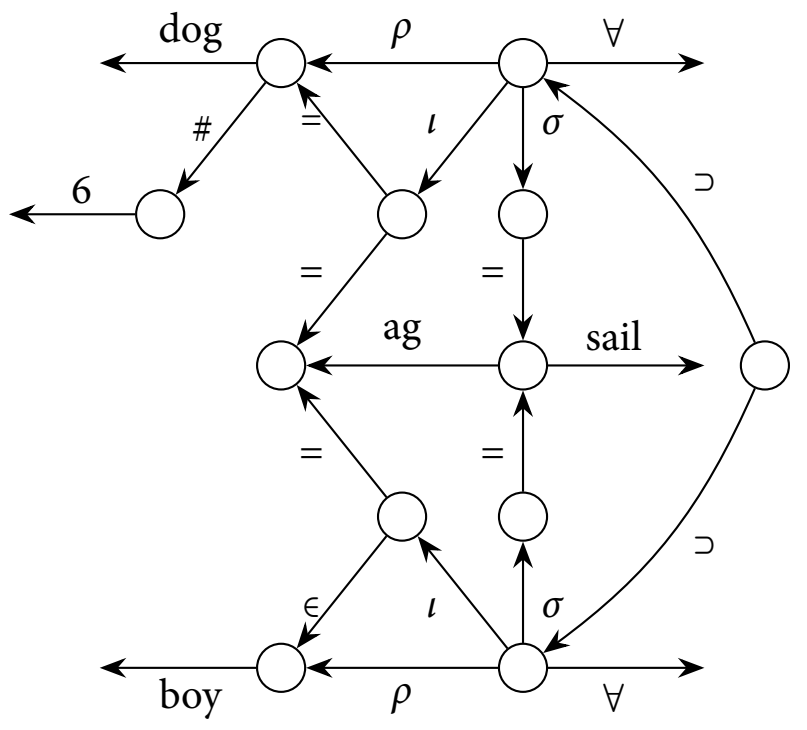


(4.57a) replaces membership with equality (= $\overrightarrow{ir}$), so in valuation $i$ duplicates $r$ instead of drawing singletons therefrom. In this connection, recall that interpretation of quantification structures does not refer to any specific label of $\overrightarrow{ir}$.

The possibilities discussed above may be strongly affected by lexical semantics. For example, if we are to perform quantifier-level conjunction in (4.58), *met* would require plural predication.

(4.58) Six dogs and less-than-three boys met.

On the other hand, *np*-conjunction, direct or indirect, might make no sense in this case. Barwise (1979, p.65) actually questions whether branching quantification with quantifiers of mixed monotonicity, say, one upward and the other downward, is interpretable at all (see Westerståhl, 1987, pp. 296ff for discussion).

Finally, we add that devices for reconciliation can also be useful in conjoining simple noun phrases. Converting one of them to a quantifier by means of a distributor, we reduce the situation to that of conjoining a noun phrase with a quantifier. For example, we can apply DIST in (4.57) to both conjuncts in (4.59), conjoin them at the quantifier level, and derive the intended reading.

(4.59) Two dogs and two cats met (= two dogs met *and* two cats met).

Thus it is fair to say non-atomic distributivity may result from distributive coordination (see Winter, 2001, sec. 6.4).

*Remark.* Conjoining two quantifiers at the *np* level effects raising of both, as noted from (4.50), and conjoining a quantifier and an indefinite as in (4.55) amounts to moving the former alone — both violating the *coordinate structure constraint* (CSC; Ross, 1967), which considers impossible both extraction of conjuncts and nonparallel subtraction from within conjuncts. Such examples and others illustrating an ambiguous scope relation between conjuncts lead Chaves (2007, secs. 3.2, 3.6) to question the relevance of the CSC to scoping (but cf. Fox, 2003, sec. 2.3.1). Even the nature of the CSC in describing overt movement like *wh*-displacement is not without debate. Kubota and Levine (2020, sec. 10.1.3) and citations therein contend that it might as well reflect the interaction of non-syntactic factors.

The standard CG account of the CSC rests on the simple requirement that conjuncts be of the same type (see Section 3.3.3.3). Should it have any merit, "conjunct extraction" in (4.50) and (4.55) (also (5.51) and (5.56) later) illustrate *np*-conjunction no less than (4.31) does. With continuized typing, "nonparallel subtraction" in (4.45) is made possible by innocent $s{\uparrow}np$-conjunction.

*4.2.3.4 Plural modification and pair-making* By *plural modification* we mean plural predication conveyed through the syntax of modification. For example, the underspecified way of relating five boys to six dogs in adoptions is the same in both (4.60a&b), and both semgraphs can be constructed equally with ease.

(4.60) a. Five boys adopted six dogs. b. Five boys who adopted six dogs.

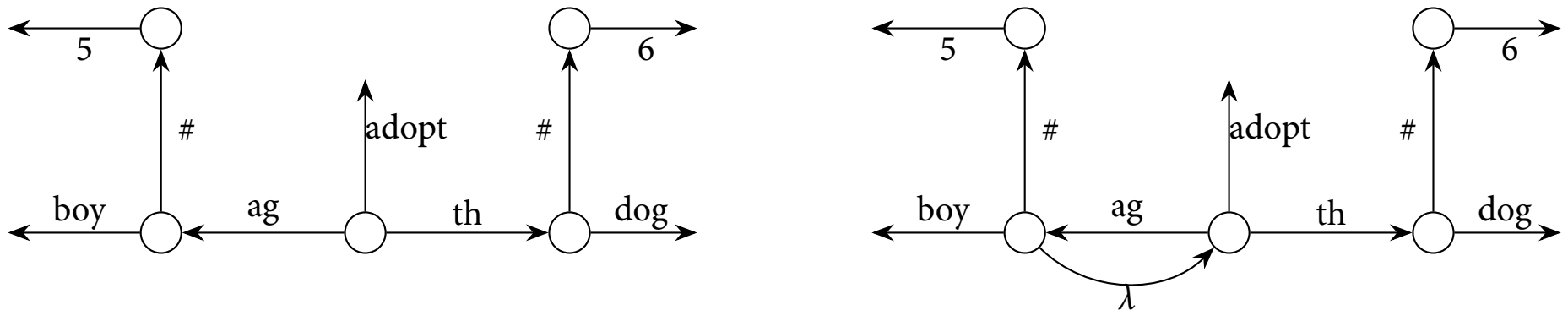

(It might be difficult to get a strictly cumulative reading of (4.60b)'s relative clause, where there are more than one adoption, but a collective reading can be readily coerced: *five boys who adopted six dogs together*.)

The question becomes interesting when plural modification meets branching quantification, as illustrated by the following example.

(4.61) Every boy and every dog who met hugged.
$np_1^2\backslash_{\&}np_2^0/_{\&}np_3^1$ (and) $np_4^0\backslash_{\$}np_5^0$ (who met)

To reduce branching to quantifier nesting, we conjoin the quantifiers at the *np* level as before. Note that the relative clause constrains the union of iterators, since $np_4$ matches $np_2$ in (4.61). The result of unification is shown in (4.62a).

(4.62) a.

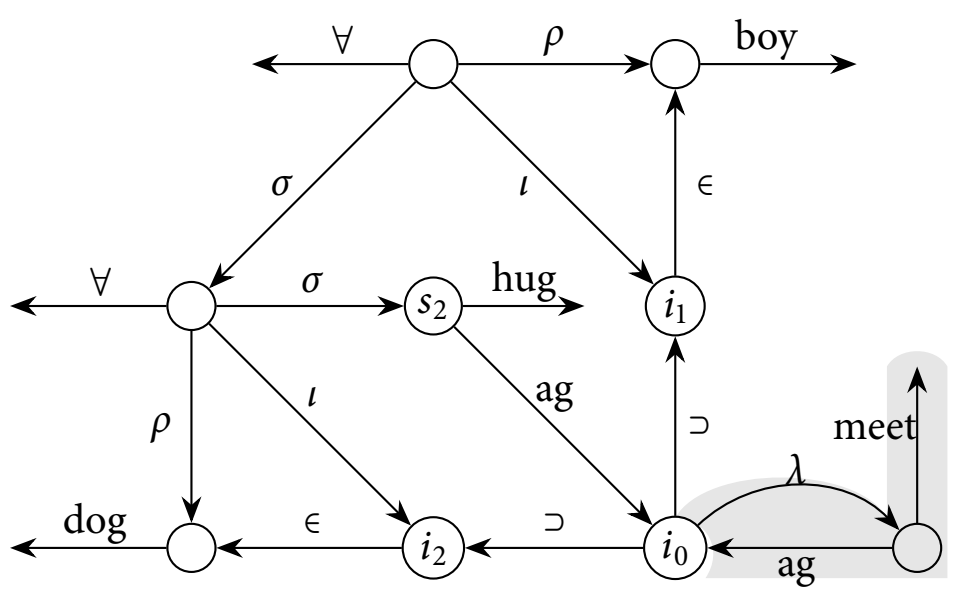


b.

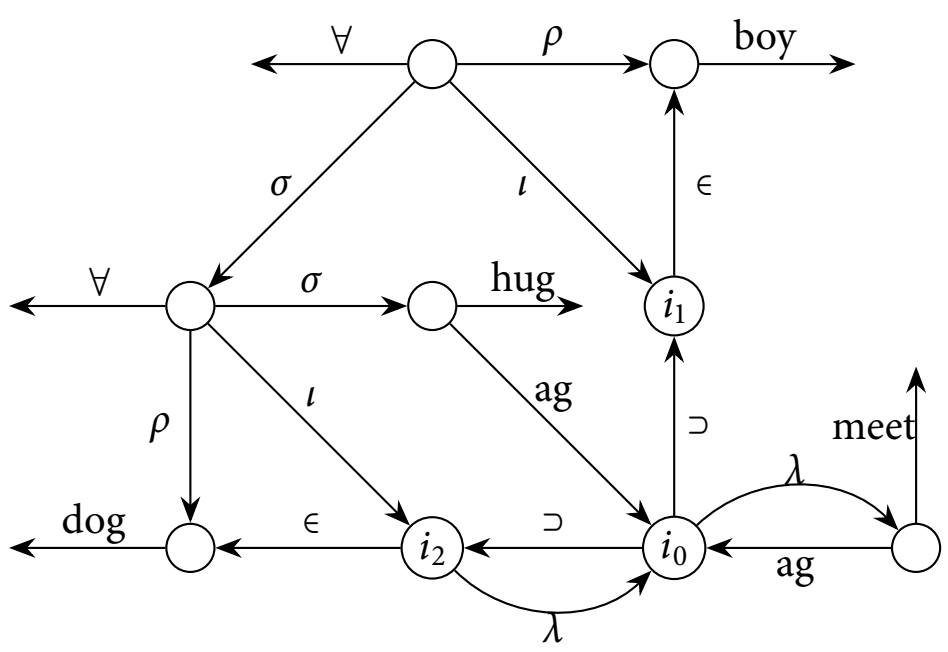


(4.62a) almost succeeds except for one problem: precisely because the shaded constraint is imposed on $i_0$ (the union of iterators), it was never reachable from either $i_1$ or $i_2$ but only from $s_2$; what is supposed to be an iterator filter becomes a scope constraint. This amounts to saying that each boy met and hugged each dog. What we want would rather be (4.62b), where the additional $\lambda\overrightarrow{i_2 i_0}$ ensures that the filter on $i_2$ selects a pair of a boy and dog only if they met.

We may attribute this addition to conjunction by introducing the following entry:

(4.63) and → $np^2\backslash_{\&}np^0/_{\&}np^1$

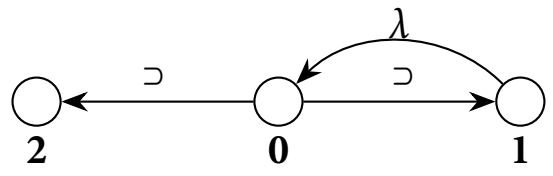


We can think of this entry as making one of the conjunct take local scope over the union, but motivating it elsewhere goes beyond the purpose of this thesis.

Examples like (4.61) came to be known as "hydras" since Link (1983, 1998, chap. 3), named after the polycephalous way a relative clause takes on conjoined tokens. They present a tension between the tradition of treating modifiers as internal to nominals (that is, a modifier combines with a noun before a determiner joins them) and the need of plural modification to stay external to conjunction. While this tension can be mitigated by continuation in syntax and separation of iterators from restrictor in quantification, (4.61) manifests a difficulty branching quantification again poses to graph traversal, besides multi-rootedness earlier.

Under the name "hydra" there is another kind of example discussed by Link (1983), one featuring noun-conjuncts sharing a determiner:

| | | | | | |
|---|---|---|---|---|---|
| (4.64) | Every | boy | and | dog | who met hugged. |
| | | $n^0$ | $n^2\backslash_{\&}n^0/_{\&}n^1$ | $n^0$ | |

(4.64) illustrates a problem independent of plural modification; with or without the relative clause, the atomic distributivity built into quantifiers raises the same question, what is being iterated over here, of which we assert meeting or hugging? As (4.64) is semantically equivalent to (4.61), a natural answer would be (unordered) boy-dog pairs, which are clearly not singletons.

Such is the problem of *pair-making*. Here we may attempt a solution that lies in the way we model semantics. Recall from Section 2.2.2 that while the domain of a model contains only first-order entities, some of them denote groups and relate to their members in events like comprisings or constitutings. Now suppose, as an ontological postulate, that for any finite set $x$ of entities, there exist a disjoint singleton $y$ and a comprising event $v$ such that $v$ has $y$ as its agent and $x$ its theme. Then we can use the entry (4.65a) to maximize $r$ to the set where each $y \in r$ (meaning $y$ is a singleton subset of $r$) comprises $x$, a union of singletons $z$ and $w$. Specifying $z$ as containing boys and $w$ as containing dogs, (4.65b) allows $i$ to draw from boy-dog pairs.

(4.65) a. and → $n^2\backslash_{\&}n^0/_{\&}n^1$

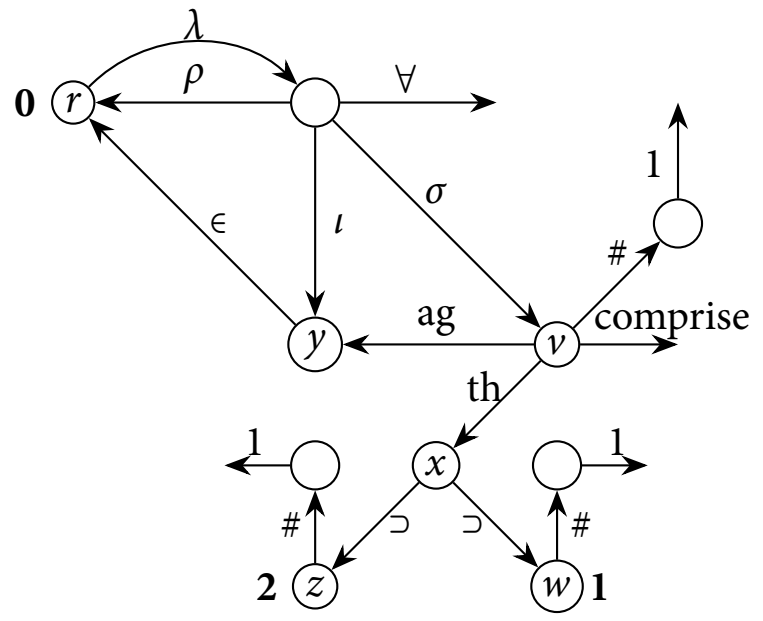


b. Every boy and dog hugged.

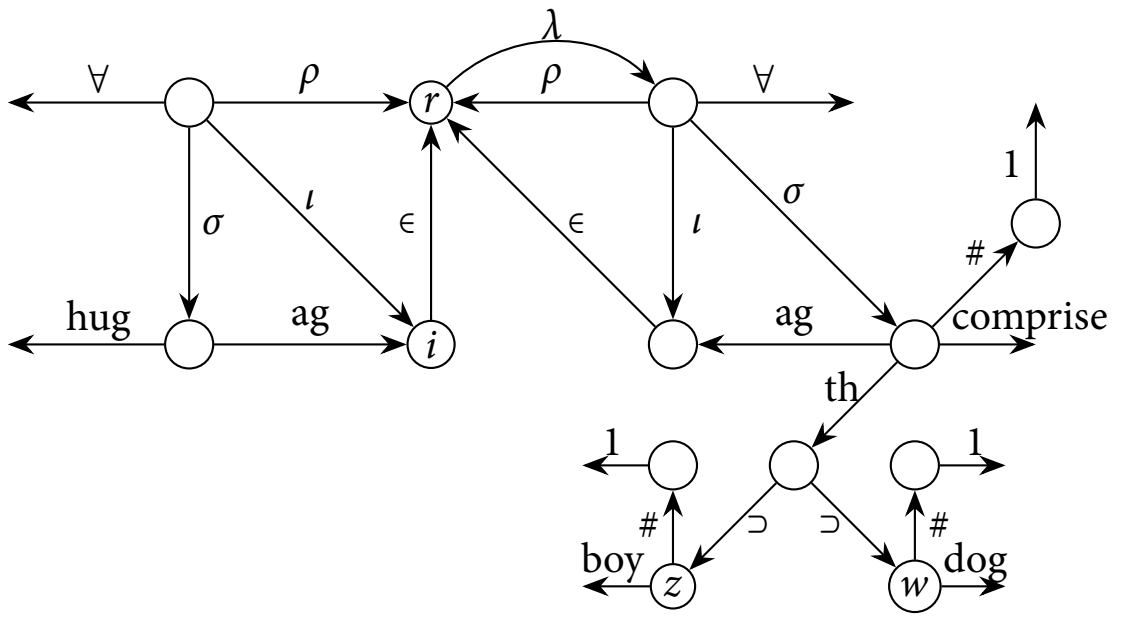


Of course, noun-conjunction can use simple union (4.30) as well. For example, (4.66) can assert enjoyment of each animal.

(4.66) Every dog and cat enjoys a good meal.

And on the other hand, with different ontological assumptions, there are other ways of making pairs. For example, if we allow for variables over sets of higher-order objects (and in that regard, non-atomic distributivity), pair-making is simply a matter of unordered Cartesian product. This is essentially the idea of Link (1983) and many subsequent works. For a recent formulation in the form of Boolean conjunction, see Champollion (2015).

# Chapter 5

# Topics in Scope

Relative scope arises as the valuation of one variable depends on that of another, and in this sense, we speak of the order of valuation that comes with quantification, intensionality, and coordination (see Section 2.2.6). Despite this unifying description, what concerns us now is how certain linguistic expressions (more precisely, the referents introduced) attain their scope from the syntax-semantics interface.

In this chapter, we discuss classic issues associated with (genuine) quantifiers and indefinites as they take scope. Central to our discussion is the proposal that the former does so through syntax but the latter through precedence resolution. The unification-based semantic construction provides a precise *and* concise language for studying interactions among quantificational expressions. Precedence resolution separate from syntax further simplifies the treatment of exceptional scoping behaviors of indefinites.

## 5.1 Scoping quantifiers

Let us start with multiple quantifiers taking scope over each other.

### 5.1.1 Scope permutation

A simple sentence like (5.1) can be verified in two ways. We can iterate through a population of boys and see if each walked most dogs, or we can find a majority of dogs (out of a population) and see if each was walked by every boy.

(5.1) Every boy walked most dogs.

The situations accepted by these two procedures then fall into two classes, respectively known as the *surface-scope* reading (*every boy* over *most dogs*) and the *non-surface-* or *inverse-scope* reading (*most dogs* over *every boy*) of the sentence. If this classification is taken seriously due to the procedural difference, then, as theorists often assume, scope ambiguity is indeed a case of ambiguity, contra underspecification in plural predication, where various situations are verified by the same procedure.

Various techniques have been developed to disambiguate scope by constructing logic formulas with nested quantifiers. Some of the textbook examples include quantifying-in (Montague, 1973),

quantifier-raising (May, 1985), Cooper storage (Cooper, 1983), flexible typing (Hendriks, 1993), surface constituency (Steedman, 2011), and continuations (Barker and Shan, 2014).

In the context of semgraphs, we have a similar goal of construction, that is, nested quantification structures. Thus the surface- and inverse-scope readings of (5.1) are given by (5.2a&b), with the valuation of $i_2$ depending on that of $i_1$ (see Section 2.3.1):

(5.2) a.

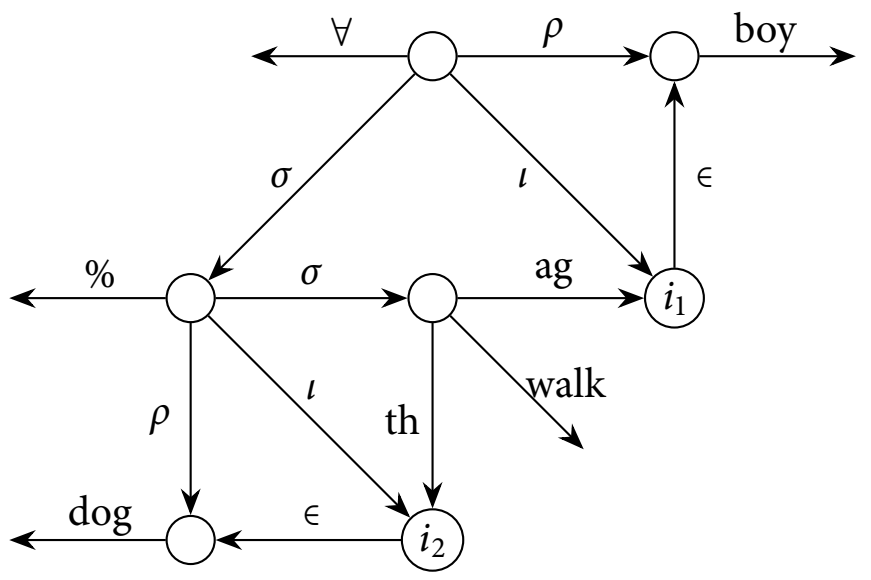


b.

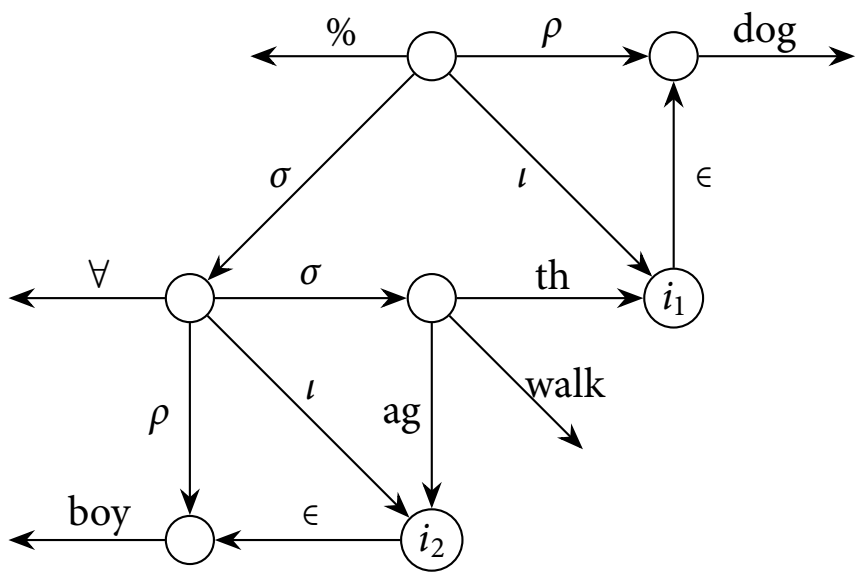


We have already constructed a similar structure in (4.50). The fact is we can use a continuized syntax like that of Barker and Shan (2014) to derive both semgraphs above, while not resorting to higher-order semantics.

To see this, consider the two ways to reduce (5.1) to the type of a sentence.

(5.3) a. **every boy** walked most dogs (∀ *boy* > % *dogs*)

$$\frac{\mathbf{s_3^0} \mid s_1^1}{np_2^2} \quad np_4^2\backslash s_5^0/np_6^1 \quad \frac{s_8^0 \mid s_7^1}{np_9^2} \quad \to \mathbf{s_3^0}$$

$$(\mathbf{s_1^1}, \mathbf{s_8^0}), (\mathbf{s_5^0}, \mathbf{s_7^1}), (np_2^2, np_4^2), (np_6^1, np_9^2)$$

b. every boy walked **most dogs** (% *dogs* > ∀ *boy*)

$$\frac{s_3^0 \mid s_1^1}{np_2^2} \quad np_4^2\backslash s_5^0/np_6^1 \quad \frac{\mathbf{s_8^0} \mid s_7^1}{np_9^2} \quad \to \mathbf{s_8^0}$$

$$(\mathbf{s_1^1}, \mathbf{s_5^0}), (\mathbf{s_3^0}, \mathbf{s_7^1}), (np_2^2, np_4^2), (np_6^1, np_9^2)$$

Although in both cases the *np* matches are identical (so the iterators of *every boy* and *most dogs* assume the same thematic roles), (5.3a) yields narrow scope for *most dogs* by matching its root ($s_8^0$) with the scope of *every boy* ($s_1^1$), whereas (5.3b) yields narrow scope for *every boy* by matching its root ($s_3^0$) with the scope of *most dogs* ($s_7^1$). The variation in *s* matches follows from the lifted application rule in Definition 3.1, which may identify either the subject (*every boy*) or the verb phrase (*walked most dogs*) as *the* tower when combining them:

(5.4) a.

$$\frac{\mathbf{s_3^0} \mid \mathbf{s_1^1}}{np_2^2}, \frac{s_8^0 \mid s_7^1}{np_4^2\backslash s_5^0} \to \dfrac{\mathbf{s_3^0} \mid \mathbf{s_1^1}}{\dfrac{s_8^0 \mid s_7^1}{s_5^0}}$$

b.

$$\frac{s_3^0 \mid s_1^1}{np_2^2}, \frac{\mathbf{s_8^0} \mid \mathbf{s_7^1}}{np_4^2\backslash s_5^0} \to \dfrac{\mathbf{s_8^0} \mid \mathbf{s_7^1}}{\dfrac{s_3^0 \mid s_1^1}{s_5^0}}$$

Successive lowering of the towers on the right produces the $s$ matches in (5.3a&b). We can see that each permutation of scope corresponds to a distinct arrangement of the continuation layers of quantifiers (e.g. $s_3^0 \mid s_1^1$ of *every boy*).

*Remark.* Nested quantification structures are similar to nested *for*-loops in programming languages: the iteration of an inner structure is performed within that of an outer structure. This is why quantifier scope cannot be reversed by changing the order of valuation. Take (5.2a) and add $\lambda\overrightarrow{q_2q_1}$ to shift rootship from $q_1$ to $q_2$:

(5.5)

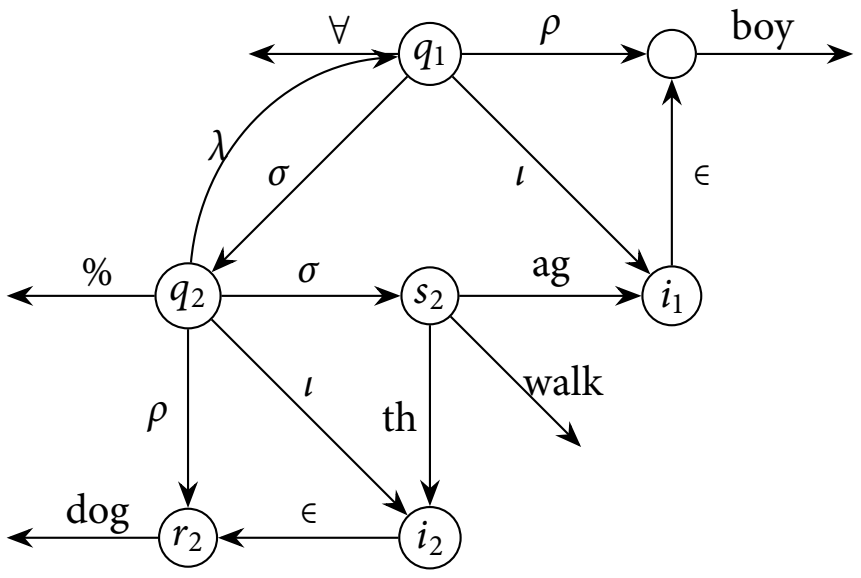


According to Definition 2.8 our semgraph interpreter ignores any out-edge of $q_2$ other than $\% \overrightarrow{q_2}$, $\rho\overrightarrow{q_2r_2}$, $\iota\overrightarrow{q_2i_2}$, $\sigma\overrightarrow{q_2s_2}$. Even if we delayed evaluating the quantification structure at $q_2$ and processed $\lambda\overrightarrow{q_2q_1}$, the latter would bring us back to (5.2a), except with the value of $q_2$ fixed across boys.

*5.1.1.1 Factorial* Without further constraints, the method described above shall generate $n!$ scope permutations for a sentence with $n > 1$ quantifiers. One can reason by induction: suppose in a syntactic derivation we can stack the continuation layers of $n-1$ quantifiers in $(n-1)!$ ways. Adding one more quantifier, we can distinguish $n$ positions in any of those $(n-1)!$ arrangements to insert another continuation layer.

We derived a classic generalization predicted by a classic scoping technique like quantifier-raising: the number of distinct scope permutations of a sentence equals the factorial of the number of the scope-takers it contains, provided they are each independently scopable. For the following example of Hobbs (1983), one would expect $5! = 120$ readings:

(5.6) In most democratic countries most politicians can fool most of the people on almost every issue most of the time.

There are, however, discussions that the classic generalization might have ignored idiosyncrasies among what are collectively known as "generalized quantifiers" (see Szabolcsi, 2010, chap. 11 for a review). In particular, downward monotone quantifiers and compound numerals are mentioned as resisting inverse scope.

For example, it is frequently cited that downward monotone quantifiers must take narrow scope under an indefinite or simple numeral subject:

(5.7) a. A boy has missed *no meal.* (Szabolcsi, 2010, p. 178)
b. Three referees read *few abstracts.* (Szabolcsi, 1997, p. 110)

But if a quantifier substitutes into the subject position, one's findings can vary:

(5.8) a. Everyone loves *no one*. (Barker and Shan, 2014, p. 110)
b. Every farmer owns *few donkeys*. (Steedman, 2011, p. 129)

According to the sources indicated, both surface- and inverse-scope readings exist in (5.8a), and in (5.8b) the inverse scope is even more accessible.

Compound numerals might make a better case. Absence of inverse scope has been reported for sentences with indefinite and quantifier subjects:

(5.9) a. Some student read *more-than-five books*. (Beghelli, 1995, p. 48)
b. Every student read *more-than-one paper*. (Szabolcsi, 2010, p. 186)

Takahashi (2006) takes these as arguing for a compositional analysis of compound numerals, a move independently motivated in Hackl (2000); Krifka (1999) among others. We will not study the scoping pattern of compound numerals along this direction, although a serious treatment of comparatives and measures in the context of semgraphs is worth future research (see Liang et al., 2013).

*5.1.1.2 Inverse linking* In the scope literature, *inverse linking* refers to the inverse scope taken by a quantifier from a post-nominal modifier. Such examples are first discussed by May (1978, 1985) and differ from usual inverse scope for having one scope taker syntactically dependent on the other, as shown by (5.10):

(5.10) Every boy from most islands sailed.

| Every boy | from | most islands | sailed. |
|---|---|---|---|
| $\dfrac{s^0 \mid s^1}{np^2}$ | | $\dfrac{s^0 \mid s^1}{np^2}$ | |

Here the intended reading amounts to saying "every boy islander sailed" holds for most islands.

Given this reading, it is fairly easy to construct (5.11b) with the preposition entry (5.11a), which serves as a filter on $i_2$, selecting a boy if he was from some island $i_1$. To obtain the desired scoping, we only need to make sure that the continuation layer of *most islands* tops the resultant tower when *every boy* combines with the prepositional modifier.

(5.11) a. from → $np^0 \backslash_{\$} np^0 / np^1$ b. Every boy from most islands sailed.

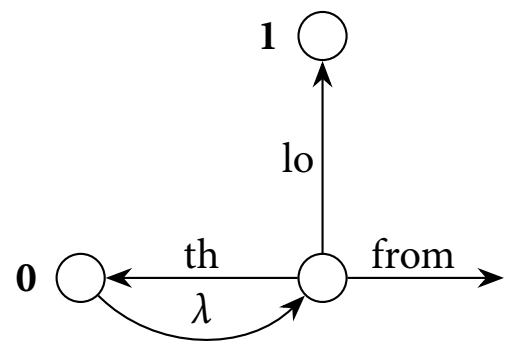


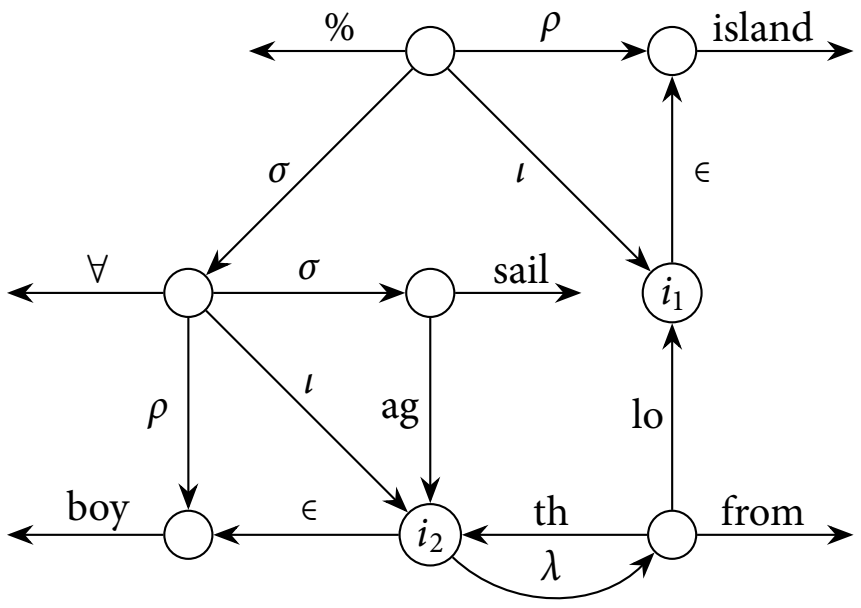


It is interesting to consider "surface linking", or the surface scope in such examples:

(5.12) Every graduate with most skills applied.

Most naturally (5.12) talks about each graduate who had most skills. This reading, however, cannot be constructed similarly to (5.11b). With (5.13a), merely placing the continuation layer of *every graduate* above that of *most skills* leads to (5.13b), which, when interpreted, turns out to be a model-independent contradiction: a valuation $h$ satisfying the iterator subgraph at $i_1$ must be defined on $i_2$, but a valuation $k \supseteq h$ satisfying the scope subgraph at $q_2$ must be undefined on $i_2$ (both subgraphs are evaluated without $i_2$ in the visiting history; see Definitions 2.5/2.6).

(5.13) a. with → $np^0\backslash_\$ np^0/np^1$ b. (*Naught*)

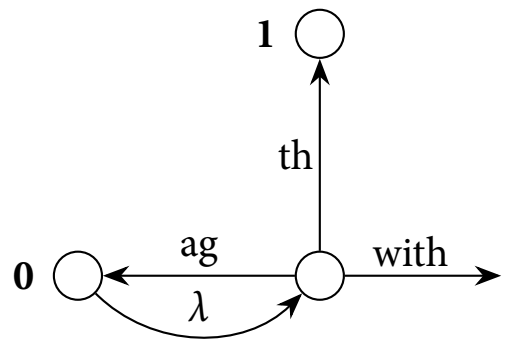


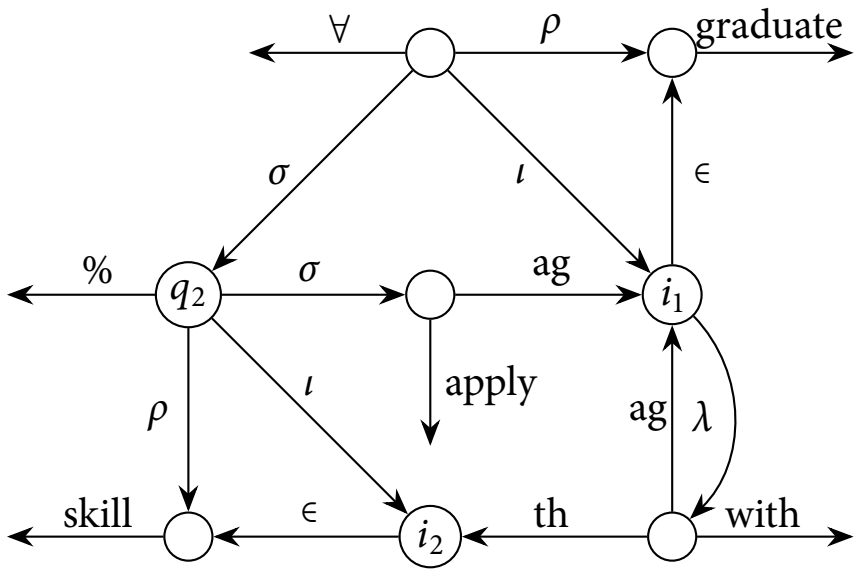


While this contradiction is akin to those which account for anaphoric accessibility in Section 5.2.4, its fundamental cause is simply that as an iterator filter, (5.13a) tries to select a graduate based on his or her relation with an individual skill instead of many.

The desired reading of (5.12) is given by (5.14b), which puts the whole narrow-scope quantifier into the iterator subgraph of the wide-scope quantifier. This structure is constructed using (5.14a).

(5.14) a. with → $np^0\backslash_\$ np^0/\dfrac{s^1 \mid s^2}{np^3}$ b. Every graduate with most skills applied.

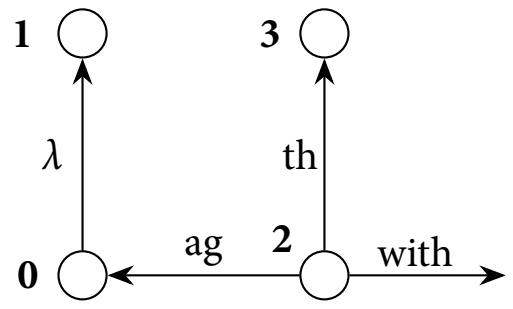


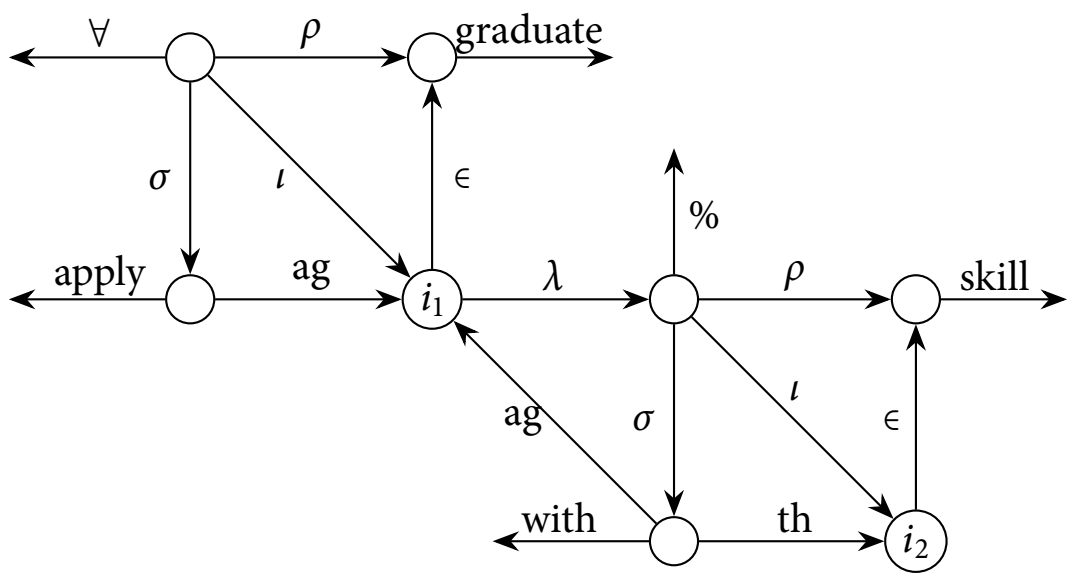


Comparing (5.13a) with (5.14a), we may say that the latter has its right denominator raised to the type of a quantifier. But a comparison of the semantics of the two entries suggests a deeper connection, since they minimally differ in how a $\lambda$-edge is positioned, or equivalently, whether a $\lambda$-cycle is lexically ready or to be formed in syntax.

To make sense of this connection, we may try to build (5.13a) and (5.14a) out of relativization, with an imaginary transitive verb WITH as follows.

(5.15) that — WITH

$np^0_1\backslash_\$ np^0_2/(s^1_3{\uparrow}np^0_4)$ $\dfrac{s^0_7{\uparrow}np^1_8 \mid_\$ s^0_5}{np^1_6}$ $np^2_9\backslash s^0_{10}/np^1_{11}$

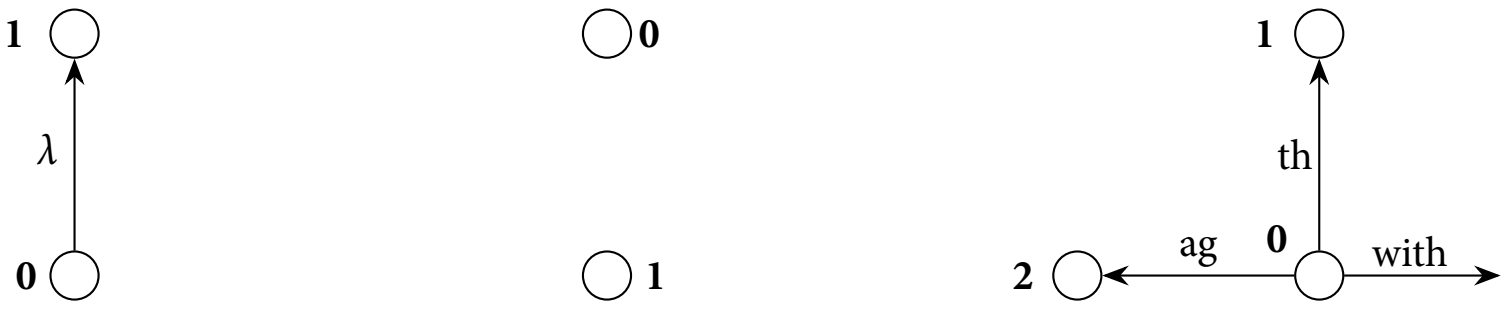

There are two ways to reduce (5.15), depending on whether or not $s_5$ cancels $s_{10}$. Though it is more convenient to validate the following matches by using a TLG (see Appendix 3.A) than by extending the CCG we have, one can nonetheless verify that (5.16a) and (5.16b) yield (5.13a) and (5.14a), respectively.

(5.16) a. $\rightarrow np^0_1\backslash_\$ np^0_2/np^0_{11}$

$(s^1_3, s^0_7), (np^0_4, np^1_8), (np^1_6, np^2_9), (\boldsymbol{s^0_5}, \boldsymbol{s^0_{10}})$

b. $\rightarrow np^0_1\backslash_\$ np^0_2/\dfrac{\boldsymbol{s^1_5} \mid \boldsymbol{s^2_{10}}}{np^3_{11}}$

$(s^1_3, s^0_7), (np^0_4, np^1_8), (np^1_6, np^2_9)$

Thus in a concrete sense, a post-nominal preposition is like a lexicalized relative clause missing an object. If we replace prepositions with incomplete relative clauses like (5.15), then surface and inverse linking will both derive from the same sequence of tokens.

Following May's (1978, sec. 2.5) suggestion, Heim and Kratzer (1998, sec. 8.5) describe a relativization technique similar to ours for deriving surface linking. In the context of a type-logical grammar, Carpenter (1997, sec. 7.5) achieves the same by raising prepositions' argument type syntactically and semantically. Our discussion above shows that the two approaches are intrinsically related. Further, if type-raising in syntax can be mirrored either by type-raising in logic or by graph editing (e.g. what transforms (5.13a) into (5.14a)), there might be an unifying theme underlying these semantic operations seemingly different in nature.

*5.1.1.3 The immobile* A continuation-based scoping method also works for quantificational expressions that are typically considered immobile in terms of quantifier raising, such as verbal negation and modals.

In the following example, negation can take scope either under or over the subject quantifier. We can be sure that (5.17a&b) represent the indicated readings by delimiting the scope subgraph of the negative quantifier in each case.

(5.17) Every cat didn't meow.

a. (∀ *cat* > ¬)

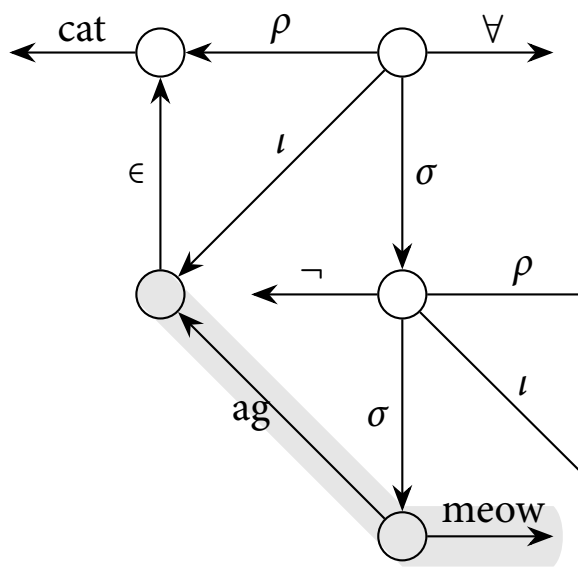


b. (¬ > ∀ *cat*)

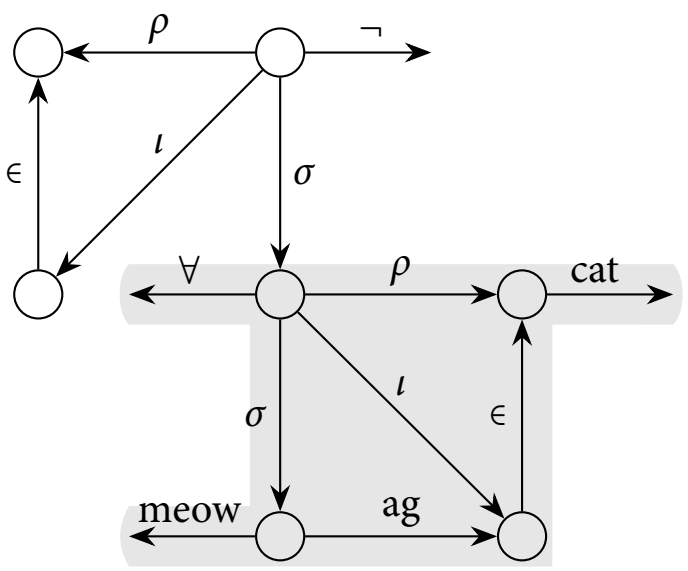


To construct these, we give the following entries for negation, a plain type for its narrow scope and a raised type for its wide scope:

(5.18) a. didn't → $np_2^2\backslash s_3^0/(np_4^2\backslash s_5^1)$ b. didn't → $\dfrac{s_6^0 \mid s_1^1}{np_2^3\backslash s_3^2/_{\$}(np_4^3\backslash s_5^2)}$

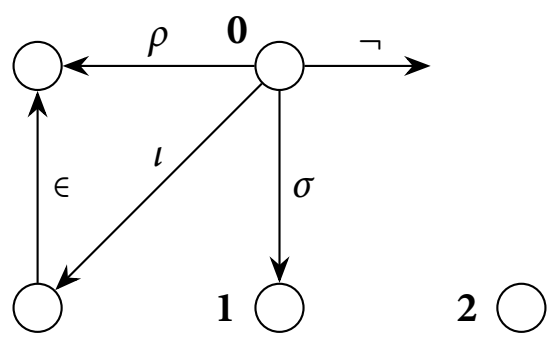


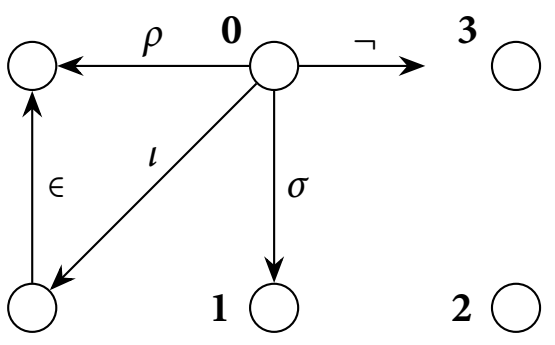


The reason that (5.18b) cannot be used to construct narrow-scope negation is that doing so ends up matching $s_1$ with $s_3$, which means fusing distinct vertices of the same graph, an operation undefined in HR algebra (recall from Section 1.6.3 that distinct sources have distinct labels). Note also that due to a tonal difference, $np_4\backslash s_5$ replicates the depths of $np_2\backslash s_3$ in (5.18b) but not in (5.18a).

Depending on their flavors, modals can bear a likewise ambiguous scope relation to quantifiers (a.o. Fintel and Iatridou, 2003). Thus (5.19) utters either a deontic necessity per adopter, or one that involves all adopters.

(5.19) Every adopter must sign.

a. (∀ *adopter* > *must*) b. (*must* > ∀ *adopter*)

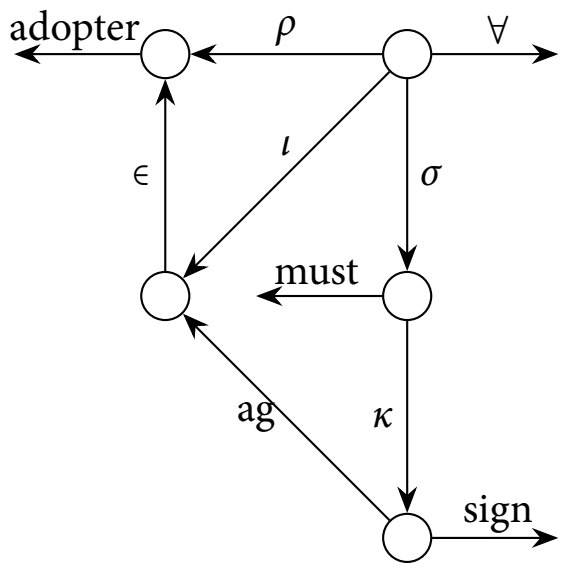


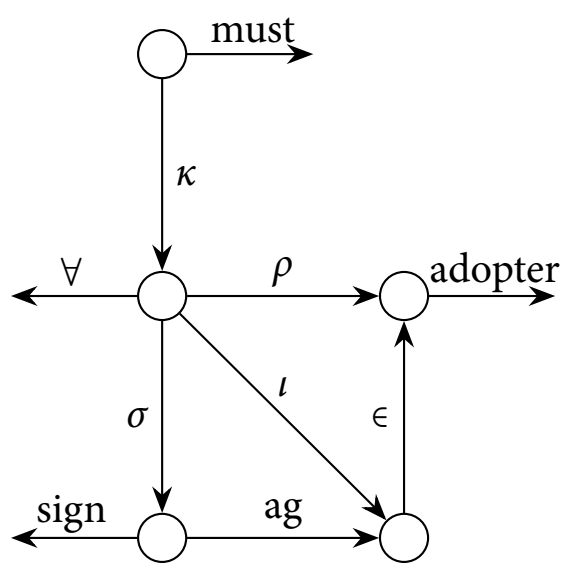


Here we can put (5.20a&b) in parallel with (5.18a&b).

(5.20) a. must → $np^2\backslash s^0/(np^2\backslash s^1)$ b. must → $\dfrac{s^0 \mid s^1}{np^3\backslash s^2/_{\$}(np^3\backslash s^2)}$

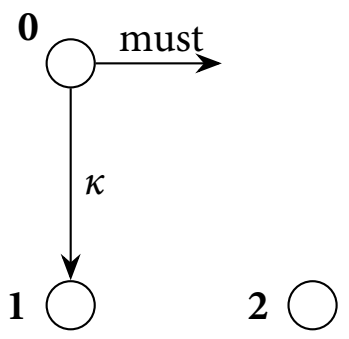


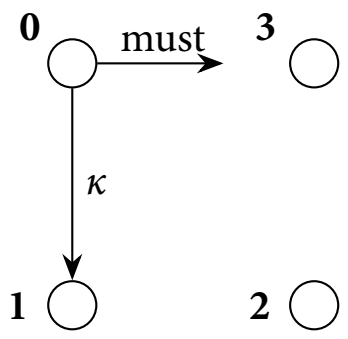


This is somewhat simplifying the matter, as the scoping options for epistemic modals, for example, are found to be more varied, sometimes subject to constraints not entirely understood. See Swanson (2010); Tancredi (2007) for discussion and Hacquard and Wellwood (2012) for a survey of epistemics embeddability in general.

A drawback of type-raised (5.18&5.20b) is that they may generate unattested wide scope for negation and modals. Take negation for example; (5.21) is not a denial that every cat promised to meow, but such wide-scope negation will arise from using (5.18b) for *not-to*.

(5.21) Every cat promised not-to meow.

$np^2\backslash s^0/(np^2\backslash s^1)$

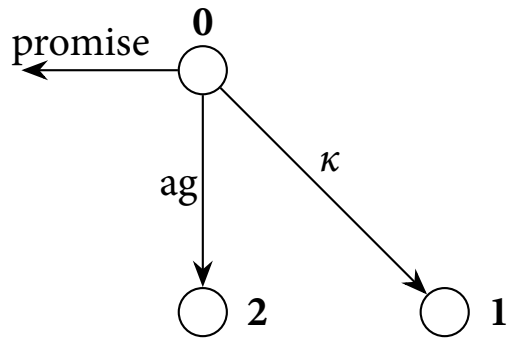


The situation is reminiscent of lexicalized scope islands discussed in Section 5.1.2: just as the finite clause introduced by certain tokens can trap certain quantifiers, the infinitive clause introduced by *promise* can trap negation. To anticipate the upcoming discussion, we can somehow mark *promise* and (5.18b), so as to block the following combination unless the tower of *not-to meow*, carrying (5.18b)'s continuation layer, is lowered:

(5.22) promised not-to meow

$$\frac{s \mid s}{np\backslash s}$$

But that is impossible, leaving only the option of deriving (5.21) with (5.18a) and confining negation to a promise.

*Remark.* A common practice in the linguistic literature treats the limited inverse scope taken by negation or modals via *reconstruction* (see Fox, 1999; Sternefeld, 2001 a.o.), which "returns" a subject quantifier to a nearby gap within the surface scope of negation or modals by either interpreting the subject at the gap or giving the gap a higher-order semantics. With semgraphs the former can be more easily reproduced; to compose (5.17b), for example, we can add a gap next to *didn't* in (5.17), give it the type and semantics of *every cat* before making the latter a single-source identity element of type $s^0/_{\$}s^0$, and retype (5.18a) as $s^0/s^1$ while dropping its 2-source.

### 5.1.2 Scope islands

The concept of scope islands serves to describe the syntactic domain up to which a quantifier may take scope. Traditionally, this domain is identified by the boundary of a finite clause (e.g. May, 1978), like the complement of an attitude in (5.23a), a relative clause in (5.23b), and the antecedent clause of a conditional in (5.23c). Each of these is followed by a nonexistent reading that illustrates the island constraint.

(5.23) a. Joe believed [that every dog swam].
(≠ For every dog $x$, Joe believed that $x$ swam.)

b. Joe saw a dog [that every boy walked].
(≠ For every boy $x$, Joe saw a dog that $x$ walked.)

c. If [every star shines], Joe sails.
(≠ For every star $x$, Joe sails if $x$ shines.)

With our continuized CCG, the scope of a quantifier is fixed when the tower containing its continuation layer is lowered. Take the main verb phrase of (5.23a), where complementizer *that* is treated as an identity element.

(5.24) believed $np_1^2\backslash s_2^0/s_3^1$ that $s_4^0/_{\$}s_5^0$ every dog $\dfrac{s_8^0 \mid s_6^1}{np_7^2}$ swam

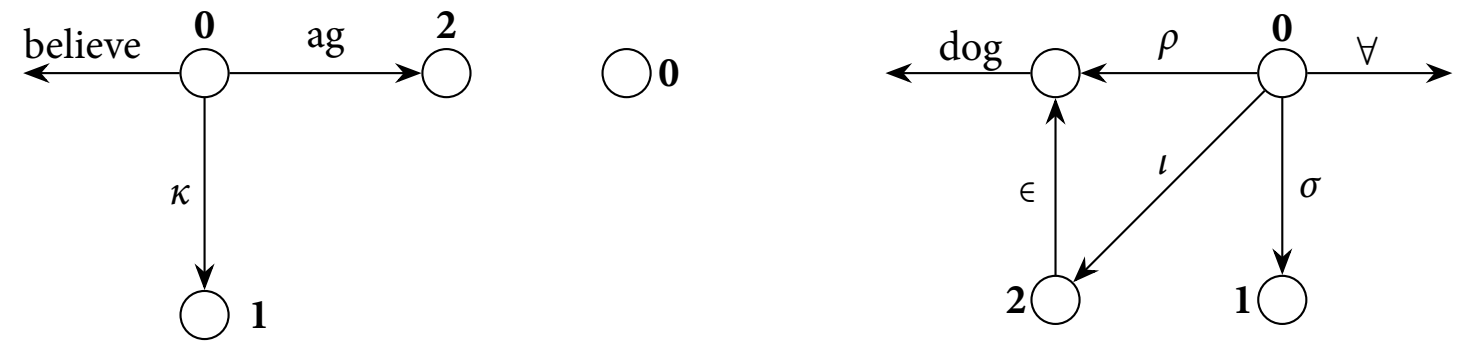


It is easy to verify that if the embedded clause is reduced to

$$\frac{s_8^0 \mid s_6^1}{s_4^0}$$

and combined with the main verb as is, *every dog* gets matrix scope, since the root ($s_2^0$) of *believed* is to equate with the scope ($s_6^1$) of the quantifier as in (5.25a). But if the tower above is lowered before it combines further, the scope of *every dog* is confined to the embedded clause, since its own root ($s_8^0$) is to equate with the $\kappa$-dependent ($s_3^1$) of *believed* as in (5.25b).

(5.25) a. (∀ *dog* > *believe*) b. (*believe* > ∀ *dog*)

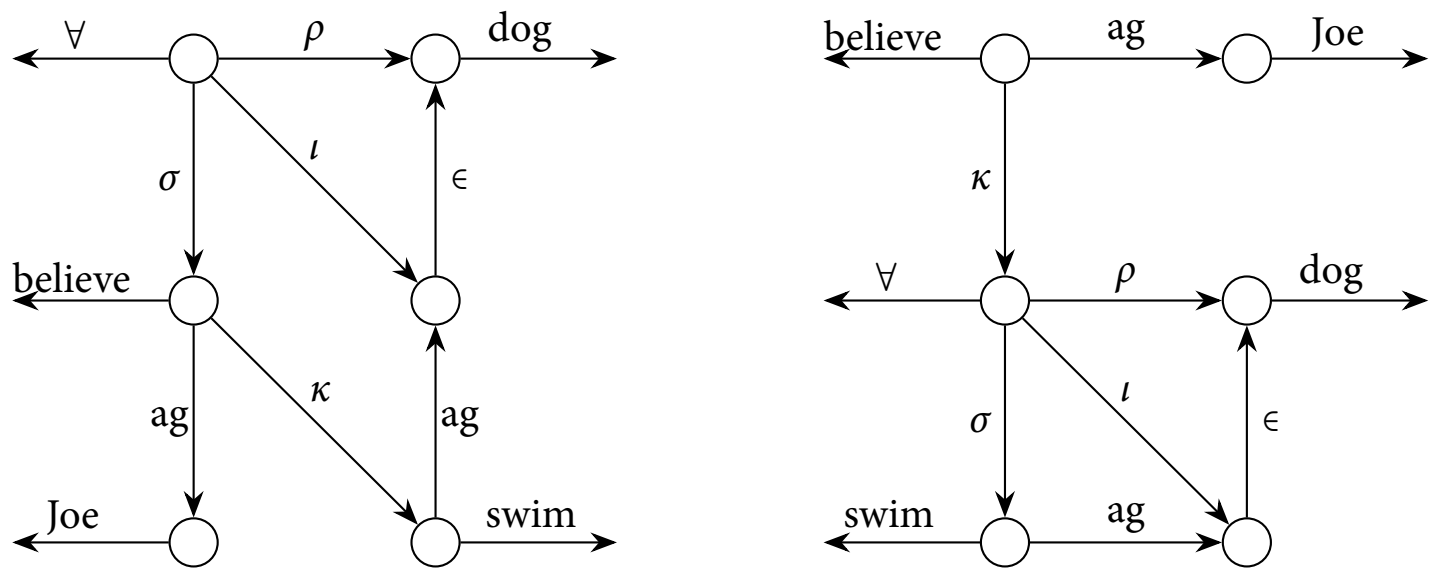


Therefore we can trap a quantifier within clausal boundaries by lowering a tower "early enough". In the case of relative clauses, we may also want to restrict the lifted application rule. This is because were the following reduction carried out, *every boy* would escape the island in (5.23b) for the reason just explained.

(5.26)

$$\begin{array}{ccc} \text{every boy} & \text{walked} & — \\ \dfrac{s_7 \mid s_5}{np_6} & np_8\backslash s_9/np_{10} & \dfrac{s_{13}\uparrow np_{14} \mid s_{11}}{np_{12}} \\ \vdots & \vdots & \vdots \end{array}$$

$$\cfrac{s_7 \mid s_5}{\cfrac{s_{13}\uparrow np_{14} \mid s_{11}}{s_9}}$$

$$\Downarrow$$

$$\ast\ \frac{\overset{\text{that}}{np_1\backslash_{\$}np_2/(s_3\uparrow np_4)} \qquad \dfrac{s_7 \mid s_5}{s_{13}\uparrow np_{14}}}{\dfrac{s_7 \mid s_5}{np_1\backslash_{\$}np_2}}$$

Thus we want to avoid matching relativizers with the non-tower occurrence in lifted application.

These intuitions can be captured if we set certain tokens as *island selectors* that prompt tower lowering. Specifically, we put a unary connective $\delta$ over the part of a selector type to be matched with an island:

(5.27) a. believe $\to np\backslash s/{}^{\delta}s$
b. that $\to np\backslash_{\$}np/{}^{\delta}(s{\uparrow}np)$
c. if $\to s/{}^{\delta}s/{}^{\delta}s$

Then by blocking lifted application

$$H,\ \frac{F \mid E}{G} \to \frac{F \mid E}{D} \qquad \frac{F \mid E}{G},\ H \to \frac{F \mid E}{D}$$

when $H,\ G \to D$ matches a pair of some $A$ occurrences, the one from $H$ marked by $\delta$, while discarding $\delta$ anywhere outside lifted application, the needed early lowering is enforced in (5.24) and the last step of (5.26) is prevented; one can verify $({}^{\delta}s_3,\ s_4)$ as the blocking match for the former, and $\left({}^{\delta}(s_3{\uparrow}np_4),\ s_{13}{\uparrow}np_{14}\right)$ as that for the latter.

Hence we arrive at a lexicalized treatment of scope islands. The idea of specifying islandhood with unary connectives is due to Morrill (1992). An in-depth application to scope islands of delimited continuations (Danvy and Filinski, 1990; Felleisen, 1988), the programming-language concept behind prompted lowering, can be found in Charlow (2014). In the context of practical parsing, White et al. (2017) showed that a careful timing of lowering curbs spurious ambiguity (i.e. structural ambiguity of no semantic consequence) and thus improves efficiency. This result gives us a concrete measure of the computation cost that a locality constraint can save.

Most recently, Barker (2021) affirmed a suspicion that has been circulating in the literature: scope islands are much more lexical than previously recognized, since certain island selectors only trap certain scope takers. In contrast with (5.23a), *every dog* can be scoped out of the complement to *ensured* (see Farkas and Giannakidou, 1996 a.o.):

(5.28) Joe ensured [that every dog swam].

To take such lexical agreement into account, we can mark the scope atom of a quantifier with a key $k$ (another unary connective to be discarded outside lifted application), reify $\delta$ above as a lookup set, and block lifted application only when $\delta$ in $H$ contains $k$:

$$H,\ \frac{F \mid {}^{k}E}{G} \to \frac{F \mid {}^{k}E}{D} \qquad \frac{F \mid {}^{k}E}{G},\ H \to \frac{F \mid {}^{k}E}{D}$$

Thus, again taking (5.26) as an example, we may say that the sequence

$$np_1\backslash_{\$}np_2/{}^{\delta}(s_3{\uparrow}np_4),\ \frac{s_7 \mid {}^{k}s_5}{s_{13}{\uparrow}np_{14}}$$

is irreducible on the grounds that $k \in \delta$, where $k$ and $\delta$ are lexically given by *that* and *every*. (Barker coded $\delta$ and $k$ numerically, based on his generalization that lookup sets are totally ordered by inclusion.)

One last question to ask is how quantifiers escape infinitive clauses. For (5.29), the paraphrased reading does exist, contra (5.23a):

(5.29) Joe believed every dog to swim.
(= For every dog $x$, Joe believed $x$ to swim.)

A lexicalized generalization about scope islands leads to the conjecture that *believed* in (5.23a) and (5.29) differ in selection properties, or rather, types. This conjecture is supported by the following examples of co-reference:

(5.30) a. **Joe**$^i$ believed that **he**$_i$/***himself**$_i$ saw a dog.
b. **Joe**$^i$ believed ***he**$_i$/**himself**$_i$ to see a dog.

Suppose *to* is an identity element (see below), as semanticists often do. Then given our discussion in Section 3.2.2, there has to be a structural difference, which we proceed to show, between the semantic contribution of *believed* in (5.23a) and that in (5.29) to result in "longer" parallel paths in the former but "shorter" ones in the latter. From atom-vertex correspondence we expect the structural difference to be somehow echoed in types.

Here we give an entry for *believed* in (5.29) that is consistent with our expectations.

(5.31) a. believe → $np^2\backslash(s^0/(np^4\backslash s^3))/np^1$ b. to → $np^1\backslash s^0/_{\$}(np^1\backslash s^0)$

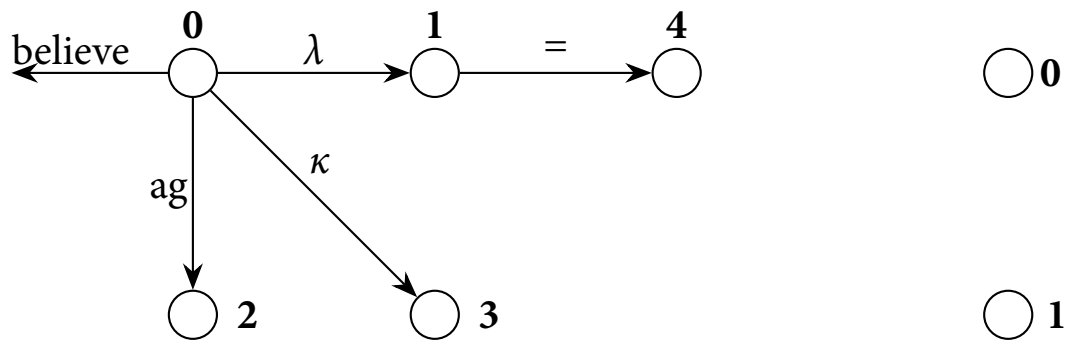


Now that no part in the type of (5.31a) is to match an island, one can easily construct (5.32) for (5.29). We leave it to the readers to verify that (5.32) represents the desired semantics.

(5.32)

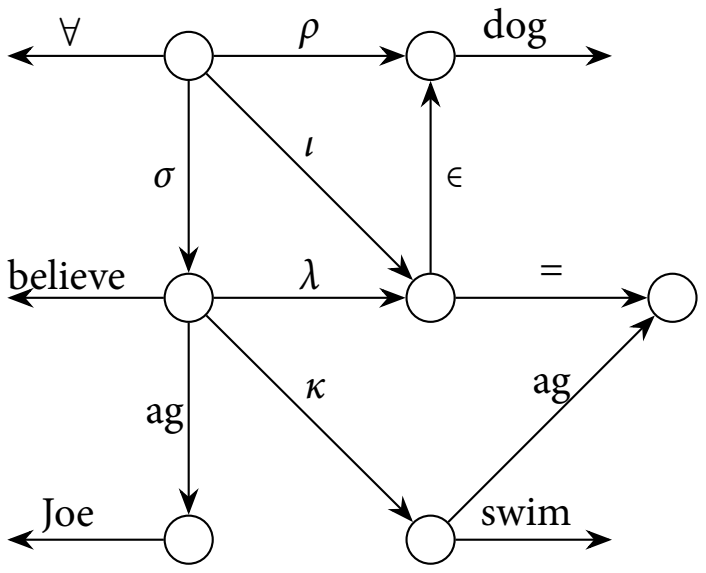


By way of digression let us consider (5.30a&b), which also motivate (5.31).

(5.33) a. **Joe**$^i$ believed that **he**$_i$ saw a dog. b. **Joe**$^i$ believed **himself**$_i$ to see a dog.

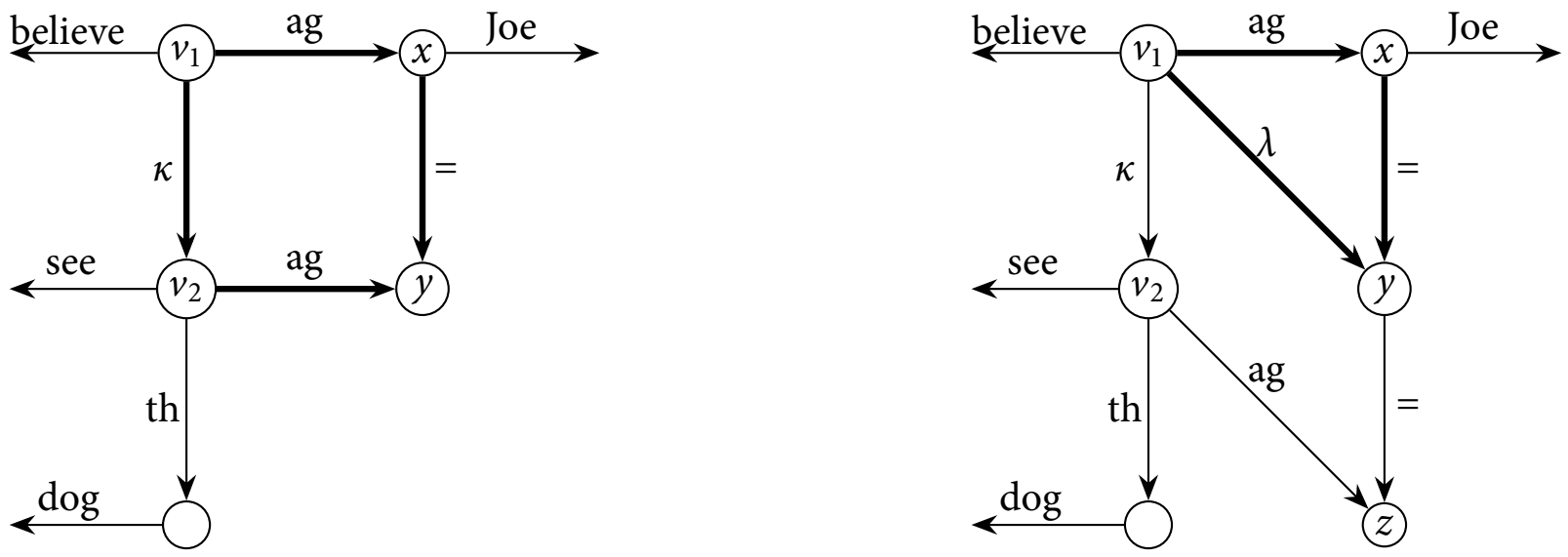

Whereas $= \overrightarrow{yz}$ in (5.33b) comes with (5.31a), $= \overrightarrow{xy}$ is contributed by the anaphor in either case. We have the relevant pair of parallel paths ($v_1 \to v_2 \to y \parallel v_1 \to x \to y$) obeying the antilocality scheme in (5.33a) and the one ($v_1 \to y \parallel v_1 \to x \to y$) obeying the locality scheme in (5.33b). As the depths in the entries for *believe* suggest, *ag* is less oblique than either $\kappa$ or $\lambda$.

Still, (5.31a) disallows an object reflexive to refer beyond an infinitive clause. One can be convinced by checking the parallel paths indicated as follows.

(5.34) a. Joe believed **Ben**$^i$ to see **himself**$_i$.

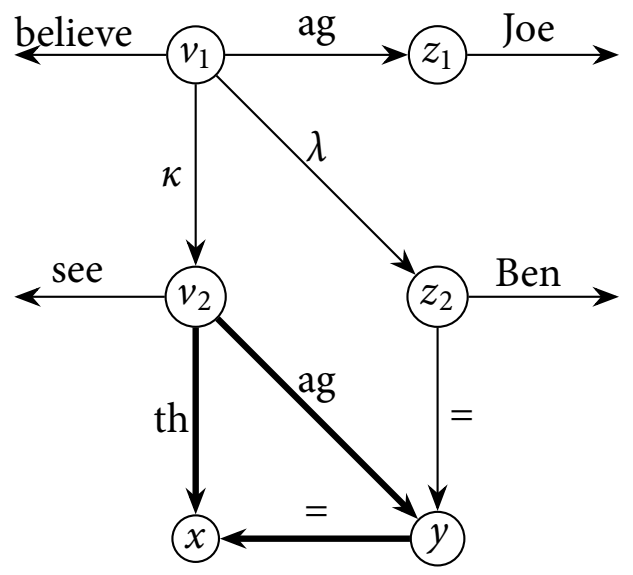


b. ***Joe**$^i$ believed Ben to see **himself**$_i$.

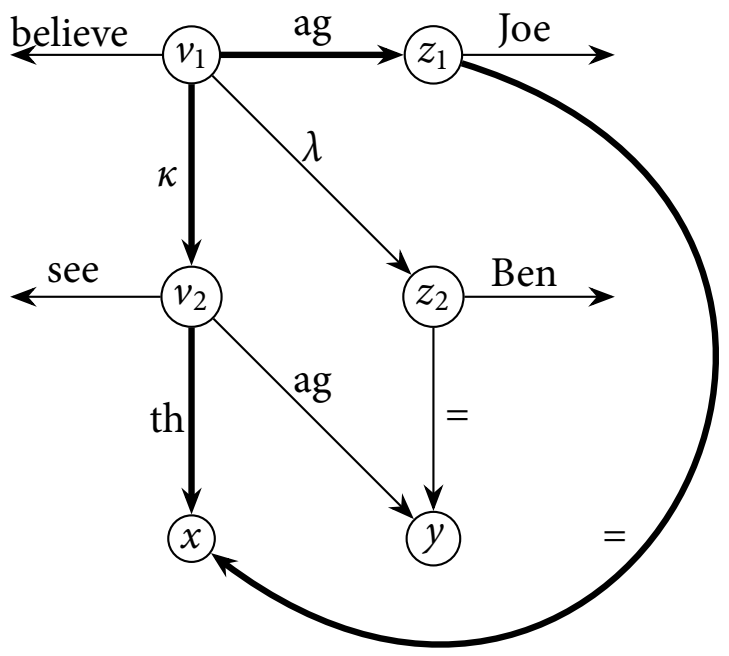


Note that in (5.34a), the tail of the equality edge introduced by *himself* equates with the 4-source of (5.31a). It is semantically equivalent for it to equate directly with the referent introduced by *Ben* ($z_2$), but as one can verify, doing so would create a pair of parallel paths violating the locality requirement. Thus *himself* is technically resolved to (a referent introduced by) *believed*.

One might have realized that (5.31a) conforms to the syntax of control, with the 4-source playing the role of PRO (see Chomsky, 1981). Yet it is neither subject- nor object-control in the traditional sense, since the 1-source, corresponding to the direct object like *Ben* in (5.34), is not a thematic dependent of the verb.

### 5.1.3 Existential vs. distributive scope

The distinction between "existential scope" and "distributive scope", a terminology due to Szabolcsi (2010, chap. 7), describes the fact that an existentially quantified plurality can take wider scope than its distributive quantificational force (see Reinhart, 1997, 2006, a.o.).

Consider the following example. In verifying its truth, there are some nuances about in what situations we expect Joe to feed dogs. First, Joe may take care of any three dogs, say, on a farm he owns, or just three specific dogs entrusted to him. Then, Joe may reward the dogs for jointly pulling a work sled that harnesses three, or he may reward them for each pulling a toy sled that harnesses one; that is, the conditional's antecedent can be interpreted either collectively or distributively (see Section 4.1).

(5.35) If three dogs pulled a sled, Joe fed them.

Thus we can distinguish at least four readings for (5.35), of which we are now interested in the two concerning the situations where each dog pulled a (toy) sled: *three dogs* may take existential scope either under or over the conditional, but in either case its distributive scope stays within the

antecedent clause, which means (5.35) never requires Joe to feed three dogs when only one of them pulled a sled.

These two readings can be constructed as (5.36b) using (5.36a) (which is for the sake of simplicity; to justify the relative obliqueness of $\iota$ and $\sigma$ and in view of postposed *if*-clauses, type *if* as $(s^2\backslash s^1)\uparrow_{\$}(s^2\backslash(s^1/s^3))\downarrow(s^0/s^2)$ and keep the source label assignment for quantificational determiners).

(5.36) a. if $\rightarrow s^0/s^2/s^1$ b. If three dogs DIST pulled a sled, Joe fed them.

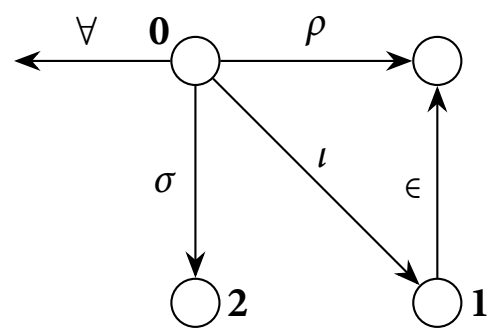


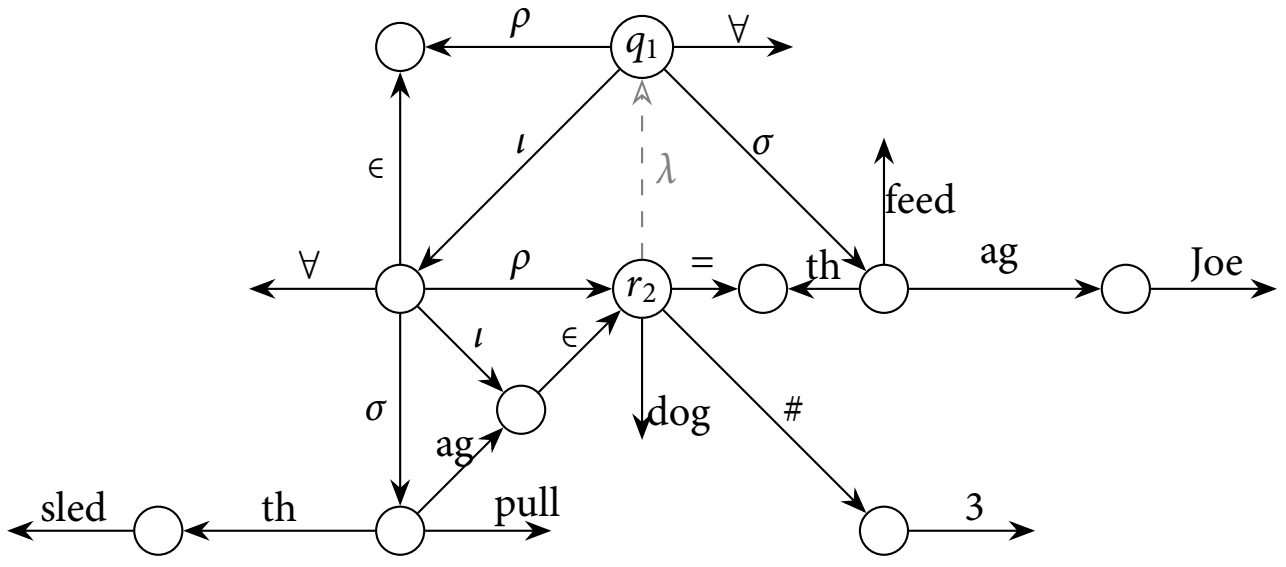


The absence and presence of $\lambda\overrightarrow{r_2 q_1}$ respectively indicate the narrow and wide existential scope of *three dogs*. Without $\lambda\overrightarrow{r_2 q_1}$ the semgraph is rooted by $q_1$, and $r_2$ can assume any set of three dogs; but if $r_2$ is resolved to precede $q_1$ (see (4.3) for the entry of numerals), then being the new root, $r_2$ is fixed to some set of three dogs before we check through iteration whether Joe fed them whenever they each pulled a sled.

Our discussion above suggests a way to treat the so-called "third reading" of quantifiers embedded in intensional contexts (von Fintel and Heim, 2011, chap. 8; Fodor, 1970). As noted in Section 5.1.2, a proposition attitude typically traps a quantifier within its complement, so the most natural reading of (5.37) attributes to Joe the belief "most huskies howled" (at each world $w$ compatible with Joe's belief, most huskies there howled).

(5.37) Joe believed that most huskies howled.

With $X$ denoting the set of huskies at the reference world, a reading (5.37) does *not* have attributes to Joe the belief "$x$ howled" for most $x$ in $X$ (this reading differs from the previous one as Joe's belief worlds may identify a different set of huskies from $X$). However, a subtle third reading (5.37) does have attributes to Joe the belief "most $x$ in $X$ howled" (conceivable in a context where Joe saw a group of huskies howling, but his opinion about their species is unknown).

*Remark.* To see that the third reading differs from the nonexistent one, consider a model where there are two worlds compatible with Joe's belief and three huskies at the reference world. The matrix

$$\begin{bmatrix} 0 & 1 & 1 \\ 1 & 0 & 1 \end{bmatrix}$$

puts 1 in its $i$'th row and $j$'th column when the $j$'th husky howled at the $i$'th world. Since at both worlds two out of three huskies howled, the third reading is true; but since one out of three huskies howled at both worlds, the nonexistent reading is false.

We may say that in the third reading of (5.37), *most huskies* takes wide existential scope but narrow distributive scope. If we keep quantificational determiners' type but also allow for the semantics in (5.38a), which adds to the restrictor a $\lambda$-edge pending precedence resolution, we can approximate the third reading in (5.38b) by resolving $r$ to precede $v$.

(5.38) a.

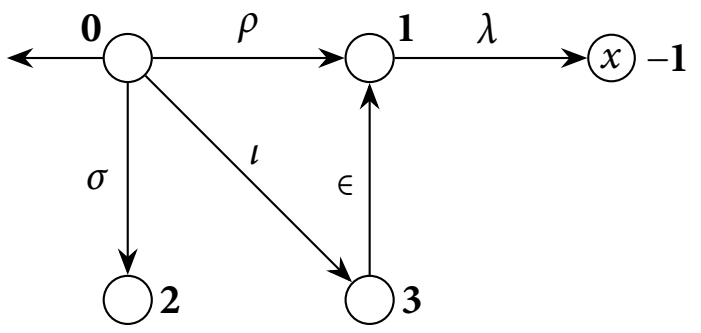

b.

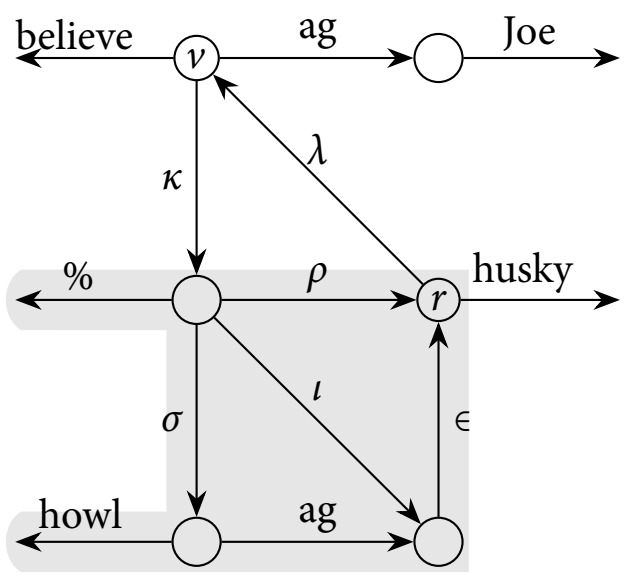

The shaded area delimits the content of Joe's belief, properly containing the iteration but excluding any out-edge of $r$. This gives *most huskies* narrow distributive scope but wide existential scope.

We are faced with two problems in this treatment. One is that $r$ might not be maximized with respect to its descriptive content as it takes scope. Recall the interpretation of a quantification structure requires the restrictor be maximal subject to satisfying the restrictor subgraph, unless it has been visited (see Definition 2.4). This is desirable for (5.36b) (otherwise $r_2$ would not be fixed), but it also means that for (5.38b), we can take any set $r$ of huskies at the reference world and verify that most of its members howled at Joe's belief worlds; in a context where five out of twenty huskies howled at every belief world of Joe's, (5.38b) can be wrongly satisfied by assigning $r$ small enough a set of huskies containing those five. To address this problem, we need to factor the maximality requirement on the restrictor out of the interpretation of quantification. We leave the implementation of this idea for the future.

The next problem is that, as a quantifier's modifier serves as its iterator filter, the latter will not trail the restrictor in scope taking. For example, the relative clause below does not take scope with $r$, but contributes to the content of Joe's belief.

(5.39) Joe believed that most huskies [who pulled a sled] howled.

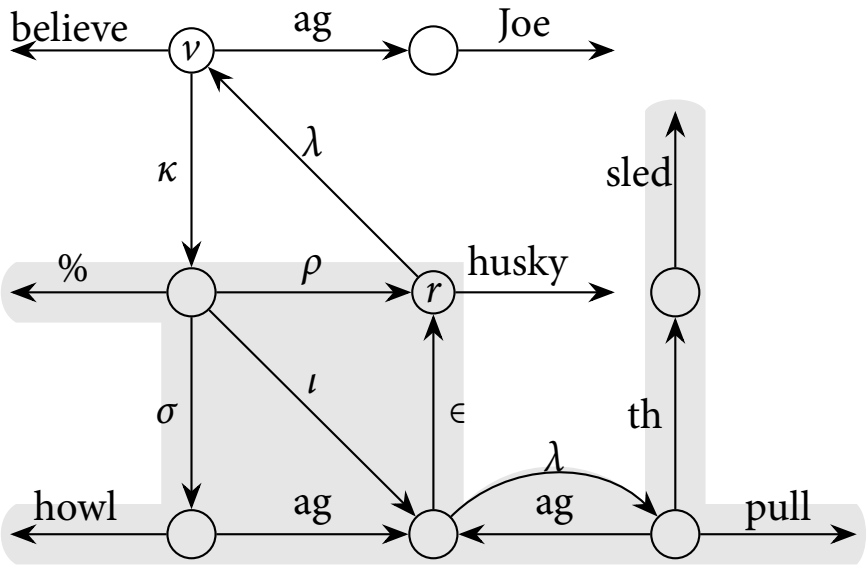

With $X$ denoting the set of huskies at the reference world, the semgraph attributes to Joe the belief "most $x$ in $X$ who pulled a sled howled". This is subtly different from attributing to Joe the belief "most $x$ in $X'$ howled", where $X' \subseteq X$ denotes the set of huskies who pulled a sled at the reference world. The second attribution might describe a possible reading of (5.39) (we can scope *a sled* by precedence resolution), but, unfortunately, it is hard to judge clearly if the first does too (but see Keshet, 2008, chap. 2; Romoli and Sudo, 2009).

In any case, for indefinites, the judgment seems robust that their modifiers form part of their descriptive contents and trail their scope. We will see how this happens in Section 5.2.1.

## 5.2 Scoping indefinites

We have encountered examples of scope-taking indefinites since Section 2.2.6. In Section 3.2.3 we introduced precedence resolution for scoping indefinites, not only because they receive a representation different from that of quantifiers, but also in view of their ability to escape almost every scope island that would trap the latter (see Brasoveanu and Farkas, 2011; Charlow, 2019; Farkas, 1981; Fodor and Sag, 1982; Reinhart, 1997, a.o.).

### 5.2.1 General remarks

Before taking up more specific questions, we remark on a simple property of the current scoping mechanism: by shifting the order of valuation, a referent is always scoped together with its descriptive constraint (with the only exception of empty disjuncts; see Section 5.2.3). This property follows directly from the definition of the semgraph interpreter, and steers us clear of a few commonly known pitfalls discussed below.

We begin with the so-called "Donald Duck problem" (a.o. Reinhart, 1997), as illustrated by the following example:

(5.40) If **a star** shines, Joe sails ($\exists$ *star* $>$ *if*)

($\neq$ for some $x$, if $x$ is a **star** and shines, Joe sails).

The incorrect paraphrase, by raising the existential only, holds in a situation where there was, say, a duck but no star at all, whereas the wide-scope-indefinite reading of (5.40) actually requires the existence of a specific star. The correct paraphrase is thus given below with a synonymous semgraph.

(5.41) a. For some **star** $x$, if $x$ shines, Joe sails.

b. 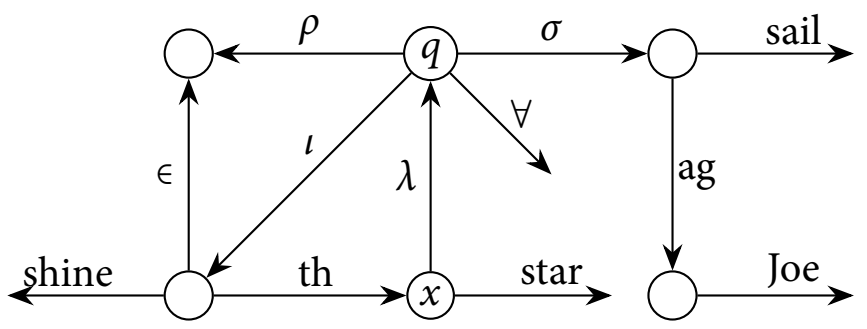

Preceding $q$ and being the root, $x$ has to be stars to satisfy (5.41b). Similarly, given the graph structure of modification, when *a dog* takes scope in (5.42), it is automatically accompanied by the relative clause and the indefinite inside, as can be seen from their absence in the content of Joe's belief (the shaded area).

(5.42) Joe believed that **a dog which saw a duck** swam. ($\exists$ *dog*, *duck* > *believe*)

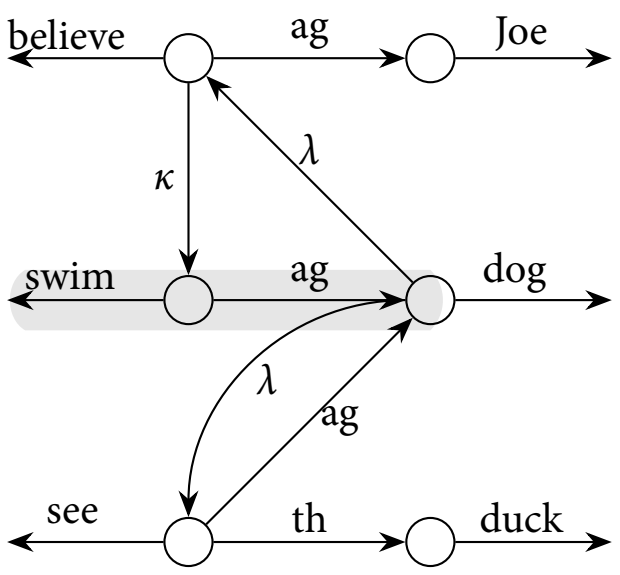

Indeed, (5.42) does not have a reading where Joe had a belief about a specific dog but no specific duck at the reference world.

Closely related to the preceding is the "empty set problem". To illustrate with (5.40) again, (5.41) is false, much as (5.43) would be, when there was no star, that is, when the predicate *star* is assigned the empty set by the reference world. This is because the rule for interpreting unary edges is so stated that $x$ be nonempty (see Definition 2.3), which is implied if we liken the article *a* to the numeral *one*.

(5.43) A star shines.

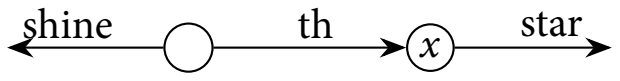

Geurts (2000) among others noted that an empty-set situation poses a problem for choice-function approaches to indefinites as presented in Reinhart (1997); Winter (1997). Roughly speaking, since the way those approaches avoid the Donald Duck problem still leaves in situ the descriptive constraints of indefinites, it is difficult for them to falsify both (5.40) and (5.43) as requried.

Finally, the descriptive constraint of a scoping indefinite may carry a pronoun bound by an outscoping quantifier. As an example, consider (5.44) (adapted from Abusch, 1994, p. 90) and a reading where the toy and its throw depend on the boy but not the dog.

(5.44) Every boy$^i$ fed every dog which caught **a toy he$_i$ threw**. ($\forall$ *boy* > $\exists$ *toy* > $\forall$ *dog*)

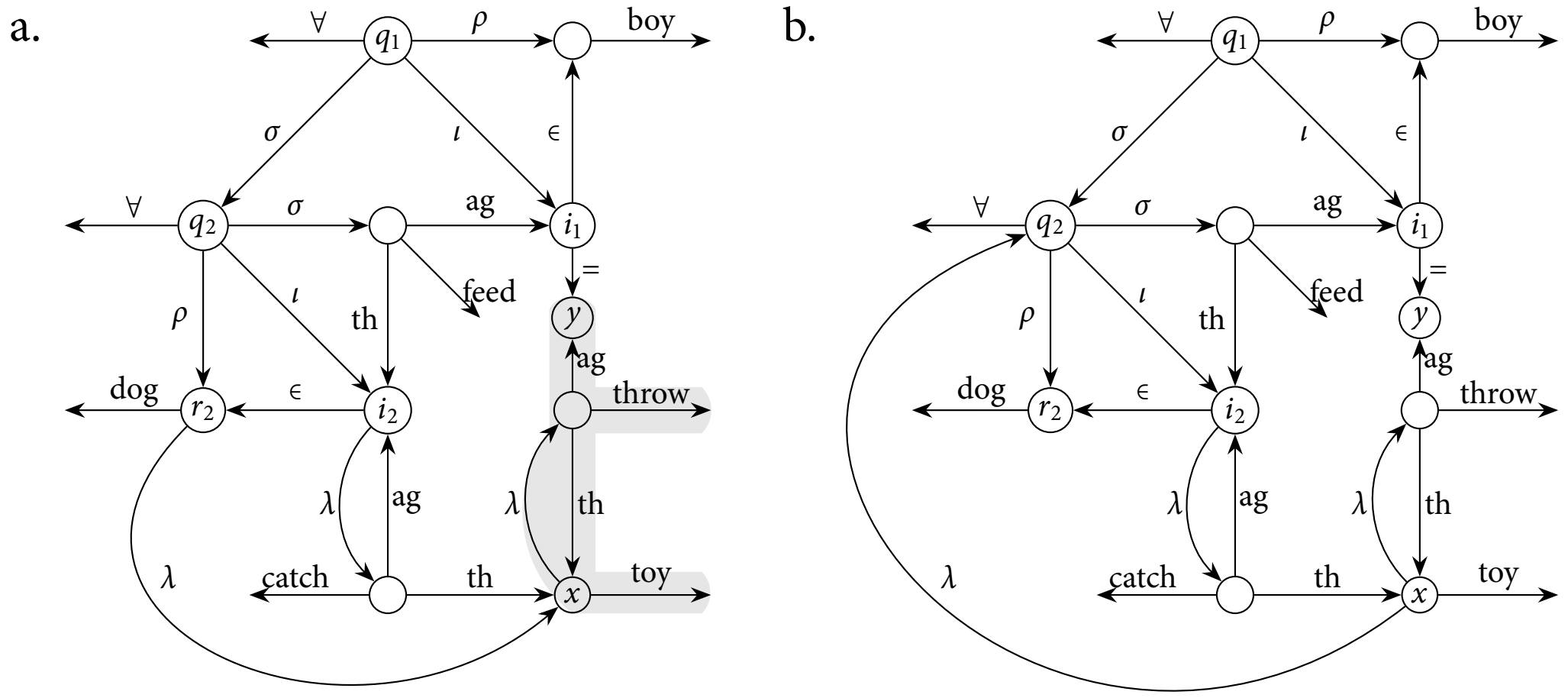

In (5.44a) we use the entry (5.38a) and resolve $r_2$ to "precede" $x$; the purpose is not to ensure $r_2$ be valuated before $x$ (it already is), but to admit $x$ into the restrictor subgraph off $r_2$ without committing $x$ to any concrete semantic dependency on $r_2$ (by virtue of $\lambda$-edges), so that the value of $x$ can be fixed across each (singleton of) dog $i_2$. But since $x$ remains valuated after each (singleton of) boy $i_1$, we obtain the desired scope relation. (Note that scoping the shaded descriptive constraint by $\lambda\overrightarrow{r_2 x}$ does not give $y$ wider scope than what it would otherwise have.) One may wonder why taking this route when the more intuitive alternative is to directly resolve $x$ to precede $q_2$, as in (5.44b). This is because doing so results in multi-rootedness. Readers may check that $\lambda\overrightarrow{x q_2}$ makes $x$ a root without canceling the rootship of $q_1$.

Examples like (5.44) and many of its variants present the “bound pronoun problem”, leading almost every description-in-situ account of indefinites to unexpected predictions of various kinds (see Abusch, 1994; Charlow, 2019; Geurts, 2000; Jäger, 2007; Kratzer, 1998). Due to its relevance to the interaction between co-reference resolution and precedence resolution in general, a special case of the bound pronoun problem, concerning how far indefinites carrying a bound pronoun can take scope (see Brasoveanu and Farkas, 2011; Schwarz, 2001), will be discussed in Section 5.2.4.

### 5.2.2 Exceptional scope

Indefinites are said to take *exceptional scope*, a term that goes back to Fodor and Sag (1982), when they seem to escape scope islands discussed in Section 5.1.2. Previously we saw that neither a relative clause, nor the complement of an attitude, nor the antecedent of a conditional traps an indefinite:

(5.45) a. Every boy fed every dog [which caught **a toy** he threw]. ($\exists$ *toy* > $\forall$ *dog*)
b. Joe believed [that **a dog** which saw a duck swam]. ($\exists$ *dog* > *believe*)
c. If [**a star** shines], Joe sails. ($\exists$ *star* > *if*)

Constructing semgraphs for such indefinites, as demonstrated in Section 3.3.4, is mostly a matter of resolving their precedence over the other scope-sensitive expression. This allows us to easily capture the oft-observed non-determinism in exceptional scope.

On the one hand, how wide or narrow an indefinite may stretch its scope is in principle unlimited. Without interaction with anaphora as in (5.44)/(5.45a) (see Section 5.2.4), the indefinite in examples like (5.46) can take either intermediate or matrix scope, as noted by Farkas (1981).

(5.46) Every boy fed every dog which caught **a toy**.

a. ($\forall$ *boy* > $\exists$ *toy* > $\forall$ *dog*)

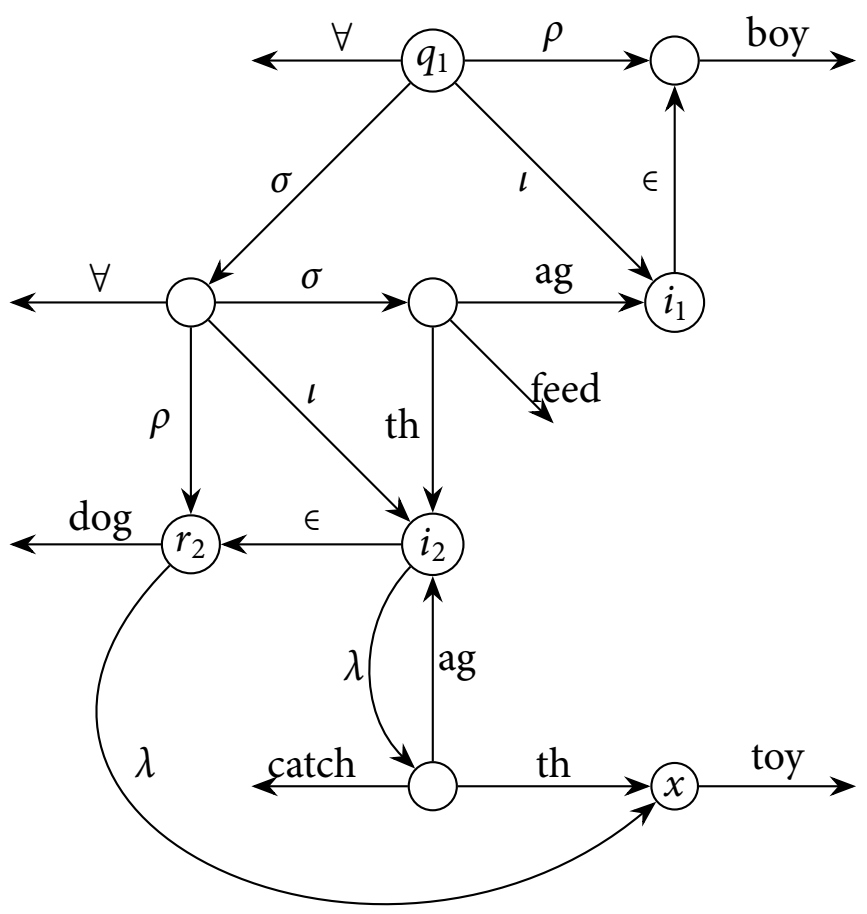

b. ($\exists$ *toy* > $\forall$ *boy* > $\forall$ *dog*)

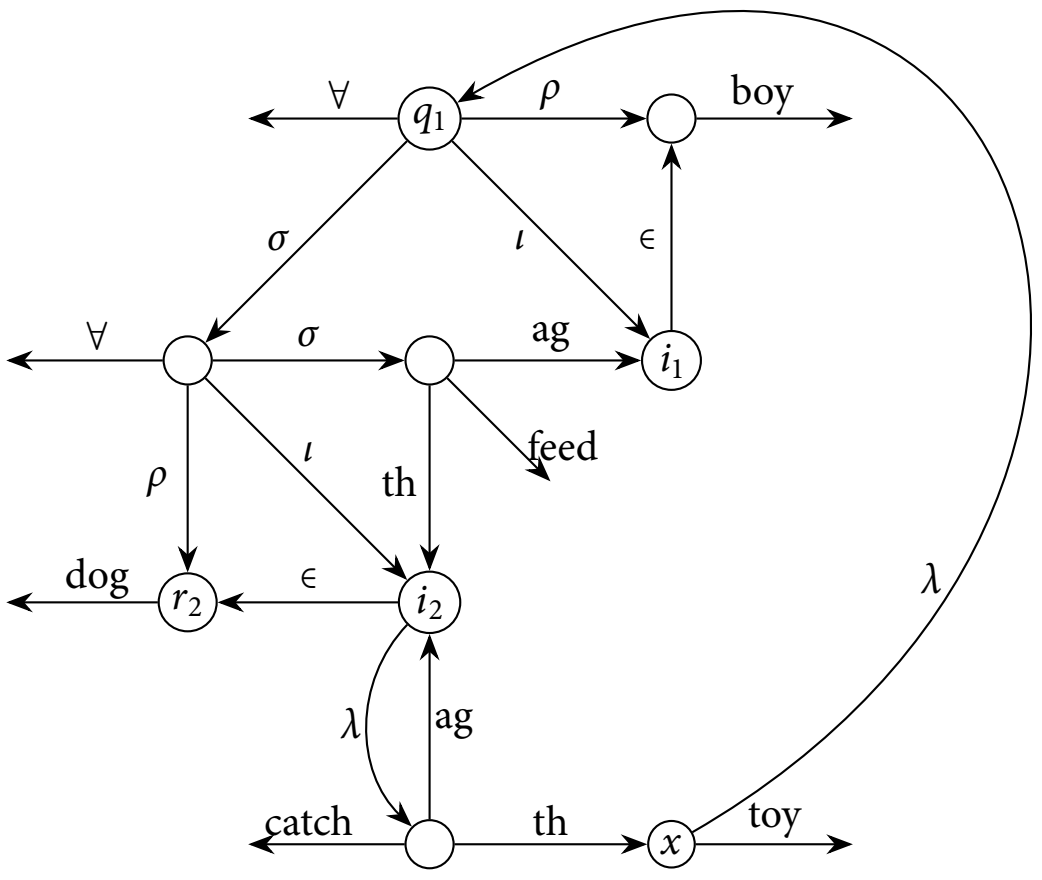

The indefinite is valuated after the iterator of *every boy* in (5.46a), but before that in (5.46b).

On the other hand, whereas modification forces one indefinite to accompany the other to take scope in (5.42)/(5.45b), Charlow (2019) showed that syntactically independent islanders can be scoped independently. In the following example, we may associate Joe's sail with a specific comet *but* no specific star, with no specific comet *but* a specific star, or with a specific comet *and* a specific star.

(5.47) If **a comet** passes **a star**, Joe sails.

Thus we can scope either indefinite by making it precede *if*:

(5.48) a. (∃ *comet* > *if* > ∃ *star*)

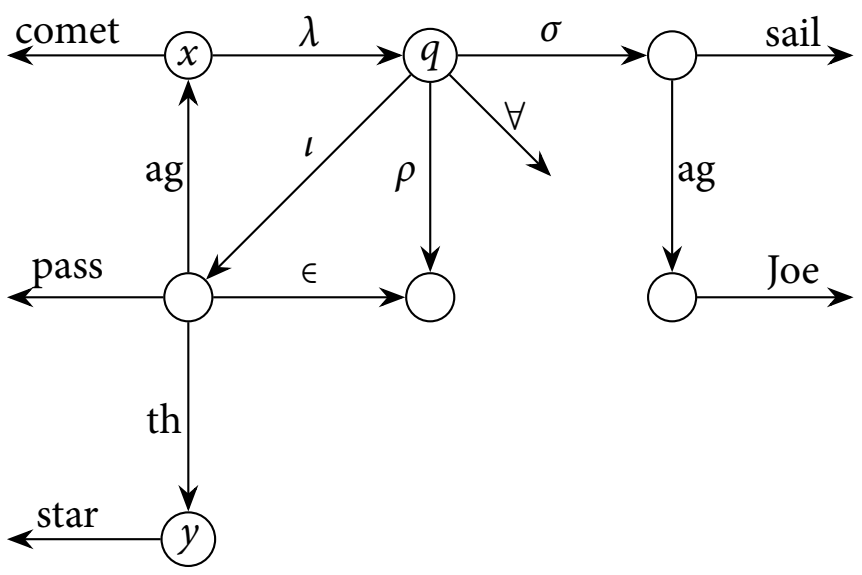

b. (∃ *star* > *if* > ∃ *comet*)

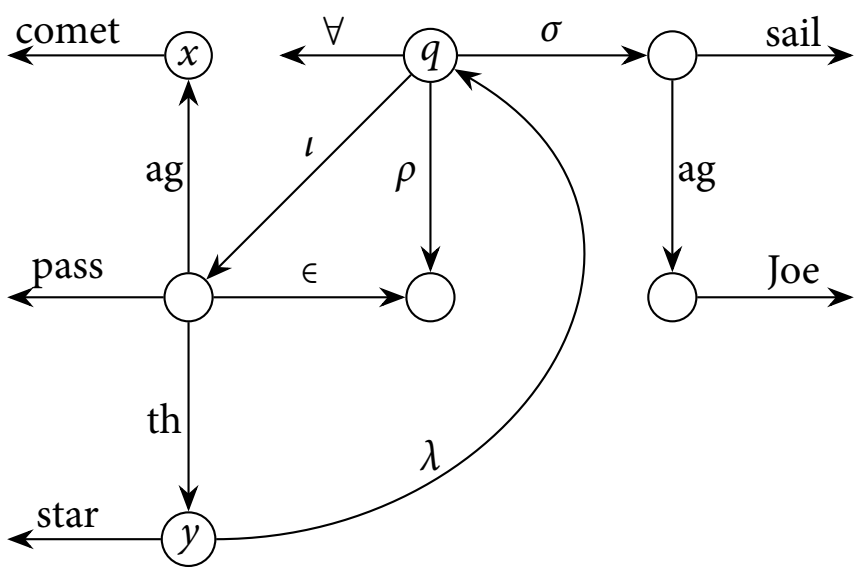

Making the other indefinite precede the one scoped as above then yields wide scope for both:

(5.49) (∃ *comet*, *star* > *if*)

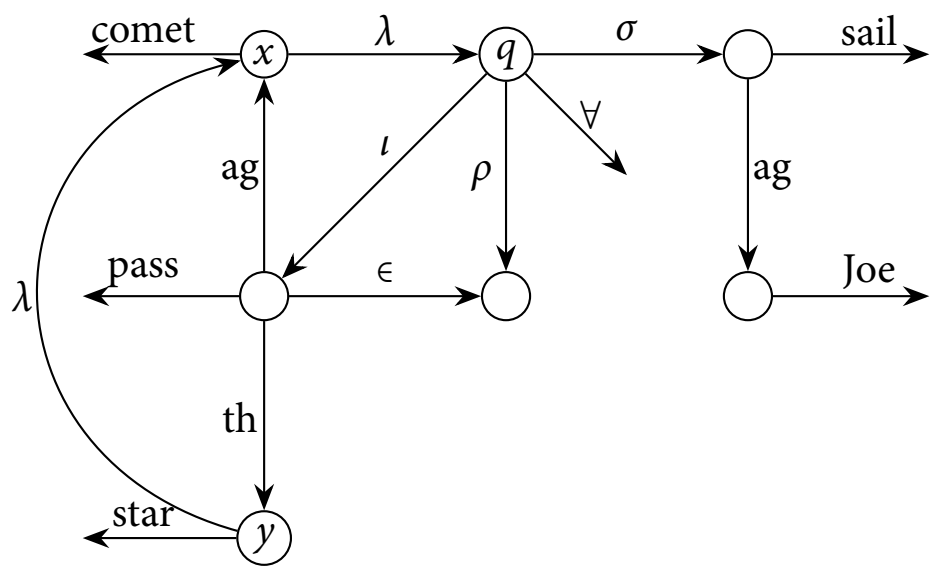

(Obviously, matrix-scope indefinites taking scope over one another make no semantic difference.)

We may even find one indefinite taking matrix scope and the other, from the same island, intermediate scope, as shown by an example adapted from Charlow (2019, p. 453).

(5.50) Every boy$^i$ smiled if **a dog** caught **a toy he$_i$ threw**. (∃ *dog* > ∀ *boy* > ∃ *toy* > *if*)

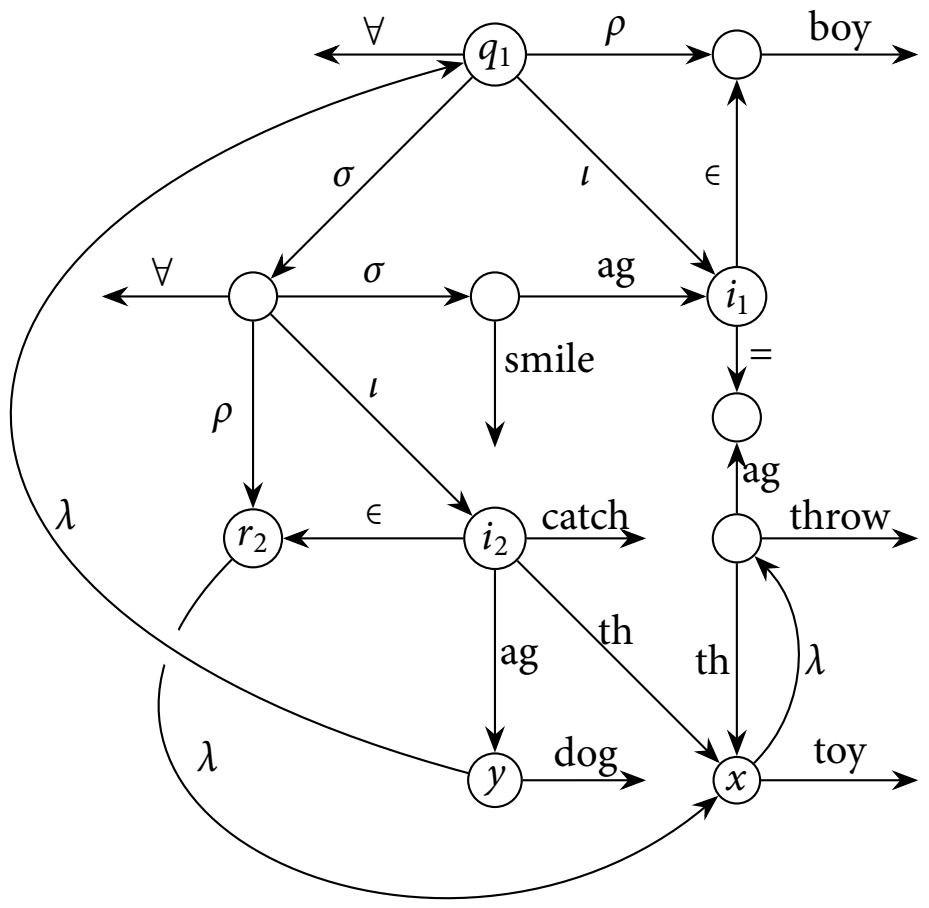

Precedence resolution here combines the patterns of (5.46a&b). As for the postposed *if*, we type it as $s^2\backslash s^0/s^1$ while keeping the semgraph in (5.36a).

### 5.2.3 Wide-scope disjunct

Towards the end of Section 2.2.6, we mentioned that there is no reason to satisfy an unchosen disjunct subgraph merely because an indefinite disjunct takes scope. Our semgraph interpreter therefore explicitly suspends processing all the out-edges of a scoping disjunct except for the $\lambda$-edge by which it takes scope. From this suspension derives what might be called "wide-scope-disjunct" readings, which differ in an interesting way from the "wide-scope-disjunction" readings discussed by Brasoveanu and Farkas (2011); Charlow (2014, 2020); Rooth and Partee (1982); Schlenker (2006).

The difference can be illustrated with the following example (adapted from Schlenker, 2006, p. 306). With the disjunction taking narrow scope, (5.51) is to be verified by checking on any pet walker.

(5.51) **Every boy who walked** a dog or$^i$ a cat **fed it**$_i$.

The wide-scope-disjunction reading, on the other hand, can be paraphrased as follows:

(5.52) Every boy who walked a dog$^i$ fed it$_i$,
*or* every boy who walked a cat$^j$ fed it$_j$.

This reading allows us to skip checking on, say, cat walkers if all dog walkers fed their dogs.

It is unclear how this reading can even be represented without duplicating the semantic resources introduced by the boldfaced tokens in (5.51) (no wonder existing accounts all build on Partee and Rooth's (1983) argument sharing technique; see the end remark of Section 4.2.2). Naturally, it will not be what we get from scoping either disjunct. Consider what happens when *a dog* takes wide scope:

(5.53)

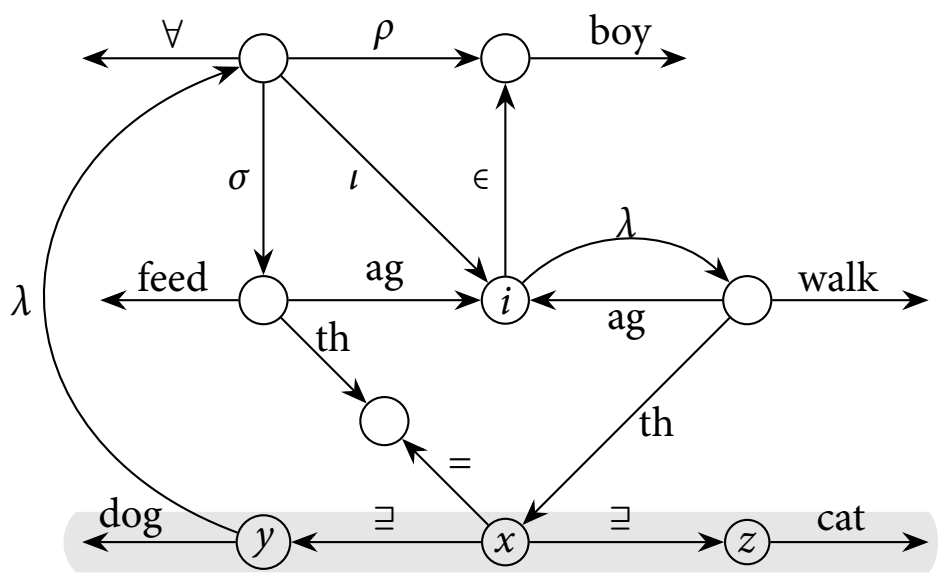

As $y$ becomes the root, the evaluation of $dog\,\overrightarrow{y}$ is suspended. Later, to satisfy the shaded subgraph contained in the iterator filter on $i$ (see Definition 2.8), the choices open to $x$ depend on how $y$ is valuated earlier: if $y$ settles on a value that satisfies $dog\,\overrightarrow{y}$, $x$ can choose between $y$ can $z$; otherwise, $x$ can only choose $z$.

As a consequence, to verify (5.53), we may need to check on all boys who walked a specific dog $y$ and all those who walked a cat (when $dog\,\overrightarrow{y}$ is satisfied), or we may only need to check on all those who walked a cat (when $dog\,\overrightarrow{y}$ is unsatisfied). That means (5.53) represents a wide-scope-disjunct reading that can be paraphrased thus:

(5.54) For some dog $y$, every boy who walked $y$ or$^i$ a cat fed it$_i$,
*or* every boy who walked a cat$^j$ fed it$_j$.

A symmetric wide-scope-disjunct reading then derives from scoping *a cat*:

(5.55) For some cat $z$, every boy who walked a dog or$^i$ $z$ fed it$_i$,
*or* every boy who walked a dog$^j$ fed it$_j$.

(5.54) and (5.55) each share one line with (5.52), while the situations satisfying their other lines arguably satisfy (5.51). Thus we have constructed not the wide-scope-disjunction reading proper, but two wide-scope-disjunct readings that jointly cover the latter.

We may similarly consider the "wide-scope-conjunct" and "wide-scope-conjunction" readings of the following sentence.

(5.56) Every boy who walked a dog and$^i$ a cat fed them$_i$.

The wide-scope-conjunction reading, generally considered impossible in the literature, can be paraphrased as follows.

(5.57) Every boy who walked a dog$^i$ fed it$_i$,
*and* every boy who walked a cat$^j$ fed it$_j$.

Following the construction of (5.53) and the interpretation rule of conjunction structures, readers should be able to verify that the two wide-scope-conjunct readings of (5.56), resulting from scoping *a dog* and *a cat*, can be paraphrased as follows.

(5.58) a. For some dog $y$, every boy who walked $y$ and$^i$ a cat fed them$_i$.
b. For some cat $z$, every boy who walked a dog and$^i$ $z$ fed them$_i$.

What distinguishes (5.58a) from (5.54), or (5.58b) from (5.55), is that the former always requires the existence of some dog, or some cat. This is because unlike the semantics of choice, the semantics of summation requires both conjunct subgraphs be satisfied, whether either conjunct takes scope or not. One can see that wide-scope-*conjunct* readings do not relate to wide-scope-*conjunction* readings in any way nearly as interesting as do wide-scope-*disjunct* readings to wide-scope-*disjunction* readings.

### 5.2.4 Interaction with co-reference

The interaction of co-reference and precedence resolutions leads to constraints on both sides.

Taking the perspective of anaphors, we may find their co-reference options constrained by the scope of indefinite antecedents. This is the subject of anaphoric accessibility dealt with in dynamic semantic theories (see Geurts, 2011 for a review), of which DRT is an example. As mentioned in Section 2.3.3, the essence of the matter consists in the way valuation dependency is managed to ensure that quantification works out as desired (see Definitions 2.5/2.6).

As a classic example, consider how a narrow-scope indefinite is inaccessible to a quantifier-external anaphor. In both (5.59a&b) *it* cannot pick up and covary with *a dog* when no specific dog is intended.

(5.59) a. Every boy who walked ***a dog**$^i$ swam and ***it**$_i$ barked. ($\forall$ *boy* > $\exists$ *dog*)

b. Joe didn't walked **a dog**$^i$ and ***it**$_i$ barked. $(\neg > \exists\ dog)$

Let us check the semgraphs implied by these invalid co-references:

(5.60) a.

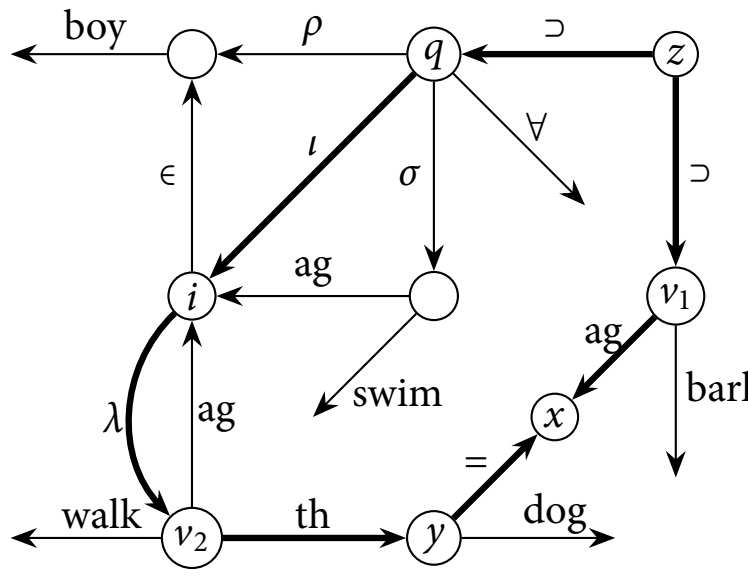

b.

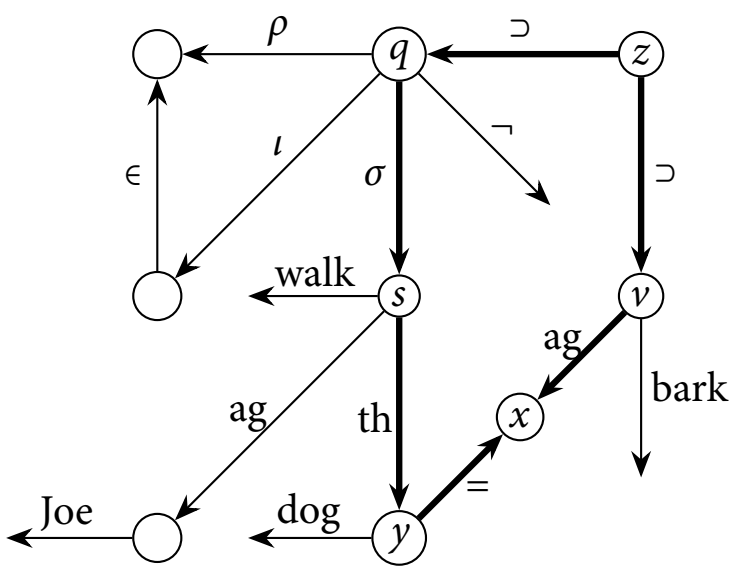

A valuation $g$ satisfying (5.60a) must be undefined on $x$, since $x$ remains an unvisited vertex of the iterator subgraph when the conjunct at $q$ is evaluated with respect to any $h \supseteq g$. Any $k \supseteq g$ satisfying the conjunct at $v$ can thus assign $x$ (the barker) any value, independently of any extension $l \supseteq h$ used for iterating over dog-walking boys. In other words, *it* in (5.59a) is interpreted as a free pronoun despite being resolved to *a dog*. A similar reasoning applies to (5.59b)/(5.60b).

One can easily verify that the co-references above would make sense if the indefinites in question take wide scope.

(5.61) a. $(\exists\ dog > \forall\ boy)$

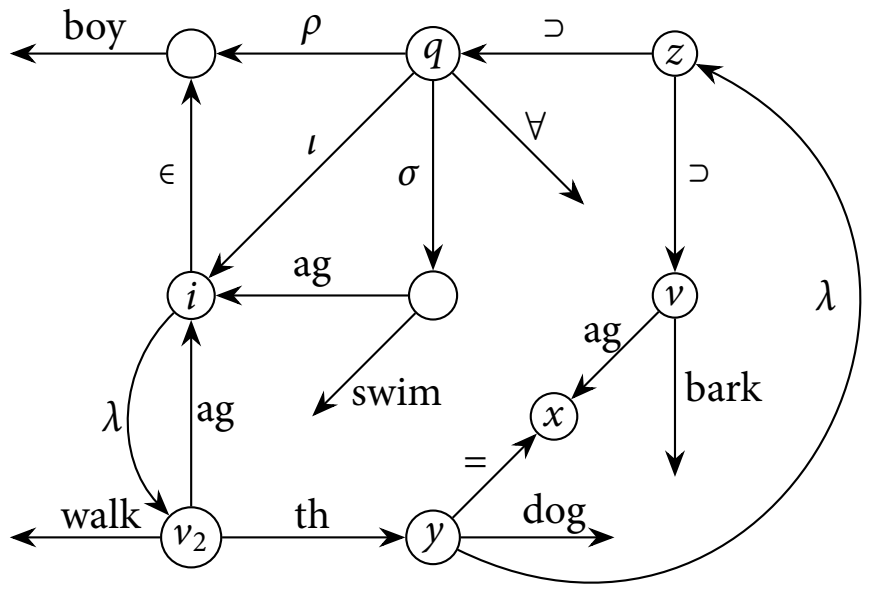

b. $(\exists\ dog > \neg)$

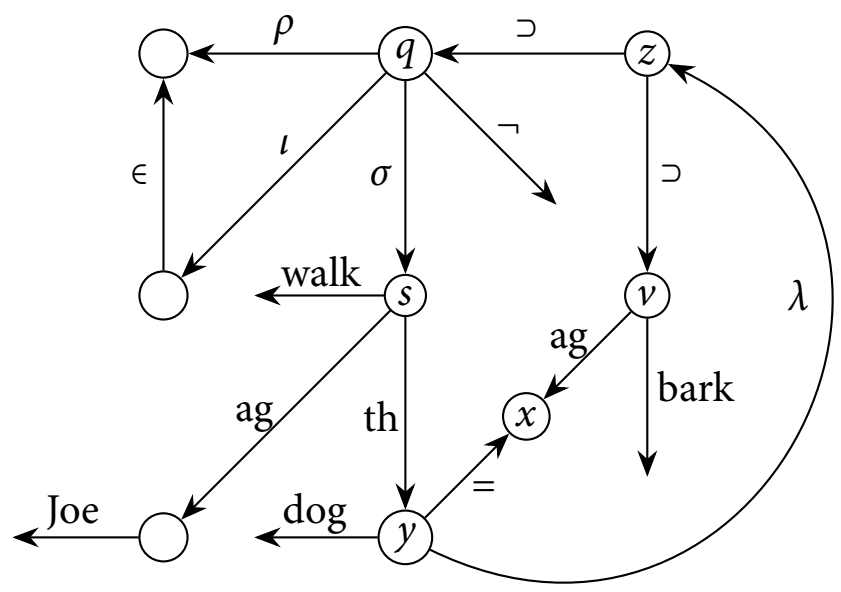

It suffices to note that by scoping $y$ over the conjunction, each conjunct is to be satisfied by a valuation defined on $x$.

*Remark.* We just attributed anaphoric accessibility to the local semantic context of a co-reference antecedent, at the cost of ignoring the fact that an anaphor may reference into, say, double negation, e.g. *it's not the case that Joe didn't sink a boat — it's over there*. But there is the possibility that the anaphor in question relies on the discourse to specify its reference, according to the end remark of Section 3.2.1. Such anaphors gain wider accessibility: *it* picks up the indefinite in *we don't have a decisive answer to that question; even if we had it ...*

Now, turning to the perspective of indefinites, we may also find their scoping options constrained by the resolution of anaphors syntactically depending on them (if any). While relevant examples bear on the bound pronoun problem introduced in Section 5.2.1, we are interested here in a specific observation discussed by Brasoveanu and Farkas (2011) among others under the name "binder roof constraint", namely, an outscoping quantifier caps the scope of an indefinite when it binds into the latter's description:

(5.62) a. **Every boy**$^i$ who walked a dog **he**$_i$ fed swam. ($\forall$ *boy* > $\exists$ *dog*)
b. **Every boy**$^i$ walked a dog **he**$_i$ fed. ($\forall$ *boy* > $\exists$ *dog*)

In both (5.62a&b), *a dog* cannot take scope over the quantifier as *he* picks up each boy; (5.62b), for example, does not mean that there was a dog that every boy walked and fed.

Let us again look at the semgraphs implied by invalid scoping. Recall that for an indefinite to get wide scope, we can either resolve the precedence of its referent, or reach it from the restrictor of the outscoped quantifier. Since both work for (5.62a&b), (5.63) illustrates one for each.

(5.63) a.

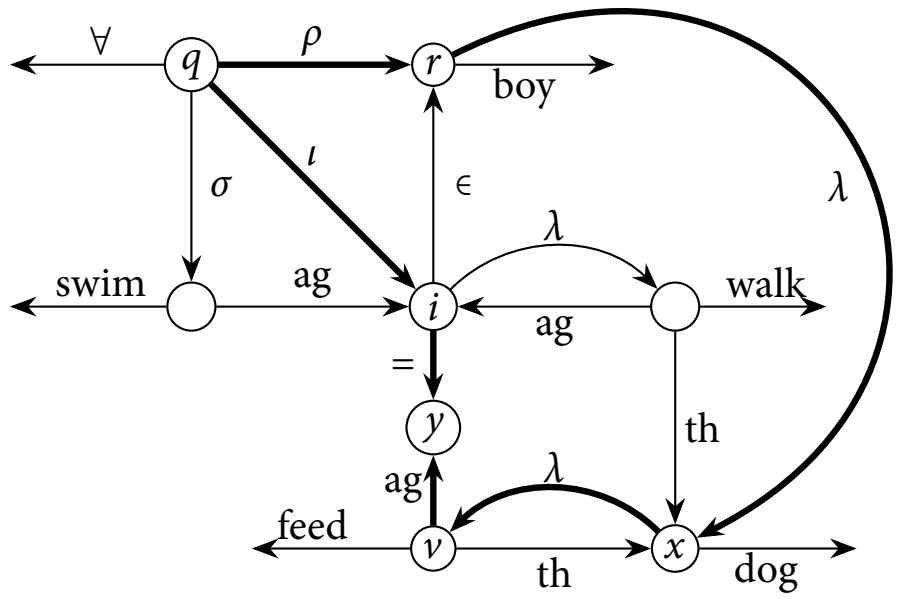

b.

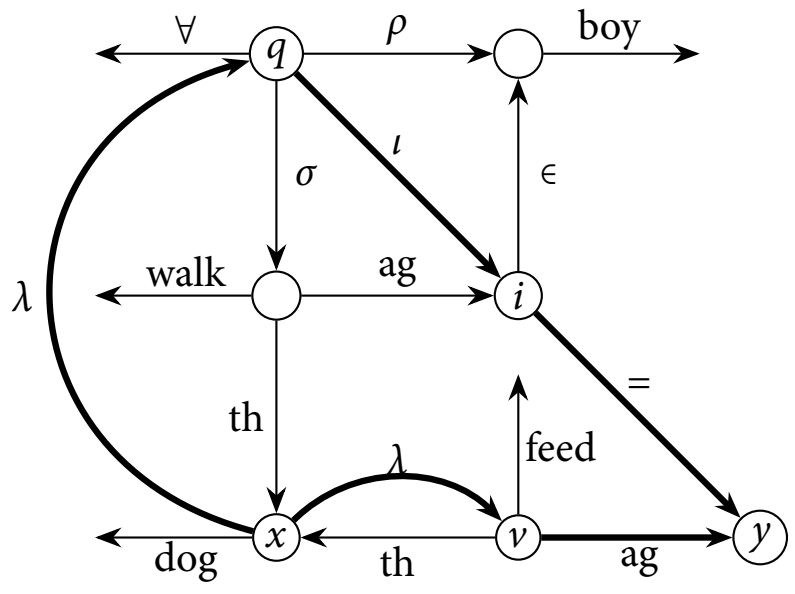

Take (5.63a). Given $\lambda\overrightarrow{rx}$, a valuation $g$ satisfying the semgraph (rooted at $q$) must satisfy the restrictor subgraph and thus be defined on $y$. But $g$ must be undefined on $y$, since $y$ remains unvisited in the iterator subgraph and lies outside the visiting history update by $r$. Hence a contradiction. One can similarly derive the contradiction in (5.63b), and in (5.44), if its indefinite were to take matrix scope.

We note that such contradictions are independent of any model or world of reference. This means that no situation satisfies the respective semgraphs, and therefore no wide-scope-indefinite reading is over-generated for the corresponding sentences. Also, since such contradictions are even independent of choice of lexical tokens (or of unary edge labels, so to speak), they seem to qualify as what Chierchia (2013, pp. 49ff) calls “grammatical triviality”, an idea attributed to Gajewski (2002) for explaining ungrammaticality.

It is clear that our accounts of anaphoric inaccessibility and the binder roof constraint proceed in roughly the same way. That (5.59) and (5.62) have similar problems with valuation dependency is reflected in a shared feature of (5.60) and (5.63): a pair of parallel paths where the one ending in the co-reference equality passes a quantification structure, but the other does not (e.g. $x \to q \to i \to y \parallel x \to v \to y$ in (5.63b)). Thus we can understand the binder roof constraint in terms of anaphoric inaccessibility caused by scoping the descriptive constraint of indefinites.

# Conclusion

Semantic representations based on directed graphs are emerging in the computational linguistics community for their expressivity and computational tractability. When equipped with a construction mechanism that composes semgraphs at the syntax-semantics interface and an interpreter that defines semgraphs' model-theoretical semantics, graph formalisms can prove a powerful tool for research in theoretical semantics.

In this thesis, we developed a graph formalism and explored its empirical applications to issues in plurality and quantification. We began in Chapter 2 by presenting a graph language that uses only monadic second-order variables but covers thematic relations; improves on previous representations of modification, co-reference, plurality, and quantification (as iteration); and introduces to semgraphs intensionality, conjunction (as summation), and disjunction (as choice). By structural induction, we defined for the first time the model-theoretical semantics of semgraphs in terms of graph traversal, where the relative scope of variables arises from their order of valuation.

In Chapter 3 we provided a unification-based mechanism for constructing semgraphs at the syntax-semantics interface. We showed that the presentation of syntax can be greatly simplified if formulated as a function, on par with non-syntactic resolutions, that computes equivalence among referents introduced by linguistic tokens. When implementing this function in categorial grammars, we proposed a partly deterministic alignment between the semgraph and syntactic type of linguistic expressions, allowing us to partly infer one from the other. We automated our syntax-semantics interface for future exploration.

We applied our graph formalism to topics in plurality and quantification in Chapters 4-5. We traced distributivity of various forms and even cumulativity with an upper bound to variants of a quantification structure that explicitly represents the domain of quantification. By representing argument sharing via equality, we addressed compositional challenges in cross-categorial conjunction while capturing the long-existing intuition that all conjunctions create some plurality. We studied the scope permutation of quantificational expressions with a continuized syntax, but, by virtue of our syntax-semantics interface, without resorting to higher-order semantics. With the non-determinism of non-syntactic resolutions, we simplified the treatment of indefinites' exceptional scoping behavior.

*Future work* This thesis leaves much work to be done and raises many questions to be answered. To start with, our graph language covers the semantic essentials usually found in textbook language fragments. The hope is that this language will remain a useful basis as one addresses the issues in plurality and quantification we left untreated or treated only with partial success, or as one extends it to explore new applications in empirical domains other than plurality and quantification. For example, we noted comparatives and measures as a domain that is of theoretical and practical interest. To express semantics pertinent to this domain, one may wonder what semantic dependencies and graph motifs need to be introduced, besides those that can be reused.

Efforts can also be invested in reinventing the infrastructure of our graph formalism. For example, the present semgraph interpreter is only defined on uniquely rooted graphs. But one may wonder in a multi-rooted semgraph, whether there is a meaningful way to deal with the different truth-conditions resulting from interpreting the graph through entering different roots; branching quantification gives an example where we want to conjoin them (Section 4.1.3), but in what situations might they illustrate underspecification?

On the other hand, the present semantic construction mechanism only builds semgraphs for linguistic expressions up to sentences, but we can surely consider how sentence semgraphs can be further merged to represent a discourse (cf. Kamp et al., 2011; Kruijff, 2003). Even for sentence-level semantic construction, at times there arises the need to not only combine semgraphs introduced by linguistic tokens, but to incorporate subgraphs copied from those tokens or from the discourse (Sections 3.2.1, 4.1.2). While graph copying itself can be implemented without difficulty as hyperedge replacement (see Courcelle, 1993; Drewes et al., 1997), the question is whether we can generally predict when to copy what, or it has to be determined case by case.

Finally, besides noticing the correlation between the binding theory constraints and the features of parallel paths created by co-reference (Section 3.2.2), we have not drawn many empirical implications from the formal properties of semgraphs. Judging from our graph language as is, however, we may ask what are the general characteristics of "well-formed" semgraphs. For example, all but few rare semgraphs in this thesis (can) have a *planar* layout, that is, they can be drawn in a plane with no two edges crossing each other. For a simple connected graph to be planar, a necessary condition is that its number $m$ of edges be linear to its number $n$ of vertices (see Diestel, 2016, sec. 4.2 for a precise formulation), whereas maximally $m$ can be quadratic to $n$. Therefore, if planarity indeed plays a role in semgraph well-formedness, we may say that natural language semantics is sparse: it places a certain number of referents in only so many semantic relations, whose number is of the same order as that of the referents; we may wonder what causes this sparsity and what predictions can be made of it.

More abstractly, we can approach semgraph well-formedness from a language-theoretical point of view. Consider the set of all the token semgraphs in some lexicon and all the semgraphs composable out of them according to syntax and non-syntactic resolutions. One may wonder how to specify a graph grammar that generates this set of semgraphs and how to characterize the complexity properties of such grammars (see Courcelle and Engelfriet, 2012). In comparison with higher-order logic, what does a representation formalism more restrictive in expressivity have to say about the limitation of natural language semantics?